%% file: paper.tex
\documentclass{tlp}
\usepackage{times}

\usepackage[latin1]{inputenc}
\usepackage{amsmath,amssymb}
\usepackage{latexsym}
\usepackage{listings}
\lstdefinelanguage{myown} %% language name
                  {morekeywords={procedure, function, for, endfor, if, endif, then, else, elseif, while, endwhile, do, return}}

\input{macro}
\usepackage{tlpmacro}
\input{fig_macro}

\title{Graphs and colorings for answer set programming}
\author[Kathrin Konczak and Thomas Linke and Torsten Schaub]%
       {Kathrin Konczak and Thomas Linke and Torsten Schaub%
         \thanks{~Affiliated with the
           School of Computing Science at
           Simon Fraser University,
           Burnaby, Canada.} \\
  Institut f\"ur Informatik,
  Universit\"at Potsdam      \\  
  Postfach 90~03~27,
  D--14439 Potsdam,
  Germany                    \\
  \{konczak,linke,torsten\}@cs.uni-potsdam.de}

  \submitted{14 August  2003}
  \revised{22 July 2004}
  \accepted{24 December  2004}

\begin{document}

\maketitle
\input{abstract}

\input{introduction}

\input{background}
\input{graphcolors}

\input{answerships}

\input{operations}
\input{wfs}
\input{discussion}

\input{conclusion}
\input{acknowledgements}

\appendix

\input{auxiliary}
\input{tools}
\input{proofs}

\bibliographystyle{acmtrans}
%% \bibliography{lit,kk}

\end{document}

%% file: macro.tex
\newcommand{\CO}[1]{\ensuremath{\mathcal{O}(#1)}}

\renewcommand{\r}{\ensuremath{r}}
\newcommand{\rp}{\ensuremath{r'}}

\newcommand{\LPif}{\leftarrow}
\newcommand{\LPnot}{\ensuremath{not}}

\newcommand{\head}[1]{\ensuremath{\mathit{head}(#1)}}
\newcommand{\body}[1]{\ensuremath{\mathit{body}(#1)}}
\newcommand{\pbody}[1]{\ensuremath{\mathit{body}^+(#1)}}
\newcommand{\nbody}[1]{\ensuremath{\mathit{body}^-(#1)}}

\newcommand{\lfp}[1]{\ensuremath{\mathit{lfp}(#1)}}

\newcommand{\bg}[1]{\ensuremath{\Gamma_{#1}}}
\newcommand{\bgv}{\ensuremath{\Pi}}
\newcommand{\bgezero}{\ensuremath{E_0}}
\newcommand{\bgeone}{\ensuremath{E_1}}

\newcommand{\subgraph}[1]{$#1$-subgraph}

\newcommand{\agrounded}{supported}
\newcommand{\anotgrounded}{unsupported}
\newcommand{\ablocked}{blocked}
\newcommand{\anotblocked}{unblocked}

\newcommand{\groundedZeroDAG}{0-DASG}

\newcommand{\col}{\ensuremath{C}}
\newcommand{\colplus}{C_\oplus}
\newcommand{\colminus}{C_\ominus}

\newcommand{\fcol}[1]{\ensuremath{C(#1)}}
\newcommand{\cplus}{\oplus}
\newcommand{\cminus}{\ominus}

\newcommand{\emptyC}{\ensuremath{(\emptyset,\emptyset)}}

\newcommand{\dlv}{\texttt{dlv}}
\newcommand{\smodels}{\texttt{smodels}}
\newcommand{\nomore}{\texttt{noMoRe}}

\newcommand{\RDGraph}{{\small\textsl{RDG}}}
\renewcommand{\groundedZeroDAG}{support graph}
\newcommand{\nafo}[0]{\mathit{not}}
\newcommand{\naf}[0]{\nafo\;}
\newcommand{\reduct}[2]{\ensuremath{#1^{#2}}}
\newcommand{\CnO}[0]{\ensuremath{\mathit{Cn}}}
\newcommand{\Cn}[1]{\ensuremath{\CnO(#1)}}
\newcommand{\To}[1]{\ensuremath{T_{#1}}}
\newcommand{\T}[2]{\To{#1}(#2)}
\newcommand{\TiO}[2]{\To{#2}^{#1}}
\newcommand{\Ti}[3]{\TiO{#1}{#2}(#3)}
\newcommand{\GR}[2]{\ensuremath{R_{#1}(#2)}} % {\ensuremath{R(#1,#2)}}

\newcommand{\fOPprop}[2]{\ensuremath{\mathcal{P}_{#1}(#2)}}  
\newcommand{\OPprop}[1]{\ensuremath{\mathcal{P}_{#1}}}  
\newcommand{\fOPpropast}[2]{\ensuremath{\mathcal{P}_{#1}^\ast(#2)}}  
\newcommand{\OPpropast}[1]{\ensuremath{\mathcal{P}^\ast_{#1}}} 

\newcommand{\fOPchoose}[4]{\ensuremath{\mathcal{C}_{#3}^{#2}(#4)}}%{\ensuremath{CP_{#2}(#1)}}
\newcommand{\OPchoose}{\ensuremath{\mathcal{C}}}
\newcommand{\AS}[2]{\ensuremath{\mathit{AS}_{#1}(#2)}}
\newcommand{\fOPMaxGround}[2]{\ensuremath{\mathcal{U}_{#1}(#2)}}
\newcommand{\OPMaxGround}{\ensuremath{\mathcal{U}}}

\newcommand{\atoms}[1]{\ensuremath{atoms(\Pi)}}
\newcommand{\atm}{\ensuremath{Atm}}
\newcommand{\fittingo}[1]{\ensuremath{\Phi_{#1}}}
\newcommand{\fitting}[3]{\fittingo{#1}(#2,#3)}

\newcommand{\fittingPos}[3]{\fittingo{#1}^{+}(#2,#3)}
\newcommand{\fittingNeg}[3]{\fittingo{#1}^{-}(#2,#3)}

 \newcommand{\weakProp}[2]{\ensuremath{\mathcal{T}^\ast_{#1}(#2)}}
 \newcommand{\weakS}[2]{\ensuremath{\mathcal{V}_{#1}(#2)}}
 \newcommand{\fweakS}{\ensuremath{\mathcal{V}}}
\newcommand{\weakChoose}[3]{\ensuremath{\mathcal{D}_{#1}^{#2}(#3)}}

\newcommand{\classC}{\ensuremath{\mathbb{C}}} %  {\ensuremath{\mathfrak{C}}}
\newcommand{\AC}[2]{\ensuremath{\mathit{AC}_{#1}(#2)}}

\newcommand{\fOPN}[2]{\ensuremath{\mathcal{N}_{#1}(#2)}}
\newcommand{\OPN}[1]{\ensuremath{\mathcal{N}_{#1}}}

\newcommand{\fOPV}[2]{\ensuremath{\mathcal{V}_{#1}(#2)}}
\newcommand{\OPV}[1]{\ensuremath{\mathcal{V}_{#1}}}

\newcommand{\fOPTast}[2]{\ensuremath{\mathcal{T}_{#1}^{\ast}(#2)}}

\newcommand{\fPUast}[2]{\ensuremath{(\OPprop{}\OPMaxGround{})^\ast_{#1}(#2)}}
\newcommand{\PUast}[1]{\ensuremath{(\OPprop{}\OPMaxGround{})^\ast_{#1}}}

\newcommand{\fPVast}[2]{\ensuremath{(\OPprop{}\OPV{})^\ast_{#1}(#2)}}

\newcommand{\GUS}[2]{\ensuremath{{\bf U}_{#1}(#2)}}

%% file: fig_macro.tex
\newcommand{\PengRDG}[6]%
{{\begin{picture}(120,180)(-15,-20)

\put(10,140){\makebox(0,0){\ensuremath{#1}}}
\put(10,140){\circle{20}}
\put(80,140){\makebox(0,0){\ensuremath{#2}}}
\put(80,140){\circle{20}}
\put(10,70){\makebox(0,0){\ensuremath{#4}}}
\put(10,70){\circle{20}}
\put(80,70){\makebox(0,0){\ensuremath{#3}}}
\put(80,70){\circle{20}}

\put(10,0){\makebox(0,0){\ensuremath{#6}}}
\put(10,0){\circle{20}}
\put(80,0){\makebox(0,0){\ensuremath{#5}}}
\put(80,0){\circle{20}}

\put(10,130){\vector(0,-1){50}}
\put(80,130){\vector(0,-1){50}}
\put(20,70){\vector(1,0){50}}
\put(70,70){\vector(-1,0){50}}
\put(20,140){\vector(1,0){50}}        %% added

\put(80,10){\vector(0,1){50}}         %% changed
\put(72.5,62.5){\vector(-1,-1){55}}
\put(10,60){\vector(0,-1){50}}
\put(20,-10){\oval(20,20)[bl]}
\put(20,-10){\oval(20,20)[br]}
\put(20,-10){\oval(20,20)[tr]}
\put(20,0){\vector(-1,0){0}}

\put(5,40){\makebox(0,0){$0$}}
\put(5,110){\makebox(0,0){$0$}}
\put(85,40){\makebox(0,0){$0$}}
\put(85,110){\makebox(0,0){$0$}}
\put(45,80){\makebox(0,0){$1$}}
\put(40,40){\makebox(0,0){$0$}}
\put(35,-10){\makebox(0,0){$1$}}
\put(45,150){\makebox(0,0){$0$}}        %% added

\end{picture}}}

%% file: abstract.tex
\begin{abstract}
We investigate the usage of rule dependency graphs and their colorings for
characterizing and computing answer sets of logic programs.
This approach provides us with insights into the interplay between rules
when inducing answer sets.
We start with different characterizations of answer sets in terms of totally
colored dependency graphs that differ in graph-theoretical aspects.
We then develop a series of operational characterizations of answer
sets in terms of operators on partial colorings.
In analogy to the notion of a derivation in proof theory,
our operational characterizations are expressed as (non-deterministically formed)
sequences of colorings,
turning an uncolored graph into a totally colored one.
In this way, we obtain an operational framework in which different
combinations of operators result in different formal properties.
Among others, we identify the basic strategy employed by the \texttt{noMoRe}
system and justify its algorithmic approach.
Furthermore, we distinguish operations corresponding to Fitting's operator as
well as to well-founded semantics.
\end{abstract}
\begin{keywords}
answer set programming, 
operational semantics, 
answer set computation, 
graph-based characterization
\end{keywords}
%%% Local Variables: 
%%% mode: latex
%%% TeX-master: "paper"
%%% End: 

%% file: introduction.tex
\section{Introduction}
\label{sec:introduction}

Graphs constitute a fundamental tool within computing science, in particular, in
programming languages, where graphs are often used for analyzing a program's
behavior.
Clearly, this also applies to logic programming.
For instance, Prolog's procedural semantics is intimately connected to the concept
of SLD-trees~\cite{lloyd87}.
For further analysis, like profiling, other types of graphs, such as call
graphs, play an important role during program development.
Similarly, in alternative semantics of logic programming, like \emph{answer set
programming}~\cite{gellif88b,gellif91a}, graphs have been used for deciding
whether answer sets exist~\cite{fages94a,bargel94a}.

We take the application of graphs even further and elaborate in this paper upon
an approach to using graphs as the underlying computational model for computing
answer sets.
To this end, we build upon and largely extend the theoretical foundations introduced
in~\cite{linke01a,ankoli02a}.
Our approach has its roots in default logic~\cite{reiter80}, where extensions
are often characterized through their (unique) set of generating default rules.
Accordingly, we are interested in characterizing answer sets by means of their
set of generating rules.
For determining whether a rule belongs to this set, we must verify that each
positive body atom is derivable and that no negative body atom is derivable.
In fact, an atom is derivable if the set of generating rules includes a rule
having the atom as its head; or conversely, an atom is not derivable if there is
no rule among the generating rules that has the atom as its head.
Consequently, the formation of the set of generating rules amounts to
resolving positive and negative dependencies among rules.
For capturing these dependencies, we take advantage of the concept of a
\emph{rule dependency graph},
wherein each node represents a rule of the underlying program
and two types of edges stand for the aforementioned positive and negative
rule dependencies, respectively.~\footnote{This type of graph was called ``block graph'' in~\cite{linke01a}.}
For expressing the applicability status of rules,
that is, whether a rule belongs to a set of generating rules or not,
we label, or as we say \emph{color}, the respective nodes in the graph.
In this way, an answer set can be expressed by a total coloring of the rule
dependency graph.
Of course, in what follows, we are mainly interested in the inverse, that is,
when does a graph coloring correspond to an answer set of the underlying
program; and, in particular, how can we compute such a total coloring.

Generally speaking,
graphs provide a formal device for making structural properties of an underlying
problem explicit.
In this guise, we start by identifying graph structures that capture structural
properties of logic programs and their answer sets.
As a result, we obtain several characterizations of answer sets in terms of
totally colored dependency graphs that differ in graph-theoretical aspects.
To a turn, we build upon these characterizations in order to develop an
operational framework for answer set formation.
The idea is to start from an uncolored rule dependency graph and to employ
specific operators that turn a partially colored graph gradually into a totally
colored one that represents an answer set.
This approach is strongly inspired by the concept of a derivation, in
particular, by that of an SLD-derivation~\cite{lloyd87}.
Accordingly, a program has a certain answer set iff there is a sequence of
operations turning the uncolored graph into a totally colored one that
provably corresponds to the answer set.

Our paper is organized as follows.
Section~\ref{sec:background} briefly summarizes concepts of logic programming.
Section~\ref{sec:graphcol} lays the formal foundations of our approach by
introducing its basic graph-theoretical instruments.
While the following Section~\ref{sec:check} addresses characterizations of
answer sets through totally colored graphs,
Section~\ref{sec:operations} deals with operational characterizations of answer
sets.
Section~\ref{sec:wfs} identifies relationships with Fitting's and well-founded
semantics. 
Section~\ref{sec:discussion} discusses the approach, in particular in the light
of related work.
We conclude our contribution in Section~\ref{sec:conclusion}.
Appendix~\ref{sec:auxiliary} and~\ref{sec:inductivedef} contain auxiliary
material, needed in the proofs given in Appendix~\ref{sec:proofs}. 

%%% Local Variables: 
%%% mode: latex
%%% TeX-master: "paper"
%%% End: 

%% file: background.tex
\section{Definitions and notation}
\label{sec:background}

We assume a basic familiarity with alternative semantics of logic
programming~\cite{lifschitz96a}.
A (normal) \emph{logic program} is a finite set of rules of the form
\begin{equation}\label{eqn:rule}
p_0\LPif p_1,\dots,p_m,\naf p_{m+1},\dots,\naf p_n,
\end{equation}
where $n\geq m\geq 0$, and each $p_i$ $(0\leq i\leq n)$ is an \emph{atom}.
The set of all atoms is denoted by $\atm$.
Given a rule $r$ of form~(\ref{eqn:rule}),
we let $\head{r}$ denote the \emph{head}, $p_0$, of $r$
and
$\body{r}$ the \emph{body},
\(
\{p_1,\dots,p_m,\ \naf p_{m+1},\dots,\naf p_n\}
\),
of $r$.
Furthermore, let
\(
\pbody{r}
=
\{p_1,\dots, p_m\}
\)
and
\(
\nbody{r}
=
\{p_{m+1},\dots, p_n\}
\).
For a program $\Pi$, we write
\(
\head{\Pi}=\{\head{r}\mid r\in\Pi\}
\).
A program is called \emph{basic} if $\nbody{r}=\emptyset$ for all its rules.
The \emph{reduct}, $\reduct{\Pi}{X}$, of a program $\Pi$ \emph{relative to} a
set $X$ of atoms is defined by
\begin{equation}
\label{eq:GLreduct}
\reduct{\Pi}{X}
=
\{\head{r}\LPif\pbody{r}\mid r\in\Pi\text{ and }\nbody{r}\cap X=\emptyset\}.
\end{equation}
A set of atoms $X$ is \emph{closed under} a basic program $\Pi$ if for any
$r\in\Pi$, $\head{r}\in X$ whenever $\pbody{r}\subseteq X$.
The smallest set of atoms which is closed under a basic program $\Pi$ is denoted
by $\Cn{\Pi}$.
With these formalities at hand,
we can define \emph{answer set semantics} for logic programs:
A set $X$ of atoms is an \emph{answer set} of a program $\Pi$
if
$\Cn{\reduct{\Pi}{X}}=X$.
We use $\AS{}{\Pi}$ for denoting the set of all answer sets of a program $\Pi$.

An alternative inductive characterization for operator \CnO\ can be obtained by
appeal to an \emph{immediate consequence operator}~\cite{lloyd87}.
Let $\Pi$ be a basic program and $X$ a set of atoms.
The operator $\To{\Pi}$ is defined as follows:
\begin{equation}
\label{eq:T}
\T{\Pi}{X} = \{\head{r}\mid r\in\Pi\text{ and }\body{r}\subseteq X\}
\ .
\end{equation}
Iterated applications of $\To{\Pi}$ are written as $\TiO{j}{\Pi}$ for
$j\geq 0$, where
\(
\Ti{0}{\Pi}{X}=X
\)
and
\(
\Ti{i}{\Pi}{X}=\T{\Pi}{\Ti{i-1}{\Pi}{X}}
\)
for $i\geq 1$.
It is well-known that
\(
\Cn{\Pi}=\bigcup_{i\geq 0}\Ti{i}{\Pi}{\emptyset}
\),
for any basic program $\Pi$.
Also, for any answer set $X$ of program $\Pi$, it holds that
\(
X=\bigcup_{i\geq 0}\Ti{i}{\reduct{\Pi}{X}}{\emptyset}
\).

Another important concept is that of the \emph{generating rules} of an answer set.
The set $\GR{\Pi}{X}$ of generating rules of a set $X$ of atoms from program
$\Pi$ is defined as
\begin{equation}
\label{eq:GR}
\GR{\Pi}{X}
=
\{r\in\Pi\mid\pbody{r}\subseteq X\text{ and }\nbody{r}\cap X=\emptyset\}
\ .
\end{equation}
In fact, one can show that a set of atoms $X$ is an answer set of a program
$\Pi$ iff $X=Cn((\GR{\Pi}{X})^\emptyset)$ (see Theorem~\ref{thm:as:consGR};
note that $\Pi^\emptyset=\{\head{r}\LPif\pbody{r}\mid r\in\Pi\}$ for any program $\Pi$).

%%% Local Variables: 
%%% mode: latex
%%% TeX-master: "paper"
%%% End: 

%% file: graphcolors.tex
\section{Graphs and colorings}
\label{sec:graphcol}

This section lays the formal foundations of our approach by introducing its basic
graph-theoretical instruments.

A \emph{graph} is a pair $(V,E)$ where $V$ is a set of \emph{vertices} and 
$E\subseteq V\times V$ a set of (directed) \emph{edges}.
A \emph{path} from $x$ to $y$ in $(V,E)$ for $x,y \in V$ is a sequence
$x_1,\ldots,x_n$ 
such that $x=x_1, y=x_n$, $(x_i,x_{i+1})\in E$ for $1\leq i < n$, and the
elements $x_i$ are pairwise disjoint. 
A set of edges $E$ contains a cycle if there is a nonempty set 
$\{x_i\mid i\in\{0,\ldots,n\}\}$ of vertices
such that $(x_i,x_{i+1})\in E$ for $i\in\{0,\ldots,n-1\}$ and $(x_n,x_0)\in E$. 
A graph $(V,E)$ is \emph{acyclic} if $E$ contains no cycles.
For $W\subseteq V$, we denote $E \cap (W \times W)$ by $E|_W$.
Also, we abbreviate the induced subgraph $G=(V\cap W,E|_W)$ of $(V,E)$ by
$G|_W$.
A \emph{labeled graph} is a graph with an associated labeling function
\(
\ell: E \to L
\)
for some set of labels $L$.
In view of our small label set $L=\{0,1\}$ (see below), we leave $\ell$ and
$L$ 
implicit and denote such labeled graphs by triples $(V,E_0,E_1)$,
where
\(
E_i=\{e\in E\mid\ell(e)=i\}
\)
for $i=0,1$.
An \emph{i-subgraph} of $(V,E_0,E_1)$ is a graph $(W,F)$ such that $W\subseteq
V$ and $F \subseteq E_i|_W$ for $i=0,1$.
\footnote{Note that an $i$-subgraph is not an induced graph.}
An {\emph{i-path}} from $x$ to $y$ in $(V,E_0,E_1)$ is a path from $x$ to $y$
in $(V,E_i)$  for $x,y\in V$ and $i=0,1$.

In the context of logic programming, we are interested in graphs reflecting
dependencies among rules.
%---------------------------------------------------------------
\begin{definition}\label{def:blockgraph:redundant}
Let $\Pi$ be a logic program.

The rule dependency graph (\RDGraph) $\bg{\Pi} = ( \bgv, \bgezero,\bgeone)$ of
$\Pi$ is a labeled  graph with
\begin{eqnarray*}
  \bgezero &=& \left\{ (\r,\rp) \mid \r,\rp \in \Pi , \head{\r} \in \pbody{\rp} \right\}\/;
  \\
  \bgeone  &=& \left\{ (\r,\rp) \mid \r,\rp \in \Pi , \head{\r} \in \nbody{\rp} \right\}\/.
\end{eqnarray*}
\end{definition}
%---------------------------------------------------------------
%
We omit the subscript $\Pi$ from $\bg{\Pi}$ whenever the underlying program is
clear from the context.
We follow~\cite{papsid94} in distinguishing between $0$- and $1$-edges.
Observe that several programs may have isomorphic \RDGraph s.
For example, 
$\Pi=\{a \LPif b, c \LPif a\}$ 
and
$\Pi'=\{a \LPif,  c \LPif a\}$
have isomorphic \RDGraph{s}.

%------------------------------------------------------------------
\begin{example}\label{ex:penguin}
Consider the logic program $\Pi_{\ref{ex:penguin}}=\{r_1,\ldots,r_6\}$,
comprising the following rules:
\[
\begin{array}{lrcl}
r_1: & p & \LPif &                \\
r_2: & b & \LPif & p              \\
r_3: & f & \LPif & b, \LPnot\; f'\\
r_4: & f'& \LPif & p, \LPnot\; f  \\
r_5: & b & \LPif & m              \\
r_6: & x & \LPif & f,f',\LPnot\; x
\end{array}
\]
The \RDGraph{} of $\Pi_{\ref{ex:penguin}}$ is given as follows:
\[
\bg{\Pi_{\ref{ex:penguin}}}
=
(
\;
\Pi_{\ref{ex:penguin}},
\
\{(r_1,r_2),(r_1,r_4),(r_2,r_3),(r_3,r_6),(r_4,r_6),(r_5,r_3)\},
\
\{(r_3,r_4),(r_4,r_3),(r_6,r_6)\}
\;
)
\]
It is depicted graphically in Figure~\ref{fig:penguin:rdg}.
%------------------------------------------------------------------
\begin{figure}[htbp]
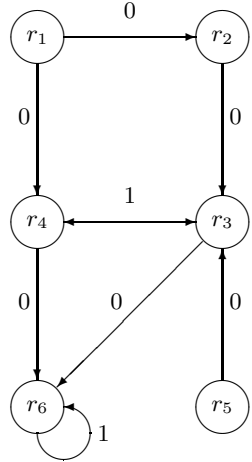

  \setlength{\unitlength}{1pt}
  \centering
  \PengRDG{r_1}{r_2}{r_3}{r_4}{r_5}{r_6}
  \caption{The \RDGraph{} of logic program $\Pi_{\ref{ex:penguin}}$.}
  \label{fig:penguin:rdg}
\end{figure}
%------------------------------------------------------------------
%
For instance,
\(
(
\{r_1,r_2,r_3,r_4\},
\{(r_1,r_2)\}
)
\)
is a $0$-subgraph of \bg{\Pi_{\ref{ex:penguin}}}
and
\(
(
\{r_5,r_6\},
\{(r_6,r_6)\}
)
\)
is a $1$-subgraph of \bg{\Pi_{\ref{ex:penguin}}}.
\end{example}

We call $\col$ a \emph{(partial)} \emph{coloring} of $\bg{\Pi}$ if $\col$
is a partial mapping $\col:\Pi \rightarrow \{\cplus,\cminus\}$.
We call $C$ a \emph{total} coloring, if $C$ is a total mapping.
Intuitively, the colors $\cplus$ and $\cminus$ indicate whether a rule is
supposedly applied or blocked, respectively.
We sometimes denote the set of all vertices colored with $\cplus$ or $\cminus$ by
$\colplus$ or $\colminus$, respectively.
That is, $\colplus=\{\r \mid \fcol{\r}=\cplus\}$ and $\colminus=\{\r \mid
\fcol{\r}=\cminus\}$.
If $\col$ is total, $(\colplus,\colminus)$ is a binary partition of $\Pi$.
That is, $\Pi=\colplus\cup\colminus$ and $\colplus\cap\colminus=\emptyset$.
Accordingly, we often identify a coloring $C$ with the pair $(\colplus,\colminus)$.
A {partial} coloring $C$ induces a pair $(\colplus,\colminus)$ of sets
such that 
$\colplus\cup\colminus\subseteq\Pi$ and $\colplus\cap\colminus=\emptyset$.
For comparing partial colorings, $C$ and $C'$, we define $C\sqsubseteq C'$, if
$\colplus\subseteq\colplus'$ and $\colminus\subseteq\colminus'$.
The ``empty'' coloring $(\emptyset,\emptyset)$ is the $\sqsubseteq$-smallest
coloring.
Accordingly, we define $C\sqcup C'$ as
$(\colplus\cup\colplus',\colminus\cup\colminus')$.%
\footnote{We use ``squared'' relation symbols, like $\sqsubseteq$ or $\sqcup$
  when dealing with partial colorings.}
We denote the set of all partial colorings of a \RDGraph\ \bg{\Pi} by
$\classC_{\bg{\Pi}}$.
For readability, we often omit the index \bg{\Pi} and simply write \classC, whenever
this is clear from the context.

If $C$ is a coloring of $\bg{\Pi}$, we call the pair $(\bg{\Pi},C)$ a
\emph{colored} \RDGraph{}.
For example, ``coloring'' the \RDGraph{} of $\Pi_{\ref{ex:penguin}}$ from
Example~\ref{ex:penguin} with
\footnote{For the sake of uniqueness, we label the coloring with the equation number.}
\begin{equation}
  \label{ex:penguin:coloring}
  C_{\ref{ex:penguin:coloring}}
  =
  (
  \{r_1,r_2\},
  \
  \{r_6\}
  )
\end{equation}
yields the colored graph given in Figure~\ref{fig:penguin:rdg:colored}.
%
%------------------------------------------------------------------
\begin{figure}[htbp]
  \setlength{\unitlength}{1pt}
  \centering
  \PengRDG{\cplus}{\cplus}{r_3}{r_4}{r_5}{\cminus}
  \caption{The (partially) colored \RDGraph{} $(\bg{\Pi_{\ref{ex:penguin}}},C_{\ref{ex:penguin:coloring}})$.}
  \label{fig:penguin:rdg:colored}
\end{figure}
%------------------------------------------------------------------
%
For simplicity, when coloring, we replace the label of a node by the respective
color.
(Cf.~Figure~\ref{fig:penguin:rdg} for the underlying uncolored graph.)
Observe that our conception of coloring is nonstandard insofar that
adjacent vertices may be colored with the same color.
We are sometimes interested in the subgraph
$\bg{\Pi}|_{\colplus\cup\colminus}$ induced by the colored nodes.
%we abbreviate it by $\bg{\Pi}|_C$.
%
Restricting \bg{\Pi_{\ref{ex:penguin}}} to the nodes colored in
Figure~\ref{fig:penguin:rdg:colored}, yields the \RDGraph{}
\(
(
\{r_1,r_2,r_6\},
\{(r_1,r_2)\},
\{(r_6,r_6)\}
)
\).

The central question addressed in this paper is how to characterize and compute the
total colorings of \RDGraph{}s that correspond to the answer sets of an underlying
program.
In fact, the colorings of interest can be distinguished in a straightforward way.
Let $\Pi$ be a logic program along with its \RDGraph{} $\bg{}$.
Then, for every answer set $X$ of $\Pi$, define an \emph{admissible coloring}%
\footnote{The term ``admissible coloring'' was coined in~\cite{bricos99};
  they were referred to as ``application colorings'' in~\cite{linke01a}.
  Note that both colorings concepts are originally defined in purely
  graph-theoretical terms.
  Here, we simply adopt this term for distinguishing colorings corresponding to
  answer sets of an underlying program (cf.~Section~\ref{sec:discussion}).}
$\col$ of \bg{} as
\begin{equation}
  \label{eq:admissiblecoloring}
  \col=(\GR{\Pi}{X},\Pi\setminus\GR{\Pi}{X})
  \ .
\end{equation}
By way of the respective generating rules,
we associate with any program a set of admissible colorings whose members are
in one-to-one correspondence with its answer sets.
Any admissible coloring is total; 
furthermore, we have $X=\head{\colplus}$.
We use $\AC{}{\Pi}$ for denoting the set of all admissible colorings of a
\RDGraph{} $\bg{\Pi}$.

For a partial coloring $\col$, we define $\AC{\Pi}{\col}$ as the set
of all admissible colorings of $\bg{\Pi}$ compatible with $\col$.
Formally, 
given the \RDGraph{}  \bg{} of a logic program $\Pi$ 
and a partial coloring \col{} of \bg{},
define
%---------------------------------------------------------------
\begin{equation}
\label{eq:def:AC}
  \AC{\Pi}{\col}
  =
  \{C'\in \AC{}{\Pi}\mid\col\sqsubseteq\col'\}
  \ .
\end{equation}
%---------------------------------------------------------------
%
Clearly, $\col_1\sqsubseteq\col_2$ implies
$\AC{\Pi}{\col_1}\supseteq\AC{\Pi}{\col_2}$.
Observe also that a partial coloring \col{} is extensible to an admissible
one $C'$, that is, $\col\sqsubseteq\col'$, iff \AC{\Pi}{\col} is non-empty.
For a total coloring $\col$, $\AC{\Pi}{\col}$ is either empty or singleton.
Regarding program $\Pi_{\ref{ex:penguin}}$ and coloring
$C_{\ref{ex:penguin:coloring}}$, 
we get
\(
\AC{\Pi_{\ref{ex:penguin}}}{C_{\ref{ex:penguin:coloring}}}
=
\AC{}{\Pi_{\ref{ex:penguin}}}
=
\{
(\{r_1,r_2,r_3\},\{r_4,r_5,r_6\}),
(\{r_1,r_2,r_4\},\{r_3,r_5,r_6\})
\}
\)
(see also Figure~\ref{fig:penguin:rdg:colored:total} below).

Accordingly, we define $\AS{\Pi}{\col}$ as the set of all answer sets $X$ of
$\Pi$ compatible with partial coloring $\col$.
%---------------------------------------------------------------
\begin{equation}
\label{eq:def:AS}
  \AS{\Pi}{\col}
  =
  \{X\in \AS{}{\Pi}
    \mid
    \colplus\subseteq\GR{\Pi}{X}\text{ and }\colminus\cap\GR{\Pi}{X}=\emptyset
  \}.
\end{equation}
%---------------------------------------------------------------
%
Note that $\head{\colplus}\subseteq X$ for any answer set $X\in\AS{\Pi}{\col}$
(cf.~Theorem~\ref{thm:X:headcolplus}).
Otherwise, similar considerations apply to \AS{\Pi}{\col} as made above for
\AC{\Pi}{\col}.
As regards program $\Pi_{\ref{ex:penguin}}$ and coloring $C_{\ref{ex:penguin:coloring}}$,
we get
\(
\AS{\Pi_{\ref{ex:penguin}}}{C_{\ref{ex:penguin:coloring}}}
=
\AS{}{\Pi_{\ref{ex:penguin}}}
=
\{\{b,p,f\},\{b,p,f'\}\}
\).
It is noteworthy that due to the one-to-one correspondence between
$\AS{\Pi}{C}$ and $\AC{\Pi}{C}$ (cf. Theorem~\ref{thm:AS:AC}), one can replace
one by the other in most subsequent results.
Often it is simply a matter of simplicity which formulation is used.

We need the following concepts for describing a rule's status of applicability
in a colored \RDGraph.
%
%---------------------------------------------------------------
\begin{definition}\label{def:acol:vertexproperties}
  Let $\bg{}=(\bgv,\bgezero,\bgeone)$ be the \RDGraph{} of logic program  $\Pi$
  and
  $\col$ be a partial coloring of $\bg{}$.
  
  For $\r\in\bgv$, we define:
  \begin{enumerate}
  \item \r\ is \agrounded\ in $(\bg{},\col)$,
    if
    \(
    \pbody{\r}\subseteq\{\head{\rp}\mid (\rp,\r)\in\bgezero,
    \rp\in\colplus\}     
    \);
  \item \r\ is \anotgrounded\ in $(\bg{},\col)$,
    if
    \(
    \{\rp\mid (\rp,\r)\in \bgezero,\head{\rp}=q\}\subseteq \colminus
    \) 
    for some $q\in\pbody{\r}$;
  \item \r\ is \ablocked\ in $(\bg{},\col)$,
    if
    $\rp \in \colplus$ for some $(\rp,\r)\in \bgeone$;
  \item \r\ is \anotblocked\ in $(\bg{},\col)$,
    if
    $\rp \in \colminus$ for all $(\rp,\r)\in \bgeone$.
  \end{enumerate}
\end{definition}
%---------------------------------------------------------------
%
For \r\ and \rp\ as given in Condition~3, we say that \r\ is \ablocked{} by \rp.
Whenever \col{} is total, a rule is unsupported or unblocked iff it is not
supported or not blocked, respectively.
Note that the qualification $(\rp,\r)\in\bgezero$ could be safely removed
from Condition~1 and~2; we left it in for stressing the symmetry among the first two
and the last two conditions.
Observe that all four properties are decidable by looking at the immediate
predecessors in the graph.
With a slightly extended graph structure, they can be expressed in
purely graph-theoretical terms, without any reference to the heads and bodies of
the underlying rules.%
\footnote{For details on these pure graph-theoretical characterization, we refer
  the reader to a companion paper~\cite{lianko02a}, dealing with the system
  \texttt{noMoRe}.}
%% UNCOMMENTED (TS) 03-04-27
%% In fact, for our theoretical purposes, the usage of the projections like \head{} and \pbody{}
%% simplifies the description considerably.
%% For instance, no special care has to be taken to distinguish facts from rules
%% whose positive body literals do not occur among the heads of the program.
%% Also, rules with multiple positive body literals some of which occur as heads in
%% several rules are handled implicitly.

For convenience,
let us introduce the following sets of rules.
%
%---------------------------------------------------------------
\begin{definition}\label{def:acol:vertexproperties:sets}
  Let $\bg{}$ be the \RDGraph{} of logic program  $\Pi$
  and
  $\col$ be a partial coloring of $\bg{}$.
  
  We define
  \begin{enumerate}
  \item 
    \(
    S(\bg{},\col)
    =
    \{\r\in\Pi\mid\r\text{ is \agrounded{} in }(\bg{},\col)\}
    \)\/;
  \item 
    \(
    \overline{S}(\bg{},\col)
    =
    \{\r\in\Pi\mid\r\text{ is \anotgrounded{} in }(\bg{},\col)\}
    \)\/;
  \item 
    \(
    B(\bg{},\col)
    =
    \{\r\in\Pi\mid\r\text{ is \ablocked{} in }(\bg{},\col)\}
    \)\/;
  \item 
    \(
    \overline{B}(\bg{},\col)
    =
    \{\r\in\Pi\mid\r\text{ is \anotblocked{} in }(\bg{},\col)\}
    \)\/.
  \end{enumerate}
\end{definition}
%---------------------------------------------------------------
%
For a total coloring \col{},
we have
$\overline{S}(\bg{},\col)=\Pi\setminus S(\bg{},\col)$
and
$\overline{B}(\bg{},\col)=\Pi\setminus B(\bg{},\col)$.
Furthermore, $S(\bg{},\col)$ and $\overline{S}(\bg{},\col)$ as well as
$B(\bg{},\col)$ and $\overline{B}(\bg{},\col)$, respectively, are disjoint.

For illustration,
consider the sets obtained regarding the colored \RDGraph{}
$(\bg{\Pi_{\ref{ex:penguin}}},C_{\ref{ex:penguin:coloring}})$, given in
Figure~\ref{fig:penguin:rdg:colored}.
\begin{equation}
\label{eq:sets:penguin}
\begin{array}{rclp{5mm}rcl}
          S (\bg{\Pi_{\ref{ex:penguin}}},C_{\ref{ex:penguin:coloring}})&=&\{r_1,r_2,r_3,r_4\}&&
\overline{S}(\bg{\Pi_{\ref{ex:penguin}}},C_{\ref{ex:penguin:coloring}})&=&\{r_5\}\\
          B (\bg{\Pi_{\ref{ex:penguin}}},C_{\ref{ex:penguin:coloring}})&=&\emptyset&&
\overline{B}(\bg{\Pi_{\ref{ex:penguin}}},C_{\ref{ex:penguin:coloring}})&=&\{r_1,r_2,r_5,r_6\}
\end{array}
\end{equation}

The following theorem shows the correspondence between properties of rules in a
logic program and properties of vertices of a \RDGraph, in the presence of an
existing answer set.
%
%--------------------------------------------------------------
\begin{theorem}\label{thm:nlp:properties}
  Let \bg{} be the \RDGraph\ of logic program $\Pi$, 
  $\col$ be a partial coloring of $\bg{}$
  and 
  $X \in \AS{\Pi}{\col}$.
  
  For $\r\in\Pi$, we have
  \begin{enumerate}
  \item $\pbody{\r} \subseteq X$,            if $\r\in         {S}(\bg{},\col)$\/;
                                             %   \r\ is \agrounded\ in $(\bg{},\col)$
  \item $\pbody{\r} \not\subseteq X$,        if $\r\in\overline{S}(\bg{},\col)$\/;
                                             %   \r\ is \anotgrounded\ in $(\bg{},\col)$  
  \item $\nbody{\r} \cap X  \not=\emptyset$, if $\r\in         {B}(\bg{},\col)$\/;
                                             %   \r\ is \ablocked\ in   $(\bg{},\col)$
  \item $\nbody{\r} \cap X  =\emptyset$,     if $\r\in\overline{B}(\bg{},\col)$\/.
                                             %   \r\ is \anotblocked\ in $(\bg{},\col)$  
\end{enumerate}
\end{theorem}
%---------------------------------------------------------------
%
For admissible colorings, we may turn the above ``if'' statements into ``iff''. 
%
%--------------------------------------------------------------
\begin{corollary}\label{cor:nlp:properties}
  Let \bg{} be the \RDGraph\ of logic program $\Pi$, 
  $\col$ be an admissible coloring of $\bg{}$ 
  and
  $\{X\}=\AS{\Pi}{\col}$.

  For $\r\in\Pi$, we have
  \begin{enumerate}
  \item $\pbody{\r} \subseteq X$            iff $\r\in         {S}(\bg{},\col)$\/;
                                            %   \r\ is \agrounded\ in $(\bg{},\col)$
  \item $\pbody{\r} \not\subseteq X$        iff $\r\in\overline{S}(\bg{},\col)$\/;
                                            %   \r\ is \anotgrounded\ in $(\bg{},\col)$
  \item $\nbody{\r} \cap X  \not=\emptyset$ iff $\r\in         {B}(\bg{},\col)$\/;
                                            %   \r\ is \ablocked\ in   $(\bg{},\col)$
  \item $\nbody{\r} \cap X  =\emptyset$     iff $\r\in\overline{B}(\bg{},\col)$\/.
                                            %   \r\ is \anotblocked\ in  $(\bg{},\col)$  
  \end{enumerate}
\end{corollary}
%---------------------------------------------------------------

The next results are important for understanding the idea of our approach.
%
%---------------------------------------------------------------
\begin{theorem}\label{thm:col:gr}
  Let $\bg{}$ be the \RDGraph{} of logic program  $\Pi$
  and
  $\col$ be a partial coloring of $\bg{}$.
  
  Then, for every $X\in\AS{\Pi}{\col}$ we have that
  \begin{enumerate}
  \item
    \(
    S(\bg{},\col)\cap\overline{B}(\bg{},\col)\subseteq\GR{\Pi}{X}
    \);
  \item
    \(
    \overline{S}(\bg{},\col)\cup B(\bg{},\col)\subseteq \Pi \setminus
    \GR{\Pi}{X} 
    \).
  \end{enumerate}

  If $\col$ is admissible, we have for $\{X\}=\AS{\Pi}{\col}$ that
  \begin{enumerate}
  \setcounter{enumi}{2}
  \item
    \(
    S(\bg{},\col)\cap\overline{B}(\bg{},\col)=\GR{\Pi}{X}
    \);
  \item
    \(
    \overline{S}(\bg{},\col)\cup B(\bg{},\col)=\Pi \setminus \GR{\Pi}{X}
    \).
  \end{enumerate}
\end{theorem}
%---------------------------------------------------------------
%
In fact, the last two equations are equivalent since $C$ is total.
Each of them can be understood as a necessary yet insufficient condition for
characterizing answer sets.
We elaborate upon sufficient graph-theoretical conditions in the next section.

Let us reconsider the partially colored \RDGraph{}
$(\bg{\Pi_{\ref{ex:penguin}}},C_{\ref{ex:penguin:coloring}})$ in
Figure~\ref{fig:penguin:rdg:colored}.
For every
\(
X\in\AS{\Pi_{\ref{ex:penguin}}}{C_{\ref{ex:penguin:coloring}}}
=
\{\{b,p,f\},\{b,p,f'\}\}
\),
we have
\[
\begin{array}{rcl}
            S (\bg{\Pi_{\ref{ex:penguin}}},C_{\ref{ex:penguin:coloring}})
  \cap
  \overline{B}(\bg{\Pi_{\ref{ex:penguin}}},C_{\ref{ex:penguin:coloring}})
  &=&
  \{r_1,r_2,r_3,r_4\}
  \cap
  \{r_1,r_2,r_5,r_6\}
  \\
  &=&
  \{r_1,r_2        \}
  \\
  &\subseteq&
  \GR{\Pi_{\ref{ex:penguin}}}{X}\/;
  \\[2mm]
  \overline{S}(\bg{\Pi_{\ref{ex:penguin}}},C_{\ref{ex:penguin:coloring}})
  \cup
            B (\bg{\Pi_{\ref{ex:penguin}}},C_{\ref{ex:penguin:coloring}})
  &=&
  \{r_5\}
  \cup
  \emptyset
  \\
  &=&
  \{r_5\}
  \\
  &\subseteq&
  \Pi \setminus \GR{\Pi_{\ref{ex:penguin}}}{X}\/.
\end{array}
\]

Regarding $\Pi=\{a \LPif{}\naf{a} \}$,
it is instructive to observe the instance of Condition~3 in Theorem~\ref{thm:col:gr} for total
coloring $C=(\{a \LPif{}\naf{a} \},\emptyset)$ and set $X=\{a\}$:
\[
  S(\bg{\Pi},C)\cap\overline{B}(\bg{\Pi},C)
  =
  \{a \LPif{}\naf{a}\}\cap\emptyset
  =
  \GR{\Pi}{X}
  \ .
\]
This demonstrates that Condition~3 is insufficient for characterizing answer sets.
In fact, observe that
\(
\colplus\neq S(\bg{\Pi},C)\cap\overline{B}(\bg{\Pi},C)
\).

%%% Local Variables: 
%%% mode: latex
%%% TeX-master: "paper"
%%% End: 

%% file: answerships.tex
\section{Deciding answersetship from colored graphs}
\label{sec:check}

The result given in Theorem~\ref{thm:col:gr} started from an existing answer
set induced 
from a given coloring.
We now develop concepts that allow us to decide whether a (total) coloring
represents an answer set by purely graph-theoretical means.

\subsection{Graph-based characterization}

To begin with, we define a graph structure accounting for the notion of
recursive support.
%
%---------------------------------------------------------------
\begin{definition}\label{def:acol:Dasg}
Let \bg{} be the \RDGraph{} of logic program $\Pi$.

We define a support graph of \bg{} as an acyclic \subgraph{0} $(V,E)$ of\/
\bg{} 
such that
\(
\pbody{\r}\subseteq\{\head{\rp}\mid (\rp,\r)\in E\}
\)
for all $\r\in V$.
\end{definition}
%---------------------------------------------------------------
%
Intuitively, support graphs constitute the graph-theoretical counterpart of
operator \textit{Cn}.
That is,
a support graph comprises dependencies among heads and positive bodies
on which potential applications of the $\To{\Pi}$ operator rely.

%
%-------------------------------------------------------------------
\begin{example}\label{ex:support:graph}
Consider program $\Pi_{\ref{ex:support:graph}}$ consisting of rules
\[  
\begin{array}{lll}
  r_1: & a & \LPif \\ 
  r_2: & b & \LPif \LPnot\; a \\ 
  r_3: & c & \LPif b\\ 
  r_4: & b & \LPif c\ .
\end{array}
\]
Among others, the \RDGraph{} of $\Pi_{\ref{ex:support:graph}}$ has support
graphs 
\(
(\emptyset,\emptyset),
(\{r_1,r_2\},\emptyset),
(\{r_2,r_3\},\{(r_2,r_3)\})
\),
and
\(
(\Pi_{\ref{ex:support:graph}},\{(r_2,r_3),(r_3,r_4)\})
\).
\end{example}
%---------------------------------------------------------------
%
Observe that the empty graph $(\emptyset,\emptyset)$ is a support graph of any
(uncolored) graph.
Self-supportedness is avoided due to the acyclicity of support graphs.
\(
(\Pi_{\ref{ex:support:graph}},\{(r_4,r_3),(r_3,r_4)\})
\)
is cyclic and hence no support graph.

Every \RDGraph{} has a unique support graph possessing a largest set of
vertices.
%
%-------------------------------------------------------------------
\begin{theorem}\label{prop:SG:existence}
  Let \bg{} be the \RDGraph{} of logic program $\Pi$.
  
  Then,
  there exists a \groundedZeroDAG{} $(V,E)$ of \bg{}
  such that
  $V'\subseteq V$ for all \groundedZeroDAG{s} $(V',E')$ of \bg{}.
\end{theorem}
%-------------------------------------------------------------------
%
For simplicity, we refer to such \groundedZeroDAG{s} as \emph{maximal
\groundedZeroDAG{s}};
all of them share the same set of vertices.
This set of vertices corresponds to the generating rules of
\Cn{\Pi^\emptyset}.
Different maximal support graphs comprise different sets of edges,
reflecting the intuition that atoms may be derivable in different
ways.
Given that the empty graph is a support graph of any (uncolored)
graph, there is always a maximal support graph.

\begin{figure}[htbp]
\setlength{\unitlength}{1pt}
\centering
{\begin{picture}(90,180)(0,-20)

%% vertices
\put(10,140){\makebox(0,0){$r_1$}}
\put(10,140){\circle{20}}
\put(80,140){\makebox(0,0){$r_2$}}
\put(80,140){\circle{20}}
\put(10,70){\makebox(0,0){$r_4$}}
\put(10,70){\circle{20}}
\put(80,70){\makebox(0,0){$r_3$}}
\put(80,70){\circle{20}}

\put(10,0){\makebox(0,0){$r_6$}}
\put(10,0){\circle{20}}
%% \put(80,0){\makebox(0,0){$r_5$}}
%% \put(80,0){\circle{20}}

%% edges
\put(10,130){\vector(0,-1){50}}
\put(80,130){\vector(0,-1){50}}
%% \put(20,70){\vector(1,0){50}}
%% \put(70,70){\vector(-1,0){50}}
\put(20,140){\vector(1,0){50}}        %% added

%% \put(80,60){\vector(0,-1){50}}
\put(72.5,62.5){\vector(-1,-1){55}}
\put(10,60){\vector(0,-1){50}}
%% \put(20,-10){\oval(20,20)[bl]}
%% \put(20,-10){\oval(20,20)[br]}
%% \put(20,-10){\oval(20,20)[tr]}
%% \put(20,0){\vector(-1,0){0}}

%% edge labeling
\put(5,40){\makebox(0,0){$0$}}
\put(5,110){\makebox(0,0){$0$}}
%% \put(85,40){\makebox(0,0){$0$}}
\put(85,110){\makebox(0,0){$0$}}
%% \put(45,80){\makebox(0,0){$1$}}
\put(40,40){\makebox(0,0){$0$}}
%% \put(35,-10){\makebox(0,0){$1$}}
\put(45,150){\makebox(0,0){$0$}}        %% added

\end{picture}}
\caption{The maximal support graph of \bg{\Pi_{\ref{ex:penguin}}}.}
\label{fig:penguin:rdg:sg}
\end{figure}
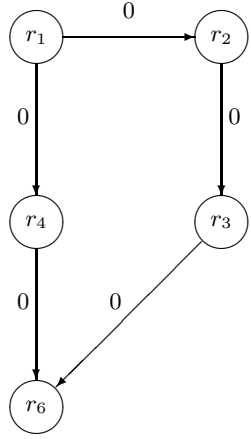

For example, the maximal support graph of the \RDGraph{} of logic program
$\Pi_{\ref{ex:penguin}}$, given in Figure~\ref{fig:penguin:rdg},
is depicted in Figure~\ref{fig:penguin:rdg:sg}.
The latter contains except for $(r_5,r_3)$ all $0$-edges of the former,
viz.~$(\bg{\Pi_{\ref{ex:penguin}}},C_{\ref{ex:penguin:coloring}})$; also 
$r_5$ is excluded since it cannot be supported (recursively).
%------------------------------------------------------------------
%------------------------------------------------------------------
%
The only maximal support graph of \bg{\Pi_{\ref{ex:support:graph}}} is 
\(
(\Pi_{\ref{ex:support:graph}},\{(r_2,r_3),(r_3,r_4)\})
\)
(cf.\ Example~\ref{ex:support:graph}).
%% , including all rules from $\Pi_{\ref{ex:support:graph}}$.
Extending $\Pi_{\ref{ex:support:graph}}$ by $r_5:b\LPif$ yields three
maximal support graphs
\(
(\Pi_{\ref{ex:support:graph}}\cup\{r_5\},\{(r_2,r_3),(r_3,r_4)\}),
(\Pi_{\ref{ex:support:graph}}\cup\{r_5\},\{(r_5,r_3),(r_3,r_4)\}),
\)
and
\(
(\Pi_{\ref{ex:support:graph}}\cup\{r_5\},\{(r_2,r_3),(r_5,r_3),(r_3,r_4)\})
\).
All of them share the same set of vertices but differ in the set of edges.

The concept of a support graph is extended to colored graphs in the following way.
%
%---------------------------------------------------------------    
\begin{definition}\label{def:acol:maxgrounded}
  Let \bg{} be the \RDGraph{} of logic program $\Pi$
  and
  $\col$ be a partial coloring of \bg{}.
  
  We define a support graph of $(\bg{},\col)$
  as a support graph $(V,E)$ of \bg{}  
  such that
  $\colplus \subseteq V$ and $\colminus \cap V=\emptyset$ for some
  $E\subseteq (\Pi \times \Pi)$.
\end{definition}
%---------------------------------------------------------------
%
Recall that $E$ consists of $0$-arcs only.
Also, note that Definition~\ref{def:acol:Dasg} and~\ref{def:acol:maxgrounded}
coincide whenever $C$ is the empty coloring.
In general,
the support graphs of $(\bg{},C)$ are exactly those support graphs
of \bg{} whose vertex set includes $\colplus$ and excludes $\colminus$.
Intuitively,
a support graph of a \emph{colored} \RDGraph{} $(\bg{},C)$ takes the
applicability status of the rules expressed by $C$ into account.
That is, it contains all rules whose positive body is derivable, given that
all rules in $\colplus$ are applicable and all rules in $\colminus$ are
inapplicable.

\begin{figure}[htbp]
\setlength{\unitlength}{1pt}
\centering
{\begin{picture}(90,90)(0,60)

%% vertices
\put(10,140){\makebox(0,0){$r_1$}}
\put(10,140){\circle{20}}
\put(80,140){\makebox(0,0){$r_2$}}
\put(80,140){\circle{20}}
\put(10,70){\makebox(0,0){$r_4$}}
\put(10,70){\circle{20}}
\put(80,70){\makebox(0,0){$r_3$}}
\put(80,70){\circle{20}}

%% \put(10,0){\makebox(0,0){$r_6$}}
%% \put(10,0){\circle{20}}
%% \put(80,0){\makebox(0,0){$r_5$}}
%% \put(80,0){\circle{20}}

%% edges
\put(10,130){\vector(0,-1){50}}
\put(80,130){\vector(0,-1){50}}
%% \put(20,70){\vector(1,0){50}}
%% \put(70,70){\vector(-1,0){50}}
\put(20,140){\vector(1,0){50}}        %% added

%% \put(80,60){\vector(0,-1){50}}
%% \put(72.5,62.5){\vector(-1,-1){55}}
%% \put(10,60){\vector(0,-1){50}}
%% \put(20,-10){\oval(20,20)[bl]}
%% \put(20,-10){\oval(20,20)[br]}
%% \put(20,-10){\oval(20,20)[tr]}
%% \put(20,0){\vector(-1,0){0}}

%% edge labeling
%% \put(5,40){\makebox(0,0){$0$}}
\put(5,110){\makebox(0,0){$0$}}
%% \put(85,40){\makebox(0,0){$0$}}
\put(85,110){\makebox(0,0){$0$}}
%% \put(45,80){\makebox(0,0){$1$}}
%% \put(40,40){\makebox(0,0){$0$}}
%% \put(35,-10){\makebox(0,0){$1$}}
\put(45,150){\makebox(0,0){$0$}}        %% added

\end{picture}}
\caption{The maximal support graph of $(\bg{\Pi_{\ref{ex:penguin}}},C_{\ref{ex:penguin:coloring}})$.}
\label{fig:penguin:rdg:colored:sg}
\end{figure}
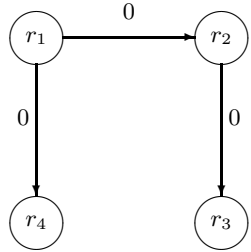

For example, the maximal support graph of the colored \RDGraph{}
$(\bg{\Pi_{\ref{ex:penguin}}},C_{\ref{ex:penguin:coloring}})$, given in
Figure~\ref{fig:penguin:rdg:colored},
is depicted in Figure~\ref{fig:penguin:rdg:colored:sg}.
The latter must include all positively colored and exclude all negatively
colored nodes of the former.
%------------------------------------------------------------------
%------------------------------------------------------------------
%
Given program $\Pi_{\ref{ex:support:graph}}$ from
Example~\ref{ex:support:graph}, a ``bad'' coloring, like
\(
C=(\{b\LPif c\},\{b\LPif\LPnot\; a\})
\),
may deny the existence of a support graph of $(\bg{},\col)$.

Given an arbitrary coloring \col, 
there is a priori no relationship among the set of rules supported by
a colored graph, viz.\ $S(\bg{},\col)$, and its support graphs.
To see this, consider the support graph
$\bg{\Pi_{\ref{ex:support:graph}}}$ of program
$\Pi_{\ref{ex:support:graph}}$ along with coloring 
\(
C=(\{r_1,r_3,r_4\},\{r_2\})
\).
While we have
\(
S(\bg{\Pi_{\ref{ex:support:graph}}},C)
=
\Pi_{\ref{ex:support:graph}}
\),
there is no support graph of $(\bg{\Pi_{\ref{ex:support:graph}}},C)$.
For one thing,
a support graph is denied since $r_3$ and $r_4$ form a circular support.
For another thing,
$r_2$ is always supported, no matter which coloring is considered,
whereas it cannot belong to a support graph once it is blocked (ie.\ colored
with $\cminus$).
This illustrates two things.
First, support graphs provide a global, recursive structure tracing the support
of a rule over 0-paths back to rules with empty positive bodies.
Unlike this, $S(\bg{},\col)$ relies only on 0-edges, capturing thus a rather
local notion of support.
Second,
support graphs take the complete applicability status expressed by a coloring
into account.
Unlike this, $S(\bg{},\col)$ may contain putatively blocked rules from $\colminus$.

As above, we distinguish maximal support graphs of colored graphs through their
maximal set of vertices.

For colored graphs,
we have the following conditions guaranteeing the existence of (maximal) support graphs.
%
%-------------------------------------------------------------------
\begin{theorem}\label{thm:max:supported:graph:existence}
  Let \bg{} be the \RDGraph{} of logic program $\Pi$
  and
  $\col$ be a partial coloring of \bg{}.

  Then, there is a (maximal) \groundedZeroDAG\ of $(\bg{},\col)$,
  if one of the following conditions holds.
  \begin{enumerate}
  \item $\AC{\Pi}{\col}\not=\emptyset$\/;
  \item $(\colplus,E)$ is a \groundedZeroDAG\ of
    $\bg{}|_{\colplus\cup\colminus}$ 
    for some $E \subseteq (\Pi\times \Pi)$.
  \end{enumerate}
\end{theorem}
%-------------------------------------------------------------------
%
The existence of a support graph implies the existence of a maximal one.
This is why we put maximal in parentheses in the preamble of
Theorem~\ref{thm:max:supported:graph:existence}.

As with Property~3 or~4 in Theorem~\ref{thm:col:gr}, respectively,
the existence of a support graph can be understood as a necessary condition for
characterizing answer sets:
%
%--------------------------------------------------------------
\begin{corollary}\label{thm:supportgraph}
  Let \bg{} be the \RDGraph{} of logic program $\Pi$ 
  and
  $\col$ be an admissible coloring of \bg{}.
  
  Then, $(\colplus,E)$ is a  support graph of $(\bg{},\col)$
  for some $E \subseteq (\Pi\times \Pi)$.
\end{corollary}
%---------------------------------------------------------------
%
In fact, taken the last result together with Property~3 or~4 in
Theorem~\ref{thm:col:gr}, respectively,
we obtain a sufficient characterization of admissible colorings
(along with their underlying answer sets).
%
%---------------------------------------------------------------
\begin{theorem}[Answer set characterization, I]\label{thm:acoloring:sets}
  Let \bg{} be the \RDGraph{} of logic program $\Pi$ and
  let $\col$ be a total coloring of \bg{}.
  
  Then, the following statements are equivalent.
  \begin{enumerate}
  \item 
    $C$ is an admissible coloring of $\bg{}$\/;
  \item 
    \(
    \colplus = S(\bg{},\col)\cap\overline{B}(\bg{},\col)
    \)
    and there is a support graph of $(\bg{},\col)$\/;
  \item 
    \(
    \colminus = \overline{S}(\bg{},\col)\cup B(\bg{},\col)
    \)
    and there is a support graph of $(\bg{},\col)$.
  \end{enumerate}
\end{theorem}
%---------------------------------------------------------------
%
Interestingly, this characterization shows that once we have established a
support graph, it doesn't matter whether we focus exclusively on the applicable
(as in Statement~2.) or on the inapplicable rules (as in 3.) for characterizing
admissible colorings.
In both cases, $\colplus$ provides the vertices of the maximal support graphs.

For illustration,
let us consider the two admissible colorings of \RDGraph{} \bg{\Pi_{\ref{ex:penguin}}},
corresponding to the two answer sets of program $\Pi_{\ref{ex:penguin}}$ (given
in Example~\ref{ex:penguin})\/:
\begin{eqnarray}
  \label{ex:penguin:coloring:total:one}
  C_{\ref{ex:penguin:coloring:total:one}}
  &=&
  (
  \{r_1,r_2,r_3\},
  \
  \{r_4,r_5,r_6\}
  )\ ;
  \\
  \label{ex:penguin:coloring:total:two}
  C_{\ref{ex:penguin:coloring:total:two}}
  &=&
  (
  \{r_1,r_2,r_4\},
  \
  \{r_3,r_5,r_6\}
  )\/.
\end{eqnarray}
The resulting colored \RDGraph{}s are depicted in Figure~\ref{fig:penguin:rdg:colored:total}.
%
%------------------------------------------------------------------
\begin{figure}[htbp]
  \setlength{\unitlength}{1pt}
  \centering
  \PengRDG{\cplus}{\cplus}{\cplus}{\cminus}{\cminus}{\cminus}
  \qquad\qquad
  \PengRDG{\cplus}{\cplus}{\cminus}{\cplus}{\cminus}{\cminus}
  \caption{The totally colored \RDGraph{}s
    $(\bg{\Pi_{\ref{ex:penguin}}},C_{\ref{ex:penguin:coloring:total:one}})$
    and
    $(\bg{\Pi_{\ref{ex:penguin}}},C_{\ref{ex:penguin:coloring:total:two}})$.}
  \label{fig:penguin:rdg:colored:total}
\end{figure}
%------------------------------------------------------------------
%
(Cf.~Figure~\ref{fig:penguin:rdg} on Page~\pageref{fig:penguin:rdg} for
the underlying uncolored graph.)
Let us detail the case of $C_{\ref{ex:penguin:coloring:total:one}}$.
We get:
\[
\begin{array}{rcl}
            S (\bg{\Pi_{\ref{ex:penguin}}},C_{\ref{ex:penguin:coloring:total:one}})
  \cap
  \overline{B}(\bg{\Pi_{\ref{ex:penguin}}},C_{\ref{ex:penguin:coloring:total:one}})
  &=&
  \{r_1,r_2,r_3,r_4\}
  \cap
  \{r_1,r_2,r_3,r_5,r_6\}
  \\
  &=&
  \{r_1,r_2,r_3    \}
  \\
  &=&
  (C_{\ref{ex:penguin:coloring:total:one}})_\cplus\/;
  \\[2mm]
  \overline{S}(\bg{\Pi_{\ref{ex:penguin}}},C_{\ref{ex:penguin:coloring:total:one}})
  \cup
            B (\bg{\Pi_{\ref{ex:penguin}}},C_{\ref{ex:penguin:coloring:total:one}})
  &=&
  \{r_5,r_6\}
  \cup
  \{r_4\}
  \\
  &=&
  \{r_4,r_5,r_6\}
  \\
  &=&
  (C_{\ref{ex:penguin:coloring:total:one}})_\cminus\/.
\end{array}
\]
The maximal support graph of
$(\bg{\Pi_{\ref{ex:penguin}}},C_{\ref{ex:penguin:coloring:total:one}})$ is
  given by 
\(
((C_{\ref{ex:penguin:coloring:total:one}})_\cplus,\{(r_1,r_2),(r_2,r_3)\})
\);
it is depicted below in Figure~\ref{fig:penguin:rdg:colored:total:one:sg:bg}.

\subsection{Capturing original concepts}

It is interesting to see how the original definition of an answer set $X$, that
is, $X=\Cn{\reduct{\Pi}{X}}$, along with its underling
constructions, viz.\ reduction {\reduct{\Pi}{X}} and the \CnO{} operator,
can be captured within our graph-based setting.

Clearly, {\reduct{\Pi}{X}} amounts to the set of unblocked rules.
%
%---------------------------------------------------------------
\begin{theorem}\label{thm:red}
  Let \bg{} be the \RDGraph{} of logic program $\Pi$.

  Furthermore, 
  let $\col$ be a total coloring of \bg{} 
  and
  $X$ be a set of atoms such that $X=\head{\colplus}$.

  Then,
  $\head{\r}\LPif \pbody{\r} \in \Pi^X$
  iff
  $\r\in \overline{B}(\bg{},\col)$
  for $\r\in \Pi$.
\end{theorem}
%---------------------------------------------------------------

The next result fixes the relationship of maximal \groundedZeroDAG{s} to the
consequence operator of basic programs.
%
%-------------------------------------------------------------------
\begin{theorem}\label{thm:max:supported:graph:C}
 Let \bg{} be the \RDGraph{} of logic program $\Pi$ and $C$ be a partial coloring
 of \bg{}.

 If $(V,E)$ is a maximal \groundedZeroDAG{} of
 $(\bg{},\col)$, then 
 $\head{V}=\Cn{\reduct{(\Pi\setminus\colminus)}{\emptyset}}$.
\end{theorem}
%-------------------------------------------------------------------
%
For the ``empty'' coloring $\col=(\emptyset,\emptyset)$, we have
$\head{V}=\Cn{\reduct{\Pi}{\emptyset}}$. 

Taking the graph-theoretical counterparts of the reduct $\Pi^X$ and the \CnO\ 
operator yields the following graph-theoretical characterization of answer sets:
%
%-------------------------------------------------------------------
\begin{theorem}[Answer set characterization, II]\label{prop:AS:maxSG}
  Let \bg{} be the \RDGraph{} of logic program $\Pi$ and
  let $\col$ be a total coloring of \bg{}.
  
  Then,
  $C$ is an admissible coloring of $\bg{}$
  iff
  $(\colplus,E)$
  is a maximal \groundedZeroDAG{} of
  $\bg{}|_{\overline{B}(\bg{},\col)}$
  for some $E\subseteq (\Pi\times\Pi)$.
\end{theorem}
%-------------------------------------------------------------------
%
Recall that $\bg{}|_{\overline{B}(\bg{},\col)}$ is the restriction of
\bg{} to the set of unblocked rules in $(\bg{},\col)$.%
\footnote{In fact, $\bg{}|_{\overline{B}(\bg{},\col)}$ could be replaced in
  Theorem~\ref{prop:AS:maxSG} by
  $\bg{}|_{S(\bg{},\col)\cap\overline{B}(\bg{},\col)}$ without changing its
  validity.}
In view of Theorem~\ref{thm:red},
the graph $\bg{}|_{\overline{B}(\bg{},\col)}$ amounts to a reduced (basic)
program $\Pi^X$ whose closure \Cn{\Pi^X} is characterized by means of a maximal
support graph (cf.~Theorem~\ref{thm:max:supported:graph:C}).

\subsection{Subgraph-based characterization}

Using ``unblocked'' rules as done in the previous characterization,
refers only implicitly to the blockage relations expressed by 1-edges.
In analogy to Definition~\ref{def:acol:Dasg},
this structure can be made explicit by the notion of a \emph{blockage graph}.
For this,
we use $\pi_2$ for projecting the second argument of a relation.~\footnote{That is, $\pi_2(R)=\{r_2\mid (r_1,r_2)\in R\}$ for a binary relation $R$.}
%-------------------------------------------------------------------
\begin{definition}\label{def:acol:onedsg}
  Let \bg{} be the \RDGraph{} of logic program $\Pi$
  and 
  $\col$ be a partial coloring of \bg{}.

  We define a blockage graph of $(\bg{},C)$ as a \subgraph{1} $(V,E)$ of\/
  \bg{}  
  such that
  \begin{enumerate}
  \item 
    $E\cap(\colplus\times\colplus) =\emptyset$;
  \item
    $\pi_2(E\cap(\colplus\times\colminus)) = \colminus \cap V$.
  \end{enumerate}
\end{definition}
%-------------------------------------------------------------------
%
In other words, the first condition says that
{there is no $(\r,\rp) \in E$ such that $\r,\rp \in \colplus$},
while the second one stipulates that
{for all $\r\in\colminus\cap V$ there exists an $\rp\in\colplus$ such that
  $(\rp,\r)\in E$}. 

Let us briefly compare the definitions of support and blockage graphs.
Support graphs capture a recursive concept, stipulating that 0-edges are
contained in 0-paths, tracing the support of rules back to rules with empty
positive body.
Unlike this, blockage graphs aim at characterizing rather local dependencies,
based on 1-edge-wise constraints.
That is, while the acyclicity of a support graph cannot be checked locally,
we may verify whether a graph is a blockage graph by inspecting one of its
1-edges after the other.

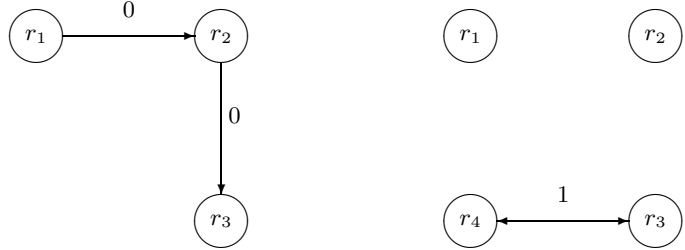
\begin{figure}[htbp]
\setlength{\unitlength}{1pt}
\centering
{\begin{picture}(90,90)(0,60)

%% vertices
\put(10,140){\makebox(0,0){$r_1$}}
\put(10,140){\circle{20}}
\put(80,140){\makebox(0,0){$r_2$}}
\put(80,140){\circle{20}}
%% \put(10,70){\makebox(0,0){$r_4$}}
%% \put(10,70){\circle{20}}
\put(80,70){\makebox(0,0){$r_3$}}
\put(80,70){\circle{20}}

%% \put(10,0){\makebox(0,0){$r_6$}}
%% \put(10,0){\circle{20}}
%% \put(80,0){\makebox(0,0){$r_5$}}
%% \put(80,0){\circle{20}}

%% edges
%% \put(10,130){\vector(0,-1){50}}
\put(80,130){\vector(0,-1){50}}
%% \put(20,70){\vector(1,0){50}}
%% \put(70,70){\vector(-1,0){50}}
\put(20,140){\vector(1,0){50}}        %% added

%% \put(80,60){\vector(0,-1){50}}
%% \put(72.5,62.5){\vector(-1,-1){55}}
%% \put(10,60){\vector(0,-1){50}}
%% \put(20,-10){\oval(20,20)[bl]}
%% \put(20,-10){\oval(20,20)[br]}
%% \put(20,-10){\oval(20,20)[tr]}
%% \put(20,0){\vector(-1,0){0}}

%% edge labeling
%% \put(5,40){\makebox(0,0){$0$}}
%% \put(5,110){\makebox(0,0){$0$}}
%% \put(85,40){\makebox(0,0){$0$}}
\put(85,110){\makebox(0,0){$0$}}
%% \put(45,80){\makebox(0,0){$1$}}
%% \put(40,40){\makebox(0,0){$0$}}
%% \put(35,-10){\makebox(0,0){$1$}}
\put(45,150){\makebox(0,0){$0$}}        %% added

\end{picture}}
\qquad\qquad\qquad\qquad
{\begin{picture}(90,90)(0,60)

%% vertices
\put(10,140){\makebox(0,0){$r_1$}}
\put(10,140){\circle{20}}
\put(80,140){\makebox(0,0){$r_2$}}
\put(80,140){\circle{20}}
\put(10,70){\makebox(0,0){$r_4$}}
\put(10,70){\circle{20}}
\put(80,70){\makebox(0,0){$r_3$}}
\put(80,70){\circle{20}}

%% \put(10,0){\makebox(0,0){$r_6$}}
%% \put(10,0){\circle{20}}
%% \put(80,0){\makebox(0,0){$r_5$}}
%% \put(80,0){\circle{20}}

%% edges
%% \put(10,130){\vector(0,-1){50}}
%% \put(80,130){\vector(0,-1){50}}
\put(20,70){\vector(1,0){50}}
\put(70,70){\vector(-1,0){50}}

%% \put(80,60){\vector(0,-1){50}}
%% \put(72.5,62.5){\vector(-1,-1){55}}
%% \put(10,60){\vector(0,-1){50}}
%% \put(20,-10){\oval(20,20)[bl]}
%% \put(20,-10){\oval(20,20)[br]}
%% \put(20,-10){\oval(20,20)[tr]}
%% \put(20,0){\vector(-1,0){0}}

%% edge labeling
%% \put(5,40){\makebox(0,0){$0$}}
%% \put(5,110){\makebox(0,0){$0$}}
%% \put(85,40){\makebox(0,0){$0$}}
%% \put(85,110){\makebox(0,0){$0$}}
\put(45,80){\makebox(0,0){$1$}}
%% \put(40,40){\makebox(0,0){$0$}}
%% \put(35,-10){\makebox(0,0){$1$}}

\end{picture}}
\caption{Support and blockage graph of
  $(\bg{\Pi_{\ref{ex:penguin}}},C_{\ref{ex:penguin:coloring:total:one}})$.}
\label{fig:penguin:rdg:colored:total:one:sg:bg}
\end{figure}

Together, both concepts provide the following characterization of answer sets.
%
%-------------------------------------------------------------------
\begin{theorem}[Answer set characterization, III]\label{thm:acol:acolorings:ver2}
  Let $\bg{}=(\bgv,\bgezero,\bgeone)$ be the \RDGraph{} of logic program $\Pi$
  and 
  let $\col$ be a total coloring of \bg{}.

  Then, $C$ is an admissible coloring of $\bg{}$
  iff 
  \begin{enumerate}
  \item there is some \groundedZeroDAG{} of $(\bg{},\col)$ 
    and
  \item
    $(S(\bg{},\col),\bgeone|_{S(\bg{},\col)})$ is a blockage graph of
    $(\bg{},C)$. 
    % $(S,\bgeone|_S)$ is a blockage graph of $(\bg{},C)$ where $S=S(\bg{},\col)$. 
  \end{enumerate}
\end{theorem}
%-------------------------------------------------------------------
%
Observe that $(\colplus,E)$ is a  \groundedZeroDAG\ of $(\bg{},C)$ for some
$E\subseteq \Pi\times\Pi$. 
Condition~2 stipulates, among other things, that all supported yet
inapplicable rules are properly blocked (cf.~Condition~2 in
Definition~\ref{def:acol:onedsg}).
The restriction to supported rules is necessary in order to eliminate
rules that are inapplicable since they are unsupported.
Note that the blockage graph in Condition~2 can also be written as 
$(\colplus\cup\colminus,E_1)|_{S(\bg{},C)}$.
For example, the  support and the blockage graph of the colored \RDGraph{}
$(\bg{\Pi_{\ref{ex:penguin}}},C_{\ref{ex:penguin:coloring:total:one}})$, given
on the left hand side in Figure~\ref{fig:penguin:rdg:colored:total},
are depicted in Figure~\ref{fig:penguin:rdg:colored:total:one:sg:bg}.
%------------------------------------------------------------------
%------------------------------------------------------------------
%
This figure nicely illustrates the subset relationship between the vertices of
the support and the blockage graph.

Without explicit mention of the blockage graph, a similar characterization can
be given in the following way.
%
%-------------------------------------------------------------------
\begin{corollary}[Answer set characterization, III$'$]\label{thm:acol:acolorings:ver1}
  Let $\bg{}=(\bgv,\bgezero,\bgeone)$ be the \RDGraph{} of logic program $\Pi$
  and 
  let $\col$ be a total coloring of \bg{}. 

  Then, $C$ is an admissible coloring of $\bg{}$
  iff
  the following conditions hold. 
  \begin{enumerate}
  \item $(\colplus,E)$ is a \groundedZeroDAG\ of $(\bg{},\col)$ for some
  $E \subseteq \Pi\times\Pi$; 
  \item for all $\r\in (\colminus\cap{S}(\bg{},\col))$ there exists an $\rp \in
    \colplus$ such that $(\rp,\r)\in \bgeone$; 
  \item for all $\r,\rp\in \colplus$ we have $(\r,\rp)\not\in \bgeone$.
%   \item there are no $\r,\rp \in \colplus$
%     such that $(\r,\rp)\in \bgeone$.
  \end{enumerate}
\end{corollary}
%---------------------------------------------------------------

%%% Local Variables: 
%%% mode: latex
%%% TeX-master: "paper"
%%% End: 

%% file: operations.tex
\section{Operational characterizations}
\label{sec:operations}

The goal of this section is to provide an operational characterization of
answer sets, based on the concepts introduced in the last section.
The idea is to start with the ``empty'' coloring $(\emptyset,\emptyset)$ and to
successively apply operators that turn a partial coloring $C$ into another one
$C'$ such that $C \sqsubseteq C'$.
This is done until finally an admissible coloring, yielding an answer set, is obtained.

\subsection{Deterministic operators}
\label{sec:operations:deterministic}

We concentrate first on operations extending partial colorings in a
deterministic way.
%
%---------------------------------------------------------------
\begin{definition}\label{def:pre:algo:conditions}
  Let $\bg{}$ be the \RDGraph{} of logic program
  $\Pi$ 
  and
  $\col$ be a partial coloring of \bg{}.
  
  Then, define $\OPprop{\bg{}}: \classC \rightarrow \classC$ as
  \(
  \fOPprop{\bg{}}{\col}
  =
  C \sqcup 
  (
             S (\bg{},\col) \cap \overline{B}(\bg{},\col)
   ,\  
   \overline{S}(\bg{},\col) \cup           B (\bg{},\col)
  )
  \ . 
  \)
\end{definition}
%---------------------------------------------------------------
% 
A partial coloring $C$ is closed under $\OPprop{\bg{}}$, if $C=\fOPprop{\bg{}}{\col}$.
Note that \fOPprop{\bg{}}{\col} does not always exist.
To see this, observe that 
\(
\fOPprop{\bg{}}{(\{a\LPif{}\naf{a}\},\emptyset)}
\)
would be
\(
(\{a\LPif{}\naf{a}\},\{a\LPif{}\naf{a}\})
\),
which is no mapping and thus no partial coloring.
Interestingly, \OPprop{\bg{}} exists on colorings expressing answer sets.
%---------------------------------------------------------------
\begin{theorem}\label{thm:P:reflexivity}
  Let \bg{} be the \RDGraph{} of logic program $\Pi$ and \col{} a partial
  coloring of \bg{}.
  
  If $\AC{\Pi}{\col}\neq\emptyset$, then $\fOPprop{\bg{}}{\col}$ exists.
\end{theorem}
%---------------------------------------------------------------
%
Note that $\fOPprop{\bg{}}{\col}$ may exist although $\AC{\Pi}{C}=\emptyset$.%
~\footnote{Since our goal is to compute members of \AC{\bg{}}{\col}, the
  precondition $\AC{\Pi}{\col}\neq\emptyset$ is sufficient for our purposes.   
  It remains future work to identify a necessary \emph{and} sufficient condition
  guaranteeing the existence of $\fOPprop{\bg{}}{\col}$.}
To see this, consider the program
\(
\Pi=\{a \LPif, c \LPif a,\LPnot\; c\}
\).
Clearly, $\AS{}{\Pi}=\emptyset$.
However, $\fOPprop{\bg{}}{(\emptyset,\emptyset)}=(\{a \LPif\},\emptyset)$ exists.

Now, we can define our principal propagation operator in the following way.
% 
%---------------------------------------------------------------
\begin{definition}\label{def:P}
  Let \bg{} be the \RDGraph{} of logic program $\Pi$ 
  and 
  \col{} a partial coloring of \bg{}.

  Then,  
  define 
  $\OPpropast{\bg{}}:\classC \rightarrow \classC$ where
  $\fOPpropast{\bg{}}{\col}$ is 
  the $\sqsubseteq$-smallest partial coloring 
  containing $C$
  and
  being closed under $\OPprop{\bg{}}$.
\end{definition}
%---------------------------------------------------------------
%
Essentially, $\fOPpropast{\bg{}}{\col}$ amounts to computing the
``immediate consequences'' from a given partial coloring $\col$.%
~\footnote{In Section~\ref{sec:wfs}, this is related to Fitting's semantics.}

Also, like $\fOPprop{\bg{}}{\col}$,  $\fOPpropast{\bg{}}{\col}$ is not
necessarily defined.
This situation is made precise next.
%
%---------------------------------------------------------------
\begin{theorem}\label{thm:P:defined}
  Let \bg{} be the \RDGraph{} of logic program $\Pi$ and \col{} a partial
  coloring of \bg{}.
  
  If $\AC{\Pi}{\col}\neq\emptyset$, then $\fOPpropast{\bg{}}{\col}$
  exists. 
\end{theorem}
%---------------------------------------------------------------
%
Under the previous conditions, we may actually characterize
$\OPpropast{\bg{}}$ in terms of iterated applications of $\OPprop{\bg{}}$;
this is detailed in Appendix~\ref{sec:inductivedef} and used in the proofs.
In fact,
the non-existence of \OPpropast{\bg{}} is an important feature since an undefined
application of \OPpropast{\bg{}} amounts to a backtracking situation at the
implementation level.
Note that $\fOPpropast{\bg{}}{(\emptyset,\emptyset)}$ always exists, even though
we may have $\AC{\Pi}{(\emptyset,\emptyset)}=\emptyset$ (because of
$\AS{}{\Pi}=\emptyset$).

We have the following result.
%---------------------------------------------------------------
\begin{corollary}\label{cor:P:empty:exists}
   Let \bg{} be the \RDGraph{} of logic program $\Pi$. 
   Then, $\fOPpropast{\bg{}}{\emptyC}$ exists.

\end{corollary}
%---------------------------------------------------------------

For illustration, consider program $\Pi_{\ref{ex:penguin}}$ in
Example~\ref{ex:penguin}.
We get:
\begin{eqnarray*}
  \fOPprop{\bg{}}{(\emptyset,\emptyset)}     &=& (\emptyset                             ,\emptyset            )
                                                 \sqcup
                                                 (\{r_1            \}\cap\{r_1,r_2,r_5\},\{r_5\}\cup\emptyset )
  \\                                         &=& (\{r_1            \},                   \{r_5\}              )
  \\
  \fOPprop{\bg{}}{(\{r_1\},\{r_5\})}         &=& (\{r_1            \},                   \{r_5\}              )
                                                 \sqcup
                                                 (\{r_1,r_2,    r_4\}\cap\{r_1,r_2,r_5\},\{r_5\}\cup\emptyset )
  \\                                         &=& (\{r_1,r_2        \},                   \{r_5\}              )
  \\
  \fOPprop{\bg{}}{(\{r_1,r_2\},\{r_5\})}     &=& (\{r_1,r_2        \},                   \{r_5\}              )
                                                 \sqcup
                                                 (\{r_1,r_2,r_3,r_4\}\cap\{r_1,r_2,r_5\},\{r_5\}\cup\emptyset )
  \\                                         &=& (\{r_1,r_2        \},                   \{r_5\}              )
\end{eqnarray*}
Hence, we obtain
\begin{eqnarray*}
\qquad\qquad\,
\fOPpropast{\bg{}}{(\emptyset,\emptyset)}    &=& (\{r_1,r_2        \},                   \{r_5\}              )
\ .
\end{eqnarray*}

Let us now elaborate more upon the formal properties of $\OPprop{\bg{}}$ and
$\OPpropast{\bg{}}$.
First, we observe that both are reflexive, that is, $C \sqsubseteq
\fOPprop{\bg{}}{C}$ and $C \sqsubseteq \fOPpropast{\bg{}}{C}$ (provided that
$\fOPprop{\bg{}}{C}$ and $\fOPpropast{\bg{}}{C}$ exists).
Furthermore, both operators are monotonic in the following sense.
%
%---------------------------------------------------------------
\begin{theorem}\label{thm:P:monoton}
  Let \bg{} be the \RDGraph{} of logic program $\Pi$ 
  and
  let $\col$ and $\col'$ be partial colorings of \bg{} such that
  $\AC{\Pi}{\col'}\neq\emptyset$.

  \begin{enumerate}
  \item   If $\col\sqsubseteq\col'$,
    then $\fOPprop{\bg{}}{\col}\sqsubseteq\fOPprop{\bg{}}{\col'}$\/;
  \item 
    If $\col\sqsubseteq\col'$,
    then $\fOPpropast{\bg{}}{\col}\sqsubseteq\fOPpropast{\bg{}}{\col'}$.
  \end{enumerate}
\end{theorem}
%---------------------------------------------------------------
%
Given that $\OPprop{\bg{}}$ is reflexive,
the last result implies that
\(
\col\sqsubseteq\fOPprop{\bg{}}{\col}\sqsubseteq\fOPprop{\bg{}}{\fOPprop{\bg{}}{\col}}
\)
whenever $\AC{\Pi}{\col}\neq\emptyset$.
In addition, $\OPpropast{\bg{}}$ clearly enjoys a restricted idempotency property,
that is,
\(
\fOPpropast{\bg{}}{\col}=\fOPpropast{\bg{}}{\fOPpropast{\bg{}}{\col}}
\)
provided that $\AC{\Pi}{\col}\neq\emptyset$.

Our next result shows that $\OPprop{\bg{}}$ and $\OPpropast{\bg{}}$ are answer
set preserving.
%
%---------------------------------------------------------------
\begin{theorem}\label{thm:P:ASpreserving}
  Let \bg{} be the \RDGraph{} of logic program $\Pi$ and \col{} be a partial
  coloring of \bg{}.
  
  Then, we have
  \begin{enumerate}
  \item $\AC{\Pi}{\col}=\AC{\Pi}{\fOPprop{\bg{}}{\col}}$\/; 
  \item $\AC{\Pi}{\col}=\AC{\Pi}{\fOPpropast{\bg{}}{\col}}$\/.
  \end{enumerate}
\end{theorem}
%---------------------------------------------------------------
%
That is,
$X\in\AS{\Pi}{\col}$
iff
$X\in\AS{\Pi}{\fOPprop{\bg{}}{\col}}$
iff
$X\in\AS{\Pi}{\fOPpropast{\bg{}}{\col}}$.

A similar result holds for the underlying support and blockage graphs.
%---------------------------------------------------------------    
\begin{theorem}\label{thm:weak:support:and:blockage:graph}
  Let $\bg{}=(\Pi,\bgezero,\bgeone)$ be the \RDGraph{} of logic program $\Pi$
  and
  $\col$ be a partial coloring of \bg{}.

  For $C'={\fOPpropast{\bg{}}{\col}}$, we have
  \begin{enumerate}
  \item if $(\colplus,E)$ is a support graph of $(\bg,\col)$,
    then 
    $(\col'_\oplus,E')$
    is a support graph of $(\bg,\col')$
    \\
    for some $E,E'\subseteq E_0$;\/
  \item 
    if $(\colplus\cup\colminus,\bgeone)|_{S(\bg{},C)}$ is a blockage graph of
    $(\bg{},C)$ and $\colplus\subseteq S(\bg{},C)$,
    \\
    then
    $(\colplus'\cup\colminus',\bgeone)|_{S(\bg{},C')}$ is a blockage graph of
    $(\bg{},C')$. 
%%     if $(\colplus \cup (\colminus \cap {S}(\bg{},C)),
%%          E_1\mid_{\colplus \cup (\colminus\cap {S}(\bg{},C))})$
%%     is a blockage 
%%     graph of 
%%     $(\bg{},C)$\\ 
%%     then
%%     $(\colplus' \cup (\colminus'\cap {S}(\bg{},C')),
%%     E_1\mid_{\colplus' \cup (\colminus'\cap {S}(\bg{},C'))})$   
%%     is a blockage graph of $(\bg{},\col')$. 
%     If $(S(\bg{},\col)\cap(\colplus\cup\colminus),E)$ is a blockage graph of
%     $(\bg,\col)$,
%     \\
%     then
%     $(S(\bg{},\col')\cap
%     (({\col'})_\oplus\cup({\col'})_\ominus),E')$   
%     is a blockage graph of $(\bg,{\col'})$
%     \\
%     for some $E,E'\subseteq (\Pi\times \Pi)$.
  \end{enumerate}
\end{theorem}
%---------------------------------------------------------------    
%
A similar result can be shown for $\OPprop{\bg{}}$. 
%
%% An analogous result for the blockage graph does not hold.
%% Observe the  program
%% \[
%% \begin{array}{lll}
%% r_1 : & a & \LPif \LPnot\; b\\
%% r_2 : & b & \LPif \LPnot\; a\\
%% r_3 : & c & \LPif b\\
%% r_4 : & d & \LPif \LPnot\; c
%% \end{array}
%% \]
%% with the partial coloring $C=(\{r_3\},\emptyset)$.
%% Then, $S(\bg{},C)=\{r_1,r_2,r_4\}$ and $S(\bg{},C)\cap (\colplus \cup
%% \colminus)=\emptyset$.
%% Thus, $(\emptyset,\emptyset)$ is a blockage graph of $(\bg{},C)$.
%% But for $\fOPpropast{\bg{}}{C}=(\{r_3\},\{r_4\})$ and 
%% $S(\bg{},\fOPpropast{\bg{}}{C})=\{r_1,r_2,r_4\}$ we notice that 
%% $(\{r_4\},\emptyset)$ is not a blockage graph of
%% $(\bg{},\fOPpropast{\bg{}}{C})$.

Finally, $\OPprop{\bg{}}$ can be used for deciding answersetship 
in the following way. 
%
%---------------------------------------------------------------
\begin{corollary}[Answer set characterization, I$'$]\label{thm:AS:iv}
  Let $\bg{}$ be the \RDGraph{} of logic program  $\Pi$ and
  let $\col$ be a total coloring of \bg{}. 
  
  Then, $C$ is an admissible coloring of $\bg{}$ 
  iff
  \(
  \fOPprop{\bg{}}{\col}=\col
  \)
  and there is a support graph of $(\bg{},\col)$.
\end{corollary}
%---------------------------------------------------------------
%
This result directly follows from Theorem~\ref{thm:acoloring:sets}.
Clearly, the result is also valid when replacing \OPprop{\bg{}} by
\OPpropast{\bg{}}. 

The following operation draws upon the maximal support graph of colored
\RDGraph{s}.
%
%---------------------------------------------------------------    
\begin{definition}\label{def:S}
  Let $\bg{}$ be the \RDGraph{} of logic program $\Pi$
  and
  $\col$ be a partial coloring of \bg{}.
  
  Furthermore,  
  let $(V,E)$ be a maximal support graph of $(\bg{},\col)$ for some
  $E\subseteq (\Pi \times \Pi)$.
  
  Then,   
  define $\OPMaxGround_{\bg{}}:\classC \rightarrow \classC$ as
  \(
  \fOPMaxGround{\bg{}}{\col}
  =
  (\colplus,\Pi\setminus V)
  \).
\end{definition}
%---------------------------------------------------------------    
%
This operator allows for coloring rules with $\cminus$ whenever it is clear
from the given partial coloring that they will remain unsupported.\footnote{The
  relation to unfounded sets is described in Corollary~\ref{cor:wfs:GUS}.}
Observe that $\Pi\setminus V =\colminus \cup (\Pi\setminus V)$.
As with $\OPpropast{\bg{}}$,
operator $\fOPMaxGround{\bg{}}{\col}$ is an extension of $\col$.
To be more precise, we have $\colplus=(\fOPMaxGround{\bg{}}{\col})_\oplus$
and
\(
\colminus\cup\overline{S}(\bg{},\col)
\subseteq
(\fOPMaxGround{\bg{}}{\col})_\ominus
\). 
Unlike $\OPpropast{\bg{}}$, however, operator $\OPMaxGround_{\bg{}}$ allows for  
coloring nodes unconnected with the already colored part of the graph.

As regards program $\Pi_{\ref{ex:penguin}}$ in Example~\ref{ex:penguin},
for instance, we obtain
\(
\fOPMaxGround{\bg{}}{(\emptyset,\emptyset)}
=
(\emptyset,\{r_5\})
\).
While this information on $r_5$ can also be supplied by \OPprop{\bg{}},
it is not obtainable for ``self-supporting 0-loops'', as in
\(
\Pi=
\{
p\LPif{}q,
q\LPif{}p
\}
\).
In this case, we obtain
\(
\fOPMaxGround{\bg{}}{(\emptyset,\emptyset)}
=
(\emptyset,\{p\LPif{}q,q\LPif{}p\})
\),
which is not obtainable through \OPprop{\bg{}}
(and \OPpropast{\bg{}}, respectively).

Although $\OPMaxGround_{\bg{}}$ cannot be defined in general, it is defined on
colorings satisfying the conditions of Theorem~\ref{thm:max:supported:graph:existence}, 
guaranteeing the existence of support graphs.
%
%-------------------------------------------------------------------
\begin{corollary}\label{cor:S:existence}
  Let $\bg{}$ be the \RDGraph{} of logic program $\Pi$
  and
  $\col$ be a partial coloring of \bg{}.

%%   \begin{enumerate}
%%   \item If $\AS{\Pi}{\col}\not=\emptyset$, then $\fOPMaxGround{\bg{}}{\col}$
%%     exists.
%%   \item
%%     If $(\colplus,E)$ is a \groundedZeroDAG\ of $\bg{}|_\col$ for some $E
%%     \subseteq (\Pi\times \Pi)$, then $\fOPMaxGround{\bg{}}{\col}$ exists.
%%   \item
    If $(\bg{},\col)$ has a \groundedZeroDAG, then
    $\fOPMaxGround{\bg{}}{\col}$ exists. 
%%   \end{enumerate}
\end{corollary}
%-------------------------------------------------------------------
%
We may actually characterize $\fOPMaxGround{\bg{}}{\col}$ in terms of iterated
applications of an operator, similar to $\To{\Pi}$;
this is detailed in Section~\ref{sec:operations:support:II} as well as in
Appendix~\ref{sec:inductivedef}.

As with $\OPpropast{\bg{}}$, 
operator $\OPMaxGround_{\bg{}}$ is reflexive, idempotent, monotonic, and answer set preserving.
%
%---------------------------------------------------------------
\begin{theorem}\label{thm:S:monoton}
  Let \bg{} be the \RDGraph{} of logic program $\Pi$ 
  and
  let \col{} and $\col'$ be partial colorings of \bg{} such that
  $\AC{\Pi}{\col}\not=\emptyset$ and $\AC{\Pi}{\col'}\not=\emptyset$.
  
  Then, we have the following properties.
  \begin{enumerate}
  \item $\col\sqsubseteq\fOPMaxGround{\bg{}}{\col}$\/;
  \item $\fOPMaxGround{\bg{}}{\col}=\fOPMaxGround{\bg{}}{\fOPMaxGround{\bg{}}{\col}}$\/;
  \item if $\col\sqsubseteq\col'$, then
    $\fOPMaxGround{\bg{}}{\col}\sqsubseteq\fOPMaxGround{\bg{}}{\col'}$.
  \end{enumerate}
\end{theorem}
%---------------------------------------------------------------
%
%---------------------------------------------------------------
\begin{theorem}\label{thm:S:hilfslemma}
  Let \bg{} be the \RDGraph{} of logic program $\Pi$ 
  and  
  \col{} be a partial coloring of \bg{}.
  
  Then, we have $\AC{\Pi}{\col}=\AC{\Pi}{\fOPMaxGround{\bg{}}{\col}}$.
\end{theorem}
%---------------------------------------------------------------
%
Note that unlike $\OPprop{\bg{}}$, operator $\OPMaxGround_{\bg{}}$ leaves the
support graph of $(\bg{},\col)$ unaffected.
Since, according to Theorem~\ref{thm:acol:acolorings:ver2}, the essential
blockage graph is composed of supported rules only, the same applies to this
graph as well.
%
%---------------------------------------------------------------    
\begin{theorem}\label{thm:weak:support:and:blockage:graph:two}
  Let $\bg{}=(\Pi,\bgezero,\bgeone)$ be the \RDGraph{} of logic program $\Pi$
  and
  $\col$ be a partial coloring of \bg{}.

  For $C'={\fOPMaxGround{\bg{}}{\col}}$, we have
  \begin{enumerate}
  \item if $(\colplus,E)$ is a support graph of $(\bg,\col)$,
    then 
    $(\colplus,E)$
    is a support graph of $(\bg,\col')$
    \\
    for some $E\subseteq E_0$;\/
  \item 
    if $(\colplus\cup\colminus,\bgeone)|_{S(\bg{},C)}$ is a blockage graph of
    $(\bg{},C)$,
    \\
    then
    $(\colplus\cup\colminus,\bgeone)|_{S(\bg{},C)}$ is a blockage graph of
    $(\bg{},C')$. 
  \end{enumerate}
\end{theorem}
%---------------------------------------------------------------    
In fact, we have in the latter case that
$(\colplus\cup\colminus,\bgeone)|_{S(\bg{},C)}=(\colplus'\cup\colminus',\bgeone)|_{S(\bg{},C')}$. 

Because $\OPMaxGround{}_{\bg{}}$ implicitly enforces the existence of a support
graph, our operators furnish yet another characterization of answer sets.
%
%---------------------------------------------------------------
\begin{corollary}[Answer set characterization, I$''$]\label{thm:answerset:PS}
  Let $\bg{}$ be the \RDGraph\ of logic program $\Pi$ and
  let $\col$ be a total coloring of $\bg{}$.

  Then,
  $C$ is an admissible coloring of $\bg{}$
  iff
  $\col=\fOPprop{\bg{}}{{\col}}$
  and
  $\col={\fOPMaxGround{\bg{}}{\col}}$.
\end{corollary}
%---------------------------------------------------------------
%
Clearly, this result is also valid when replacing \OPprop{\bg{}} by \OPpropast{\bg{}}.

Note that the last result is obtained from Corollary~\ref{thm:AS:iv} by
replacing the requirement of the existence of a support graph by
$\col={\fOPMaxGround{\bg{}}{\col}}$.  
However, the last condition cannot guarantee that all supported unblocked rules
belong to $\colplus$.
For instance, $(\emptyset,\{a\LPif\})$ has an empty support graph;
hence $(\emptyset,\{a\LPif\})={\fOPMaxGround{\bg{}}{(\emptyset,\{a\LPif\})}}$.
That is, the trivially supported fact $a\LPif$ remains in $\colminus$.
In our setting, such a miscoloring is detected by operator
$\OPprop{\bg{}}$.
That is, $\fOPprop{\bg{}}{(\emptyset,\{a\LPif\})}$ does not exist, since it
would yield $(\{a\LPif\},\{a\LPif\})$, which is no partial coloring. 

\subsection{Basic operational characterization}
\label{sec:operations:characterization}

We start by providing a very general operational characterization that possesses
a maximum degree of freedom.

To this end, we observe that Corollary~\ref{thm:AS:iv} and~\ref{thm:answerset:PS},
respectively, can serve as a straightforward \emph{check} for deciding whether a
given total coloring constitutes an answer set.
A corresponding \emph{guess} can be provided through an operator capturing a
non-deterministic (don't know) choice.
%
%---------------------------------------------------------------    
\begin{definition}\label{def:pre:algo:choose:ver2}
  Let $\bg{}$ be the \RDGraph{} of logic program $\Pi$
  and
  $\col$ be a partial coloring of \bg{}.
  
  For $\circ \in \{\cplus,\cminus\}$,
  define $\OPchoose^{\circ}_{\bg{}}:\classC \rightarrow \classC$
  as
  \begin{enumerate}
  \item $\fOPchoose{\Pi}{\cplus}{\bg{}}{\col}=(\colplus \cup\{\r\},\colminus)$
    \qquad for some $\r\in\Pi\setminus (\colplus\cup\colminus)$;
  \item $\fOPchoose{\Pi}{\cminus}{\bg{}}{\col}=(\colplus,\colminus\cup\{\r\})$
    \qquad for some $\r\in\Pi\setminus (\colplus\cup\colminus)$.
  \end{enumerate}
\end{definition}
%---------------------------------------------------------------
%
We use $\OPchoose_{\bg{}}^\circ$ whenever the distinction between
$\fOPchoose{\Pi}{\cplus}{\bg{}}{\col}$ and $\fOPchoose{\Pi}{\cminus}{\bg{}}{\col}$
is of no importance.
Strictly speaking, $\OPchoose_{\bg{}}^\circ$ is also parametrized with $r$;
we leave this implicit to abstract from the actual choice. 
In fact, whenever both operators $\fOPchoose{\Pi}{\cplus}{\bg{}}{\col}$ and
$\fOPchoose{\Pi}{\cminus}{\bg{}}{\col}$ are available, the choice of $r$
is only a \emph{``don't care''} choice, while that among
${\cplus}$ and ${\cminus}$ is the crucial \emph{``don't know''} choice.
Intuitively, this is because all rules must be colored either way;
it is the attributed color that is of prime importance for the existence of
an answer set.

Combining the previous guess and check operators yields our first operational
characterization of admissible colorings (along with its underlying answer sets).
%
%---------------------------------------------------------------
\begin{theorem}[Operational answer set characterization, I]\label{thm:pre:algo:main:ver1}
  Let \bg{} be the \RDGraph{} of logic program $\Pi$ and
  let $\col$ be a total coloring of \bg{}.

  Then,
  $C$ is an admissible coloring of $\bg{}$
  iff 
  there exists a sequence $(\col^i)_{0 \leq i \leq n}$ with the following
  properties: 
  \begin{enumerate}
  \item $\col^0=(\emptyset,\emptyset)$\/;
  \item
    \(
    \col^{i+1}=\fOPchoose{\Pi}{\circ}{\bg{}}{\col^i}     
    \)
    for some $\circ \in \{\cplus,\cminus\}$
    and
    $0\leq i < n\/$;
  \item $\col^n=\fOPprop{\bg{}}{{\col^n}}$\/;
  \item $\col^n={\fOPMaxGround{\bg{}}{\col^n}}$\/;
  \item $\col^n=\col$.
  \end{enumerate}
\end{theorem}
%---------------------------------------------------------------
%
In what follows, we refer to such sequences also as \emph{coloring sequences}.
Note that all sequences satisfying conditions 1-5 of
Theorem~\ref{thm:pre:algo:main:ver1} 
are \emph{successful} in the sense that their last element corresponds to an
existing answer set.
If a program has no answer set, then no such sequence exists.

Although this straightforward guess and check approach may not be of great
implementation value, it supplies us with an initial skeleton for the coloring
process that we refine in the sequel.
In particular, this characterization stresses the basic fact that we possess
complete freedom in forming a coloring sequence as long as we can guarantee that
the resulting coloring is a fixed point of $\OPprop{\bg{}}$ and
$\OPMaxGround_{\bg{}}$. 
It is worth mentioning that this simple approach is inapplicable when
fixing $\circ$ to either $\oplus$ or $\ominus$
(cf.~Section~\ref{sec:operations:unicoloring} below).

We observe the following properties.
%
%---------------------------------------------------------------
\begin{theorem}\label{thm:collection:OASC:i}
  Given the same prerequisites as in Theorem~\ref{thm:pre:algo:main:ver1},
  let $(\col^i)_{0 \leq i \leq n}$ be a sequence satisfying conditions 1-5 
  in Theorem~\ref{thm:pre:algo:main:ver1}.

  Then, we have the following properties for $0\leq i\leq n$.
  \begin{enumerate}
  \item $\col^i$ is a partial coloring\/;
  \item $\col^i\sqsubseteq \col^{i+1}$\/;
  \item $\AC{\Pi}{\col^i}\supseteq\AC{\Pi}{\col^{i+1}}$\/;
  \item $\AC{\Pi}{\col^i}\not= \emptyset$;
  \item $(\bg{},\col^i)$ has a (maximal) support graph.
  \end{enumerate}
\end{theorem}
%---------------------------------------------------------------
%
All these properties represent invariants of the consecutive colorings.
While the first three properties are provided by operator
$\mathcal{C}_{\bg{}}^\circ$ in choosing among uncolored rules only, the last
two properties are actually enforced by the final coloring
$\col^n$, 
that is, the ``check'' expressed by conditions~3--5 in
Theorem~\ref{thm:pre:algo:main:ver1}.
In fact, sequences only enjoying conditions 1 and~2 in
Theorem~\ref{thm:pre:algo:main:ver1}, fail to satisfy
Property~4 and~5 in general.
In practical terms, this means that computations of successful sequences may be
led numerous times on the ``garden path'' before termination.

As it is well-known,
the number of choices can be significantly reduced by applying deterministic
operators.
To this end, given a partial coloring $\col$,
define $(\OPprop{}\OPMaxGround)_{\bg{}}^\ast(\col)$ as the
$\sqsubseteq$-smallest partial coloring containing $\col$ 
and being closed under $\OPprop{\bg{}}$ and $\OPMaxGround_{\bg{}}$.~\footnote{An iterative characterization of 
$(\OPprop{}\OPMaxGround)_{\bg{}}^\ast(\col)$ is given in
Appendix~\ref{sec:inductivedef}.} 
%
%---------------------------------------------------------------
\begin{theorem}[Operational answer set characterization, II]\label{thm:pre:algo:main:ver2}
  Let \bg{} be the \RDGraph{} of logic program $\Pi$ and
  let $\col$ be a total coloring of \bg{}.

  Then,
  $C$ is an admissible coloring of $\bg{}$
  iff 
  there exists a sequence $(\col^i)_{0 \leq i \leq n}$ with the following
  properties: 
  \begin{enumerate}
  \item $\col^0=(\OPprop{}\OPMaxGround)_{\bg{}}^\ast((\emptyset,\emptyset))$\/;
  \item
    \(
    \col^{i+1}
    =
    (\OPprop{}\OPMaxGround)_{\bg{}}^\ast
    (\fOPchoose{\Pi}{\circ}{\bg{}}{\col^i})    
    \)
    for some $\circ\in\{\cplus,\cminus\}$
    and
    $0\leq i < n$\/;
  \item $\col^n=\col$.
  \end{enumerate}
\end{theorem}
%---------------------------------------------------------------
%
On the one hand, the continuous applications of $\OPprop{\bg{}}$ and
$\OPMaxGround_{\bg{}}$ extend colorings after each choice.
On the other hand, this proceeding guarantees that each partial coloring
$\col^i$ is closed under $\OPprop{\bg{}}$ and $\OPMaxGround_{\bg{}}$.

Regarding correctness and completeness, however, it is clear in view of
Theorem~\ref{thm:pre:algo:main:ver1} that any number of iterations of
$\OPprop{\bg{}}$ and $\OPMaxGround_{\bg{}}$ can be executed after
$\OPchoose_{\bg{}}^\circ$ as long as $(\OPprop{}\OPMaxGround)_{\bg{}}^\ast$ is
the final operation leading to $C^n$ in Theorem~\ref{thm:pre:algo:main:ver2}.

\begin{figure}[htbp]
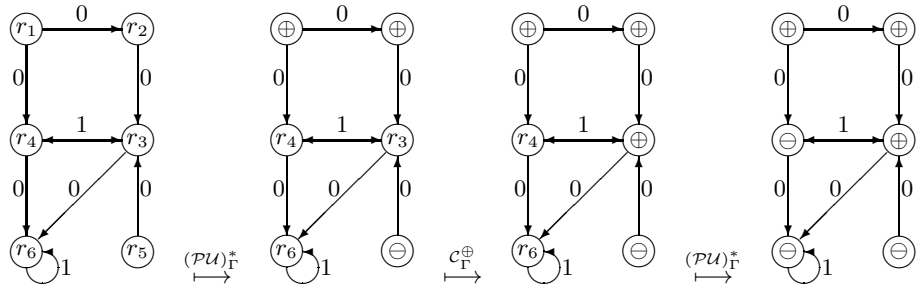

  \centering
  \setlength{\unitlength}{.6pt}
  \PengRDG{r_1}{r_2}{r_3}{r_4}{r_5}{r_6}
  $\stackrel{(\OPprop{}\OPMaxGround)_{\bg{}}^\ast}{\longmapsto}$
  \PengRDG{\cplus}{\cplus}{r_3}{r_4}{\cminus}{r_6}
  $\stackrel{\OPchoose_{\bg{}}^{\cplus}}{\longmapsto}$
  \PengRDG{\cplus}{\cplus}{\cplus}{r_4}{\cminus}{r_6}
  $\stackrel{(\OPprop{}\OPMaxGround)_{\bg{}}^\ast}{\longmapsto}$
  \PengRDG{\cplus}{\cplus}{\cplus}{\cminus}{\cminus}{\cminus}
  \caption{A coloring sequence.}
  \label{fig:penguin:coloring:sequence}
\end{figure}

For illustration, consider the coloring sequence $(\col^0,\col^1)$
obtained for answer set $\{b,p,f'\}$ of program $\Pi_{\ref{ex:penguin}}$ in
Example~\ref{ex:penguin}:
\[
\begin{array}{lclcl}
\col^0&=&(\OPprop{}\OPMaxGround)_{\bg{}}^\ast((\emptyset,\emptyset))
      &=&(\{p\LPif,b \LPif p\},\{b\LPif m\})
\\
\col^1&=& (\OPprop{}\OPMaxGround)_{\bg{}}^\ast(\fOPchoose{\Pi}{\cplus}{\bg{}}{\col^0})  
      &=&(\{p\LPif,b\LPif p,f'\LPif p,\LPnot\; f\},
\\    & &
      & & \qquad\{b\LPif m, f\LPif b,\LPnot\; f', x\LPif f,f',\LPnot\; x\})
\end{array}
\]
The decisive operation in this sequence is the application of
$\OPchoose^{\cplus}_{\bg{}}$ leading to $\col(f\LPif b,\LPnot\; f')=\cplus$.
Note that in this simple example all propagation is accomplished by operator \OPprop{\bg{}}. 
We have illustrated the formation of the sequence in Figure~\ref{fig:penguin:coloring:sequence}.
The same final result is obtained when choosing $\OPchoose^{\cminus}_{\bg{}}$
such that $\col(f' \LPif  p, \LPnot\; f)=\cminus$.
This illustrates that several coloring sequences may lead to the same answer set.

As with Corollary~\ref{cor:P:empty:exists}, we have the following result.
%
%---------------------------------------------------------------
\begin{corollary}\label{cor:PUast:empty:exists}
  Let \bg{} be the \RDGraph{} of logic program $\Pi$. 
  Then, $\fPUast{\bg{}}{\emptyC}$ exists.

\end{corollary}
%---------------------------------------------------------------

The usage of continuous propagations leads to further invariant properties.
%
%---------------------------------------------------------------
\begin{theorem}\label{thm:collection:OASC:ii}
  Given the same prerequisites as in Theorem~\ref{thm:pre:algo:main:ver2},
  let $(\col^i)_{0 \leq i \leq n}$ be a sequence satisfying conditions 1-3 
  in Theorem~\ref{thm:pre:algo:main:ver2}.

  Then, we have the following:
  \begin{enumerate}
  \item[1.]\hspace*{-5pt}--5. as given in Theorem~\ref{thm:collection:OASC:i}\/;
    \setcounter{enumi}{5}
  \item
    \(
    \colplus^{i+1} \supseteq S(\bg{},\col^i)\cap\overline{B}(\bg{},\col^i)
    \)\/;
  \item 
    \(
    \colminus^{i+1} \supseteq \overline{S}(\bg{},\col^i)\cup
    B(\bg{},\col^i) 
    \)\/.
  \end{enumerate}
\end{theorem}
%---------------------------------------------------------------
%
Taking the last two properties together with Condition~5 from
Theorem~\ref{thm:collection:OASC:i}, we see that propagation gradually enforces
exactly the attributes on partial colorings, expressed in
Theorem~\ref{thm:acoloring:sets} for admissible colorings.

Given that we obtain only two additional properties, one may wonder whether
exhaustive propagation truly pays off.
In fact, its great value becomes apparent when looking at the properties of
prefix sequences, not necessarily leading to a successful end.
%
%---------------------------------------------------------------
\begin{theorem}\label{thm:collection:OASC:ii:sub}
  Given the same prerequisites as in Theorem~\ref{thm:pre:algo:main:ver2},
  let $(\col^j)_{0 \leq j \leq m}$ be a sequence satisfying Condition~1 and~2
  in Theorem~\ref{thm:pre:algo:main:ver2}.

  Then, we have the following.
  \begin{enumerate}
  \item[1.]\hspace*{-5pt}--3., 5. as given in
    Theorem~\ref{thm:collection:OASC:i}\/; 
    \setcounter{enumi}{5}
  \item[6.]\hspace*{-5pt}--7.
    as given in Theorem~\ref{thm:collection:OASC:ii}\/.
  \end{enumerate}
\end{theorem}
%---------------------------------------------------------------
%
Using exhaustive propagations, 
we observe that except for Property~4  all properties
possessed by successful
sequences, are shared by (possibly unsuccessful) prefix sequences.

The next results make the aforementioned claim on the effect of deterministic
operators on admissible prefix sequences more precise:
%
%---------------------------------------------------------------
\begin{theorem}\label{thm:comparison:i}
  Let \bg{} be the \RDGraph{} of logic program $\Pi$.

  Let $m$ be the number of sequences over \classC{} satisfying
  conditions~1-4 in Theorem~\ref{thm:pre:algo:main:ver1}
  and
  let $n$ be the number of sequences over \classC{} satisfying
  conditions 1 and 2 in Theorem~\ref{thm:pre:algo:main:ver2}.

  Then, we have $n\leq m$.
\end{theorem}
%---------------------------------------------------------------
%
Moreover, successful sequences are usually shorter when using propagation.
%
%---------------------------------------------------------------
\begin{theorem}\label{thm:comparison:ii}
  Let \bg{} be the \RDGraph{} of logic program $\Pi$
  and
  $C$ be an admissible coloring of \bg{}.

  Let $(\col^i)_{0 \leq i \leq m}$ and $(\col^j)_{0 \leq j \leq n}$ be the
  shortest sequences obtained for $C$ according to
  Theorem~\ref{thm:pre:algo:main:ver1} and
  Theorem~\ref{thm:pre:algo:main:ver2}, 
  respectively.

  Then, we have $n\leq m$.
\end{theorem}
%---------------------------------------------------------------
%
The same result can be shown for the longest sequences.

\subsection{Unicoloring operational characterization}
\label{sec:operations:unicoloring}

As mentioned above, the use of two coloring operators is essential for our
initial operational characterization given in Theorem~\ref{thm:pre:algo:main:ver1}.
In fact, this is obsolete when using continuous propagation, as done in
Theorem~\ref{thm:pre:algo:main:ver2}.
That is, rather than using two coloring operators, we may actually use only one
of them, and leave the attribution of the complementary color to propagation operators.
To be more precise, this amounts to replacing Condition~2 in
Theorem~\ref{thm:pre:algo:main:ver2} either by
\begin{center}
  \it
  2.$^+$ \
  \(
  \col^{i+1}
  =
  (\OPprop{}\OPMaxGround)_{\bg{}}^\ast
  (\fOPchoose{\Pi}{\cplus}{\bg{}}{\col^i})    
  \)
  \quad
  or
  \qquad
  2.$^-$ \
  \(
  \col^{i+1}
  =
  (\OPprop{}\OPMaxGround)_{\bg{}}^\ast
  (\fOPchoose{\Pi}{\cminus}{\bg{}}{\col^i})    
  \)
  \qquad
  for $0\leq i < n$\/.
\end{center}
Then, we have the following result.
%
%---------------------------------------------------------------
\begin{corollary}[Operational answer set characterization, II$^+$/II$^-$]\label{thm:pre:algo:main:ver2:pm}

  Theorem~\ref{thm:pre:algo:main:ver2} still holds,
  when replacing Condition~2 either by 2.$^+$ or 2.$^-$.
\end{corollary}
%---------------------------------------------------------------
%
Note that the possible set of (prefix) sequences is smaller than that obtained
from the ``bi-coloring'' in 
Theorem~\ref{thm:pre:algo:main:ver2}, simply, because choices are restricted
to a single color.
On the other hand, we see no advantage of either approach regarding the length
of successful sequences.
Also, since the choice of the color is fixed, the choice of the rule to be
colored becomes a ``don't know'' choice.

The last type of characterization is based on exhaustive propagation.
Hence, we are interested in the question how much propagation is sufficient
for compensating the possibility of choosing among two colors.
The next result gives an answer for the case of unicoloring with
$\OPchoose^{\cplus}_{\bg{}}$.
%
%---------------------------------------------------------------
\begin{theorem}[Operational answer set characterization, III$^+$]\label{thm:sequence:plus}
  Let \bg{} be the \RDGraph{} of logic program $\Pi$ and 
  let $\col$ be a total coloring of \bg{}.

  Then,
  $C$ is an admissible coloring of $\bg{}$
  iff 
  there exists a sequence $(\col^i)_{0 \leq i \leq n}$ with the following
  properties: 
  \begin{enumerate}
  \item $\col^0=(\emptyset,\emptyset)$\/;
  \item
    \(
    \col^{i+1}=\fOPchoose{\Pi}{\cplus}{\bg{}}{\col^i}     
    \)
    for
    $0\leq i < n-1$\/;
  \item $\col^n={\fOPMaxGround{\bg{}}{\fOPpropast{\bg{}}{\col^{n-1}}}}$\/;
  \item $\col^n=\col$.
  \end{enumerate}
\end{theorem}
%---------------------------------------------------------------
%
The actual propagation is done in Condition~3.
That is, the consecutive application of $\OPMaxGround_{\bg{}}$ and
$\OPpropast{\bg{}}$ allows for coloring all remaining rules in ${\col^{n-1}}$ with
$\ominus$.

Note that the application order of $\OPpropast{\bg{}}$ and
$\OPMaxGround_{\bg{}}$ 
in Condition~3 cannot be reversed.
To see this, reconsider program $\Pi_{\ref{ex:support:graph}}$ in
Example\ref{ex:support:graph},
consisting of rules
\[
\begin{array}{lll}
r_1: & a & \LPif \\ 
r_2: & b & \LPif \LPnot\; a \\ 
r_3: & c & \LPif b\\ 
r_4: & b & \LPif c\ .
\end{array}
\]
and observe that its only answer set $\{a\}$ corresponds to coloring
\(
(\{r_1\},\{r_2,r_3,r_4\})
\).
In order to obtain this according to Theorem~\ref{thm:sequence:plus},
we must color rule $r_1$ with $\cplus$.
We then get
\(
{\fOPpropast{\bg{}}{(\{r_1\},\emptyset)}}
=
(\{r_1\},\{r_2\})
\)
and finally
\(
{\fOPMaxGround{\bg{}}{(\{r_1\},\{r_2\})}}
=
(\{r_1\},\{r_2,r_3,r_4\})
\).
Switching the last operators yields
\(
\fOPMaxGround{\bg{}}{(\{r_1\},\emptyset)}=(\{r_1\},\emptyset)
\)
and
\(
{\fOPpropast{\bg{}}{(\{r_1\},\emptyset)}}
=
(\{r_1\},\{r_2\})
\)
and we fail to obtain the desired result.

Interestingly, the usage of the operator $\OPpropast{\bg{}}$ is sufficient when
coloring with $\cminus$ only.
%
%---------------------------------------------------------------
\begin{theorem}[Operational answer set characterization, III$^-$]\label{thm:sequence:minus}
  Let \bg{} be the \RDGraph{} of logic program $\Pi$ and 
  let $\col$ be a total coloring of \bg{}.

  Then,
  $C$ is an admissible coloring of $\bg{}$
  iff 
  there exists a sequence $(\col^i)_{0 \leq i \leq n}$ with the following
  properties: 
  \begin{enumerate}
  \item $\col^0=(\emptyset,\emptyset)$\/;
  \item
    \(
    \col^{i+1}=\fOPchoose{\Pi}{\cminus}{\bg{}}{\col^i}     
    \)
    for
    $0\leq i < n-1$\/;
  \item $\col^n={\fOPpropast{\bg{}}{\col^{n-1}}}$\/;
  \item $\col^n=\col$.
  \end{enumerate}
\end{theorem}
%---------------------------------------------------------------
%

Unicoloring offers another perspective on strategies employing
bicoloring:
The coloration with the second color may be regarded as some sort of
``lemmatization'' avoiding duplicate solutions rather than a genuine choice.

\subsection{Support-driven operational characterization}
\label{sec:operations:support}

The number of possible choices encountered during a computation is of crucial
importance.
We have already seen above that propagation has great computational advantages.
What else may cut down the number of choices?

Looking at the graph structures underlying an admissible coloring
(cf.~Theorem~\ref{thm:acol:acolorings:ver2}),
we observe that support graphs capture a global  --- since recursive ---
structure, 
while blockage graphs aim at a rather local structure, based on arc-wise
constraints.
Consequently, it seems advisable to prefer choices maintaining support
structures over those maintaining blockage relations,
since the former have more global repercussions than the latter.

To this end, we develop in this section a strategy that is based on a choice
operation restricted to supported rules.
%
%---------------------------------------------------------------    
\begin{definition}\label{def:weakChoose}
  Let $\bg{}$ be the \RDGraph{} of logic program $\Pi$
  and
  $\col$ be a partial coloring of \bg{}.
  
  For $\circ \in \{\cplus,\cminus\}$,
  define $\mathcal{D}^{\circ}_{\bg{}}:\classC \rightarrow \classC$
  as
  \begin{enumerate}
  \item $\weakChoose{\bg{}}{\cplus}{\col}=(\colplus \cup\{\r\},\colminus)$
    \qquad for some $\r\in S(\bg{},C) \setminus (\colplus\cup\colminus)$;
  \item $\weakChoose{\bg{}}{\cminus}{\col}=(\colplus,\colminus\cup\{\r\})$
    \qquad for some $\r\in S(\bg{},C) \setminus (\colplus\cup\colminus)$.
  \end{enumerate}
\end{definition}
%---------------------------------------------------------------
%
Compared to operators $\OPchoose_{\bg{}}^\circ$, the latter restrict
their choice to supported rules.
Verifying whether a rule is supported can then be done in a local fashion by
looking at the immediate predecessors in the \RDGraph{}.
With this little additional effort,
the number of colorable rules is smaller than that encountered when applying
$\OPchoose_{\bg{}}^\circ$.
The benefit of support-driven characterizations is that the
length of coloring sequences is bound by the number of supported rules.
Depending on how the non-determinism of $\mathcal{D}_{\bg{}}^\circ$ is dealt
with algorithmically,
this may either lead to a reduced depth of the search tree or a reduced
branching factor.

In fact, in a successful coloring sequence $(\col^i)_{0 \leq i \leq n}$,
all rules in $(C^n)_\cplus$ must belong to an encompassing support graph and thus be
supported.
Hence, by means of \weakChoose{\bg{}}{\cplus}{\col} (along with
$\OPpropast{\bg{}}$) the supportedness of each set $\colplus^i$ can be made invariant.
Consequently, such a proceeding allows for establishing the
existence of support graphs, as stipulated in
Theorem~\ref{thm:acoloring:sets}, so to speak ``on the fly''.
To a turn, this allows for a much simpler approach to the task(s) previously
accomplished by operator $\OPMaxGround_{\bg{}}$.
We discuss two such simplifications in what follows.

\subsubsection{Support-driven operational characterization I}
\label{sec:operations:support:I}

Given that the existence of support graphs is guaranteed, one may
actually completely dispose of operator $\OPMaxGround_{\bg{}}$ and color in a
final step all uncolored rules with $\cminus$.
%
%---------------------------------------------------------------
\begin{definition}\label{def:N}
  Let \bg{} be the \RDGraph{} of logic program $\Pi$
  and
  \col\ a partial coloring of \bg{}.

  Then,  
  define 
  $\OPN{\bg{}}:\classC \rightarrow \classC$ as
  \(
  \fOPN{\bg{}}{C}= (\colplus, \Pi\setminus \colplus)
  \).
\end{definition}
%---------------------------------------------------------------

Roughly speaking,
the idea is then to ``actively'' color only supported rules and rules blocked by
supported rules; all remaining rules are then unsupported and ``thrown'' into
$C_\cminus$ in a final step.
This is made precise in the following characterization.
%
%---------------------------------------------------------------
\begin{theorem}[Operational answer set characterization, IV] \label{thm:sequence:D:plusminus}
  Let \bg{} be the \RDGraph{} of logic program $\Pi$ and
  let $\col$ be a total coloring of \bg{}.

  Then,
  $C$ is an admissible coloring of $\bg{}$
  iff 
  there exists a sequence $(\col^i)_{0 \leq i \leq n}$ with the following
  properties: 
  \begin{enumerate}
  \item $\col^0=(\emptyset,\emptyset)$\/;
  \item
    \(
    \col^{i+1}
    =
    \weakChoose{\bg{}}{\circ}{\col^i}
    \)
    where
    $\circ \in  \{ \cplus,\cminus\}$
    and
    $0\leq i < n-1$\/;
  \item $\col^n=\fOPN{\bg{}}{\col^{n-1}}$\/;
  \item $\col^n=\fOPprop{\bg{}}{\col^n}$\/;
  \item $\col^n=\col$.
  \end{enumerate}
\end{theorem}
%---------------------------------------------------------------
%
We note that there is a little price to pay for turning $\OPMaxGround_{\bg{}}$
into $\mathcal{N}_{\bg{}}$, expressed by the test on the final total coloring in
Condition~4.
Without it, one could use $\mathcal{N}_{\bg{}}$ to obtain a total coloring by
coloring rules with $\cminus$ in an arbitrary way.%

We obtain the following properties for the previous type of coloring sequences.
%
%---------------------------------------------------------------
\begin{theorem}\label{thm:collection:sequence:D:plusminus}
  Given the same prerequisites as in Theorem~\ref{thm:sequence:D:plusminus},
  let $(\col^i)_{0 \leq i \leq n}$ be a sequence satisfying conditions~1-5
  in Theorem~\ref{thm:sequence:D:plusminus}.

  Then, we have the following.
  \begin{enumerate}
  \item[1.]\hspace*{-5pt}--5. as given in Theorem~\ref{thm:collection:OASC:i}\/;
    \setcounter{enumi}{7}
  \item $(\colplus^i,E)$ is a support graph of $(\bg{},C^i)$ for some
  $E\subseteq \Pi\times \Pi$.
%%   \item[9.] $(S(\bg{},C^i)\cap (\colplus^i\cup\colminus^i),E)$ is a blockage
%%   graph of $\bg{}$ where $E=\bgeone|_{S(\bg{},C^i)\cap
%%   (\colplus^i\cup\colminus^i)}$. 
%%   \item[6.]\hspace*{-5pt}--7.
%%    \comment{K:8+9 gelten nicht wegen choice operator}
%%     as given in Theorem~\ref{thm:collection:OASC:ii}\/.
  \end{enumerate}
\end{theorem}
%---------------------------------------------------------------
%
Condition~8 makes the aforementioned claim on the supportedness of each rule in
$\colplus^i$ explicit.
In contrast to the coloring sequences enjoying Condition~5 only, the sequences
formed by means of $\mathcal{D}_{\bg{}}^\circ$ guarantee that each $\colplus^i$
forms a support graph.

In fact, there is some overlap among operator $\mathcal{D}_{\bg{}}^{\cminus}$ and
$\mathcal{N}_{\bg{}}$.
To see this, consider $\Pi=\{a\LPif{}, b\LPif{}\naf{a}\}$.
We must initially apply $\mathcal{D}_{\bg{}}^{\cplus}$ to obtain
\(
(\{a\LPif\},\emptyset)
\)
from the empty coloring.
Then, however, there are two possibilities for obtaining total coloring
\(
(\{a\},\{b\LPif{}\naf{a}\})
\),
either by applying $\mathcal{D}_{\bg{}}^{\cminus}$
or by applying $\mathcal{N}_{\bg{}}$.
In fact, in view of this,
$\mathcal{N}_{\bg{}}$ allows us to dispose of $\mathcal{D}_{\bg{}}^{\cminus}$:
%
%---------------------------------------------------------------
\begin{corollary}[Operational answer set characterization, IV$^+$]\label{thm:sequence:D:plus}

  Theorem~\ref{thm:sequence:D:plusminus} still holds,
  when replacing Condition~2 by
  \begin{itemize}
  \item [2.$^+$]
    \(
    \col^{i+1}
    =
    \weakChoose{\bg{}}{\cplus}{\col^i}
    \)
    where
    $0\leq i < n-1$\/.
  \end{itemize}
\end{corollary}
%---------------------------------------------------------------
%
Observe that there is no characterization using $\mathcal{D}_{\bg{}}^{\cminus}$
and $\mathcal{N}_{\bg{}}$ because this leaves no possibility for coloring rules
with $\cplus$.
%
%---------------------------------------------------------------
\begin{theorem}\label{thm:collection:sequence:D:plus}
  Given the same prerequisites as in Corollary~\ref{thm:sequence:D:plus},
  let $(\col^i)_{0 \leq i \leq n}$ be a sequence satisfying conditions~1-5
  in Corollary~\ref{thm:sequence:D:plus}.

  Then, we have the following.
  \begin{enumerate}
  \item[1.]\hspace*{-5pt}--5. as given in Theorem~\ref{thm:collection:OASC:i}\/;
    \setcounter{enumi}{7}
  \item[8.] as given in Theorem~\ref{thm:collection:sequence:D:plusminus}\/;
  \item[9.] 
  $(\colplus^i \cup\colminus^i,E)|_{S(\bg{},C^i)}$ is a blockage
  graph of $(\bg{},C^i)$.
%%   \item[6.]\hspace*{-5pt}--7.
%%    \comment{K:8+9 gelten nicht wegen choice operator}
%%     as given in Theorem~\ref{thm:collection:OASC:ii}\/.
  \end{enumerate}
\end{theorem}
%---------------------------------------------------------------
%
Interestingly, only support-driven unicoloring by means of
$\mathcal{D}_{\bg{}}^\cplus$ can guarantee the consecutive existence of blockage
graphs.
This is because $\mathcal{D}_{\bg{}}^\cplus$ warrants, first, that
$\colplus^i\subseteq{S(\bg{},C^i)}$ and thus all blocking rules are taken into
account and, second, that rules are not arbitrarily colored with $\cminus$ but
rather guarded by operator $\OPprop{\bg{}}$.~\footnote{This is more apparent in Corollary~\ref{thm:sequence:DP:plus} below,
  when using $\OPpropast{\bg{}}$ for propagation as well.}

As discussed above, there is some overlap in the characterization expressed
in Theorem~\ref{thm:sequence:D:plusminus}.
Interestingly, this can be eliminated by adding propagation operator
$\OPpropast{\bg{}}$ to the previous characterization.
This results in coloring sequences corresponding to the basic strategy used in the
\texttt{noMoRe} system~\cite{ankoli02a}.
%
%---------------------------------------------------------------
\begin{theorem}[Operational answer set characterization, V]\label{thm:sequence:DP:plusminus}
  Let \bg{} be the \RDGraph{} of logic program $\Pi$ and 
  let $\col$ be a total coloring of \bg{}.

  Then,
  $C$ is an admissible coloring of $\bg{}$
  iff 
  there exists a sequence $(\col^i)_{0 \leq i \leq n}$ with the following
  properties: 
  \begin{enumerate}
  \item $\col^0=\fOPpropast{\bg{}}{(\emptyset,\emptyset)}$\/;
  \item
    \(
    \col^{i+1}
    =
    \fOPpropast{\bg{}}{\weakChoose{\bg{}}{\circ}{\col^i}}
    \)
    where
    $\circ \in \{\cplus,\cminus\}$
    and
    $0\leq i < n-1$\/;
  \item $\col^n=\fOPN{\bg{}}{\col^{n-1}}$\/;
  \item $\col^n=\fOPprop{\bg{}}{\col^n}$\/;
  \item $\col^n=\col$.
  \end{enumerate}
\end{theorem}
%---------------------------------------------------------------
%

For illustration, consider the coloring sequence $(\col^0,\col^1,\col^2)$
obtained for answer set $\{b,p,f'\}$ of program $\Pi_{\ref{ex:penguin}}$ in
Example~\ref{ex:penguin}:
\[
\begin{array}{lclcl}
\col^0&=&\fOPpropast{\bg{}}{(\emptyset,\emptyset)}
      &=&(\{p\LPif,b \LPif p\},\{b\LPif m\})
\\
\col^1&=& \fOPpropast{\bg{}}{\weakChoose{\bg{}}{\cplus}{\col^0}}
      &=&(\{p\LPif,
            b\LPif p,
            f'\LPif p, \LPnot\; f
          \},
\\    & &
      & & \qquad
          \{b\LPif m, 
            f\LPif b,\LPnot\; f', 
            x\LPif f,f',\LPnot\; x
          \}
         )
\\
\col^2&=& \fOPN{\bg{}}{\col^1}&=&\col^1
\end{array}
\]
This sequence is similar to the one obtained for $\Pi_{\ref{ex:penguin}}$
with the characterization given in Theorem~\ref{thm:pre:algo:main:ver2}.
All propagation is accomplished by operator \OPprop.
However, operator $\mathcal{D}_{\bg{}}^{\cplus}$ is faced with less
choices than $\mathcal{C}_{\bg{}}^{\cplus}$ (in
Theorem~\ref{thm:pre:algo:main:ver2}) because only two among the three 
uncolored rules are supported.

For another example, consider program $\Pi_{\ref{ex:support:graph}}$ in
Example~\ref{ex:support:graph}.
We get a coloring sequence $(C^0,C^1)$, where
\[
\begin{array}{lclcl}
\col^0&=&\fOPpropast{\bg{}}{(\emptyset,\emptyset)}&=&(\{a\LPif\},\{b\LPif\LPnot\; a\})
\\
\col^1&=&\fOPN{\bg{}}{(\{a\LPif\},\{b\LPif\LPnot\; a\})}&=&(\{a\LPif\},\{b\LPif\LPnot\; a,c \LPif b,b \LPif c\})
\ .
\end{array}
\]
Note that operator $\mathcal{D}_{\bg{}}^{\circ}$ is inapplicable to $C^0$,
since
\(
S(\bg{},C^0) \setminus (\colplus^0\cup\colminus^0)
\)
is empty.
In this situation, $\mathcal{C}_{\bg{}}^{\circ}$ would be applicable to color
either of the two uncolored rules in $\Pi\setminus (\colplus^0\cup\colminus^0)$.
In the final step, the two unfounded rules are directly colored by operator
\OPN{\bg{}} without any further efforts.

Indeed, the strategy of \texttt{noMoRe} applies $\mathcal{P}_{\bg{}}^{\ast}\circ\mathcal{D}_{\bg{}}^{\circ}$ as long as there are supported rules.
Once no more uncolored supported rules exist, operator $\mathcal{N}_{\bg{}}$ is
called.
Finally, \OPprop{\bg{}} is applied but only to those rules colored previously by
$\mathcal{N}_{\bg{}}$.~\footnote{Efficient propagation operations, only working on
  certain subsets of the input, are discussed in a companion paper.}
At first sight,
this approach may seem to correspond to a subclass of the coloring sequences
described above, in the sense that \texttt{noMoRe} enforces a maximum number of
transitions described in Condition~2.
To see that this is not the case, we observe the following property.~\footnote{It is worth mentioning that in practice the application of
  \OPprop{\bg{}} in Condition~4 can be restricted to those rules colored by
  $\mathcal{N}_{\bg{}}$ in Condition~3.}
%
%---------------------------------------------------------------
\begin{theorem}\label{thm:sequence:DP:plusminus:N}
  Given the same prerequisites as in Theorem~\ref{thm:sequence:DP:plusminus},
  let $(\col^i)_{0 \leq i \leq n}$ be a sequence satisfying conditions~1-5
  in Theorem~\ref{thm:sequence:DP:plusminus}.

  Then, we have
  \(
  (\fOPN{\bg{}}{\col^{n-1}}_\cminus\setminus\col^{n-1}_\cminus)\subseteq\overline{S}(\bg{},\col)   
  \).
\end{theorem}
%---------------------------------------------------------------
%
That is, no matter which (supported) rules are colored $\cminus$ by
$\mathcal{D}_{\bg{}}^{\cminus}$, operator $\OPN{\bg{}}$ only applies to
unsupported ones.
It is thus no restriction to enforce the consecutive application of
$\OPpropast{\bg{}}$ and $\mathcal{D}_{\bg{}}^{\circ}$ until no more supported
rules are available.
In fact, it is the interplay of the two last operators that guarantees this
property.
For instance, looking at $\Pi=\{a,b\LPif{}\naf{a}\}$, we see that we directly
obtain the final total coloring because
\(
(\{a\},\{b\LPif{}\naf{a}\})
=
\fOPpropast{\bg{}}{\weakChoose{\bg{}}{\cplus}{(\emptyset,\emptyset)}}
\),
without any appeal to $\OPN{\bg{}}$.
Rather it is \OPpropast{\bg{}} that detects that $b\LPif{}\naf{a}$ belongs to
the set of blocked rules.
Generally speaking, $\mathcal{D}_{\bg{}}^{\cplus}$ consecutively chooses the
generating rules of an answer set, finally gathered in
\(
\colplus = S(\bg{},\col)\cap\overline{B}(\bg{},\col)
\).
Every rule in $B(\bg{},\col)$ is blocked by some rule in $\colplus$.
So whenever a rule $r$ is added by $\mathcal{D}_{\bg{}}^{\cplus}$ to $\colplus$,
operator \OPpropast{\bg{}} adds all rules blocked by $r$ to $
\colminus$.
In this way, $\OPpropast{\bg{}}$ and $\mathcal{D}_{\bg{}}^{\cplus}$ gradually
color all rules in $S(\bg{},\col)\cap\overline{B}(\bg{},\col)$ and
$B(\bg{},\col)$, so that all remaining uncolored rules, subsequently
treated by $\OPN{\bg{}}$, must belong to
$\overline{S}(\bg{},\col)$.~\footnote{Some rules of $\overline{S}(\bg{},\col)$
  are already detected and added to $\colminus$ by \OPpropast{\bg{}}.}

Furthermore, we obtain the following properties.
%
%---------------------------------------------------------------
\begin{theorem}\label{thm:collection:sequence:DP:plusminus}
  Given the same prerequisites as in Theorem~\ref{thm:sequence:DP:plusminus},
  let $(\col^i)_{0 \leq i \leq n}$ be a sequence satisfying conditions~1-5
  in Theorem~\ref{thm:sequence:DP:plusminus}.

  Then, we have the following.
  \begin{enumerate}
  \item[1.]\hspace*{-5pt}--5. as given in Theorem~\ref{thm:collection:OASC:i}\/;
    \setcounter{enumi}{5}
   \item[6.]\hspace*{-5pt}--7.
     as given in Theorem~\ref{thm:collection:OASC:ii}\/;
  \item[8.] as given in Theorem~\ref{thm:collection:sequence:D:plusminus}.
  \end{enumerate}
\end{theorem}
%---------------------------------------------------------------

Finally, for the sake of completeness, we give the unicoloring variant along
with its properties.
%
%---------------------------------------------------------------
\begin{corollary}[Operational answer set characterization, V$^+$]\label{thm:sequence:DP:plus}

  Theorem~\ref{thm:sequence:DP:plusminus} still holds,
  when replacing Condition~2 by
  \begin{itemize}
  \item [2.$^+$]
    \(
    \col^{i+1}
    =
    \fOPpropast{\bg{}}{\weakChoose{\bg{}}{\cplus}{\col^i}}
    \)
    where
    $0\leq i < n-1$\/.
  \end{itemize}
\end{corollary}
%---------------------------------------------------------------
%---------------------------------------------------------------
\begin{theorem}\label{thm:collection:sequence:DP:plus}
  Given the same prerequisites as in Corollary~\ref{thm:sequence:DP:plus},
  let $(\col^i)_{0 \leq i \leq n}$ be a sequence satisfying conditions~1-5
  in Corollary~\ref{thm:sequence:DP:plus}.

  Then, we have the following.
  \begin{enumerate}
  \item[1.]\hspace*{-5pt}--5. as given in Theorem~\ref{thm:collection:OASC:i}\/;
    \setcounter{enumi}{5}
  \item[6.]\hspace*{-5pt}--7.
    as given in Theorem~\ref{thm:collection:OASC:ii}\/;
  \item[8.]\hspace*{-5pt}--9.
     as given in Theorem~\ref{thm:collection:sequence:D:plusminus}
     and~\ref{thm:collection:sequence:D:plus}\/. 

%%   \item[8.] $(\colplus^i,E)$ is a support graph of $(\bg{},C^i)$ for some
%%   $E\subseteq \Pi\times \Pi$\/;
%%   \item[9.] $(S(\bg{},C^i)\cap (\colplus^i\cup\colminus^i),E)$ is a blockage
%%   graph of $\bg{}$ where $E=\bgeone|_{S(\bg{},C^i)\cap
%%   (\colplus^i\cup\colminus^i)}$. 
  \end{enumerate}
\end{theorem}
%---------------------------------------------------------------
%
The previous characterization also satisfies Theorem~\ref{thm:sequence:DP:plusminus:N}.

\subsubsection{Support-driven operational characterization II}
\label{sec:operations:support:II}

Although the previous approach has turned out to be of practical value as the
core inference strategy of the \texttt{noMoRe} system,
it is still improvable.
This is because the early detection of unsupported rules may allow for better
propagation results and thus fewer choices.
To see this, consider program $\Pi_{\ref{ex:V}}=\{r_1,r_2,r_3\}$, where
\footnote{For the sake of uniqueness, we label the program with the equation number.}
\begin{equation}\label{ex:V}
\begin{array}{lrcl}
r_1: & p & \LPif & \LPnot\; q \\
r_2: & q & \LPif & r,\LPnot\; p  \\
r_3: & r & \LPif & q.
\end{array}
\end{equation}
This program has the answer set $\{p\}$ represented by the
admissible coloring
\(
(\{r_1\},\{r_2,r_3\})
\).
Without an operator like $\OPMaxGround_{\bg{}}$ one would need a choice operation
for detecting the unfounded rules $r_2$ and $r_3$.
Following Theorem~\ref{thm:sequence:DP:plusminus}, we get the coloring sequence:
\[
\begin{array}{lclcl}
\col^0&=&\fOPpropast{\bg{}}{(\emptyset,\emptyset)}
      &=&(\emptyset,\emptyset)
\\
\col^1&=& \fOPpropast{\bg{}}{\weakChoose{\bg{}}{\cplus}{\col^0}}
      &=&(\{r_1\},\emptyset)
\\
\col^2&=& \fOPN{\bg{}}{\col^1}&=&(\{r_1\},\{r_2,r_3\})
\end{array}
\]
Although the support-driven choice operator $\mathcal{D}_{\bg{}}^{\circ}$
is only faced with a single alternative, as opposed to the three alternatives
encountered by operator $\mathcal{C}_{\bg{}}^{\circ}$,
it would be clearly advantageous to solve this example by propagation only.
This motivates a variant of operator $\OPMaxGround_{\bg{}}$ that takes advantage
of the support-driven strategy pursued by $\mathcal{D}_{\bg{}}^{\circ}$.

Given the \RDGraph{}  \bg{} of a logic program $\Pi$ 
and a partial coloring \col{} of \bg{},
define $\mathcal{T}_{\bg{}}:\classC \rightarrow \classC$ as
\[
\mathcal{T}_{\bg{}}(\col)=
(\colplus \cup (S({\bg{}},{\col})\setminus \colminus),\colminus)  
\]
and
define $\weakProp{\bg{}}{\col}$ as the $\sqsubseteq$-smallest partial
coloring containing $\col$ and being closed under $\mathcal{T}_{\bg{}}$.

The next result shows that $\mathcal{T}^\ast_{\bg{}}$ extends a given
support graph to a maximal one.
%
%---------------------------------------------------------------    
\begin{theorem}\label{thm:weakP:supportgraph}
  Let $\bg{}$ be the \RDGraph{} of logic program $\Pi$
  and
  $\col$ be a partial coloring of \bg{}.

  If $(\colplus,E)$ is a support graph of $(\bg,\col)$,
  then 
  $((\weakProp{\bg{}}{\col})_\oplus,E')$ is a maximal support graph of
  $(\bg{},\col)$ 
  for some $E,E'\subseteq (\Pi\times \Pi)$.
\end{theorem}
%---------------------------------------------------------------    
%
Recall that by definition $\col\sqsubseteq\weakProp{\bg{}}{\col}$.

With this, we can define the following incremental (and constructive) variant
of $\OPMaxGround_{\bg{}}$: 
%
%---------------------------------------------------------------
\begin{definition}\label{def:weakS}
  Let \bg{} be the \RDGraph{} of logic program $\Pi$
  and
  \col\ a partial coloring of \bg{}.

  Then,  
  define 
  $\mathcal{V}_{\bg{}}:\classC \rightarrow \classC$ as
  $\weakS{\bg{}}{\col}=(\colplus,\Pi\setminus V)$ 
  where $V=\weakProp{\bg{}}{\col}_\cplus$.
\end{definition}
%---------------------------------------------------------------
%
Observe that $\Pi\setminus V =\colminus \cup (\Pi\setminus V)$.
It is instructive to compare the latter definition with that of
$\OPMaxGround_{\bg{}}$, given in Definition~\ref{def:S}.
In fact, $\OPMaxGround_{\bg{}}$ can be obtained by means of
$\mathcal{T}^\ast_{\bg{}}$ by defining $V$ in Definition~\ref{def:S} as
\(
V=\weakProp{\bg{}}{(\emptyset,\emptyset)}_\cplus
\)
subject to the condition $\colplus\subseteq V$ and $\colminus\cap V=\emptyset$.

The next result tells us when both operators coincide.
%
%---------------------------------------------------------------    
\begin{corollary}\label{thm:weakS:S}
  Let $\bg{}$ be the \RDGraph{} of logic program $\Pi$
  and
  $\col$ be a partial coloring of \bg{}.

  If $(\colplus,E)$ is a support graph of $(\bg{},\col)$ for some 
  $E \subseteq (\Pi \times \Pi)$,
  then 
  $\fOPMaxGround{\bg{}}{\col}=\weakS{\bg{}}{\col}$.

\end{corollary}
%---------------------------------------------------------------    
%
Given that $\mathcal{D}_{\bg{}}^\circ$ adds only supported rules to a coloring \col,
it gradually extends the underlying support graph around $\colplus$.
Hence, we may replace operator $\OPMaxGround_{\bg{}}$ by $\OPV{\bg{}}$ whenever
using choice operator $\mathcal{D}_{\bg{}}^\circ$.
This is impossible when using operator $\OPchoose_{\bg{}}^\circ$, since its
choice may not be supported and thus destroy the invariant support property
expressed in Property~8 in Theorem~\ref{thm:collection:sequence:D:plusminus}.
Unlike $\OPMaxGround_{\bg{}}$,
operator $\OPV{\bg{}}$ takes the support status of all rules in $\colplus$ for
granted.
This allows $\OPV{\bg{}}$ to focus on uncolored rules, viz.~$\Pi\setminus
(\colplus\cup\colminus)$.
Such an assumption is not made by $\OPMaxGround_{\bg{}}$, which (possibly)
reestablishes the support status of rules in $\colplus$; it is thus proned to
consider all rules in $\Pi\setminus\colminus$.
Technically, this is reflected by the fact that the computation of $\OPV{\bg{}}$ by means
of $\mathcal{T}_{\bg{}}$ may start from $C$, while the one of\/
$\OPMaxGround_{\bg{}}$ by $\mathcal{T}_{\bg{}}$ must start out with the empty
coloring.
In all, $\OPV{\bg{}}$ does not lead to fewer choices than
$\OPMaxGround_{\bg{}}$,
it has rather the computational advantage of avoiding redundant computations.

Now, we are ready to give support-oriented counterparts of the
operational characterizations given in the two previous subsections.
Most of them are obtainable by simply replacing operators 
$\OPchoose_{\bg{}}^\circ$ and $\OPMaxGround_{\bg{}}$
by 
$\mathcal{D}_{\bg{}}^\circ$ and $\fweakS_{\bg{}}$, respectively.
The first exception is Theorem~\ref{thm:pre:algo:main:ver1} where the
replacement of  
$\OPchoose_{\bg{}}^\circ$ by $\mathcal{D}_{\bg{}}^\circ$ leaves no way of
coloring unsupported rules with $\cminus$.
(This provides further evidence for the necessity of $\mathcal{N}_{\bg{}}$ in
Theorem~\ref{thm:sequence:D:plusminus}.)

For defining a support-oriented counterpart of Theorem~\ref{thm:pre:algo:main:ver2},
we first need the following propagation operator:
As with $(\OPprop{}\OPMaxGround)_{\bg{}}^\ast$, given a partial coloring $\col$,
we define $(\OPprop{}\fweakS)_{\bg{}}^\ast(\col)$ 
as the
$\sqsubseteq$-smallest partial coloring containing $\col$ 
and being closed under $\OPprop{\bg{}}$ and $\fweakS_{\bg{}}$.~\footnote{A characterization in terms of iterated applications is given in
  Appendix~\ref{sec:inductivedef}.} 
%
%---------------------------------------------------------------
\begin{theorem}[Operational answer set characterization, VI]\label{thm:pre:algo:main:ver4}
  Let \bg{} be the \RDGraph{} of logic program $\Pi$ and
  let $\col$ be a total coloring of \bg{}.

  Then,
  $C$ is an admissible coloring of $\bg{}$
  iff 
  there exists a sequence $(\col^i)_{0 \leq i \leq n}$ with the following
  properties: 
  \begin{enumerate}
  \item $\col^0=(\OPprop{}\fweakS)_{\bg{}}^\ast((\emptyset,\emptyset))$\/;
  \item
    \(
    \col^{i+1}
    =
    (\OPprop{}\fweakS)^\ast_{\bg{}}(\weakChoose{\bg{}}{\circ}{\col^i})    
    \)
    where
    $\circ \in \{\cplus,\cminus\}$
    and
    $0\leq i < n$\/;
  \item $\col^n=\col$.
  \end{enumerate}
\end{theorem}
%---------------------------------------------------------------
%

For illustration, consider program $\Pi_{\ref{ex:V}}$ in (\ref{ex:V}).
We get a coloring sequence $(C^0)$, where
\[
\begin{array}{lclcl}
\col^0&=&(\OPprop{}\fweakS)_{\bg{}}^\ast((\emptyset,\emptyset))&=&(\{p\LPif\LPnot\; q\},\{q\LPif r,\LPnot\; p,r\LPif q\})
\ .
\end{array}
\]
Or to be more precise,
\[
\begin{array}{lclcl}
&&\fweakS_{\bg{}}((\emptyset,\emptyset))&=&(\emptyset,\{q\LPif r,\LPnot\; p,r\LPif q\})
\\
\col^0&=&\OPprop{\bg{}}((\emptyset,\{q\LPif r,\LPnot\; p,r\LPif q\}))&=&(\{p\LPif\LPnot\; q\},\{q\LPif r,\LPnot\; p,r\LPif q\})
\ .
\end{array}
\]

Analogously, we obtain for program $\Pi_{\ref{ex:support:graph}}$ in
Example~\ref{ex:support:graph} a singular
coloring sequence $(C^0)$, where
\[
\begin{array}{lclcl}
\col^0&=&(\OPprop{}\fweakS)_{\bg{}}^\ast((\emptyset,\emptyset))&=&(\{a\LPif\},\{b\LPif\LPnot\; a,c \LPif b,b \LPif c\})
\ .
\end{array}
\]

We note the following invariant properties of the sequence.
%
%---------------------------------------------------------------
\begin{theorem}\label{thm:collection:OASC:iv}
  Given the same prerequisites as in Theorem~\ref{thm:pre:algo:main:ver4},
  let $(\col^i)_{0 \leq i \leq n}$ be a sequence satisfying conditions 1-3 
  in Theorem~\ref{thm:pre:algo:main:ver4}.

  Then, we have the following.
  \begin{enumerate}
  \item[1.]\hspace*{-5pt}--5. as given in Theorem~\ref{thm:collection:OASC:i}\/;
    \setcounter{enumi}{5}
  \item[6.]\hspace*{-5pt}--7. as given in
  Theorem~\ref{thm:collection:OASC:ii}\/;
    \setcounter{enumi}{7}
  \item[8.] as given in
  Theorem~\ref{thm:collection:sequence:DP:plus}.
   
  \end{enumerate}
\end{theorem}
%---------------------------------------------------------------
%

In analogy to what we have shown in Section~\ref{sec:operations:unicoloring},
we may replace Condition~2 in Theorem~\ref{thm:pre:algo:main:ver4} either by
\begin{center}
  \it
  2.$^+$ \
  \(
  \col^{i+1}
  =
  (\OPprop{}\fweakS)_{\bg{}}^\ast
  (\weakChoose{\bg{}}{\cplus}{\col^i})    
  \)
  \quad
  or
  \qquad
  2.$^-$ \
  \(
  \col^{i+1}
  =
  (\OPprop{}\fweakS)_{\bg{}}^\ast
  (\weakChoose{\bg{}}{\cminus}{\col^i})    
  \)
  \qquad
  for $0\leq i < n$\/.
\end{center}
Then, we have the following.
%
%---------------------------------------------------------------
\begin{corollary}[Operational answer set characterization, VI$^+$/VI$^-$]\label{thm:pre:algo:main:ver4:pm}

  Theorem~\ref{thm:pre:algo:main:ver4} still holds,
  when replacing Condition~2 either by 2.$^+$ or 2.$^-$.
\end{corollary}
%---------------------------------------------------------------
%
Also, all results obtained in Section~\ref{sec:operations:support:I}
remain true, when replacing  $\mathcal{N}_{\bg{}}$ by  $\fweakS_{\bg{}}$.

\subsection{Summary}
\label{sec:summary}

In view of the many different operational characterizations, we summarize in
Table~\ref{tab:ops:summary} their approach to the formation of coloring
sequences and their major properties.
For this, we distinguish between the formation of coloring sequences
$(\col^i)_{0 \leq i \leq n}$ and the test on their final element $C^n$.
For describing the formation process in a compact way, we confine ourselves to
operators and denote their composition by $\circ$.
To be precise, given operators $O_1$ and $O_2$, we write $(O_1 \circ O_2)$
instead of $(\lambda x.O_1(O_2(x)))$.
Furthermore, we use $O^k$ to denote $k$ consecutive applications of $O$ for some
arbitrary positive integer $k$.

For instance, Theorem~\ref{thm:pre:algo:main:ver1} captures sequences that are
obtained by $k=n$ applications of $\OPchoose_{\bg{}}^\circ$; this is indicated
in Table~\ref{tab:ops:summary} by ${[\OPchoose_{\bg{}}^\circ]^k}$.
The two final tests on $C^n$, viz. $\col^n=\fOPprop{\bg{}}{{\col^n}}$ and
$\col^n={\fOPMaxGround{\bg{}}{\col^n}}$, are pointed out by
${\OPprop{\bg{}}},{\OPMaxGround_{\bg{}}}$.
\begin{table}[htbp]
  \renewcommand{\arraystretch}{1.5}
  \newcommand{\op}[7]{#1&{#2}&\ref{#3}&\ensuremath{#4}&\ensuremath{#5}&#6&#7}
  \centering
  \begin{tabular}{l|llrlll}
    &Section&Theorem&Formation&Check&Properties&Properties
    \\
    &&&Process&&&Prefix sequences
    \\
    \hline
    \op{I}{\ref{sec:operations:characterization}}
       {thm:pre:algo:main:ver1}
       {[\OPchoose_{\bg{}}^\circ]^k}
       {{\OPprop{\bg{}}},{\OPMaxGround_{\bg{}}}}
       {1--5}
       {1--3}
    \\
    \op{II}{}%{sec:operations:characterization}
       {thm:pre:algo:main:ver2}
       {[(\OPprop{}\OPMaxGround)_{\bg{}}^\ast\circ\OPchoose_{\bg{}}^{\circ}]^k\circ(\OPprop{}\OPMaxGround)_{\bg{}}^\ast}
       {-}
       {1--7}
       {1--3,5--7}
    \\
    \op{II$^+$}{\ref{sec:operations:unicoloring}}
       {thm:pre:algo:main:ver2:pm}
       {[(\OPprop{}\OPMaxGround)_{\bg{}}^\ast\circ\OPchoose_{\bg{}}^{\cplus }]^k\circ(\OPprop{}\OPMaxGround)_{\bg{}}^\ast}
       {-}
       {1--7}
       {1--3,5--7}
    \\
    \op{II$^-$}{}%{sec:operations:unicoloring}
       {thm:pre:algo:main:ver2:pm}
       {[(\OPprop{}\OPMaxGround)_{\bg{}}^\ast\circ\OPchoose_{\bg{}}^{\cminus}]^k\circ(\OPprop{}\OPMaxGround)_{\bg{}}^\ast}
       {-}
       {1--7}
       {1--3,5--7}
    \\
    \op{III$^+$}{}%{sec:operations:unicoloring}
       {thm:sequence:plus}
       {\OPMaxGround_{\bg{}}\circ\OPpropast{\bg{}}\circ[\OPchoose_{\bg{}}^{\cplus }]^k}
       {-}
       {1--5}
       {1--3,5}
    \\
    \op{III$^-$}{}%{sec:operations:unicoloring}
       {thm:sequence:minus}
       {                         \OPpropast{\bg{}}\circ[\OPchoose_{\bg{}}^{\cminus }]^k}
       {-}
       {1--5}
       {1--3,5}
    \\
    \op{IV}{\ref{sec:operations:support:I}}
       {thm:sequence:D:plusminus}
       {\mathcal{N}_{\bg{}}\circ[\mathcal{D}_{\bg{}}^\circ]^k}
       {{\OPprop{\bg{}}}}
       {1--5,8}
       {1--3,5,8}
    \\
    \op{IV$^+$}{}%{sec:operations:support:I}
       {thm:sequence:D:plus}
       {\mathcal{N}_{\bg{}}\circ[\mathcal{D}_{\bg{}}^\cplus]^k}
       {\OPprop{\bg{}}}
       {1--5,8--9}
       {1--3,5,8--9}
    \\
    \op{V}{}%{sec:operations:support:I}
       {thm:sequence:DP:plusminus}
       {\mathcal{N}_{\bg{}}\circ[\OPpropast{\bg{}}\circ\mathcal{D}_{\bg{}}^\circ]^k\circ\OPpropast{\bg{}}}
       {\OPprop{\bg{}}}
       {1--8}
       {1--3,5--8}
   \\
    \op{V$^+$}{}%{sec:operations:support:I}
       {thm:sequence:DP:plus}
       {\mathcal{N}_{\bg{}}\circ[\OPpropast{\bg{}}\circ\mathcal{D}_{\bg{}}^\cplus]^k\circ\OPpropast{\bg{}}}
       {\OPprop{\bg{}}}
       {1--9}
       {1--3,5--9}
    \\
    \op{VI}{\ref{sec:operations:support:II}}
       {thm:pre:algo:main:ver4}
       {[(\OPprop{}\fweakS)_{\bg{}}^\ast\circ\mathcal{D}_{\bg{}}^{\circ}]^k\circ(\OPprop{}\fweakS)_{\bg{}}^\ast}
       {-}
       {1--8}
       {1--3,5--8}
    \\
    \op{VI$^+$}{}%{sec:operations:support:II}
       {thm:pre:algo:main:ver4:pm}
       {[(\OPprop{}\fweakS)_{\bg{}}^\ast\circ\mathcal{D}_{\bg{}}^{\cplus }]^k\circ(\OPprop{}\fweakS)_{\bg{}}^\ast}
       {-}
       {1--9}
       {1--3,5--9}
    \\
    \op{VI$^-$}{}%{sec:operations:support:II}
       {thm:pre:algo:main:ver4:pm}
       {[(\OPprop{}\fweakS)_{\bg{}}^\ast\circ\mathcal{D}_{\bg{}}^{\cminus}]^k\circ(\OPprop{}\fweakS)_{\bg{}}^\ast}
       {-}
       {1--8}
       {1--3,5--8}
  \end{tabular}
  \caption{Summary of operational characterizations.}
  \label{tab:ops:summary}
\end{table}

We observe that all successful coloring sequences enjoy properties 1-5.
Properties~6 and~7 rely on exhaustive propagations, at least with
operator~$\OPpropast{\bg{}}$.
While Property~8 is guaranteed in all support-driven characterizations,
Property~9 is only warranted when unicoloring with
$\mathcal{D}^{\cplus}_{\bg{}}$.
In fact, we see that exhaustive propagations moreover enforce that the
respective properties (except for Property~4 and in one case Property~5) are
also enjoyed by prefix sequences.
That is, sequences satisfying the first two conditions, thus sharing the format
\(
[O_2]^k\circ O_1
\)
for some combination of operators $O_i$ for $i=1,2$.
This is interesting from a computational point of view, since the more
properties are enforced on partial colorings, the smaller is the overall search
space.
For brevity, we refrain from giving explicit theorems on prefix sequences (in
addition to Theorem~\ref{thm:collection:OASC:ii:sub}) in this paper,
since their proofs are obtained in a straightforward way.

Finally, let us summarize the computational complexity of the various operators.
%
%---------------------------------------------------------------
\begin{theorem}\label{thm:complexity:operators}
  Let \bg{} be the \RDGraph{} of logic program $\Pi$ and
  let $\col$ be a partial coloring of \bg{}.

  If $n$ is the number of rules in $\Pi$,
  then
  \begin{enumerate}
  \item $\fOPprop{\bg{}}{\col}$ is computable in $\CO{n}$,
  \item $\fOPpropast{\bg{}}{\col}$ is computable in $\CO{n}$,
  \item $\fOPMaxGround{\bg{}}{\col}$ is computable in $\CO{n}$, 
  \item $\fPUast{\bg{}}{\col}$ is computable in $\CO{n^2}$,
  \item $\fOPV{\bg{}}{\col}$ is computable in $\CO{n}$, 
  \item $\fPVast{\bg{}}{\col}$ is computable in $\CO{n^2}$,
  \item $\fOPN{\bg{}}{\col}$ is computable in $\CO{n}$.

  \end{enumerate}
 
\end{theorem}
%---------------------------------------------------------------
%
In view of this result,
we can decide in polynomial time whether a coloring sequence is in accord with a
particular characterization.
Hence, a successful coloring sequence can be generated in nondeterministic polynomial time.

%%% Local Variables: 
%%% mode: latex
%%% TeX-master: "paper"
%%% End: 

%% file: wfs.tex
\section{Fitting and Well-founded Semantics}
\label{sec:wfs}

Unlike answer sets semantics,
other approaches rely on 3-valued models (or partial models).
Such a model consists of three parts: 
a set of  true   atoms,
a set of false   atoms, and
a set of unknown atoms.
%% Given that the union of these three sets is $\atm$, it is sufficient to specify
%% two of the three sets for determining a 3-valued interpretation.
Accordingly, a 3-valued interpretation $I$ is a pair $(X,Y)$ where $X$ and
$Y$ are sets of atoms with $X\cap Y=\emptyset$. 
That is, $L\in X$ means that $L$ is true in $I$,
while $L\in Y$ means that $L$ is false in $I$.
Otherwise, $L$ is considered to be unknown in $I$.
The most prominent among these semantics are due to
Fitting~\cite{fitting02fixpoint,ross92} and Van 
Gelder~\cite{vangelder91wellfounded}.
In contrast to answers sets semantics, both aim at characterizing skeptical
conclusions comprised in a single 3-valued model of the underlying program.
Interestingly, this 3-valued model provides an approximation of answer sets
semantics in the sense that all atoms true in a 3-valued model belong to all
answer sets of a given program,
and all false atoms are excluded from all answer sets.

Among both 3-valued semantics, less conclusions are obtained in Fitting's
semantics.
It can be defined by means of the following operator.
%
%---------------------------------------------------------------
\begin{definition}\label{def:fitt}
  Let $\Pi$ be a logic program and let $X,Y$ be sets of atoms. 
  
  We define
  \begin{eqnarray*}
    \fittingPos{\Pi}{X}{Y}
    &=&
    \{\head{\r} \mid \r\in \Pi,\pbody{\r}\subseteq X,\nbody{\r}\subseteq Y\}
    \\
    \fittingNeg{\Pi}{X}{Y}
    &=&
    \{q \mid \text{ for all } \r\in\Pi,
    \\&&\quad
    \text{ if }  \head{\r}=q \text {, then }
    (\pbody{\r}\cap Y \not=\emptyset\text{ or }\nbody{\r}\cap X \not=\emptyset)\}
    \ .
  \end{eqnarray*}
\end{definition}
%---------------------------------------------------------------
%
The pair mapping
\(
\fitting{\Pi}{X}{Y}=(\fittingPos{\Pi}{X}{Y},\fittingNeg{\Pi}{X}{Y})
\)
is often referred to as Fitting's operator~\cite{fages94a}.
Given that $\fittingo{\Pi}$ is monotonic,
we may start with $(\emptyset,\emptyset )$ and iterate $\fittingo{\Pi}$ until
its least fixpoint is reached.
Iterated applications of $\fittingo{\Pi}$ are written as $\fittingo{\Pi}^i$ for
$i\geq 0$, where
\(
\fittingo{\Pi}^0(X,Y)=(X,Y)
\)
and
\(
\fittingo{\Pi}^{i+1}(X,Y)=\fittingo{\Pi}\fittingo{\Pi}^i(X,Y)
\)
for $i\geq 0$.
We denote this least fixpoint by $\lfp{\fittingo{\Pi}}$.
We have $\lfp{\fittingo{\Pi}}=\fittingo{\Pi}^\omega(\emptyset,\emptyset)=\fittingo{\Pi}^n(\emptyset,\emptyset)$ for some
$n < \omega$ because $\Pi$ is finite.

For relating this operator to the ones on \RDGraph{}s, we need the following
definition.
%
%---------------------------------------------------------------
\begin{definition}\label{def:C:XY}
  Let \bg{} be the \RDGraph{} of logic program $\Pi$
  and
  let $C$ be a partial coloring of $\bg{}$.
  
  We define
  \begin{eqnarray*}
    X_C&=&\{\head{r} \mid  r\in \colplus\},
    \\
    Y_C&=&\{q \mid \text{for all } r\in\Pi, 
           \text{ if } \head{r}=q, \text{ then } r\in\colminus \}.
  \end{eqnarray*}
\end{definition}
%---------------------------------------------------------------
%
The pair $(X_C,Y_C)$ is a  3-valued interpretation of $\Pi$.
%% We define $(X,Y) \subseteq (X',Y')$ if $X\subseteq X'$ and $Y\subseteq Y'$ for
%% 3-valued interpretations.
%% %

We have the following result.
%
%---------------------------------------------------------------
\begin{theorem}\label{thm:wfs:fitting:P:empty}
  Let \bg{} be the \RDGraph{} of logic program $\Pi$.
  
  If $\col=\fOPpropast{\bg{}}{(\emptyset,\emptyset)}$, 
  then $\fittingo{\Pi}^\omega(\emptyset,\emptyset)=( X_C,Y_C )$.

%%   If  $\fittingo{\Pi}^\omega=\langle X,Y \rangle$  and
%%   $\fOPpropast{\bg{}}{(\emptyset,\emptyset)}=\col$,
%%   then
%%   $X=\head{\colplus}$ and $Y=\atm \setminus \head{\Pi\setminus
%%   \colminus}$. 
%%
%%    If $\fOPpropast{\bg{}}{(\emptyset,\emptyset)}=\col$,
%%    $X=\head{\colplus}$ and $Y=\atm \setminus \head{\Pi\setminus
%%    \colminus}$,
%%    then $\fittingo{\Pi}^\omega=\langle X,Y \rangle$. 
%%
\end{theorem} 
%---------------------------------------------------------------
%
Note that $\fOPpropast{\bg{}}{(\emptyset,\emptyset)}$ as well as
$\fittingo{\Pi}^\omega(\emptyset,\emptyset)$ always exists
(Cf.\ Corrollary~\ref{cor:P:empty:exists}).

For capturing other semantics, the construction $\Cn{\Pi^X}$ is sometimes
regarded as an operator $C_{\Pi}(X)$.
The anti-monotonicity of $C_{\Pi}$ implies that $C^2_{\Pi}$ is monotonic. 
As shown in~\cite{vangelder93}, 
different semantics are obtained by distinguishing different groups of
(alternating) fixpoints of $C^2_{\Pi}(X)$.
For instance, given a program $\Pi$,
the least fixed point of $C^2_\Pi$ is known to amount to its
\emph{well-founded semantics}.
Answer sets of $\Pi$ are simply fixed points of $C^2_\Pi$ that are
also fixed points of $C_{\Pi}$.
The well-founded model can be characterized in terms of the least fixpoint of
operator $C^2_\Pi$. 
That is,
the well-founded model of a program $\Pi$ is given by the 3-valued
interpretation
\(
(\lfp{C^2_\Pi},\atm\setminus C_\Pi{\lfp{C^2_\Pi}})
\).
Hence, it is sufficient to consider the least fixpoint of $C^2_\Pi$, since it
determines the well-founded model. 
We therefore refer to the least fixpoint of $C^2_\Pi$ as the
\emph{well-founded set} of $\Pi$.
The set $\atm\setminus C_\Pi{\lfp{A_\Pi}}$ is usually referred to as the
\emph{unfounded} set of $\Pi$.

Concerning 3-valued interpretations we obtain the following definition
 of unfounded sets (according
 to~\cite{vangelder91wellfounded}).
%---------------------------------------------------------------
\begin{definition}\label{def:unfounded set}
  Let $\Pi$ be a logic program and let $Z$ be a set of atoms. 
  
  Furthermore, let $(X,Y)$ be a 3-valued interpretation.

  Then, $Z$ is an unfounded set  of $\Pi$ wrt $(X,Y)$ if each $q\in Z$ 
 satisfies the following condition:
 For each $r\in\Pi$ with $\head{r}=q$, one of the following conditions hold:
 \begin{enumerate}
 \item $\pbody{r}\cap Y \not=\emptyset$ or
       $\nbody{r}\cap X\not=\emptyset$;
 \item there exists a $p\in \pbody{r}$ such that $p\in Z$.
 \end{enumerate}

\end{definition}
%---------------------------------------------------------------
 The greatest unfounded set of $\Pi$ wrt $(X,Y)$, denoted $\GUS{\Pi}{X,Y}$, is the  union of all sets that are unfounded wrt $(X,Y)$.

To begin with, we fix the relationship among greatest unfounded sets and
maximal support graphs.
%
%---------------------------------------------------------------
 \begin{theorem}\label{thm:wfs:GUS}
   Let \bg{} be the \RDGraph{} of logic program $\Pi$ 
   and $C$ be a partial coloring of \bg{} such that
   $\colminus\subseteq \overline{S}(\bg{},C)\cup B(\bg{},C)$. 

   Furthermore, let $(V,E)$ be a maximal support graph of $(\bg{},C)$ for some
   $E\subseteq \Pi\times\Pi$.

   Then, $(\atm\setminus \head{V})$ is the greatest unfounded set of $\Pi$ wrt   $(X_C,Y_C)$.

 \end{theorem}
%---------------------------------------------------------------
%
This result can be expressed in terms of operator $\OPMaxGround_{\bg{}}$.
%
%---------------------------------------------------------------
 \begin{corollary}\label{cor:wfs:GUS}
   Let \bg{} be the \RDGraph{} of logic program $\Pi$ 
   and $C$ be a partial coloring of \bg{} such that
   $\colminus\subseteq \overline{S}(\bg{},C)\cup B(\bg{},C)$. 

   If $C'=\fOPMaxGround{\bg{}}{C}$, then
   $(\atm\setminus  \head{\Pi\setminus\colminus'})$ is the greatest unfounded
   set of $\Pi$ wrt $(X_C,Y_C)$.
 \end{corollary}
%---------------------------------------------------------------
%
Finally, we can express the well-founded semantics in terms of our operators in
the following way.
%
%---------------------------------------------------------------
\begin{theorem}\label{thm:wfs:main}
  Let \bg{} be the \RDGraph{} of logic program $\Pi$. 

  If $\col=(\OPprop{}\OPMaxGround)^\ast_{\bg{}}((\emptyset,\emptyset))$, 
  then $( X_C,Y_C )$ is the well-founded model of $\Pi$.
\end{theorem}
%---------------------------------------------------------------
%
Hence, by Theorem~\ref{thm:complexity:operators}, the well-founded model is
computable in our approach in quadratic time in size of $\Pi$.

%%% Local Variables: 
%%% mode: latex
%%% TeX-master: "paper"
%%% End: 

%% file: discussion.tex
\section{Discussion and related work}
\label{sec:discussion}\label{sec:relatedwork}

Our approach has its roots in earlier work~\cite{linsch98a,linsch00a},
proposing \emph{block graphs} as a tool for query-answering in default
logic~\cite{reiter80} and the underlying existence of extensions problem.
Roughly speaking,
these graphs are closely related to the blockage graphs introduced in
Section~\ref{sec:check} because both possess a single type of edges
indicating blockage relations.
Inspired by the distinction between supporting and blocking arcs made
in~\cite{papsid94},
this has led to our approach to characterizing and computing answer sets;
it provides the theoretical foundations of the \texttt{noMoRe} answer set
programming system~\cite{ankoli02a}.

In this paper,
we put forward the simple concept of a rule dependency graph (\RDGraph) for
capturing the interplay between rules inducing answer sets.%
\footnote{The definition of the \RDGraph{} differs from that in~\cite{linke01a},
whose practically motivated restrictions turn out to be superfluous from a
theoretical perspective.}$^,$%
\footnote{Due to the aforementioned historical development, the \RDGraph{} was
  in~\cite{linke01a,lianko02a} still referred to as ``block graph''.
  We abandon the latter term in order to give the same status to
  support and blockage relations.}
Many other forms of dependency graphs can be found in the literature.
For instance, dependency graphs (DGs) among predicate symbols were proposed
in~\cite{apblwa87} for defining \emph{stratified}  programs.
DGs among atoms can be defined
analogously~\cite{przymusinski88a}:
The nodes of such graphs are atoms appearing in a program $\Pi$;
edges are distinguished similar to
Definition~\ref{def:blockgraph:redundant}, viz.\
\(
E_0=\{(\head{r},p)\mid r\in\Pi,p\in\pbody{r}\}
\)
and
\(
E_1=\{(\head{r},p)\mid r\in\Pi,p\in\nbody{r}\}
\).
Originally, these edges are referred to as being positive and negative,
respectively.
Accordingly, a cycle is said to be negative if it contains some negative
edge.
With this, a program is \emph{stratified} if its DG  
does not contain a negative cycle.
Stratified logic programs have a unique answer set~\cite{gellif88b}.
Many other properties, obtained from the structure of the DG, 
were identified for investigating the consistency of Clark's
completion~\cite{clark78}.
In fact, it is shown in~\cite{fages94a} that most of them also guarantee the
existence of answer sets.
As discussed in~\cite{constan01a}, DGs do not allow for capturing answer set
semantics of logic programs.
The difficulty is that there are syntactically and semantically different
programs having the same DG.

Among the more recent literature, we find~\cite{dimtor96}, where 
rule dependency graphs are defined for \emph{reduced} \emph{negative} programs.
A program is \emph{negative} if it includes only rules $r$ where
$\pbody{r}=\emptyset$.
Informally, a program is \emph{reduced} if different rules 
$h_1\LPif B,\dots,h_k\LPif B$
with same body $B$
are merged into one rule 
$h_1\wedge\dots\wedge h_k\LPif B$
where the head is a conjunction of atoms.
When restricting our attention to negative programs with unique bodies,
the graphs of~\cite{dimtor96} amount to \RDGraph{s} restricted to $1$-edges.
The following interesting results are shown in~\cite{dimtor96} for reduced
programs:
Stable models~\cite{gellif88b}, 
partial stable models~\cite{saczan90a},
and well-founded semantics~\cite{vangelder91wellfounded} 
of reduced negative programs
correspond to kernels, semi-kernels and the initial acyclic
part of the corresponding \RDGraph{}, respectively.
General programs are dealt with by program transformations, turning general
programs to reduced negative ones. 

Another interesting and closely related graph-theoretical approach is described
in~\cite{bricos99,codapr02a,cospro03a}.
Although their primary focus lies on special negative programs, referred to as
\emph{kernel programs} (see below) their dependency graph can be defined in a
general way.
This approach relies on extended dependency graphs (EDG), whose nodes are given
by the multi-set of rule heads together with atoms not appearing as heads.
For a program $\Pi$, the set of vertices amounts to the set
\(
\{(\head{r},r)\mid r\in\Pi\}\cup\{(u,u)\mid u\not\in\head{\Pi}\}
\).
There is a positive edge $(u,v)$ in the EDG if 
$u\in\pbody{r}$ and $v=\head{r}$ for some rule $r\in\Pi$;
there is a negative edge $(u,v)$ in the EDG if
$u\in\nbody{r}$. 
Hence EDGs and \RDGraph{s} are generally different.
For example, take program 
\(
\{r_a,r_b\}
=
\{a \LPif \LPnot\;b, \; b \LPif\LPnot\;a, \LPnot\;c\}
\). 
Then the EDG of this program has three nodes $a,b$ and $c$ and
three negative edges $(b,a)$, $(a,b)$ and $(c,b)$,
whereas the \RDGraph{} has two nodes $r_a$ and $r_b$ and two edges 
$(r_a,r_b)$ and $(r_b,r_a)$.
\emph{Kernel programs} are negative programs subject to the condition that each
head atom must also appear as a body atom in the program and all atoms in the
program must be undefined in its well-founded model.
According to~\cite{bricos99}, every normal program 
can be transformed into some equivalent kernel program.
In analogy to~\cite{dimtor96}, general programs are then dealt with through
program transformations.
Interestingly, it is shown in~\cite{codapr02a} that EDGs and \RDGraph{s} are isomorphic
for kernel programs.
EDGs are used in~\cite{bricos99} to study properties of logic programs, 
like existence of answer sets.
Furthermore, colorings of EDGs are used in~\cite{bricos99} for characterizing
answer sets of kernel programs.
For kernel programs, these colorings correspond to the ones studied in
Section~\ref{sec:check} (because of the aforecited isomorphism).
Interestingly, \cite{bricos99} defines admissible colorings in terms of their
``complements''.
That is, informally in our terminology, a coloring \col{} is
\emph{non-admissible} if for some $u$, either
(i) $u\in\colplus{}$, and for some $(u,v)\in E_1$, $v\in\colplus{}$,
or
(ii) $u\in\colminus{}$ and for all $(u,v)\in E_1$, $v\in\colminus{}$.
Then, \col{} is admissible if it is not non-admissible.
This definition is only concerned with blockage and thus applies to
negative programs only.
On the other hand, it is closely related to the concept of a blockage graph
(cf.~Definition~\ref{def:acol:onedsg}).
That is, in the case of negative programs, the blockage graph amounts to an
admissibly colored EDG.
To see this, compare the negation of Condition~2 and~3 in
Corollary~\ref{thm:acol:acolorings:ver1} with Condition~(i) and~(ii) above
(while setting ${S}(\bg{},\col)$ to $\Pi$).
A blockage graph may thus be regarded as a natural extension of the concept of
blockage used in \cite{bricos99} from negative to general logic programs.

In all, the major difference between the two latter approaches~\cite{dimtor96,bricos99}
and ours boils down to the indirect or direct treatment of positive body
atoms, respectively.
While our techniques are developed for full-fledged logic programs,
\cite{dimtor96,bricos99} advocate an initial transformation to negative programs,
on which their methods are primarily defined.
(Note that general logic programs cannot be reduced to negative ones in a
modular way~\cite{janhunen00a}.)
Rather all our characterizations of answer sets in Section~\ref{sec:check}
stress the duality between supporting and blocking relations among rules.
Finally, it is noteworthy that all three approaches address several rather
different problems.
While we are primarily interested in operational characterizations, the two
other approaches address fundamental problems such as the existence of answer
sets.
In particular, they elaborate upon the graph-theoretical concept of negative
programs. 
In this way, all approaches are nicely complementary to each other and therefore
do largely benefit from each other.
An overview over different graphs associated with logic programs can be found
in~\cite{constan01a}.

We have introduced support graphs for capturing the inferential dependencies
among rule heads and positive body atoms.
In fact, support graphs may be seen as a (rule-oriented) materialization of the
notion of a \emph{well-supported} interpretation, or more precisely, its underlying
well-founded partial order (cf.~\cite{fages94a}):
An interpretation $X$ is well-supported if there is a strict well-founded partial
order $\prec$ on $X$ such that for every $p\in X$ there is some
$r\in\GR{\Pi}{X}$ with $p=\head{r}$ and $q\prec p$ for every $q\in\pbody{r}$.
An edge $(r_1,r_2)$ in a support graph of a colored \RDGraph{} corresponds to the pairs
$\head{r_2}\prec\head{r_1}$ for $q\in\pbody{r_2}$.

A major goal of our paper is to provide operational characterizations of
answer sets that allow us to bridge the gap between formal yet static characterizations of
answer sets and algorithms for computing them.
For instance, in the seminal paper~\cite{niesim96a} describing the approach
underlying the \texttt{smodels} system,
the characterization of answer sets is given in terms of so-called \emph{full-sets}
and their computation is directly expressed in terms of procedural algorithms.
Our operational semantics aims at offering an intermediate stage that facilitates
the formal elaboration of computational approaches.
Our approach is strongly inspired by the concept of a derivation, in
particular, that of an SLD-derivation~\cite{lloyd87}.
This attributes our coloring sequences the flavor of a derivation in a family of
calculi, whose respective set of inference rules correspond to the selection of
operators.
A resolution calculus for skeptical stable model semantics is given
in~\cite{bonatti01a}.
Interestingly, this calculus is not derived from credulous inference;
also, it does not need the given program to be instantiated before reasoning.
Gentzen-style calculi for default logic (and thus implicitly also for logic
programming) can be found in~\cite{bonatti96a,bonoli97a}.

Regarding our modeling of operations,
it is worth mentioning that one could also use total operations instead of
partial ones.
For instance, instead of defining $\classC$ as a set of
partial mappings,
one could consider the set of a binary partitions of $\Pi$ \emph{plus}
$(\Pi,\Pi)$ as the representative for inconsistent colorings.
For instance, $\OPprop{\bg{}}$ could then be defined as a mapping
\(
\OPprop{\bg{}}: \classC \rightarrow \classC\cup\{(\Pi,\Pi)\}
\),
where $(\Pi,\Pi)$ is obtained whenever $\OPprop{\bg{}}$ ``detects'' an
inconsistency.
Although such an approach seems natural in a logical setting, involving
deductive closure,
we put forward partial mappings in our abstract operational setting.
In this way, an answer set exists iff there \emph{exists} a corresponding coloring
sequence.
Accordingly, there is no answer set iff there is no coloring sequence.
This nicely corresponds to the concept of a derivation and notably avoids the
distinction between coloring sequences leading to admissible and inconsistent
colorings.

We have furthermore shown in Section~\ref{sec:wfs} that particular operations
correspond to Fitting's and well-founded
semantics~\cite{fitting02fixpoint,vangelder91wellfounded,przymusinski90wellfounded,dematr00a}.
A stepwise characterization of both semantics was proposed
in~\cite{brass98transformationbased} by defining a confluent rewriting system.
The rewrite of a program corresponds to a 3-valued interpretation
in which all facts in the rewritten program are true and all atoms not appearing
among the heads of the rewrite are false.
All other atoms are undefined.
The program transformations $\mapsto_P$ and $\mapsto_S$ delete atoms from
the positive and negative part of the body,
if they are true in the associated 3-valued interpretation.
If a rule is transformed into a fact through these transformations, then this
corresponds in our approach to coloring this rule with $\cplus$ by $\OPprop{\bg{}}$.
Analogously, the transformations $\mapsto_N$ and $\mapsto_F$ delete rules which
have atoms in their body being false in the 3-valued interpretation.
These transformations correspond to coloring non-applicable rules with $\cminus$.
In this way, the iterative application of these 4 transformations yields the
least fixpoint of Fitting's operator.
In view of Theorem~\ref{thm:wfs:fitting:P:empty}, these 4 transformations have
the same effect as operator $\OPprop{\bg{}}$.
Another program transformation, viz.~$\mapsto_L$, is introduced
in~\cite{brass98transformationbased} for $0$-loop detection; thus
allowing for computing the greatest unfounded set. 
This transformation is similar to operator $\OPMaxGround_{\bg{}}$.
Taken together, $\mapsto_P,\mapsto_S,\mapsto_N,\mapsto_F,\mapsto_L$ form a
confluent rewriting system, whose final rewrite corresponds to the well-founded
model of the initial logic program.

Although we leave algorithmic and implementation issues to a companion
paper, in particular those, dealing with our system
\texttt{noMoRe},
some remarks relating our approach to the ones underlying the answer set
programming systems \texttt{dlv}~\cite{eifalepf99a,dlv,dlv03a} and
\texttt{smodels}~\cite{smodels,niesim96a,siniso02a} are in order.
A principal difference manifests itself in how choices are performed.
While the two latter's choice is based on atoms occurring (negatively) in the
underlying program,
our choices are based on its rules.
The former approach is per se advantageous whenever multiple rules share a
common head.
This is compensated in the \texttt{noMoRe} system by additional propagation
rules eliminating a rule from the inference process, once its head has been
derived in an alternative way.
From a general perspective,
a rule-based choice can be regarded as a compound choice on atoms.
That is, assigning a rule $r$ a positive applicability status (via $\cplus$)
corresponds to assigning all atoms in $\head{r}\cup\pbody{r}$ the value true and
all atoms in $\nbody{r}$ the value false.
Conversely, assigning $r$ a negative status of applicability (viz.~$\cminus$)
corresponds to assigning at least one atom in $\pbody{r}$ the value false or one
in $\nbody{r}$ the value true.
Only if all rules with the same head are colored with $\cminus$, a rule head can
be assigned false.
While this type of choice is realized by $\OPchoose^{\circ}_{\bg{}}$,
the one by $\mathcal{D}^{\circ}_{\bg{}}$ is more restrictive since all atoms in
$\pbody{r}$ are known to be true.
%% \comment{UNCOMMENTED MATERIAL!!}
%% In general, however,
%% it is an open question how the size of the search space is affected by choosing
%% literals as opposed to rules.\footnote{Thanks to Miros{\l }aw Truszczy{\'n}ski
%%   for raising this question.}
%
An advantage of the approach based on choice operator $\mathcal{D}^{\circ}_{\bg{}}$ is that we can guarantee the support of rules on
the fly.
The elimination of unsupported rules can then either be restricted to
uncolored rules, as done with operator $\fweakS_{\bg{}}$,
or even done in a final step without further detection efforts by appeal to operator \OPN{\bg{}}.
Interestingly, the choice operator of \texttt{dlv} has also a
support-driven flavor: When choosing a (negative) body literal $q$, one of the
qualifying conditions is the existence of a rule $r$ such that $q\in\nbody{r}$
and $\pbody{r}\subseteq I$, where $I$ is the current partial
assignment.\footnote{Formally, $I$ is a four-valued interpretation. 
Two further conditions qualify the (disjunctive) head and the negative body
literals of the rule $r$ depending on their truth values.}${}^,$%
\footnote{No support is taken into account by \texttt{dlv} when choosing a (positive) head literal.}
Unlike this, support checking is a recurring operation in the
\texttt{smodels} system, similar to operator $\OPMaxGround_{\bg{}}$.
On the other hand, this approach ensures that the \texttt{smodels} algorithm
runs in linear space complexity, while a graph-based approach needs quadratic
space in the worst case (due to its number of edges).
Interestingly, \texttt{smodels}' implementation relies on a rule-head dependency
graph, in which rules and atoms are connected via pointers.
Such an investment in space pays off once one is able to exploit the
additional structural information offered by a graph.
First steps in this direction are made in~\cite{linke03a}, where graph
compressions are described that allow for conflating entire subgraphs into
single nodes.
Propagation is more or less done similarly in all three approaches.
That is,
all these systems follow the strategy coined in~\cite{sunava95a},
namely
``Answer Sets = Well-founded Semantics + Branch and Bound''.
\texttt{smodels} relies on computing well-founded semantics, whereas \texttt{dlv}
uses Fitting's  or well-founded semantics, depending on whether
(in our terminology) the program's \RDGraph{} contains 0-cycles or not~\cite{cafalepf01}.
Also, both systems use back-propagation mechanisms.
In \texttt{dlv}, this allows to mark atoms as being eventually true.
Operators capturing \texttt{dlv}'s propagation operations are given
in~\cite{faber02a}.
Among them, operator $T_\Pi$ amounts to our Operator $\OPprop{\Pi}$; others
address the aforementioned back-propagation and propagation of ``eventually
true'' atoms.
In addition to the propagation operators discussed in Section~\ref{sec:operations},
\nomore{} also uses different types of back-propagation~\cite{lianko02a},
including special treatment of integrity constraints,
as well as
operations for ignoring rules once their head has been established.
What truly distinguishes \nomore{}'s propagation operations is their
support-preserving way in conjunction with choice operator
$\mathcal{D}^{\circ}_{\bg{}}$ (cf.~Property~8 in
Theorem~\ref{thm:collection:sequence:D:plusminus},
\ref{thm:collection:sequence:D:plus},
\ref{thm:collection:sequence:DP:plusminus},
\ref{thm:collection:sequence:DP:plus}, and
\ref{thm:collection:OASC:iv}).

\input{experiments}

The fact that \texttt{dlv} deals with disjunctive programs makes many of its
special features inapplicable in our setting of normal logic programs.
Alternative approaches can be found in~\cite{linzha04a,cmodels}, where answer sets
are computed by means of SAT solvers.
Algorithms and implementation techniques for computing well-founded semantics
can be found among others in~\cite{saswwa96,lontru01a}.

%%% Local Variables: 
%%% mode: latex
%%% TeX-master: "paper"
%%% End: 

%% file: experiments.tex
\newcommand{\experiment}[3]{#1&#2&#3}
\newcommand{\operatordescription}{
\multicolumn{3}{c|}{V} &
\multicolumn{3}{c|}{VI} &
\multicolumn{3}{c|}{VI$+h_{sm}$} &
\multicolumn{3}{c|}{}
}
\newcommand{\resultdescription}{chs & ass & time & chs & ass & time & chs & ass & time & chs & ass & time}
%
%%%%%%%%%%%%%%%%%%%%%%%%%%%%%%%%%%%%%%%%%%%%%%%%%%%%%%%%%%%%%%%%%%%%%%%
\begin{table}[h!]
\begin{center}
{\footnotesize
\begin{tabular}{|c|@{}r@{}r@{}r|@{}r@{}r@{}r|@{}r@{}r@{}r|@{}r@{}r@{}r|}
\hline
 & 
\multicolumn{9}{c|}{\nomore} &
\multicolumn{3}{c|}{\smodels} 
\\ \hline
 & 
\operatordescription \\ 
$n$ &
\resultdescription
\\ \hline
{3}&\experiment{1}{39}{0.0}&\experiment{1}{39}{0.0}&\experiment{1}{175}{0.0}&\experiment{1}{45}{0.001}\\
{4}&\experiment{13}{240}{0.0}&\experiment{12}{241}{0.01}&\experiment{5}{710}{0.03}&\experiment{5}{381}{0.0}\\
{5}&\experiment{74}{1430}{0.06}&\experiment{64}{1420}{0.08}&\experiment{23}{3918}{0.23}&\experiment{26}{2757}{0.002}\\
{6}&\experiment{468}{9935}{0.59}&\experiment{385}{9687}{0.71}&\experiment{119}{24046}{1.88}&\experiment{305}{34202}{0.019}\\
{7}&\experiment{3370}{78803}{5.77}&\experiment{2676}{75664}{6.88}&\experiment{719}{0.18$M\!\!\!$}{16.98}&\experiment{4814}{0.53$M\!\!\!$}{0.319}\\
{8}&\experiment{27480}{0.70$M\!\!\!$}{61.84}&\experiment{21259}{0.67$M\!\!\!$}{72.99}&\experiment{5039}{1.47$M\!\!\!$}{173.81}&\experiment{86364}{9.17$M\!\!\!$}{6.29}\\
{9}&\experiment{0.25$M\!\!\!$}{6.97$M\!\!\!$}{730}&\experiment{0.19$M\!\!\!$}{6.55$M\!\!\!$}{849}&\experiment{40319}{14$M\!\!\!$}{3639}&\experiment{1.86$M\!\!\!$}{197$M\!\!\!$}{159}\\
\hline
\end{tabular}
}
\caption{Results for computing \textbf{all} answer sets of the Hamiltonian cycle
  problem on complete graphs with $n$ nodes.
  \textbf{Abbreviations}: \emph{chs} for choices, \emph{ass} for assigments,
  \emph{time} for time in seconds, and $M$ abbreviates millions.}\label{tab:all:k} 
\end{center}
\end{table}
%%%%%%%%%%%%%%%%%%%%%%%%%%%%%%%%%%%%%%%%%%%%%%%%%%%%%%%%%%%%%%%%%%%%%%%
\begin{table}[h!]
\begin{center}
{\footnotesize
\begin{tabular}{|c|@{}r@{}r@{}r|@{}r@{}r@{}r|@{}r@{}r@{}r|@{}r@{}r@{}r|}
\hline
 & 
\multicolumn{9}{c|}{\nomore} &
\multicolumn{3}{c|}{\smodels} 
\\ \hline
 & 
\operatordescription \\
$n$ &
\resultdescription
\\ \hline
{7}&\experiment{15}{386}{0.03}&\experiment{15}{386}{0.04}&\experiment{5}{1075}{0.13}&\experiment{30}{4701}{0.005}\\
{8}&\experiment{21}{639}{0.06}&\experiment{21}{639}{0.07}&\experiment{6}{1626}{0.22}&\experiment{8}{2941}{0.003}\\
{9}&\experiment{28}{1000}{0.11}&\experiment{28}{1000}{0.14}&\experiment{7}{2368}{0.4}&\experiment{48}{12555}{0.009}\\
{10}&\experiment{36}{1494}{0.19}&\experiment{36}{1494}{0.24}&\experiment{8}{3337}{0.75}&\experiment{1107}{193287}{0.155}\\
{11}&\experiment{45}{2148}{0.31}&\experiment{45}{2148}{0.57}&\experiment{9}{4571}{1.45}&\experiment{18118}{2.81$M\!\!\!$}{2.613}\\
{12}&\experiment{55}{2991}{0.53}&\experiment{55}{2991}{1.01}&\experiment{10}{6110}{2.47}&\experiment{0.39$M\!\!\!$}{56.6$M\!\!\!$}{60}\\
{13}&\experiment{66}{4054}{1.13}&\experiment{66}{4054}{1.62}&\experiment{11}{7996}{3.87}&\experiment{5.30$M\!\!\!$}{721$M\!\!\!$}{866}\\
{14}&\experiment{78}{5370}{1.92}&\experiment{78}{5370}{2.47}&\experiment{12}{10273}{5.62}&\experiment{---}{---}{$>$2h}\\
{15}&\experiment{91}{6974}{2.83}&\experiment{91}{6974}{3.47}&\experiment{13}{12987}{8.06}&\experiment{---}{---}{$>$2h}\\
{16}&\experiment{105}{8903}{4.01}&\experiment{105}{8903}{4.86}&\experiment{14}{16186}{11.07}&\experiment{---}{---}{$>$2h}\\
{17}&\experiment{120}{11196}{5.49}&\experiment{120}{11196}{6.58}&\experiment{15}{19920}{14.79}&\experiment{---}{---}{$>$2h}\\
{18}&\experiment{136}{13894}{7.15}&\experiment{136}{13894}{8.62}&\experiment{16}{24241}{19.74}&\experiment{---}{---}{$>$2h}\\
\hline
\end{tabular}
}
\caption{Results for computing \textbf{one} answer set of the Hamiltonian cycle problem on
  complete graphs with $n$ nodes ($M$ abbreviates millions).}\label{tab:one:k} 
\end{center}
\end{table}
%%%%%%%%%%%%%%%%%%%%%%%%%%%%%%%%%%%%%%%%%%%%%%%%%%%%%%%%%%%%%%%%%%%%%%%
\begin{table}[h!]
\begin{center}
{\footnotesize
\begin{tabular}{|c|@{}r@{}r@{}r|@{}r@{}r@{}r|@{}r@{}r@{}r|@{}r@{}r@{}r|}
\hline
 & 
\multicolumn{9}{c|}{\nomore} &
\multicolumn{3}{c|}{\smodels} 
\\ \hline
 & 
\operatordescription
\\ 
$n$ &
\resultdescription
\\ \hline
{4}&\experiment{289}{5132}{0.22}&\experiment{56}{1357}{0.09}&\experiment{8}{2006}{0.16}&\experiment{2}{1532}{0.001}\\
{4}&\experiment{18}{395}{0.02}&\experiment{14}{332}{0.02}&\experiment{2}{515}{0.03}&\experiment{1}{735}{0.001}\\
{4}&\experiment{396}{5468}{0.29}&\experiment{179}{3776}{0.28}&\experiment{11}{1834}{0.18}&\experiment{18}{10393}{0.007}\\
{4}&\experiment{22}{366}{0.02}&\experiment{21}{364}{0.02}&\experiment{6}{1000}{0.08}&\experiment{3}{1489}{0.002}\\
{4}&\experiment{118}{2223}{0.14}&\experiment{47}{1227}{0.11}&\experiment{13}{2763}{0.29}&\experiment{20}{13525}{0.008}\\
{5}&\experiment{3765}{54264}{3.06}&\experiment{37}{733}{0.08}&\experiment{14}{2721}{0.29}&\experiment{6}{4291}{0.004}\\
{5}&\experiment{1113}{13535}{1.08}&\experiment{93}{1779}{0.22}&\experiment{32}{4929}{1.1}&\experiment{2306}{1.08$M\!\!\!$}{0.775}\\
{5}&\experiment{207}{3340}{0.17}&\experiment{74}{1450}{0.14}&\experiment{11}{2611}{0.29}&\experiment{4}{2956}{0.003}\\
{5}&\experiment{1195}{14780}{1.14}&\experiment{191}{5390}{0.79}&\experiment{29}{7773}{1.64}&\experiment{201}{98546}{0.069}\\
{5}&\experiment{4535}{72129}{4.91}&\experiment{505}{15690}{2.12}&\experiment{54}{17425}{3.66}&\experiment{82}{60317}{0.038}\\
{6}&\experiment{359}{6563}{0.52}&\experiment{89}{2174}{0.47}&\experiment{346}{76915}{25.46}&\experiment{0.17$M\!\!\!$}{95$M\!\!\!$}{76}\\
{6}&\experiment{1228}{20970}{1.64}&\experiment{261}{6238}{1.38}&\experiment{4419}{1.06$M\!\!\!$}{335}&\experiment{0.24$M\!\!\!$}{214$M\!\!\!$}{152}\\
{6}&\experiment{1.71$M\!\!\!$}{33$M\!\!\!$}{3278}&\experiment{0.10$M\!\!\!$}{3.18$M\!\!\!$}{935}&\experiment{12608}{4.2$M\!\!\!$}{1537}&\experiment{--}{--}{$>$2h}\\
{6}&\experiment{233}{4937}{0.38}&\experiment{158}{3908}{0.96}&\experiment{1161}{0.27$M\!\!\!$}{94}&\experiment{14}{11841}{0.01}\\
{6}&\experiment{3237}{41286}{2.78}&\experiment{499}{8336}{1.94}&\experiment{57}{9162}{2.62}&\experiment{38}{38277}{0.025}\\
\hline
\end{tabular}
}
\caption{Results for computing \textbf{one} answer set of the Hamiltonian cycle
  problem on clumpy graphs with $n$ clumps ($M$ abbreviates millions).}\label{tab:one:clumpy} 
\end{center}
\end{table}
%%%%%%%%%%%%%%%%%%%%%%%%%%%%%%%%%%%%%%%%%%%%%%%%%%%%%%%%%%%%%%%%%%%%%%%
%
Finally, let us underpin the potential of our approach by some indicative
empirical results.
For this purpose, we have chosen the Hamiltonian cycle problem on two different
types of graphs, namely \emph{complete} and so-called \emph{clumpy graphs}, as put
forward in~\cite{warsch04a}.
This choice is motivated by the fact that 
--- unlike most of the other known benchmark examples ---
Hamiltonian problems 
naturally lead to non-tight encodings and thus comprise more complex
support structures.%
\footnote{Tight encodings for Hamiltonian problems were proposed in~\cite{linzhao03}.}
Each clumpy graph has a given number of clumps (sets of nodes) where
nodes are connected with more edges than between clumps.
That is, edges in clumpy graphs are distributed less uniform and solving
Hamiltonian cycle problems becomes more difficult.
All benchmark examples are included in the distribution of the \nomore\ system~\cite{nomore}.

All presented experiments have been done under \texttt{Linux} (kernel
2.6) on a \texttt{Intel Pentium 4} processor with $2.26$GHz and $512$MB main
memory.  
We have used \nomore\ V1.0~\cite{nomore}
%
%% \footnote{\nomore\ is available at \textsf{http://www.cs.uni-potsdam.de/\~{}linke/nomore}.}
under \texttt{Eclipse} Prolog with \RDGraph{s} (flag \textsf{asp\_r} set)
and \smodels\ version 2.27~\cite{smodels}.
Although we do not report it here, we have run the whole test series with
\dlv{} as well.
We have not included the results because \dlv{} outperforms
\smodels\ as well as \nomore{} on all problem instances as regards time.
%
%% For example, \dlv\ needed 1.7 seconds for computing all Hamiltonian cycles of
%% a complete graph with 8 nodes. 
%
Furthermore, \dlv\ does not report assignments and it has a different concept of
choices than \smodels{} and \nomore.

Tables~\ref{tab:all:k} and~\ref{tab:one:k} summarize
the results for computing \emph{all} and \emph{one} answer set for Hamiltonian cycle
problems on complete graphs with $n$ nodes, respectively.
Observe that, although the Hamiltonian cycle problem for complete
graphs is easy solvable for humans, it is difficult for ASP solvers.
The computation of all solutions reflects the system behavior on excessive
backtracking.\footnote{We also ran test series on non-answer-set instances,
  namely, Hamiltonian cycle problems
  on bipartite graphs;
  the overall result was the same as obtained with the series reported here.}
Table~\ref{tab:one:clumpy} gives results for computing Hamiltonian
cycles on clumpy graphs with $n$ clumps; we have tested five
different instances for each $n$.
Each table reports the number of choices (chs), the number of
assignments%
\footnote{For each change of the truth value of a atom and for each change of the color
of a node in a \RDGraph{} one assignment is counted in \smodels\ and \nomore, respectively.}
(ass) and the consumed time in seconds (time) for
each operational characterization of the \nomore\ and \smodels\ systems.
The first two operational characterizations V and VI correspond to the
respective rows in Table~\ref{tab:ops:summary}, whereas VI$+h_{sm}$ indicates
a modified version of VI comprising an smodels-like heuristic including lookahead.
Observe that we compare a Prolog and a C++ implementation and thus the
resulting time measurements should not be overrated. 
Instead the number of needed choices and assignments give a better indication
for the relation of \nomore\ and \smodels. 
Concerning choices and assignments, Table~\ref{tab:all:k} shows that for
computing all Hamiltonian cycles of complete graphs the \nomore\
strategy VI$+h_{sm}$ performs better than \smodels. 
Further evidence for this is given in Table~\ref{tab:one:k} where we 
are able to observe this phenomenon even by looking at time measurements 
(for $n\le 11$).

The results for clumpy graphs in Table~\ref{tab:one:clumpy}
demonstrate that \nomore\ behaves more uniform on those
examples than \smodels. 
For examples in the first three instances with six clumps
\nomore\ needs at most 4419 choices whereas \smodels\
needs at least 170000 choices. 
Furthermore, on the third instance \smodels\ did not even get
an answer set after more than two hours.
On the other hand, even if there are examples where \smodels\ needs less choices
than \nomore\ we did not recognize such extreme outliers for \nomore\ as we did
for \smodels\  during our tests.

Comparing the three \nomore\ strategies among each other,
we observe that the extension of strategy VI by an smodels-like heuristics
including lookahead usually decreases the number of choices (except for clumpy
graphs with $n=6$).
However, this does not necessarily lead to better performance in time,
since the lookahead increases the number of assignments.
Similarly, we observe that strategy VI always makes fewer choices than
strategy V.
Again, this advantage does not always pay off as regards time, since the
verification of support by operator $\fweakS_{\bg{}}$ is more time consuming
than applying operator \OPN{\bg{}}.

We stress that these experiments have a purely indicative character;
a systematic experimental evaluation is given in a companion paper.
Nonetheless our experiments show prospects insofar as even a moderately
optimized Prolog implementation of our strategies outperforms a state-of-the-art
system on certain benchmarks.
Even though many other benchmark problems are still solved much faster by the
state-of-the-art solvers,
a comparison of the number of choices reveals that our approach has no
substantial disadvantages.
%% , provided that some heuristics including lookahead is used.

%%% Local Variables: 
%%% mode: latex
%%% TeX-master: "paper"
%%% End: 

%% file: conclusion.tex
\section{Conclusion}
\label{sec:conclusion}

We have elaborated upon rule dependency graphs (\RDGraph{s}) and their colorings for
characterizing and computing answer sets of logic programs.
While \RDGraph{s} determine the possible interplay among rules inducing answer sets,
its colorings fix their concrete application status.

We have started by identifying graph structures that capture structural
properties of logic programs and their answer sets.
As a result, we obtain several characterizations of answer sets in terms of
totally colored dependency graphs.
All characterizations reflect the dichotomy among the notions of support and
blockage.
In fact, once a ``recursive support'' is established, this dichotomy
allows for characterizing answer sets in terms of their generating or their
non-generating rules.
The notion of ``recursive support'' is captured by the graph-theoretical concept
of a support graph, whose counterpart is given by the blockage graph.
Unlike the basic set-theoretic concepts,
these subgraphs do not reflect the aforementioned dichotomy in a fully symmetric
way.
This is because support graphs capture a global --- since recursive ---
structure,
whilst blockage graphs aim at a rather local structure, based on arc-wise
constraints.
Taken together, both subgraphs provide another characterization of answer sets.
Interestingly, their existence is incrementally enforced whenever appropriate
propagation operations are used during the coloring process.

To a turn, we build upon these basic graph-theoretical characterizations for
developing an operational framework for non-deterministic answer set formation.
The goal of this framework is to offer an intermediate stage between declarative
characterizations of answer sets and corresponding algorithmic specifications.
We believe that this greatly facilitates the formal elaboration of computational
approaches.
The general idea is to start from an uncolored \RDGraph{} and to employ specific
operators that turn a partially colored graph gradually in a totally colored one,
finally representing an answer set.
To this end, we have developed a variety of deterministic and non-deterministic
operators.
Different coloring sequences (enjoying different formal properties) are
obtained by selecting different combinations of operators.
Among others, we distinguish unicoloring and support-driven operational
characterizations.
In particular, we have identified the basic strategies employed by the
\texttt{noMoRe} system
as well as operations yielding Fitting's and well-founded semantics.
Taken together, the last results show that \texttt{noMoRe}'s principal
propagation operation amount to applying Fitting's operator, when
using  strategy IV in Table~\ref{tab:ops:summary} or computing well-founded
semantics, when applying strategy VI.
Notably, the explicit detection of 0-loops can be avoided by employing a
support-driven choice operation.
More recent developments within \texttt{noMoRe}, such as
back-propagation, heuristics, and implementation details are
dealt with in a companion paper.
Generally speaking, \texttt{noMoRe} is conceived as a parametric system that
allows for choosing different strategies.
The \texttt{noMoRe} system is available at
\texttt{http://www.cs.uni-potsdam.de/\ensuremath{\sim}linke/nomore}.

In fact, our operational framework can be seen as a ``theoretical toolbox'' that
allows for assembling specific strategies for answer set formation.
The algorithmic realization of our coloring sequences in terms of backtracking
algorithms is rather straightforward.
The interesting step is of course the implementation of choice operators.
In principle, a choice is made from a set of rules, each of which may be
attributed a (different) color.
This leaves room for different implementations, inducing differently shaped
search trees.
A prototypical platform offering the described spectrum of operations can be
downloaded at
\texttt{http://www.cs.uni-potsdam.de/\ensuremath{\sim}konczak/system/gcasp};
a corresponding implementation in C++ is currently under development. 
In all, our elaboration has laid the basic formal foundation for computing
answer sets by means of \RDGraph{s} and their colorings.
Current and future work mainly deals with further exploitation of the structural
information offered by a graph-based approach.
In another paper, we show how preferences among rules are easily
incorporated as a third type of edges~\cite{koscli03a,koscli03b}.
Other work includes graph compressions allowing for collapsing entire subgraphs
into single nodes~\cite{linke03a}.
Last but not least, our approach seems to be well-suited for debugging and profiling
purposes.
First, given that it relies on rules, the objects of computation are the same as
the descriptive objects within the problem specification.
Second, the underlying graph allows for a very natural visualization of
computations.
This has already led to a \nomore{}-specific profiler, described in~\cite{bolisc04a}.

%%% Local Variables: 
%%% mode: latex
%%% TeX-master: "paper"
%%% End: 

%% file: acknowledgements.tex
\paragraph{Acknowledgements}
We would like to thank
Christian Anger,
Philippe Besnard,
Stefania Costantini,
Martin Gebser,
Susanne Grell,
Andr\'e Neumann,
Tomi Janhunen,
Vladimir Sarsakov,
Mirek Truszczy{\'n}ski,
and the
referees of~\cite{kolisc03a,kolisc03b} and in particular the ones of the given
paper for many helpful discussions and constructive suggestions.

The authors were supported by the German Science Foundation (DFG)
under grant FOR~375/1 and SCHA 550/6,~TP~C and by the IST programme of the
European Commission, Future and Emerging Technologies under the IST-2001-37004
WASP project.

%%% Local Variables: 
%%% mode: latex
%%% TeX-master: "paper"
%%% End: 

%% file: auxiliary.tex
\section{Auxiliary results}
\label{sec:auxiliary}

In this section we want to provide theorems which are needed in the proofs.
The first theorem gives an characterisation of answer sets in terms of
generating rules which corresponds to $\colplus$ in our approach.
%--------------------------------------------------------------
\begin{theorem}\label{thm:AS:AC}
  Let $\bg{}$ be the \RDGraph{} of logic program  $\Pi$, let $X$ be a set of
  atoms, and
  let $\col$ be a partial coloring of \bg{}. 

  Then, $X\in \AS{\Pi}{C}$ iff
  $(\GR{\Pi}{X},\Pi\setminus\GR{\Pi}{X})\in\AC{\Pi}{C}$. 
\end{theorem}
%---------------------------------------------------------------
%
%--------------------------------------------------------------
\begin{theorem}\label{thm:as:consGR}
Let $\Pi$ be a logic program and $X$ be a set of atoms.

Then, $X$ is an answer set of $\Pi$ iff $X=Cn((\GR{\Pi}{X})^\emptyset)$.
\end{theorem}
%---------------------------------------------------------------
%
If we have an answer set, the set of generating rules possesses an enumeration
which will provide the support graph of a (colored) \RDGraph. 
%
%--------------------------------------------------------------
\begin{theorem}\label{thm:gr:grounded}
Let $\Pi$ be a logic program and $X$ an answer set of $\Pi$.
Then, there exists an enumeration $\langle r_i \rangle_{i \in I}$ of
\GR{\Pi}{X} such 
that for all $i \in I$ we have $\pbody{r_i} \subseteq \head{\{r_j \mid j <
  i\}}$.
\end{theorem}
%---------------------------------------------------------------
Given an answer set $X\in \AS{\Pi}{C}$ of a partial coloring we observe that
$\head{\colplus}\subseteq X$.
Furthermore, if $C$ is total, the heads of all rules in $\colplus$ generates
the answer set $X$.
%--------------------------------------------------------------
\begin{theorem}\label{thm:X:headcolplus}
Let $\bg{}$ be the \RDGraph\ of logic program $\Pi$ and $\col$ be a partial
coloring of $\bg{}$.
Furthermore let $X\in \AS{\Pi}{\col}$.

Then, $\head{\colplus}\subseteq X$.

If \col\ is admissible, then $\head{\colplus}=X$.

\end{theorem}
%---------------------------------------------------------------
An analog is observed with the set $\colminus$.
If for some atom $q$ all rules with $q$ as head are in the set $\colminus$
then $q$ is not in the answer set $X\in \AS{\Pi}{C}$.

%--------------------------------------------------------------
\begin{theorem}\label{cor:X:headcolplus}
Let $\bg{}$ be the \RDGraph\ of logic program $\Pi$ and $\col$ be a partial
coloring of $\bg{}$.
Furthermore, let $X\in \AS{\Pi}{\col}$ and $p\in \atm$.

If $\{\r\in \Pi \mid \head{\r}=p\}\subseteq \colminus$ then $p\not\in X$.

\end{theorem}
%---------------------------------------------------------------

%% %--------------------------------------------------------------
%% \begin{Theorem}(thm:U:unsupported)
%% Let $\bg{}$ be the \RDGraph\ of logic program $\Pi$ and $\col$ be a partial
%% coloring of $\bg{}$.
%% Furthermore let $X\in \AS{\Pi}{\col}$ be an answer set such that
%% $\colplus=\GR{\Pi}{X}$, $\head{\colplus}=X$ and $\colminus=\emptyset$.

%% Then $\fOPMaxGround{\bg{}}{\col}=(\colplus,\{\r\mid \pbody{\r}\not \subseteq
%% X\})$. 

%% \end{Theorem}
%% %---------------------------------------------------------------

%% %--------------------------------------------------------------
%% \begin{Theorem}(thm:P:blocked)
%% \comment{K: to reformulate}
%% Let $\bg{}$ be the \RDGraph\ of logic program $\Pi$ and $\col$ be a partial
%% coloring of $\bg{}$.
%% Furthermore let $X\in \AS{\Pi}{\col}$ be an answer set such that
%% $\colplus=\GR{\Pi}{X}$, $\head{\colplus}=X$ and $\colminus=\{\r\mid
%% \pbody{\r}\not\subseteq X\}$.

%% Then $\fOPprop{\bg{}}{\col}$ is a total coloring such that
%%   $\fOPprop{\bg{}}{\col}=(\colplus,\colminus\cup\{\r\mid \nbody{\r}\cap
%%   X\not= \emptyset\})$. 

%% \end{Theorem}
%% %---------------------------------------------------------------

Given an answer set $X\in \AS{\Pi}{C}$ for a partial coloring where
$\colplus=\GR{\Pi}{X}$ and $\colminus=\emptyset$ then all rules which are
blocked are obtained by the operator $\OPprop{\bg{}}$.
%---------------------------------------------------------------
\begin{theorem}\label{thm:sequence:P:blocked}
  Let \bg{} be the \RDGraph{} of logic program $\Pi$, $X$ be an answer set
  of $\Pi$
  and $\col$ be a partial coloring of \bg{}
  such that $\colplus=\GR{\Pi}{X}$ and $\colminus=\emptyset$.
  Furthermore, let $\col^X$ be a total coloring of $\bg{}$ such that $\{X\}=
  \AS{\Pi}{\col^X}$. 

  Then, $\fOPprop{\bg{}}{\col}=(\colplus,\colminus')$
  where
  $B(\bg{},\col^X) \subseteq \colminus'$.

\end{theorem}
%---------------------------------------------------------------

The next Theorem about monotonicity of 3-valued interpretations is used in the proofs of Section~\ref{sec:wfs}.
%---------------------------------------------------------------
  \begin{theorem}\label{cor:I:monoton}
    Let \bg{} be the \RDGraph{} of logic program $\Pi$ 
    and $C,C'$ be  partial colorings of \bg{}. 

   If $C\sqsubseteq C'$ then $(X_C,Y_C) \subseteq (X_{C'},Y_{C'})$.
  \end{theorem}
%---------------------------------------------------------------

%%% Local Variables: 
%%% mode: latex
%%% TeX-master: "paper"
%%% End: 

%% file: tools.tex
\section{Inductive definitions}
\label{sec:inductivedef}

In this section we want to give the inductive definitions of our operators
given in Section~\ref{sec:operations}.
We write $i<\omega$ for $i$ being a finite natural number greater or equal
than $0$.

According to $\OPpropast{\bg{}}$, we define $P(\col)$ as  
\(
P(\col) = \bigsqcup_{i<\omega} P^i(\col)
\)
where 
\begin{enumerate}
\item $P^0(\col)=\col$ and
\item
  \(
  P^{i+1}(\col)
  =
%   P^i(\col)
%   \sqcup
  \fOPprop{\bg{}}{P^i(\col)}
  \)
  for $i<\omega$.
\end{enumerate}
Clearly, $P^i(C) \sqsubseteq P^{i+1}(C)$ for all $i < \omega$.
By using $\OPprop{\bg{}}$ in every iteration step, we have that 
$P^i(C)$ is always a partial coloring for $i < \omega$. 
Note that $P(C)$ not always exists.
To see this, observe that 
\(
P^1(\{a\LPif{}\naf{a}\},\emptyset)
\)
would be
\(
(\{a\LPif{}\naf{a}\},\{a\LPif{}\naf{a}\})
\)
which is not a partial coloring.

%---------------------------------------------------------------
\begin{theorem}\label{thm:inductive:prop}
  Let \bg{} be the \RDGraph{} of logic program $\Pi$ and \col{} be a partial
  coloring of \bg{}.

%   If $\AS{\Pi}{\col}\not=\emptyset$ then 
  Then,
  \begin{enumerate}
  \item if $\AC{\Pi}{C}\not=\emptyset$ then $P(C)$ exists,
  \item $P(\col)$ is a  partial coloring,
% \item if $\AS{\Pi}{C}\not=\emptyset$ then $\AS{\Pi}{P(\col)}\not=\emptyset$,
%  \item if $X\in \AS{\Pi}{C}$ then $X \in \AS{\Pi}{P(\col)}$,
%%   \item If $X\in \AS{\Pi}{C}$ then $X \in \AS{\Pi}{P^i(\col)}$ for all $i <
%%   \omega$ and so $X \in \AS{\Pi}{P(\col)}$ 
  \item $\AC{\Pi}{C}=\AC{\Pi}{P(C)}$ and for all $i < \omega$ we have $X\in
  \AS{\Pi}{C}$ iff $X \in \AS{\Pi}{P^i(\col)}$ iff $X \in \AS{\Pi}{P(\col)}$,
  \item $\col \sqsubseteq P(C)$,
  \item $P(C)$ closed under $\OPprop{\bg{}}$,
  \item $P(C)$ is the $\sqsubseteq$-smallest partial coloring closed under
  $\OPprop{\bg{}}$, and
  \item $P(C)=\fOPpropast{\bg{}}{\col}$.
  \end{enumerate}
\end{theorem}
%---------------------------------------------------------------

According to $\mathcal{T}^\ast_{\bg{}}$, 
we define 
$T(\col)$ as 
\(
T(\col) = \bigsqcup_{i<\omega} T^i(\col)
\)
where 
\begin{enumerate}
\item $T^0(\col)=\col$ and
\item
  \(
  T^{i+1}(\col)
  =
  \mathcal{T}_{\bg{}}(T^i(\col))
  \)
  for $i<\omega$. 
\end{enumerate}
Clearly, $T^i(C) \sqsubseteq T^{i+1}(C)$ for all $i < \omega$. 

%---------------------------------------------------------------
\begin{theorem}\label{thm:inductive:propT}
  Let \bg{} be the \RDGraph{} of logic program $\Pi$ and \col{} be a partial
  coloring of \bg{}.

  Then,
  \begin{enumerate}
  \item $T(\col)$ is a partial coloring,
  \item $\col \sqsubseteq T(C)$,
  \item $T(C)$ closed under $\mathcal{T}_{\bg{}}$,
  \item $T(C)$ is the $\sqsubseteq$-smallest partial coloring closed under
  $\mathcal{T}_{\bg{}}$, and
  \item $T(C)=\weakProp{\bg{}}{\col}$.

  \end{enumerate}
 \end{theorem}
%---------------------------------------------------------------

According to $(\OPprop{}\OPMaxGround{})^\ast_{\bg{}}$,
we define $PU(\col)$ as  
\(
PU(\col) = \bigcup_{i<\omega} PU^i(\col)
\)
where 
\begin{enumerate}
\item $PU^0(\col)=\col$ and
\item
  \(
  PU^{i+1}(\col)
  =
%%   PU^i(\col)
%%   \sqcup
%%   \fOPprop{\bg{}}{PU^i(\col)}
%%   \sqcup
%%   \fOPMaxGround{\bg{}}{PU^i(\col)}
  \fOPMaxGround{\bg{}}{\fOPprop{\bg{}}{PU^i(C)}}
  \)
  for $i<\omega$.
\end{enumerate}
Clearly, $PU^i(C) \sqsubseteq PU^{i+1}(C)$ for all $i < \omega$. 

%---------------------------------------------------------------
\begin{theorem}\label{thm:inductive:propPU}
  Let \bg{} be the \RDGraph{} of logic program $\Pi$ and \col{} be a partial
  coloring of \bg{}.

  Then,
  \begin{enumerate}
  \item If $\AC{\Pi}{C}\not=\emptyset$ then $PU(C)$ exists,
  \item $PU(\col)$ is a partial coloring,
%  \item if $X\in \AS{\Pi}{C}$ then $X\in\AS{\Pi}{PU(C)}$,
  \item $\AC{\Pi}{C}=\AC{\Pi}{PU(C)}$ and for all $i < \omega$ we have $X\in
  \AS{\Pi}{C}$ iff $X \in \AS{\Pi}{PU^i(\col)}$ iff $X \in \AS{\Pi}{PU(\col)}$,
  \item $\col \sqsubseteq PU(C)$,
  \item $PU(C)$ closed under $\OPprop{\bg{}}$ and $\OPMaxGround_{\bg{}}$,
  \item $PU(C)$ is the $\sqsubseteq$-smallest partial coloring closed under
  $\OPprop{\bg{}}$ and $\OPMaxGround_{\bg{}}$, and
   \item $PU(C)=(\OPprop{}\OPMaxGround{})^\ast_{\bg{}}(\col)$.

  \end{enumerate}
 \end{theorem}
%---------------------------------------------------------------

% %---------------------------------------------------------------
% \begin{Theorem}(thm:inductive:propPU:PU)
%   Let \bg{} be the \RDGraph{} of logic program $\Pi$ and \col{} a partial
%   coloring of \bg{}.

%    If $\AS{\Pi}{\col}\neq\emptyset$, then 
%    $PU(C)=(\OPprop{}\OPMaxGround{})^\ast_{\bg{}}(\col)$.
% \end{Theorem}
% %---------------------------------------------------------------

Analogous, we define $PV(\col)$ as  
\(
PV(\col) = \bigcup_{i<\omega} PV^i(\col)
\)
where 
\begin{enumerate}
\item $PV^0(\col)=\col$ and
\item
  \(
  PV^{i+1}(\col)
  =
%%   PV^i(\col)
%%   \sqcup
%%   \fOPprop{\bg{}}{PV^i(\col)}
%%   \sqcup
%%   \weakS{\bg{}}{PV^i(\col)}
  \weakS{\bg{}}{\fOPprop{\bg{}}{PV^i(C)}}
  \)
  for $i<\omega$.
\end{enumerate}
Clearly, $PV^i(C) \sqsubseteq PV^{i+1}(C)$ for all $i < \omega$. 
%---------------------------------------------------------------
\begin{theorem}\label{thm:inductive:propPV}
  Let \bg{} be the \RDGraph{} of logic program $\Pi$ and \col{} be a partial
  coloring of \bg{}.

  Then,
  \begin{enumerate}
  \item If $\AC{\Pi}{C}\not=\emptyset$ then $PV(C)$ exists,
  \item $PV(\col)$ is a partial coloring,
  \item $\AC{\Pi}{C}=\AC{\Pi}{PV(C)}$ and for all $i < \omega$ we have $X\in
  \AS{\Pi}{C}$ iff $X \in \AS{\Pi}{PV^i(\col)}$ iff $X \in \AS{\Pi}{PV(\col)}$,
  \item $\col \sqsubseteq PV(C)$,
  \item $PV(C)$ closed under $\OPprop{\bg{}}$ and $\fweakS_{\bg{}}$,
  \item $PV(C)$ is the $\sqsubseteq$-smallest partial coloring closed under
  $\OPprop{\bg{}}$ and $\fweakS_{\bg{}}$, and
  \item $PV(C)=(\OPprop{}\fweakS)^\ast_{\bg{}}(\col)$.
  \end{enumerate}
 \end{theorem}
%---------------------------------------------------------------

% %---------------------------------------------------------------
% \begin{Theorem}(thm:inductive:propPV:PV)
%   Let \bg{} be the \RDGraph{} of logic program $\Pi$ and \col{} a partial
%   coloring of \bg{}.

%    If $\AS{\Pi}{\col}\neq\emptyset$, then 
%    $PV(C)=(\OPprop{}\fweakS)^\ast_{\bg{}}(\col)$.
% \end{Theorem}
% %---------------------------------------------------------------

%%% Local Variables: 
%%% mode: latex
%%% TeX-master: "paper"
%%% End: 

%% file: proofs.tex
\section{Proofs}
\label{sec:proofs}
We write $i<\omega$ for $i$ being a finite natural number greater or equal than
$0$.

%------------------Section 3 ----------------------------------
 \subsection{Section~\ref{sec:graphcol}}            
%% SHORT
%% changed for admissible colorings: YES
   \input{Proofs/ShortVersion/thm_nlp_properties}                 %1  
   \input{Proofs/ShortVersion/cor_nlp_properties}                 %2  
   \input{Proofs/ShortVersion/thm_col_gr}                         %3 

%% LONG
%% changed for admissible colorings y/n
%%    \input{Proofs/LongVersion/thm_nlp_properties}                 %1 okTL  
%%    \input{Proofs/LongVersion/cor_nlp_properties}                 %2 okTL 
%%    \input{Proofs/LongVersion/thm_col_gr}                         %3 okTL 

%------------------Section 4 ----------------------------------
 \subsection{Section~\ref{sec:check}}               

%% SHORT
%% changed for admissible colorings y/n
%%   \input{Proofs/ShortVersion/prop_SG_existence}                   %1  ok
%%   \input{Proofs/ShortVersion/thm_max_supported_graph_existence}   %2  ok
%%   \input{Proofs/ShortVersion/thm_supportgraph}                    %3  ok
%%   \input{Proofs/ShortVersion/thm_acoloring_sets}                  %4  ok
%%   \input{Proofs/ShortVersion/thm_red}                             %5  ok
%%   \input{Proofs/ShortVersion/thm_max_supported_graph_C}           %6  ok
%%   \input{Proofs/ShortVersion/prop_AS_maxSG}                       %7  ok
%%   \input{Proofs/ShortVersion/thm_acol_acolorings_ver2}            %8  ok
%%   \input{Proofs/ShortVersion/thm_acol_acolorings_ver1}            %9  ok

%% changed for admissible colorings YES
   \input{Proofs/LongVersion/prop_SG_existence}                   %1  okTL
   \input{Proofs/ShortVersion/thm_max_supported_graph_existence}   %2  okTL
   \input{Proofs/LongVersion/thm_supportgraph}                    %3  okTL
   \input{Proofs/LongVersion/thm_acoloring_sets}                  %4  okTL
   \input{Proofs/LongVersion/thm_red}                             %5  okTL
   \input{Proofs/LongVersion/thm_max_supported_graph_C}           %6  okTL
   \input{Proofs/LongVersion/prop_AS_maxSG}                       %7  okTL
   \input{Proofs/LongVersion/thm_acol_acolorings_ver2}            %8  okTL
   \input{Proofs/LongVersion/thm_acol_acolorings_ver1}            %9  okTL

%------------------Section 5 ----------------------------------
\subsection{Section~\ref{sec:operations}}          
%% LONG
%% changed for admissible colorings YES
\input{Proofs/LongVersion/thm_P_reflexivity}                   %1 ok (TL)
\input{Proofs/LongVersion/thm_P_defined}                       %2 ok (TL)
%% cor 5.3
\input{Proofs/LongVersion/cor_P_empty_exists}                  %%NEU

\input{Proofs/ShortVersion/thm_P_monoton}                       %3 ok (TL)
\input{Proofs/LongVersion/thm_P_ASpreserving}                  %4 ok (TL)
\input{Proofs/LongVersion/thm_weak_support_and_blockage_graph} %5 ok (TL)
\input{Proofs/LongVersion/thm_AS_iv}                           %6 ok (Tl)
\input{Proofs/LongVersion/cor_S_existence}                     %7 ok (TL)
\input{Proofs/LongVersion/thm_S_monoton}                       %8 ok (TL)
\input{Proofs/LongVersion/thm_S_hilfslemma}                    %9 ok (TL)
\input{Proofs/LongVersion/thm_weak_support_and_blockage_graph_two} %10 (TL)
\input{Proofs/LongVersion/thm_answerset_PS}                    %11 ok (TL)
\input{Proofs/LongVersion/thm_pre_algo_main_ver1}              %12 ok (TL)
\input{Proofs/LongVersion/thm_collection_OASC_i}               %13 ok (TL)   
\input{Proofs/LongVersion/thm_pre_algo_main_ver2}              %14 ok(?) 
%% cor 5.16
\input{Proofs/LongVersion/cor_PUast_empty_exists}               %%NEU

\input{Proofs/LongVersion/thm_collection_OASC_ii}              %15 ok (TL)
\input{Proofs/LongVersion/thm_collection_OASC_ii_sub}          %16 ok (TL)
\input{Proofs/LongVersion/thm_comparison_i}                    %17 ok
\input{Proofs/LongVersion/thm_comparison_ii}                   %18 ok
\input{Proofs/LongVersion/thm_pre_algo_main_ver2_plus_minus}   %19 ok?
\input{Proofs/LongVersion/thm_sequence_plus}                   %20 ok (reproof
                                %show 2; AC(C')\not=\emptyset)
\input{Proofs/LongVersion/thm_sequence_minus}                  %21 ok 
\input{Proofs/LongVersion/thm_sequence_D_plusminus}            %22 ok
\input{Proofs/LongVersion/thm_collection_sequence_D_plusminus} %23 ok
\input{Proofs/LongVersion/thm_sequence_D_plus}                 %24 ok
\input{Proofs/LongVersion/thm_collection_sequence_D_plus}      %25 ok
\input{Proofs/LongVersion/thm_sequence_DP_plusminus}           %26 ok
\input{Proofs/LongVersion/thm_sequence_DP_plusminus_N}         %27 ok
\input{Proofs/LongVersion/thm_collection_sequence_DP_plusminus}%28 ok
\input{Proofs/LongVersion/thm_sequence_DP_plus}                %29 ok
\input{Proofs/LongVersion/thm_collection_sequence_DP_plus}     %30 ok
\input{Proofs/LongVersion/thm_weakP_supportgraph}              %31 ok
\input{Proofs/LongVersion/thm_weakS_S}                         %32 ok
\input{Proofs/LongVersion/thm_pre_algo_main_ver4}              %33 ok
\input{Proofs/LongVersion/thm_collection_OASC_iv}              %34 ok
\input{Proofs/LongVersion/thm_pre_algo_main_ver4_pm}           %35 ok
\input{Proofs/LongVersion/thm_complexity_operators}            %% NEU

%------------------Section 6 ----------------------------------
%% LONG
%% changed for admissible colorings YES
\subsection{Section~\ref{sec:wfs}}
\input{Proofs/LongVersion/thm_wfs_fitting_P_empty}             %1 
\input{Proofs/LongVersion/thm_wfs_GUS}                         %2 
\input{Proofs/LongVersion/cor_wfs_GUS}                         %3 
\input{Proofs/LongVersion/thm_wfs_main}                        %4

%------------------Appendix  ----------------------------------
\subsection{Appendix}                                 

%--------------------- A -----------------------------------
%% LONG
%% changed for admissible colorings YES
\subsubsection{Section~\ref{sec:auxiliary}}
 \input{Proofs/LongVersion/thm_AS_AC}                 % A1
 \input{Proofs/LongVersion/thm_as_consGR}             % A2  
 \input{Proofs/LongVersion/thm_gr_grounded}           % A3  
 \input{Proofs/LongVersion/thm_X_headcolplus}         % A4  
 \input{Proofs/LongVersion/cor_X_headcolplus}         % A5  
 \input{Proofs/LongVersion/thm_sequence_P_blocked}    % A6   
 \input{Proofs/LongVersion/cor_I_monoton}             % A7 

%--------------------- B -----------------------------------
%% LONG
%% changed for admissible colorings y/n
\subsection{Section~\ref{sec:inductivedef}}
\input{Proofs/LongVersion/thm_inductive_prop}         % B1  
\input{Proofs/LongVersion/thm_inductive_propT}        % B2    
\input{Proofs/LongVersion/thm_inductive_propPU}       % B3  
\input{Proofs/LongVersion/thm_inductive_propPV}       % B4

%%%%%% old things
%\subsection{Section~\ref{sec:original}}
%\input{Proofs/thm_coloring}
%\input{Proofs/cor_acoloring_sets_TL}
% \input{Proofs/cor_S_exists}                        %sec 4 cor 6   
% \input{Proofs/thm_extension_AS}                     % A7 -> 4.4
%%%  \input{Proofs/cor_acoloring_sets}                  %zu 5addieren TS fragen
% %\input{Proofs/thm_pre_algo_main_comparison}        %13
% %\input{Proofs/thm_weakS_S}                         %14 
% %\input{Proofs/cor_algo_main_minimal}              %
% %\input{Proofs/thm_collection}                     %
%%%\input{Proofs/prop_extension_defined}              %sec 4 1  ok, an TS
%  \input{Proofs/cor_max_supported_graph_existence}   %3  sec 3, now in Thm3.2
%%%%%%%% old T/V operator
% \input{Proofs/thm_pre_algo_main_ver3}              %23
% \input{Proofs/thm_comparison_iii}                  %24
% \input{Proofs/thm_pre_algo_main_nomore}            %27
%%%%%% old auxiliary
% \input{Proofs/thm_U_unsupported}                    % A5   ok, an TS
% \input{Proofs/thm_P_blocked}                        % A6  neu machen
%\input{Proofs/thm_inductive_propPU_PU}                
%\input{Proofs/thm_inductive_propPV_PV}                % A14

%%% Local Variables: 
%%% mode: latex
%%% TeX-master: "paper"
%%% End: 

%% file: Proofs/ShortVersion/thm_nlp_properties.tex
%---------------------------------------------------------------
\begin{proof}{thm:nlp:properties}
Let $\bg{}$ be the \RDGraph\ of logic program $\Pi$,
$\col$ be a partial 
coloring 
of $\bg{}$ and $X \in \AS{\Pi}{\col}$.
By Equation~(\ref{eq:def:AS}), we have $\colplus\subseteq  \GR{\Pi}{X}$ and by
Theorem~\ref{thm:X:headcolplus} 
$\head{\colplus}\subseteq X$.

{\bf 1:}
Let $\r\in S(\bg{},\col)$.
By definition, for all $p \in \pbody{\r}$ there exists an $\rp \in \Pi$ such
that 
$(\rp,\r)\in \bgezero$, $p=\head{\rp}$ and $\rp\in\colplus$.
From $\head{\colplus}\subseteq X$, we can reconclude that for each $p \in
\pbody{\r}$, we have $p \in X$, and thus $\pbody{\r}\subseteq X$.

Conditions~2-4 follow analogous to Condition~1 by
Theorem~\ref{thm:X:headcolplus} and~\ref{cor:X:headcolplus}. 
\end{proof}
%----------------------------------------------------------------

%%% Local Variables: 
%%% mode: latex
%%% TeX-master: t
%%% End: 

%% file: Proofs/ShortVersion/cor_nlp_properties.tex
\begin{proof}{cor:nlp:properties}
Let $\bg{}$ be the \RDGraph\ of logic program $\Pi$,
$\col$ be an admissible
coloring 
of $\bg{}$, and $\{X\} = \AS{\Pi}{\col}$.
By Equation~(\ref{eq:def:AS}) we have $\colplus = \GR{\Pi}{X}$ and 
$\head{\colplus} = X$ (Theorem~\ref{thm:X:headcolplus}).  
By Theorem~\ref{thm:nlp:properties} we only have to show ``$\Rightarrow$''.

{\bf 1 "$\Rightarrow$":}
Let be $\pbody{\r}\subseteq X=\head{\colplus}$ for $\r \in \Pi$.
Then, for each $p \in \pbody{\r}$ there exists an $\rp \in \Pi$ such that
$\rp\in\colplus$ and $p=\head{\rp}$.
From the definition of the \RDGraph\ we can conclude that $\r \in
S(\bg{},\col)$. 

Conditions~2-4 follow analogous.
\end{proof}
%----------------------------------------------------------------

%% file: Proofs/ShortVersion/thm_col_gr.tex
\begin{proof}{thm:col:gr}
This theorem follows by Theorem~\ref{thm:nlp:properties},
Corollary~\ref{cor:nlp:properties}, and Equation~(\ref{eq:GR}). 
%%   Let $\bg{}$ be the \RDGraph{} of logic program  $\Pi$,
%%   $\col$ be a partial coloring of $\bg{}$ and 
%%   $X \in \AS{\Pi}{\col}$ be an answer set.
%
%% {\bf Show~1:}
%% Let $\r\in S(\bg{},\col)\cap\overline{B}(\bg{},\col)$.
%% %
%% By Theorem~\ref{thm:nlp:properties} we have
%% $\pbody{\r}\subseteq X$ and $\nbody{\r} \cap X=\emptyset$.
%% %
%% Thus, $\r\in \GR{\Pi}{X}$.
%
%% Analogously, we have condition~2 and  conditions~3 and~4 follow in the same way
%% by Corollary~\ref{cor:nlp:properties}.  
\end{proof}

%%% Local Variables: 
%%% mode: latex
%%% TeX-master: t
%%% End: 

%% file: Proofs/LongVersion/prop_SG_existence.tex
\begin{proof}{prop:SG:existence}
  Let $\bg{}=(\Pi,E_0,E_1)$ be the \RDGraph{} of logic program $\Pi$.
  Let
  $(V'',E)$ and $(V',E')$ be  support graphs of $\bg{}$ for some $E,E'\subseteq
  (\Pi\times \Pi)$ such that $V'' \not=V'$.

  We show that there exists a support graph $(V,E)$ of $\bg{}$
  for some $E\subseteq  (\Pi\times \Pi)$ such that $V''\subseteq V$
  and $V'\subseteq V$.
  Since $\Pi$ is finite and by finiteness of the number of support graphs, we
  then conclude that there exists a maximal support graph of \bg{}.

Trivially, according to Definition~\ref{def:acol:Dasg}, there exists an
enumeration  
$\langle r_i\rangle_{0 \leq i \leq n}$  of $V''$ such that
$\pbody{r_i}\subseteq\{\head{r_j}\mid j < i\}$ and there exists an enumeration  
 $\langle r'_i\rangle_{0 \leq i \leq m}$ of $V'$ such that
$\pbody{r'_i}\subseteq\{\head{r'_j}\mid j < i\}$.

The idea of the construction of $(V,E)$ is as follows:
we merge the vertices of $V''$ and $V'$ to $V$ and additionally include all
vertices 
into $V$ whose positive body is derivable by the heads of the rules
belonging to $V$. 
More precisely, $V$ should be closed under all rules whose positive body
is derivable by the heads of rules in $V$.

Next, we define an enumeration $\langle s_i \rangle_{0 \leq i \leq k}$ of
rules in 
$\Pi$ inductively as follows. 
Let 
$s_0=r\in\Pi$ where $\pbody{r}=\emptyset$ and
%$s_0=r_0$ and 
%% \[
%% s_i = \left\{   
%%   \begin{array}{ll}
%%     r_j & \text{ for } j \leq n \text{ such that } 
%%           \pbody{r_j}\subseteq\{\head{s_l}\mid l < i\}
%%           \text{ or }\\
%%     r'_j & \text{ for } j \leq m \text{ such that } 
%%           \pbody{r'_j}\subseteq\{\head{s_l}\mid l < i\}\\
%%     r    & r\in\Pi, \pbody{r}\subseteq \{\head{s_l}\mid l < i\}
%%   \end{array}
%%       \right.
%% \]
\[
s_i = r\in\Pi \text{ such that } \pbody{r}\subseteq \{\head{s_l}\mid l < i\}
\]
for $0 \leq i \leq k$ and for some maximal $k$ such that there exists no
$\rp\in \Pi\setminus \{s_0,\ldots,s_k\}$ where $\pbody{\rp}\subseteq
\{\head{s_l}\mid l \leq k\}$.
Note that then the enumeration $\langle s_i \rangle_{0 \leq i \leq k}$ is
maximal wrt vertices.
Furthermore, we have $n\leq k,m \leq k$ and $n+m - |V''\cap V'|\leq k$.

Next, we show that for  $V= \{s_i \mid 0\leq i \leq k\}$ the following
conditions are fulfilled: 
\begin{enumerate}
\item $V'' \subseteq V$,
\item $V' \subseteq V$,
\item for all $r\in \Pi$ if $\pbody{r}\subseteq \{\head{r'}\mid r'\in
  V\}$ then $r\in V$, and
\item $(V,E)$ is a support graph of $\bg{}$ for some
  $E\subseteq (\Pi\times\Pi)$.
\end{enumerate}
Condition~3 states that all rules whose positive body can be derived by the
heads of the rules in $V$ are included in $V$.
Observe that if there is no $r\in\Pi$ such that $\pbody{r}=\emptyset$ we have
that then $V'=\emptyset$ and $V''=\emptyset$ and hence, $V''=V'$ which is a
contradiction to the assumption $V''\not= V'$.

{\bf 1+2:}
Note that for non-empty $V''$ and non-empty $V'$ we have $\pbody{r_0}=\emptyset$
and $\pbody{r'_0}=\emptyset$ by construction of the enumerations of $V''$ and
$V'$.
Hence, there exist $0\leq l,l' \leq k$ such that $r_0=s_l$ and $r'_0=s_{l'}$.
Thus, we conclude by induction that $V'' \subseteq \{s_0,\ldots,s_k\}$
(Condition~1) and 
$V'\subseteq \{s_0,\ldots,s_k\}$ (Condition~2) by construction of $V$.

{\bf 3:}
Condition~3 is fulfilled, since in we include as many rules as possible into
$\langle s_i \rangle_{0  \leq i \leq k}$ ($k$ being maximal). 

{\bf 4:}
For $0 < i \leq k$, we define 
\[
E^i= \bigcup \{(\rp,s_i)\mid 
\rp\in\{s_0,\ldots,s_{i-1}\}\}\cap E_0
\]
and $E = \bigcup_{0 \leq i \leq k} E^i$.
Clearly, $(V,E)$ is \subgraph{0}\  of $\bg{}$ by construction.
Furthermore, $(V,E)$ is acyclic since there are only edges
$(s_j,s_i)$ where $j<i$ for $j,i\in \{0,\ldots,k\}$.
Also, we obtain $r\in V$ whenever $\pbody{r}\subseteq \{\head{r'}\mid
(r',r) \in E\}$.
Thus, $(V,E)$ is a support graph of $\bg{}$.

Hence, there exists a \groundedZeroDAG{} $(V,E)$ of \bg{}
  such that
  $V'\subseteq V$ for all \groundedZeroDAG{s} $(V',E')$ of \bg{}.
\end{proof}
%%% Local Variables: 
%%% mode: latex
%%% TeX-master: t
%%% End: 

%% file: Proofs/ShortVersion/thm_max_supported_graph_existence.tex
\begin{proof}{thm:max:supported:graph:existence}
  Let $\bg{}=(\Pi,\bgezero,\bgeone)$ be the \RDGraph{} of logic program $\Pi$
  and
  $\col$ be a partial coloring of \bg{}.

We abbreviate $\bg{}|_{\colplus\cup \colminus}$ with $\bg{}|_C$.
%% We have to prove
%%   \begin{enumerate}
%%   \item   If $\AC{\Pi}{\col}\not=\emptyset$,
%%   then there is a (maximal) \groundedZeroDAG\ of $(\bg{},\col)$.
%%   \item If $(\colplus,E)$ is a \groundedZeroDAG\ of $\bg{}|_\col$
%%   for some $E \subseteq (\Pi\times \Pi)$,
%%   then there is a (maximal) \groundedZeroDAG\ of $(\bg{},\col)$.
%%   \end{enumerate}

\paragraph{1:}
If $\AC{\Pi}{\col}\not=\emptyset$ then we have $\AS{\Pi}{\col}\not=\emptyset$ by
Theorem~\ref{thm:AS:AC}. 
Let be $X\in \AS{\Pi}{C}$. 
Then, we have by definition that $\colplus\subseteq \GR{\Pi}{X}$ and
$\colminus 
\cap \GR{\Pi}{X}=\emptyset$ hold.
We want to construct a support graph $(\GR{\Pi}{X},E)$ of $(\bg{},C)$ for some
$E\subseteq (\Pi\times\Pi)$.
Then, there exists a maximal support graph of $(\bg{},C)$.

By Theorem~\ref{thm:gr:grounded} we have an enumeration  $\langle r_i
\rangle_{i \in I}$ of \GR{\Pi}{X} such 
that for all $i \in I$ we have $\pbody{r_i} \subseteq \head{\{r_j \mid j <
  i\}}$. 

%% Let $V^0=\{r_0\}$ and $E^0=\emptyset$.
%% Assume, $V^i\subseteq \Pi$ and $E^i\subseteq \Pi\times\Pi$ are constructed for
%% some $i < \omega$. 
%% Define
%% \(
%% V^{i+1}=V^i\cup\{r_i\}
%% \)
%% and
%% \(
%% E^{i+1}=E^i\cup E^{r_i}
%% \)
%% where
%% \[
%% E^{r_i}=\bigcup_{p\in \pbody{r_i}} \{(\rp,\r_i) \mid \rp\in V^i, \head{\rp}=p\}.
%% \]
%% Then, $(V,E)= \left( \bigcup_{i < \omega} V^i, \bigcup_{i< \omega} E^i
%% \right)$.
Define $V=\bigcup_{i\in I} \{r_i\}$ and
\(
E=\{(r_j,r_i)\mid j < i\}\cap \bgezero
\).
Clearly, $V=\GR{\Pi}{X}$ by construction and thus we have $\colplus \subseteq V$,
$\colminus \cap V =\emptyset$, and $(V,E)$ is a support graph of $\bg{}$.
Furthermore, $(V,E)$ is a support graph of $(\bg{},C)$.
The existence of a maximal one follows by
Theorem~\ref{prop:SG:existence}. 

\paragraph{2:}
Let $(\colplus,E')$ be a support graph of $\bg{}|_\col$ for some
$E' \subseteq (\Pi \times \Pi)$.
Then, $(\colplus,E')$ is a support graph of $(\bg{},C)$.
By Theorem~\ref{prop:SG:existence}, there exists a (maximal) support graph of
$(\bg{},C)$.
%% %% With Definition~\ref{def:acol:maxgrounded} we have to show, that there
%% exists 
%% %% a (maximal) support graph $(V,E)$ of $\bg{}$ such that $\colplus
%% \subseteq V$ and 
%% %% $\colminus \cap V=\emptyset$.
%
%% Now, we want to extend $(\colplus,E')$ to a support graph $(V,E)$ of $\bg{}$
%% such that $\colplus \subseteq V$ and $\colminus \cap V=\emptyset$.
%% Let 
%% \[(V^0,E^0)=(\colplus,E').\]
%% Assume, that we have constructed $(V^i,E^i)$ for $0 < i < \omega$.
%% We define, 
%% \[(V^{i+1},E^{i+1})=(V^i \cup \{r_{i+1}\},E^i\cup E^{r_{i+1}})\]
%%  where 
%% $\r_{i+1} \in \Pi \setminus (V^i\cup \colminus)$, $\pbody{\r_{i+1}}\subseteq
%% V^i$ and 
%% \[E^{r_{i+1}}= \bigcup_{p\in \pbody{\r_{i+1}}} \{(\rp,\r_{i+1})\mid \rp\in V^i,
%% \head{\rp}=p\}.\]
%% Define $(V,E)=\left(\bigcup_{i < \omega} V^i,\bigcup_{i<\omega}
%%   E^i\right)$. 
%% %% Clearly, we have $\colplus \subseteq V$, $\colminus \cap V=\emptyset$ by
%% %% construction of $V^i$ for all $i<\omega$.
%% %% Furthermore, $E$ contains no cycles because $E'$ contains further
%% %% edges are only from $V^i$ to $V^{i+1}$ for $i < \omega$ and we have
%% %% $\pbody{\r}\subseteq \{\head{\rp} \mid (\rp,\r)\in E\}$ for all $\r\in V$. 
%% Thus, $(V,E)$ is a support graph of $(\bg{},\col)$.
%% Furthermore, this support graph is maximal wrt vertices by the construction
%% of $V$.
\end{proof}
%%% Local Variables: 
%%% mode: latex
%%% TeX-master: "~/tex/Papers/GraphLP/Grundlagen/Long/paper"
%%% End: 

%% file: Proofs/LongVersion/thm_supportgraph.tex
\begin{proof}{thm:supportgraph}
Let \bg{} be the \RDGraph\ of logic program $\Pi$ and
 $\col$ be an admissible coloring.
By $\{C\}=\AC{\Pi}{C}\not=\emptyset$ and
Theorem~\ref{thm:max:supported:graph:existence}, there exists a support graph
of $(\bg{},C)$.
According to Definition~\ref{def:acol:maxgrounded}, this must be
$(\colplus,E)$  for some $E\subseteq \Pi\times\Pi$, since $C$ is a total
 coloring. 
Furthermore, $(\colplus,E)$ is a maximal support graph of $(\bg{},C)$. 
%% Because $\col$ is total, a support graph of $\bg{}|_\col$ is a support graph
%% of $\bg{}$.
%%
%% We have to show, that $(\colplus,E)$ is a  support graph of
%% $\bg{}|_\col=\bg{}$ for some $E \subseteq 
%% (\Pi\times \Pi)$. 
%% By Theorem~\ref{thm:max:supported:graph:existence}, $(\colplus,E)$ is then a
%% maximal support graph of $(\bg{},\col)$ because $\col$ is a total coloring. 
%%
%% Analogous to Theorem~\ref{thm:max:supported:graph:existence} we can
%% construct a support graph $(\colplus,E)$ for some $E \subseteq
%% (\Pi\times \Pi)$ starting from an enumeration  of $\GR{\Pi}{X}=\colplus$. 
%% \comment{reproof it!}
% Through theorem~\ref{thm:gr:grounded} there exists an enumeration $\langle r_i
% \rangle_{i \in I}$ of
% $\GR{\Pi}{X}$ such that for all $i \in I$ we have $\pbody{r_i} \subseteq
% \head{\{r_j \mid j <   i\}}$.
% This property gives us directly the existence of an $E\subseteq (\Pi\times
% \Pi)$ such that  $(\colplus,E)$ is a support graph.
\end{proof}
%%% Local Variables: 
%%% mode: latex
%%% TeX-master: t
%%% End: 

%% file: Proofs/LongVersion/thm_acoloring_sets.tex
\begin{proof}{thm:acoloring:sets}
  Let $\bg{}$ be the \RDGraph{} of logic program  $\Pi$ and
  let $\col$ be a total coloring of \bg{}.

\paragraph{"$1 \Rightarrow 2$":}
Since $C$ is an admissible coloring, there
exists an answer set $X$ of $\Pi$ such 
that $\{X\}=\AS{\Pi}{C}$.
By Theorem~\ref{thm:max:supported:graph:existence} and $\{C\}=\AC{\Pi}{C}$
there exists a 
support graph of $(\bg{},\col)$.
$\colplus =S(\bg{},\col)\cap\overline{B}(\bg{},\col)$
holds by Theorem~\ref{thm:col:gr} and $\colplus=\GR{\Pi}{X}$.
%% \subparagraph{Show "$\subseteq$":}
%% For all $\r\in \GR{\Pi}{X}=\colplus$ we have $\pbody{\r}\subseteq X$ and
%% $\nbody{\r}\cap X =\emptyset$ by the definition of $\GR{\Pi}{X}$.
%% Therefore by Corollary~\ref{cor:nlp:properties} we have 
%% $\colplus \subseteq S(\bg{},\col)\cap\overline{B}(\bg{},\col)$.
%% \subparagraph{Show "$\supseteq$":}
%% Let $\r \in S(\bg{},\col)\cap\overline{B}(\bg{},\col)$ then by
%% Corollary~\ref{cor:nlp:properties} and by definition of $\GR{\Pi}{X}$ we have
%% that $\r\in \GR{\Pi}{X}=\colplus$. 
%% Therefore, $ S(\bg{},\col)\cap\overline{B}(\bg{},\col)\subseteq \colplus$.

\paragraph{"$2 \Rightarrow 1$":}
We have to show that $C$ is an admissible coloring that is
$\colplus=\GR{\Pi}{X}$ holds, where $X$ is an answer set of $\Pi$.

Let $X$ be the set of atoms such that $X=\head{\colplus}$, then
\begin{eqnarray*}
  \GR{\Pi}{X} & = & \{r \mid \pbody{r}\subseteq X, \nbody{r}\cap X
  =\emptyset\}\\
  &=& \{r\mid \pbody{r}\subseteq \head{\colplus}, \nbody{r}\cap \head{\colplus}
  =\emptyset\}\\
  & = & \{r\mid r\in S(\bg{},C), r\in\overline{B}(\bg{},C))\}\\
  & = & \colplus.
\end{eqnarray*}
By Theorem~\ref{thm:max:supported:graph:C} and $(\colplus,E)$ is a (maximal)
support 
graph of $(\bg{},C)$ for some $E\subseteq \Pi\times\Pi$
(Corollary~\ref{thm:supportgraph}), we have  
\begin{eqnarray*}
  \head{\colplus}& = & \Cn{(\Pi\setminus\colminus)^\emptyset}\\
   & = & \Cn{\colplus^\emptyset}\\
   & = & \Cn{(\GR{\Pi}{X})^\emptyset}.
\end{eqnarray*}
By $X=\head{\colplus}=\Cn{(\GR{\Pi}{X})^\emptyset}$ we have by
Theorem~\ref{thm:as:consGR} that $X$ is an answer set.
Hence, $C$ is an admissible coloring.

%% We have to show that $X$ is an answer set, that is $\Cn{\Pi^X}=X$.
%% With Theorem~\ref{thm:as:consGR} it is enough to show that
%% $X=\Cn{(\GR{\Pi}{X})^\emptyset}$. 
%% Because $\col$ is a total coloring we have $\GR{\Pi}{X}=\colplus=\Pi\setminus
%% \colminus$.
%% Because we have a maximal support graph $(V,E)$ of $(\bg{},\col)$ for some
%% $E\subseteq (\Pi \times \Pi)$ we can conclude
%% $\head{V}=\Cn{(\Pi\setminus\colminus)^\emptyset}$ by
%% Theorem~\ref{thm:max:supported:graph:C}.  
%% Furthermore, by Definition~\ref{def:acol:maxgrounded} we must have $V=\colplus$ because $\col$ is a total coloring.
%% Thus,
%% $X=\head{\colplus}=\Cn{(\Pi\setminus\colminus)^\emptyset}=\Cn{(\GR{\Pi}{X})^\emptyset}$  
%% and therefore $X$ is an answer set.

% For this proof we use theorem~\ref{thm_coloring} and the
% definition~\ref{def:acoloring} of an \acoloring.

% Clearly, we have that \r\ is supported in $(\bg{},\col)$ if \r\ is grounded
% wrt \bg{}. 

% \paragraph{"$\Rightarrow$":}
% It is easy to see that we have $\colplus =
% S(\bg{},\col)\cap\overline{B}(\bg{},\col)$.
% It remains to show, that there exists a maximal support graph.
% But through theorem~\ref{thm:supportgraph} the existence is ensured.

% \paragraph{"$\Leftarrow$":}
% Through the existence of the support graph we ensure that there exists a
% grounded-0-graph $(V,E)$ for every $\r\in \colplus$ such that $V \subseteq
% \colplus$.
% The other properties follow immediately and $\col$ is an \acoloring.

\paragraph{"$2 \Leftrightarrow 3$":}
%% It is enough to show that
%% $\colplus=S(\bg{},\col) \cap \overline{B}(\bg{},\col)$ iff $\colminus =
%% \overline{S}(\bg{},\col)\cup B(\bg{},\col)$.
This holds by $\col$ is a total coloring and by Theorem~\ref{thm:col:gr}.
\end{proof}
%%% Local Variables: 
%%% mode: latex
%%% TeX-master: t
%%% End: 

%% file: Proofs/LongVersion/thm_red.tex
\begin{proof}{thm:red}
Let \bg{}  be the \RDGraph\ of logic program $\Pi$, $\col$ be a total coloring 
of \bg{} and $X$ be a set of atoms such that $X=\head{\colplus}$.

We obtain for $r\in \Pi$,
\begin{eqnarray*}
  r\in  \overline{B}(\bg{},\col) 
  & \text{ iff } &  \nbody{\r}\cap \head{\colplus} =\emptyset\\
  & \text{ iff } & \nbody{\r}\cap X=\emptyset\\
  & \text{ iff } &  \head{\r} \LPif \pbody{\r} \in \Pi^X.      
\end{eqnarray*}
\end{proof}
%%% Local Variables: 
%%% mode: latex
%%% TeX-master: t
%%% End: 

%% file: Proofs/LongVersion/thm_max_supported_graph_C.tex
\begin{proof}{thm:max:supported:graph:C}
  Let $\bg{}$ be the \RDGraph{} of logic program $\Pi$,
  $C$ be a partial coloring
  and 
  $(V,E)$ be a maximal \groundedZeroDAG{} of $(\bg{},\col)$.

We have to show that $\head{V}=Cn((\Pi\setminus\colminus)^\emptyset)$ holds.
More precisely, we have to show that $\head{V}$ is the smallest set of atoms
which is closed
under $(\Pi\setminus)^\emptyset$.

First, we show that $\head{V}$ is closed
under $(\Pi\setminus)^\emptyset$.
That is, for any $r\in (\Pi\setminus \colminus)^\emptyset$ we have
$\head{r}\in \head{V}$ whenever $\pbody{r}\subseteq \head{V}$.
Let be $r\in (\Pi\setminus\colminus)^\emptyset$.
If we have $\pbody{r}\subseteq \head{V}$ then we have $\head{r}\in\head{V}$
since the support graph $(V,E)$ is maximal.
Thus, $\head{V}$ is closed under $(\Pi\setminus\colminus)^\emptyset$.

Second, we show that $\head{V}$ is the smallest set of atoms which is closed
under $(\Pi\setminus\colminus)^\emptyset$.
Assume that $\head{V}$ is not the smallest set of atoms which is closed under
$(\Pi\setminus\colminus)^\emptyset$.
Then, there exists a $q\in \head{V}$ such that $\head{V}\setminus q$ is closed
under $(\Pi\setminus\colminus)^\emptyset$.
We have $\head{V}\setminus q = \head{V\setminus Q}$ where $Q=\{r\mid
\head{r}=q, r\in V\}$.
Moreover, we have for all $r^+\in (\Pi\setminus\colminus)^\emptyset$ that
$\head{r^+}\in \head{V\setminus Q}$ whenever $\pbody{r^+}\subseteq
\head{V\setminus Q}$. 
Let be $r'\in Q$.
By $r'\in V$ we have that $r$ is a vertex in the maximal support graph of
$(\bg{},C)$. 
Hence, by Definition~\ref{def:acol:maxgrounded} we have that
$\pbody{r'}\subseteq \head{V}$.
Moreover, we have $\pbody{r'}\subseteq \head{V\setminus Q}$ since
$Q=\{r\mid\head{r}=q, r\in V\}$ and $(V,E)$ is acyclic.
Thus, we have that $\head{r'}\in \head{V\setminus Q}$ by $\head{V}\setminus q$
being closed under 
$(\Pi\setminus)^\emptyset$. 
But then we have $q=\head{r'}\in \head{r}\setminus q$ which is a
contradiction.
Hence, $\head{V}$ is the smallest set of atoms which is closed under $(\Pi\setminus\colminus)^\emptyset$.
\end{proof}
%%% Local Variables: 
%%% mode: latex
%%% TeX-master: t
%%% End: 

%% file: Proofs/LongVersion/prop_AS_maxSG.tex
\begin{proof}{prop:AS:maxSG}
  Let $\bg{}$ be the \RDGraph{} for logic program $\Pi$ and 
  $\col$ be a total coloring of $\bg{}$.

\paragraph{"$\Rightarrow$"}
Let $C$ be an admissible coloring of $\bg{}$.
By Theorem~\ref{thm:acoloring:sets} we have that 
\(
\colplus = S(\bg{},\col)\cap\overline{B}(\bg{},\col)
\)
and there exists a support graph of $(\bg{},\col)$.
By $C$ is total we have that $(\colplus,E)$ is a support graph of $(\bg{},C)$
for some $E\subseteq \Pi\times\Pi$.
Hence, $(\colplus,E)$ is a support graph of $\bg{}$.
Since $\colplus\subseteq \overline{B}(\bg{},C)$
we have that
$(\colplus,E)$ is a support graph of $\bg{}|_{\overline{B}(\bg{},C)}$.
Next, we want to show the maximality of $(\colplus,E)$.
Assume, $(\colplus,E)$ is not a maximal support graph of
$\bg{}|_{\overline{B}(\bg{},C)}$. 
Then, there exists an $r\in \colminus$ such that
$(\colplus\cup\{r\},E')$ is a support graph of
$\bg{}|_{\overline{B}(\bg{},C)}$ for some $E'\subseteq \Pi\times\Pi$.
But then, $r\in \overline{B}(\bg{},C)$ and $r\in S(\bg{},C)$ by
Definition~\ref{def:acol:Dasg} of a support graph of
$\bg{}|_{\overline{B}(\bg{},C)}$.
Hence, $r\in S(\bg{},C)\cap \overline{B}(\bg{},C)=\colplus$ 
But this is a contradiction to $r\in\colminus$.
Thus, $(\colplus,E)$ is a maximal support graph of
$\bg{}|_{\overline{B}(\bg{},C)}$.

\paragraph{"$\Leftarrow$"}
Let $(\colplus,E)$ be a maximal support graph of
  $\bg{}|_{\overline{B}(\bg{},\col)}$ for some $E\subseteq (\Pi\times\Pi)$.
By Theorem~\ref{thm:acoloring:sets} we have to show that
\(
\colplus = S(\bg{},\col)\cap\overline{B}(\bg{},\col)
\)
and there exists a support graph of $(\bg{},\col)$.

Since $(\colplus,E)$ is a support graph of $\bg{}|_{\overline{B}(\bg{},\col)}$
we have that $(\colplus,E)$ is a also support graph of $\bg{}$.
Furthermore, $(\colplus,E)$ is a support graph of $(\bg{},C)$ since $C$ is a
total coloring.

It remains to show that
\(
\colplus = S(\bg{},\col)\cap\overline{B}(\bg{},\col)
\)
holds.

\subparagraph{``$\subseteq$''}
Assume that there exists an $r\in\colplus$ such that
$r\not\in S(\bg{},\col)\cap\overline{B}(\bg{},\col)$.
That is, $r\in \overline{S}(\bg{},C)\cup B(\bg{},C)$ holds.
Since $(\colplus,E)$ is a support graph of
  $\bg{}|_{\overline{B}(\bg{},\col)}$,
$r\in B(\bg{},C)\cap \colplus$ is not possible.
Assume that $r\in\overline{S}(\bg{},\col)\cap \colplus$.
But by Definition~\ref{def:acol:Dasg} and $r\in\colplus$ which is in a support
graph of $\bg{}|_{\overline{B}(\bg{},\col)}$, we  have that
$r\in S(\bg{},C)$.
But this is a contradiction and hence, we have
$\colplus \subseteq S(\bg{},\col)\cap\overline{B}(\bg{},\col)$.

\subparagraph{``$\supseteq$''}
Assume there exists an $r\in S(\bg{},\col)\cap\overline{B}(\bg{},\col)$ such
that $r\not\in \colplus$ that is $r\in\colminus$ holds.
If $r\in \overline{B}(\bg{},\col)$ then $r$ is a vertex of the graph
$\bg{}|_{\overline{B}(\bg{},\col)}$.
If furthermore $r\in S(\bg{},\col)$ then $r$ is a vertex of the maximal
support graph of $\bg{}|_{\overline{B}(\bg{},\col)}$.
Since $(\colplus,E)$ is a maximal support graph of
$\bg{}|_{\overline{B}(\bg{},\col)}$, we  have $r\in\colplus$. 
\end{proof}
%%% Local Variables: 
%%% mode: latex
%%% TeX-master: t
%%% End: 

%% file: Proofs/LongVersion/thm_acol_acolorings_ver2.tex
\begin{proof}{thm:acol:acolorings:ver2}
  Let  $\bg{}=(\bgv,\bgezero,\bgeone)$ be the \RDGraph{} of a logic program
  $\Pi$ and
  let $\col$ be a total coloring of \bg{}.

\paragraph{``$\Rightarrow$''}
Let $C$ be an admissible coloring of $\bg{}$, then by Theorem~\ref{thm:acoloring:sets} there
exists a support graph of $(\bg{},\col)$.
Hence, it remains to show, that $(S,\bgeone|_S)$ is a blockage graph of
$(\bg{},C)$ where  $S=S(\bg{},\col)$.

For all $\r\in \colplus$ we have $r\in\overline{B}(\bg{},\col)$ by
Theorem~\ref{thm:acoloring:sets}.
Thus, Condition~1 for a blockage graph in Definition~\ref{def:acol:onedsg}
holds. 

Let be $\r\in S\cap \colminus$. 
By $\colminus=\overline{S}(\bg{},C)\cup B(\bg{},\col)$ we have that $r\in
B(\bg{},C)$. 
%% We have to show, that $r$ is blocked by some $\rp\in\colplus$.
%% We have $\pbody{\r}\subseteq X$ and $\r\not\in
%% \GR{\Pi}{X}$ by Corollary~\ref{cor:nlp:properties} and the definition of
%% $\GR{\Pi}{X}$.  
%% Thus, $\nbody{\r}\cap X \not=\emptyset$ by the definition of the \RDGraph\ and
%% \r\ is blocked by some $\rp\in 
%% \colplus $ (Corollary~\ref{cor:nlp:properties}).
Hence, Condition~2 from Definition~\ref{def:acol:onedsg} holds.
Thus, $(S,\bgeone|_S)$ is a blockage graph of
$(\bg{},C)$.

\paragraph{``$\Leftarrow$''}
By Theorem~\ref{thm:acoloring:sets} and the existence of a support graph of
$(\bg{},C)$  it remains to show that $\colplus =
S(\bg{},\col)\cap\overline{B}(\bg{},\col)$. 
Since $C$ is total, the support graph of $(\bg{},C)$ must be $(\colplus,E)$
for some $E\subseteq \Pi\times\Pi$.
\subparagraph{"$\subseteq$":}
Let be $\r\in \colplus$. Then, we have $\r\in S(\bg{},\col)$ by $(\colplus,E)$ is
a support graph of $(\bg{},\col)$ for some $E\subseteq (\Pi\times\Pi)$ and by
Definition~\ref{def:acol:maxgrounded}.
Assume, we have $\r\in B(\bg{},\col)$ then there exists an $\rp\in \colplus$ such that
$(\rp,\r)\in E_1$.
But this is a contradiction to Condition~1 in
Definition~\ref{def:acol:onedsg} of a blockage graph.
Thus, we have $\r\in \overline{B}(\bg{},\col)$ and hence, we have
$\colplus\subseteq S(\bg{},\col)\cap\overline{B}(\bg{},\col)$.
\subparagraph{"$\supseteq$":}
Let be $\r\in S(\bg{},\col)\cap\overline{B}(\bg{},\col)$.
We have to show that $\r\in \colplus$.
Assume that we have $\r\in \colminus$.
Then, by Condition~2 in Definition~\ref{def:acol:onedsg}
there exists an $\rp\in  \colplus$ such that $\r$ is blocked by $\rp$.
That's a contradiction and thus we have $\r\in \colplus$.
\end{proof}
%%% Local Variables: 
%%% mode: latex
%%% TeX-master: t
%%% End: 

%% file: Proofs/LongVersion/thm_acol_acolorings_ver1.tex
\begin{proof}{thm:acol:acolorings:ver1}
  Let  $\bg{}=(\bgv,\bgezero,\bgeone)$ be the \RDGraph{} of a logic program
  $\Pi$ and
  let $\col$ be a total coloring of \bg{}. 

To show this corollary we show the equivalence to the conditions given in
Theorem~\ref{thm:acol:acolorings:ver2}.
Clearly, a support graph of $(\bg{},C)$ is  $(\colplus,E)$ for some
$E\subseteq \Pi\times\Pi$ by Definition~\ref{def:acol:maxgrounded} since $C$
is total. 
Furthermore, conditions~2 and~3 of this
corollary are equivalent to 
conditions~1 and~2 in Definition~\ref{def:acol:onedsg} of a  blockage graph.
%% the existence of a
%% blockage graph:
%% \[(S,\bgeone|_S) \text{ is a blockage graph of }$(\bg{},C)$\text{ where }
%%   S=S(\bg{},\col)\]
%% iff
%% \begin{description}
%%   \item[C~1:] for all $\r\in (\colminus\cap{S}(\bg{},\col))$ there exists an
%%     $\rp \in \colplus$ such that $(\rp,\r)\in \bgeone$; 
%%   \item[C~2:] there are no $\r,\rp \in \colplus$
%%     such that $(\r,\rp)\in \bgeone$.
%% \end{description}
%% We will remark that all rules $\r \in \Pi\setminus S(\bg{},\col)$ are colored
%% with $\cminus$.
%% Otherwise $(\colplus,E)$ is no support graph of $(\bg{},\col)$, because an
%% \anotgrounded\ rule could not be in a support graph.
%% Condition C~2 corresponds directly to condition~1 of a blockage graph in
%% Definition~\ref{def:acol:onedsg}. 
%% Even under regard that $\colplus\subseteq S(\bg{},\col)$ because
%% $(\colplus,E)$ is a support graph of $(\bg{},\col)$.
%% Condition C~1 corresponds directly to condition~1 of a blockage graph in
%% Definition~\ref{def:acol:onedsg} by $\colplus \subseteq S(\bg{},\col)$ and
%% thus $\colplus \cap S(\bg{},\col)=\colplus$. 
\end{proof}
%%% Local Variables: 
%%% mode: latex
%%% TeX-master: t
%%% End: 

%% file: Proofs/LongVersion/thm_P_reflexivity.tex
\begin{proof}{thm:P:reflexivity}
  Let \bg{} be the \RDGraph{} of logic program $\Pi$ and \col{} a partial
  coloring of \bg{}.
If $\AC{\Pi}{C}\not=\emptyset$ then $\AS{\Pi}{C}\not=\emptyset$ by
  Theorem~\ref{thm:AS:AC}.
Let $X\in \AS{\Pi}{\col}$ be an answer set of $\Pi$.
Then $\colplus \subseteq \GR{\Pi}{X}$ and $\colminus \cap
  \GR{\Pi}{X}=\emptyset$. 

Let be $S=S(\bg{},\col),B=B(\bg{},\col),\overline{S}=\overline{S}(\bg{},\col)$
and $\overline{B}=\overline{B}(\bg{},\col)$.

For showing that $\fOPprop{\bg{}}{C}$ exists, we have to prove that
$\fOPprop{\bg{}}{C}$ is a partial coloring.
We prove this by Theorem~\ref{thm:col:gr} and by the fact that $C$ is a
partial coloring.
We observe
\begin{eqnarray*}
  \fOPprop{\bg{}}{C}_\cplus \cap \fOPprop{\bg{}}{C}_\cminus 
&=& \left( \colplus \cup (S \cap \overline{B})\right) \cap 
\left( \colminus \cup \overline{S} \cup B \right) \\
& = & (\colplus\cap \colminus) \cup (\colplus\cap\overline{S}) \cup 
      (\colplus\cap B)\\
&   &  \cup \;(S \cap \overline{B}\cap\colminus) \cup
      (S \cap \overline{B}\cap \overline{S})\cup
      (S \cap \overline{B}\cap B) \\
& = & (\colplus\cap\overline{S}) \cup (\colplus\cap B) \cup 
      (S \cap \overline{B}\cap\colminus) \\
& =& \emptyset.
\end{eqnarray*}
The last equality follows by Theorem~\ref{thm:col:gr}.
Thus, $\fOPprop{\bg{}}{C}$ is a partial coloring and hence,
$\fOPprop{\bg{}}{C}$ exists.
\end{proof}
%%% Local Variables: 
%%% mode: latex
%%% TeX-master: "~/tex/Papers/GraphLP/Grundlagen/Long/paper"
%%% End: 

%% file: Proofs/LongVersion/thm_P_defined.tex
\begin{proof}{thm:P:defined}
This follows directly from Theorem~\ref{thm:inductive:prop} by the existence of
$P(C)$. 
\end{proof}

%%% Local Variables: 
%%% mode: latex
%%% TeX-master: "~/tex/Papers/GraphLP/Grundlagen/paper"
%%% End: 

%% file: Proofs/LongVersion/cor_P_empty_exists.tex
\begin{proof}{cor:P:empty:exists}[Scetch]
   Let \bg{} be the \RDGraph{} of logic program $\Pi$. 
   We have to prove that $\fOPpropast{\bg{}}{\emptyC}$ exists.
   By Theorem~\ref{thm:inductive:prop} it is enough to show that $P(\emptyC)$
   exists. 

   Clearly, $P^0(\emptyC) = \emptyC$ exists.
   Assuming that $P^i(\emptyC)$ exists, we have to show 
   $\fOPprop{\bg{}}{P^i(\emptyC)}_\cplus\cap
   \fOPprop{\bg{}}{P^i(\emptyC)}_\cminus =\emptyset$ 
   stating that $P^{i+1}(\emptyC)$ exists.
   But this holds if
   \begin{eqnarray}
     \label{eq:proof:Pemptyexists:1}
    P^i(\emptyC)_\cplus \cap \overline{S}(\bg{},P^i(\emptyC)) &=&  \emptyset\\ 
     \label{eq:proof:Pemptyexists:2}
    P^i(\emptyC)_\cplus \cap B(\bg{},P^i(\emptyC)) &=&  \emptyset\\ 
     \label{eq:proof:Pemptyexists:3}
    P^i(\emptyC)_\cminus \cap S(\bg{},P^i(\emptyC)) \cap
    \overline{B}(\bg{},P^i(\emptyC)) &=&  \emptyset
   \end{eqnarray}
   We obtain $S(\bg{},C) \subseteq S(\bg{},C')$ and $\overline{B}(\bg{},C)
   \subseteq \overline{B}(\bg{},C')$ for all partial colorings $C,C'$ such
   that $C\sqsubseteq C'$.
   Hence, we have $P^i(\emptyC)_\cplus \subseteq S(\bg{},P^i(\emptyC)) \cap
   \overline{B}(\bg{},P^i(\emptyC))$. 
   Thus, Equation~\ref{eq:proof:Pemptyexists:1} and analogously
   equations~\ref{eq:proof:Pemptyexists:2} and~\ref{eq:proof:Pemptyexists:3}
   hold. 
   For this reason we have that $P^{i+1}(\emptyC)$ exists and, by induction,
   that $\fOPpropast{\bg{}}{\emptyC}$ exists.
%%Proof by induction over $i$ by using Theorem~\ref{thm:inductive:prop}.
\end{proof}

%% file: Proofs/ShortVersion/thm_P_monoton.tex
\begin{proof}{thm:P:monoton}
  Let \bg{} be the \RDGraph{} of logic program $\Pi$ 
  and
  let $\col$ and $\col'$ be partial colorings of \bg{} such that
  $\AC{\Pi}{\col'}\neq\emptyset$.
Furthermore, let be $\col\sqsubseteq\col'$.
Then, we also have $\AC{\Pi}{C}\not=\emptyset$ by definition.

\paragraph{$\fOPprop{\bg{}}{\col}\sqsubseteq\fOPprop{\bg{}}{\col'}$:}
It is easy to see that $\col \sqsubseteq \col'$ implies
\begin{enumerate}
\item $S(\bg{},\col) \subseteq S(\bg{},\col')$,
\item $\overline{S}(\bg{},\col) \subseteq \overline{S}(\bg{},\col')$,
\item $B(\bg{},\col) \subseteq B(\bg{},\col')$, and
\item $\overline{B}(\bg{},\col) \subseteq \overline{B}(\bg{},\col')$.
\end{enumerate}
Thus, $\fOPprop{\bg{}}{\col}\sqsubseteq\fOPprop{\bg{}}{\col'}$.

\paragraph{$\fOPpropast{\bg{}}{\col}\sqsubseteq\fOPpropast{\bg{}}{\col'}$:} 
This follows by showing $P(C)\sqsubseteq P(\col')$ through an induction proof and by Theorem~\ref{thm:inductive:prop}.
%% By Theorem~\ref{thm:inductive:prop} property~7 it is enough to show
%% that 
%% $P(C)\sqsubseteq P(\col')$. 
%% Thus, it remains to show, that $P^i(C) \sqsubseteq P^i(C')$ for all $i <
%% \omega$. 
%% We prove this by induction over $i$.
%% For $i=0$ we have $P^0(C)=C \sqsubseteq C' = P^0(C')$.
%% Assume that $P^i(C) \sqsubseteq P^i(C')$ for some $ i < \omega$.
%% We have to show that $P^{i+1}(C) \sqsubseteq P^{i+1}(C')$.
%% But $P^{i+1}(C)=\fOPprop{\bg{}}{P^i(C)}\sqsubseteq
%% \fOPprop{\bg{}}{P^i(C')}=P^{i+1}(C')$ holds by the first half of this theorem.
\end{proof}
%%% Local Variables: 
%%% mode: latex
%%% TeX-master: t
%%% TeX-master: "~/tex/Papers/GraphLP/Grundlagen/Long/paper"
%%% End: 

%% file: Proofs/LongVersion/thm_P_ASpreserving.tex
\begin{proof}{thm:P:ASpreserving}
  Let \bg{} be the \RDGraph{} of logic program $\Pi$ and \col{} be a partial
  coloring of \bg{}.

\paragraph{1:}
$\col \sqsubseteq \fOPprop{\bg{}}{\col}$ implies $\AC{\Pi}{C} \supseteq
\AC{\Pi}{\fOPprop{\bg{}}{\col}}$. 
Thus, it remains to show $\AC{\Pi}{C} \subseteq
\AC{\Pi}{\fOPprop{\bg{}}{\col}}$. 
If $\AC{\Pi}{C}=\emptyset$ then $\AC{\Pi}{\fOPprop{\bg{}}{\col}}=\emptyset$ by
$\col \sqsubseteq \fOPprop{\bg{}}{\col}$ and by Equation~(\ref{eq:def:AC}).

Let be $C'=(\GR{\Pi}{X},\Pi\setminus\GR{\Pi}{X}) \in \AC{\Pi}{C}$ an
admissible coloring of $\bg{}$ for some answer set $X$ of $\Pi$. 
We have to show that $\fOPprop{\bg{}}{\col}_\cplus \subseteq \GR{\Pi}{X}$ and
$\fOPprop{\bg{}}{\col}_\cminus \cap \GR{\Pi}{X} = \emptyset$.
But this holds by Theorem~\ref{thm:col:gr} and $\colplus \subseteq
\GR{\Pi}{X}$ and $\colminus \cap \GR{\Pi}{X}=\emptyset$.
\paragraph{2:}
Holds by Theorem~\ref{thm:inductive:prop}.
% Abbreviatory we write $C'$ instead of $P^i(C)$.
% Let $X \in \AS{\Pi}{C}$ then $X \in \AS{\Pi}{C'}$ by
% Theorem~\ref{thm:inductive:prop}.
% Then, $C'_\cplus \subseteq
% \GR{\Pi}{X}$ and $C'_\cminus \cap \GR{\Pi}{X} =\emptyset$.
% We have to show that $\AS{\Pi}{C} \subseteq \AS{\Pi}{P^{i+1}(C)}$.
% We have  
% \begin{eqnarray*}
% P^{i+1}(\col) & = & \fOPprop{\bg{}}{\col'} \\
%               & = & ( \col'_\cplus \cup (S(\bg{},\col') \cap
%               \overline{B}(\bg{},\col') ),
%                       \col'_\cminus \cup \overline{S}(\bg{},\col') \cup
%               B(\bg{},\col')).
% \end{eqnarray*}
% It remains to show, that $S(\bg{},\col') \cap
%               \overline{B}(\bg{},\col') \subseteq \GR{\Pi}{X}$ and 
% $(\overline{S}(\bg{},\col') \cup B(\bg{},\col')) \cap \GR{\Pi}{X}=\emptyset$.
% But this holds by Theorem~\ref{thm:col:gr}.
%% Thus, $\AS{\Pi}{C} \subseteq \AS{\Pi}{P^{i+1}(C)}$ and for this reason 
%% $\AS{\Pi}{C} \subseteq \AS{\Pi}{P(C)}$.
\end{proof}

%%% Local Variables: 
%%% mode: latex
%%% TeX-master: "~/tex/Papers/Preference/Acolorings/paper"
%%% End: 

%% file: Proofs/LongVersion/thm_weak_support_and_blockage_graph.tex
\begin{proof}{thm:weak:support:and:blockage:graph}
  Let $\bg{}=(\Pi,\bgezero,\bgeone)$ be the \RDGraph{} of logic program $\Pi$
  and
  $\col$ be a partial coloring of \bg{}.
Furthermore, let be $C'=\fOPprop{\bg{}}{C}$.

\paragraph{1:}
Let $(\colplus,E)$ be a support graph of $(\bg,\col)$
    for some $E\subseteq (\Pi\times \Pi)$.
We have to show that $(({\fOPpropast{\bg{}}{\col}})_\oplus,E')$
    is a support graph of $(\bg,{\fOPpropast{\bg{}}{\col}})$
    for some $E'\subseteq (\Pi\times \Pi)$.
We show this by construction of $(({\fOPpropast{\bg{}}{\col}})_\oplus,E')$ and
    by Theorem~\ref{thm:inductive:prop}.

Let $V^0=P^0(C)_\cplus=\colplus$ and $E^0=E$.
Assume that we have constructed the sets $V^i\subseteq \Pi$ and $E^i \subseteq
(\Pi \times \Pi)$ for some $i < \omega$ such that $P^i(C)_\cplus = V^i$.
Now, we want to construct the sets $V^{i+1}\subseteq \Pi$ and
$E^{i+1}\subseteq (\Pi\times\Pi)$.
We define
\begin{eqnarray*}
  V^{i+1} &= & V^i \cup \fOPprop{\bg{}}{P^i(C)}_\cplus\\
          &= & V^i \cup (S(\bg{},P^i(C))
              \cap \overline{B}(\bg{},P^i(C)))  \cup P^i(C)_\cplus\\
          &= &V^i \cup (S(\bg{},P^i(C))
              \cap \overline{B}(\bg{},P^i(C))),
\end{eqnarray*}
\[
E^{i+1} = E^i \cup \left( \{(r',r) \mid r'\in V^i, r\in V^{i+1}\} \cap
  \bgezero \right),
\]
\(
V=\bigcup_{i<\omega} V^i,
\)
and
\(
E'= \bigcup_{i<\omega} E^i.
\)
%% \[
%% E^{i+1}=E^i \cup E^{P^i(C)}
%% \]
%% where 
%% \[
%% E^{P^i(C)}= \bigcup_{\r_i\in S(\bg{},P^i(C)) \cap \overline{B}(\bg{},P^i(C)) }
%%             E^{r_i}
%% \]
%% such that
%% \[
%% E^{r_i} = \bigcup_{p\in \pbody{r_i}} \left\{ (\rp,\r_i) \mid 
%%   \head{\rp}=p \text{ and } \rp\in V^i  \right\}.
%% \]
%% We define
%% \[
%% (V,E')= \left( \bigcup_{i < \omega} V^i, \bigcup_{i < \omega} E^i \right).
%% \]
We have to show
\begin{description}
\item[(1a):] $V= P(C)_\cplus$ and
\item[(1b):] $(V,E')$ is a support graph of $(\bg{},P(C))$.
\end{description}
Then, by Theorem~\ref{thm:inductive:prop} we can conclude that 
$(({\fOPpropast{\bg{}}{\col}})_\oplus,E')$
    is a support graph of $(\bg,{\fOPpropast{\bg{}}{\col}})$
    for some $E'\subseteq (\Pi\times \Pi)$
\subparagraph{(1a):}
This holds by construction of $V$.
\subparagraph{(1b):}
 Clearly, $(V,E')$ is acyclic because $E$ is acyclic and we have only edges from
 $V^i$ to $V^{i+1}$ for all $i < \omega$.
 Furthermore, $(V,E')$ is a \subgraph{0} of $\bg{}$ by construction.
 Now, we have to show that for all $r\in V$ we have
 $\pbody{r}\subseteq\{\head{\rp}\mid (\rp,\r)\in E'\}$.
 For $r\in V^0$ this holds by $(\colplus,E)$ is a support graph of
 $(\bg{},\col)$.
 For $r \in V^i \setminus V^{i-1}$ for some $i > 0$ we have $\r \in
 S(\bg{},P^i(C)) \cap \overline{B}(\bg{},P^i(C))$.
 Thus, if we have $\r\in S(\bg{},P^i(C))$ and by
 Definition~\ref{def:acol:vertexproperties} 
 and construction of $E^{i+1}$ we have
 $\pbody{r}\subseteq\{\head{\rp}\mid (\rp,\r)\in E^{i+1}\}$.
 Thus, $(\fOPpropast{\bg{}}{\col}_\oplus,E')$ is a support graph of $\bg{}$.
 \paragraph{2:}
 Let $(\colplus\cup\colminus,\bgeone)|_{S(\bg{},C)}$ be a blockage graph of
    $(\bg{},C)$ and $\colplus\subseteq S(\bg{},C)$.
We have to show that
$(\colplus'\cup\colminus',\bgeone)|_{S(\bg{},C')}$
    is a blockage graph of $(\bg{},\col')$
where $C'={\fOPpropast{\bg{}}{\col}}$.
That is, we have to show that
\begin{description}
\item[(2a):] for all $r,\rp\in\colplus' \cap S(\bg{},C')$ we
  have $(r,\rp)\not\in E_1\mid_{S(\bg{},C')}$,
\item[(2b):] for all $r\in\colminus'\cap S(\bg{},C')$ there
  exists an $\rp\in \colplus'\cap S(\bg{},C')$ such that
  $(\rp,\r)\in E_1\mid_{S(\bg{},C')}$.
\end{description}
But both conditions hold by $C'$ being closed under $\fOPprop{\bg{}}{C}$.
\end{proof}
%%% Local Variables: 
%%% mode: latex
%%% TeX-master: "~/tex/Papers/Preference/Acolorings/paper"
%%% End: 

%% file: Proofs/LongVersion/thm_AS_iv.tex
\begin{proof}{thm:AS:iv}
  Let $\bg{}$ be the \RDGraph{} of logic program  $\Pi$ and
  let $\col$ be a total coloring of \bg{}.

\paragraph{"$\Rightarrow$":}
By Theorem~\ref{thm:acoloring:sets} it remains to show that $C=\fOPprop{\bg{}}{C}$.
We have 
\(
\fOPprop{\bg{}}{C}=C \sqcup   (
            S (\bg{},\col) \cap \overline{B}(\bg{},\col)
  ,\  
  \overline{S}(\bg{},\col) \cup           B (\bg{},\col)
  )
.\)
By Theorem~\ref{thm:acoloring:sets} we have 
\(
C=(
            S (\bg{},\col) \cap \overline{B}(\bg{},\col)
  ,\  
  \overline{S}(\bg{},\col) \cup           B (\bg{},\col)
  )
\)
and thus, $C=\fOPprop{\bg{}}{C}$.
\paragraph{"$\Leftarrow$":}
By Theorem~\ref{thm:acoloring:sets} it remains to show, that $\colplus= S (\bg{},\col) \cap
\overline{B}(\bg{},\col)$.
By $C=\fOPprop{\bg{}}{C}$ we have  $\colplus \supseteq S (\bg{},\col) \cap
\overline{B}(\bg{},\col)$.
Now, we show $\colplus \subseteq S (\bg{},\col) \cap
\overline{B}(\bg{},\col)$.
Let $\r\in \colplus$.
If $\r\in \overline{S}(\bg{},C)$ or $\r\in B(\bg{},C)$, then $\r\in \colminus$ by
$C=\fOPprop{\bg{}}{C}$, but this is a contradiction.
Thus, $\r\in S(\bg{},C) \cap \overline{B}(\bg{},C)$.
For this reason, we have $\colplus= S (\bg{},\col) \cap
\overline{B}(\bg{},\col)$.
\end{proof}
%%% Local Variables: 
%%% mode: latex
%%% TeX-master: t
%%% End: 

%% file: Proofs/LongVersion/cor_S_existence.tex
\begin{proof}{cor:S:existence}
 Let $\bg{}$ be the \RDGraph{} of logic program $\Pi$ and
  $\col$ be a partial coloring of \bg{}.
%% \paragraph{Show 1:}
%% Through Theorem~\ref{thm:max:supported:graph:existence} there exists a
%% maximal support graph of $(\bg{},\col)$.
%% Thus, $\fOPMaxGround{\bg{}}{\col}$ exists.
%% \paragraph{Show 2:}
%%   Let $(\colplus,E)$ be a \groundedZeroDAG\ of $\bg{}|_\col$
%%   for some $E \subseteq (\Pi\times \Pi)$.
%% We have to show, that $\fOPMaxGround{\bg{}}{\col}$ exists, that is that there
%% exists a maximal support graph $(V,E')$ of $(\bg{},\col)$ for some $E'
%% \subseteq (\Pi \times \Pi)$.
%% But this holds through Theorem~\ref{thm:max:supported:graph:existence}.
%% \paragraph{Show 3:}
If there exists a support graph of $(\bg{},C)$, then there always exists a 
maximal support graph of $(\bg{},C)$ (Theorem~\ref{prop:SG:existence}).
Thus, $\fOPMaxGround{\bg{}}{\col}$ exists.
\end{proof}
%%% Local Variables: 
%%% mode: latex
%%% TeX-master: t
%%% End: 

%% file: Proofs/LongVersion/thm_S_monoton.tex
%---------------------------------------------------------------
\begin{proof}{thm:S:monoton}
  Let \bg{} be the \RDGraph{} of logic program $\Pi$ 
  and
  let \col{} and $\col'$ be partial colorings of \bg{} such that
  $\AC{\Pi}{\col}\not=\emptyset$, $\AC{\Pi}{\col'}\not=\emptyset$.
%% If $\AC{\Pi}{C}\not=\emptyset$ then $\AS{\Pi}{C}\not=\emptyset$ by
%%   Theorem~\ref{thm:AS:AC}.
%% The same holds with $C'$.

Note that for $\col$ and $\col'$, $\fOPMaxGround{\bg{}}{\col}$ and
  $\fOPMaxGround{\bg{}}{\col'}$ exist by Corollary~\ref{cor:S:existence} and
  Theorem~\ref{thm:max:supported:graph:existence}. 

\paragraph{1:}
$C\sqsubseteq \fOPMaxGround{\bg{}}{\col}$ holds by definition of
$\OPMaxGround_{\bg{}}$. 
\paragraph{2:}
Let $(V,E)$ be a maximal support graph of $(\bg{},C)$ for some $E\subseteq
\Pi\times\Pi$. 
Then, we have
\(
\fOPMaxGround{\bg{}}{C}=(\colplus,
\Pi\setminus V)
\).
We claim that $(V,E)$ is also a maximal support graph of
$(\bg{},\fOPMaxGround{\bg{}}{C})$.
To see this, observe that
\begin{enumerate}
\item $(V,E)$ is a  support graph of $\bg{}$,
\item $\fOPMaxGround{\bg{}}{C}_\cplus \subseteq V$, and
\item $V \cap (\fOPMaxGround{\bg{}}{C})_\cminus = V \cap (\Pi\setminus
  V)=\emptyset$.
\end{enumerate}
Furthermore,  $(V,E)$ is maximal since $V \cup
  (\fOPMaxGround{\bg{}}{C})_\cminus= V \cup (\Pi\setminus V)=\Pi$ and
$V \cap (\fOPMaxGround{\bg{}}{C})_\cminus=\emptyset$.
Therefore, we have
\begin{eqnarray*}
  \fOPMaxGround{\bg{}}{\fOPMaxGround{\bg{}}{C}} & =& 
           (\fOPMaxGround{\bg{}}{C}_\cplus, \Pi\setminus V)\\
          &=& (\colplus, \Pi\setminus V)\\
          &=& \fOPMaxGround{\bg{}}{C}.
\end{eqnarray*}

%% We have to show
%% \begin{description}
%% \item[(a)] $\fOPMaxGround{\bg{}}{C}_\cplus =
%%   \fOPMaxGround{\bg{}}{\fOPMaxGround{\bg{}}{C}}_\cplus $
%% \item[(b)] $\fOPMaxGround{\bg{}}{C}_\cminus =
%%   \fOPMaxGround{\bg{}}{\fOPMaxGround{\bg{}}{C}}_\cminus$
%% \end{description}
%% \subparagraph{Show~(a):} This holds by definition of $\OPMaxGround_{\bg{}}$. 
%% \subparagraph{Show~(b):}

%% % By, $ \fOPMaxGround{\bg{}}{C} \sqsubseteq
%% % \fOPMaxGround{\bg{}}{\fOPMaxGround{\bg{}}{C}}$ we have 
%% % \(
%% %  \fOPMaxGround{\bg{}}{C}_\cminus \subseteq
%% %   \fOPMaxGround{\bg{}}{\fOPMaxGround{\bg{}}{C}}_\cminus
%% % \).
%% % Thus, it remains to show that 
%% % \(
%% %  \fOPMaxGround{\bg{}}{C}_\cminus \supseteq
%% %   \fOPMaxGround{\bg{}}{\fOPMaxGround{\bg{}}{C}}_\cminus
%% % \).

%% Let $(V,E)$  be a maximal support graph of $(\bg{},C)$.
%% Then, $\fOPMaxGround{\bg{}}{C}_\cminus=\colminus \cup (\Pi\setminus V)$.
%% Assume,
%% $\fOPMaxGround{\bg{}}{C}\not=\fOPMaxGround{\bg{}}{\fOPMaxGround{\bg{}}{C}}$.
%% By $\fOPMaxGround{\bg{}}{C'}_\cplus = C'_\cplus$ for any partial coloring $C'$
%% we must have
%% $\fOPMaxGround{\bg{}}{C}_\cminus\not=\fOPMaxGround{\bg{}}{\fOPMaxGround{\bg{}}{C}}_\cminus$.
%% $\fOPMaxGround{\bg{}}{C}_\cminus \subseteq
%% \fOPMaxGround{\bg{}}{\fOPMaxGround{\bg{}}{C}}_\cminus$ holdy by property~1 in
%% this theorem.
%% Thus, there exists an $\r\in
%% \fOPMaxGround{\bg{}}{\fOPMaxGround{\bg{}}{C}}_\cminus$ such that 
%% $\r\not\in \fOPMaxGround{\bg{}}{C}_\cminus$.
%% Then, $\r\in V$ and $r\not\in V'$ where $(V',E')$ is a maximal support graph
%% of $(\bg{},\fOPMaxGround{\bg{}}{C})$.
\paragraph{3:}
We have to show that
$\fOPMaxGround{\bg{}}{\col}_\cplus\subseteq\fOPMaxGround{\bg{}}{\col'}_\cplus$
 and
$\fOPMaxGround{\bg{}}{\col}_\cminus\subseteq\fOPMaxGround{\bg{}}{\col'}_\cminus$. 
Since  $\col\sqsubseteq\col'$  we have
$\fOPMaxGround{\bg{}}{\col}_\cplus\subseteq\fOPMaxGround{\bg{}}{\col'}_\cplus$.
Because of $\col\sqsubseteq\col'$, we must have for a maximal support graph
$(V',E')$ of $(\bg{},\col')$ and for a maximal support graph
$(V,E)$ of $(\bg{},\col)$ that $V' \subseteq V$ for some $E,E'\in (\Pi\times
\Pi)$. 
Thus,
$\fOPMaxGround{\bg{}}{\col}_\cminus\subseteq\fOPMaxGround{\bg{}}{\col'}_\cminus$
holds by definition of $\OPMaxGround_{\bg{}}$.
\end{proof}
%---------------------------------------------------------------

%%% Local Variables: 
%%% mode: latex
%%% TeX-master: t
%%% End: 

%% file: Proofs/LongVersion/thm_S_hilfslemma.tex
\begin{proof}{thm:S:hilfslemma}
  Let \bg{} be the \RDGraph{} of logic program $\Pi$ 
  and  
  \col{} be a partial coloring of \bg{}.

Clearly, we have $\col \sqsubseteq \fOPMaxGround{\bg{}}{\col}$ by
Thm.~\ref{thm:S:monoton}. 
Thus, we have $\AC{\Pi}{\fOPMaxGround{\bg{}}{\col}}\subseteq \AC{\Pi}{\col}$.
It remains to show that $\AC{\Pi}{\col}\subseteq
\AC{\Pi}{\fOPMaxGround{\bg{}}{\col}}$. 
If $\AC{\Pi}{\col}=\emptyset$ then
$\AC{\Pi}{\fOPMaxGround{\bg{}}{\col}}=\emptyset$ since $\col \sqsubseteq
\fOPMaxGround{\bg{}}{\col}$ (Theorem~\ref{thm:S:monoton}).
Let be $C'\in \AC{\Pi}{\col}$, where $C'=(\GR{\Pi}{X},\Pi\setminus
\GR{\Pi}{X})$ for some answer set $X$ of $\Pi$.
Let $(V,E)$ be a maximal support graph of $(\bg{},\col)$ for some $E \subseteq
(\Pi \times \Pi)$.
The existence of $(V,E)$ is ensured by
Theorem~\ref{thm:max:supported:graph:existence}. 
We have to show, 
\begin{enumerate}
\item $\fOPMaxGround{\bg{}}{C}_\cplus \subseteq \GR{\Pi}{X}$, and
\item $\fOPMaxGround{\bg{}}{C}_\cminus \cap \GR{\Pi}{X} = \emptyset$.
\end{enumerate}

\paragraph{1:}
This holds by $\fOPMaxGround{\bg{}}{C}_\cplus=\colplus \subseteq \GR{\Pi}{X}$.
\paragraph{2:}
We have $\fOPMaxGround{\bg{}}{C}_\cminus =\Pi \setminus V$.
Assume that there exists an $r\in (\Pi\setminus V) \cap \GR{\Pi}{X}$.
By $r\in \GR{\Pi}{X}$ we have $r\in S(\bg{},C')$ for $C'\in \AC{\Pi}{C}$.
But this is a contradiction to $r\in (\Pi\setminus V)$ since all rules in
$\Pi\setminus V$ can never be supported.
Thus,  $\fOPMaxGround{\bg{}}{C}_\cminus \cap \GR{\Pi}{X} = \emptyset$ holds.
% Then, for all rules $\r \in \Pi\setminus V$ there is no possibility to get
% $\pbody{\r}\subseteq X$ by definition of a maximal support graph.
% %
% Thus, $\pbody{\r}\not\subseteq X$ for all $\r\in \Pi\setminus V$ and \r\ is
% \anotgrounded\ in $(\bg{},\col)$ (by Corollary~\ref{cor:nlp:properties}) and
% thus $\r\not\in \GR{\Pi}{X}$ by definition of $\GR{\Pi}{X}$. 
% %
% Because \r\ is colored with $\cminus$ in $\fOPMaxGround{\bg{}}{\col}$ we have
% also $X\in \AS{\Pi}{\fOPMaxGround{\bg{}}{\col}}$ by
% $\fOPMaxGround{\bg{}}{\col}_\cminus \cap \GR{\Pi}{X}=\emptyset$. 
\end{proof}
%%% Local Variables: 
%%% mode: latex
%%% TeX-master: t
%%% End: 

%% file: Proofs/LongVersion/thm_weak_support_and_blockage_graph_two.tex
\begin{proof}{thm:weak:support:and:blockage:graph:two}
  Let $\bg{}=(\Pi,\bgezero,\bgeone)$ be the \RDGraph{} of logic program $\Pi$
  and
  $\col$ be a partial coloring of \bg{}.
  Furthermore, let $C'={\fOPMaxGround{\bg{}}{\col}}$.
  Condition~1 holds trivially since $C \sqsubseteq C'$ and $\colminus'\cap
  \colplus =\emptyset$.
  Condition~2 holds by $\colplus'=\colplus$ and by $(\colminus'\setminus
  \colminus)\cap S(\bg{},C)=\emptyset$.
\end{proof}

%%% Local Variables: 
%%% mode: latex
%%% TeX-master: "~/tex/Papers/GraphLP/Grundlagen/paper"
%%% End: 

%% file: Proofs/LongVersion/thm_answerset_PS.tex
\begin{proof}{thm:answerset:PS}
  Let $\bg{}$ be the \RDGraph\ of logic program $\Pi$ and let 
  $\col$ be a total coloring of \bg{}.

To show this theorem we show the equivalence to Corollary~\ref{thm:AS:iv}.

\paragraph{"$\Rightarrow$"}
It remains to show that   $\col={\fOPMaxGround{\bg{}}{\col}}$.
By  Corollary~\ref{thm:AS:iv}, there exists a (maximal) support graph of
$(\bg{},\col)$. 
That must be $(\colplus,E)$ for some $E \subseteq (\Pi \times \Pi)$ since
  \col\ is a total coloring.
Thus $\fOPMaxGround{\bg{}}{\col}=(\colplus,\Pi\setminus
  \colplus)=C$. 
% \subparagraph{Show $\colinit \sqsubseteq \col$:}
% It is enough to show, that $\colinitplus \sqsubseteq \colplus$.
% Let $\r\in\{\rp\mid \pbody{\rp}=\emptyset, (\Pi \times \{\rp\})\cap E_1
%   =\emptyset\}$ then \r\ is unblocked in $(\bg{},\col)$, $\emptyset
%   =\pbody{\r}\subseteq X$ and $\nbody{\r}\cap 
%   X=\emptyset$ by Corollary~\ref{cor:nlp:properties}.
% Thus, $\r\in \GR{\Pi}{X}=\colplus$ by the definition of $\GR{\Pi}{X}$ and thus
%   $\colinit \sqsubseteq \col$.

\paragraph{"$\Leftarrow$"}
It remains to show that there exists a maximal support graph of
$(\bg{},\col)$.
But this exists by the existence of $\OPMaxGround_{\bg{}}$.
\end{proof}
%%% Local Variables: 
%%% mode: latex
%%% TeX-master: t
%%% TeX-master: t
%%% End: 

%% file: Proofs/LongVersion/thm_pre_algo_main_ver1.tex
\begin{proof}{thm:pre:algo:main:ver1}
  Let \bg{} be the \RDGraph{} of logic program $\Pi$ and
  let $\col$ be a total coloring of \bg{}.

\paragraph{"$\Rightarrow$":}
Let $C$ be an admissible coloring of $\bg{}$.
We have to show that there exists an sequence $(\col^i)_{0 \leq i \leq n}$
satisfying the given conditions.
Let $m=|\colplus|$ and $\langle r_i \rangle_{0 \leq i < m}$ be an
arbitrary enumeration of $\colplus$.
Analogous let $\langle r_j \rangle_{m \leq j < n}$ be an
arbitrary enumeration of $\Pi \setminus \colplus=\colminus$.
For $0 \leq i < m$ take $\col^{i+1}=\fOPchoose{\Pi}{\cplus}{\bg{}}{\col^i}=
(\colplus^i \cup \{r_i\},\colminus^i)$ and 
for $m \leq i < n $ take $\col^{i+1}=\fOPchoose{\Pi}{\cminus}{\bg{}}{\col^i}=
(\colplus^i,\colminus^i\cup \{r_i\})$.
Thus, $\col^n$ is a total coloring and conditions~3 and~4 are fulfilled by
Corollary~\ref{thm:answerset:PS}. 

\paragraph{"$\Leftarrow$":}
By conditions~3,~4 and~5 we have $\col=\fOPprop{\bg{}}{{\col}}$ and
$\col={\fOPMaxGround{\bg{}}{\col}}$.
Thus, $C$ is an admissible coloring of $\bg{}$ by
Corollary~\ref{thm:answerset:PS}. 
\end{proof}
%%% Local Variables: 
%%% mode: latex
%%% TeX-master: "~/tex/Papers/Preference/Acolorings/paper"
%%% End: 

%% file: Proofs/LongVersion/thm_collection_OASC_i.tex
\begin{proof}{thm:collection:OASC:i}
  Let \bg{} be the \RDGraph{} of logic program $\Pi$ and
  let $\col$ be a total coloring of \bg{}.

  Let $(\col^i)_{0 \leq i \leq n}$ be a sequence satisfying 
conditions 1-5 
  in Theorem~\ref{thm:pre:algo:main:ver1}.

\paragraph{$1$:}
We have $\col^i$ is a partial coloring for $0\leq i \leq n$ by
Definition~\ref{def:pre:algo:choose:ver2}. 
\paragraph{$2$:}
By construction of $\col^i$ and by Definition~\ref{def:pre:algo:choose:ver2}
we have $\col^i \sqsubseteq \col^{i+1}$.
\paragraph{$3$:}
We have $\col^i\sqsubseteq \col^{i+1}$ and thus we have
$\AC{\Pi}{\col^i}\supseteq\AC{\Pi}{\col^{i+1}}$.
\paragraph{$4$:}
Since we have $C=C^n\sqsupseteq C^i$ for admissible coloring $C$, we have
$C' \in \AC{\Pi}{C^i}\not=\emptyset$.
\paragraph{$5$:}
Because $\AC{\Pi}{\col^i}\not= \emptyset$ we have by
Theorem~\ref{thm:max:supported:graph:existence} that there exists a (maximal)
support graph of $(\bg{},\col^i)$.
\end{proof}
%%% Local Variables: 
%%% mode: latex
%%% TeX-master: t
%%% End: 

%% file: Proofs/LongVersion/thm_pre_algo_main_ver2.tex
\begin{proof}{thm:pre:algo:main:ver2}
  Let \bg{} be the \RDGraph{} of logic program $\Pi$ and 
  let $\col$ be a total coloring of \bg{}.

\paragraph{"$\Leftarrow$":}
Since $(\OPprop{}\OPMaxGround)_{\bg{}}^\ast$ is closed under $\OPprop{\bg{}}$ and
$\OPMaxGround_{\bg{}}$ we have $\col=\fOPprop{\bg{}}{\col}$ and
$\col=\fOPMaxGround{\bg{}}{\col}$ and thus,
we have that $C$ is an admissible coloring of $\bg{}$ by Corollary~\ref{thm:answerset:PS}.

\paragraph{"$\Rightarrow$":}
Let $C=(\GR{\Pi}{X},\Pi\setminus \GR{\Pi}{X})$ be an admissible coloring of
$\bg{}$ for 
answer set $X$ of $\Pi$. 
We have to show that there exists a sequence $(\col^i)_{0 \leq i \leq n}$
satisfying the given conditions.
Let $\col^0= (\OPprop{}\OPMaxGround)_{\bg{}}^\ast ((\emptyset,\emptyset))$.
Assume, $C^i$ is defined for some $0\leq i$.
Let be $\r\in \Pi \setminus (\colplus^i \cup \colminus^i)$.
If we have $\r\in \GR{\Pi}{X}$ then we take
    \(
    \col^{i+1}
    =
    (\OPprop{}\OPMaxGround)_{\bg{}}^\ast
    (\colplus^i \cup \{r\},\colminus^i)    
    \).
Otherwise, if we have $\r\in \Pi \setminus \GR{\Pi}{X}$ then we take
    \(
    \col^{i+1}
    =
    (\OPprop{}\OPMaxGround)_{\bg{}}^\ast
    (\colplus^i ,\colminus^i\cup\{\r\})    
    \).
By construction there exists an $n<\omega$ such that
$\Pi=\colplus^n\cup\colminus^n$. 
Furthermore, by Theorem~\ref{thm:P:ASpreserving},~\ref{thm:S:hilfslemma}, and
by construction of each $C^{i+1} (for 0\leq i < n)$, we have that
$C^n=C=(\GR{\Pi}{X},\Pi\setminus\GR{\Pi}{X})$.  
\end{proof}
%%% Local Variables: 
%%% mode: latex
%%% TeX-master: "~/tex/Papers/Preference/Acolorings/paper"
%%% End: 

%% file: Proofs/LongVersion/cor_PUast_empty_exists.tex
\begin{proof}{cor:PUast:empty:exists}
Proof by induction over $i$ by using Theorem~\ref{thm:inductive:propPU}.
\end{proof}

%% file: Proofs/LongVersion/thm_collection_OASC_ii.tex
\begin{proof}{thm:collection:OASC:ii}
  Let \bg{} be the \RDGraph{} of logic program $\Pi$ and 
  let $\col$ be a total coloring of \bg{}.

  Let $(\col^i)_{0 \leq i \leq n}$ be a sequence satisfying conditions 1-3 
  in Theorem~\ref{thm:pre:algo:main:ver2}.

\paragraph{1-5:}
Hold analogous to Theorem~\ref{thm:collection:OASC:i} by
Definition~\ref{def:pre:algo:choose:ver2} and by
Theorem~\ref{thm:inductive:propPU}. 

%% By definition of $(\OPprop{}\OPMaxGround)_{\bg{}}^\ast$ we have that $\col^i$
%% is a partial coloring for all $i \in \{0,\ldots, n\}$.
%% \paragraph{Show 2:}
%% $C^i \sqsubseteq C^{i+1}$ holds by definition of
%% $(\OPprop{}\OPMaxGround{})_{\bg{}}^\ast$. 
%% \paragraph{Show 3:}
%% We have $\col^i\sqsubseteq \col^{i+1}$ and thus we have
%% $\AS{\Pi}{\col^i}\supseteq\AS{\Pi}{\col^{i+1}}$ by definition of
%% $\AS{\Pi}{.}$.  
%% \paragraph{Show 4:}
%% Because $\AS{}{\Pi}\not=\emptyset$ we have $\{X\}=\AS{\Pi}{C}$ by
%% $\GR{\Pi}{X}=\colplus$, $\colminus \cap 
%% \GR{\Pi}{X}=\emptyset$ and by $\col$ is a total coloring.
%% Thus by $\AS{\Pi}{\col^i}\supseteq\AS{\Pi}{\col^{i+1}}$ for $0\leq i <n$ and
%% $\AS{\Pi}{\col^n}=\AS{\Pi}{\col}=\{X\}$ we have
%% $\AS{\Pi}{\col^i}\not=\emptyset$ for $0 \leq i \leq n$.
%% \paragraph{Show 5:}
%% Because $\AS{\Pi}{\col^i}\not= \emptyset$ we have through
%% Theorem~\ref{thm:max:supported:graph:existence} that there exists a maximal
%% support graph of $(\bg{},\col^i)$.
\paragraph{6+7:}
These conditions hold by definition of $\OPprop{\bg{}}$:
\[
\begin{array}{l}
S(\bg{},C^i)\cap \overline{B}(\bg{},C^i) \subseteq \fOPprop{\bg{}}{C^i}_\cplus
  \subseteq
  (\OPprop{}\OPMaxGround)_{\bg{}}^\ast(\fOPchoose{\Pi}{\circ}{\bg{}}{\col^i})_\cplus 
  =C^{i+1}_\cplus, \text{ and } \\ 
\overline{S}(\bg{},C^i)\cup B(\bg{},C^i) \subseteq
  \fOPprop{\bg{}}{C^i}_\cminus 
  \subseteq
  (\OPprop{}\OPMaxGround)_{\bg{}}^\ast(\fOPchoose{\Pi}{\circ}{\bg{}}{\col^i})_\cminus 
  =C^{i+1}_\cminus . 
\end{array}
\]
\end{proof}
%%% Local Variables: 
%%% mode: latex
%%% TeX-master: "~/tex/Papers/Preference/Acolorings/paper"
%%% End: 

%% file: Proofs/LongVersion/thm_collection_OASC_ii_sub.tex
\begin{proof}{thm:collection:OASC:ii:sub}
Condition~1--3, 6--7 hold analogous to Theorem~\ref{thm:collection:OASC:ii}.
By $C^i$ is closed under $\OPMaxGround_{\bg{}}$, there exists a support graph
of $(\bg{},C^i)$.
Hence, Condition~5 holds.
\end{proof}

%%% Local Variables: 
%%% mode: latex
%%% TeX-master: "~/tex/Papers/GraphLP/Grundlagen/paper"
%%% End: 

%% file: Proofs/LongVersion/thm_comparison_i.tex
\begin{proof}{thm:comparison:i}
  Let \bg{} be the \RDGraph{} of logic program $\Pi$.

First, we show that every sequence over \classC{} satisfying
  conditions 1 and 2 in Theorem~\ref{thm:pre:algo:main:ver2} induces a 
sequence over \classC{} satisfying
  conditions~1-4 in Theorem~\ref{thm:pre:algo:main:ver1}.

Let $(\col^i)_{i \in J}$ be a sequence satisfying conditions 1 and 2
in Theorem~\ref{thm:pre:algo:main:ver2}.
We obtain an enumeration $\langle r_i \rangle_{i \in I}$ of $\Pi$ such that we
have: 
if $r_i \in PU^k(C^l)$, $r_j \in PU^e(C^f)$ where $l < f$ or $l=f$ and $k \leq
e$ then $i < j$.
Thus, we obtain a sequence  $(\col^{'i})_{i \in J'}$ such that
$\col^{'0}=(\emptyset,\emptyset)$ and 
\[
\col^{'i+1}= \left\{  
  \begin{array}{ll}
    (\col^{'i}_\cplus \cup \{r_i\}, \col^{'i}_\cminus) & \text{ if }
    r_i\in\GR{\Pi}{X} \\
    (\col^{'i}_\cplus , \col^{'i}_\cminus\cup \{r_i\}) & \text{ if }
    r_i\not\in\GR{\Pi}{X}.
  \end{array}
             \right.
\]
$(\col^{'i})_{i \in J'}$ clearly satisfies conditions~1-4 in
Theorem~\ref{thm:pre:algo:main:ver1} by $C^i$ is closed under $\OPprop{\bg{}}$
and $\OPMaxGround_{\bg{}}$ for every $i\in J$ in
Theorem~\ref{thm:pre:algo:main:ver2}.   

Second, 2 different sequences from Theorem~\ref{thm:pre:algo:main:ver2}
induces 2 different sequences in Theorem~\ref{thm:pre:algo:main:ver1}. 
Hence, $n \leq m$.
\end{proof}
%%% Local Variables: 
%%% mode: latex
%%% TeX-master: "~/tex/Papers/GraphLP/Grundlagen/paper"
%%% End: 

%% file: Proofs/LongVersion/thm_comparison_ii.tex
\begin{proof}{thm:comparison:ii}
  Let \bg{} be the \RDGraph{} of logic program $\Pi$
  and
  let $C$ be an admissible coloring of $\bg{}$.
  
  Let $(\col^i_1)_{0 \leq i \leq m}$ and $(\col^{j}_2)_{0 \leq j \leq n}$ be
  the 
  shortest sequences obtained for $C$ according to
  Theorem~\ref{thm:pre:algo:main:ver1} and
  Theorem~\ref{thm:pre:algo:main:ver2}, 
  respectively.
Since
$|C^{i+1}_1 \setminus C^i_1|=1$ for all $0 \leq i < m$
and
$|C^{j+1}_2 \setminus C^j_2|\geq 1$ for all $0 \leq j < n$, we have that
$n \leq m $ holds where $m=|\Pi|$.
\end{proof}
%%% Local Variables: 
%%% mode: latex
%%% TeX-master: "~/tex/Papers/GraphLP/Grundlagen/paper"
%%% End: 

%% file: Proofs/LongVersion/thm_pre_algo_main_ver2_plus_minus.tex
\begin{proof}{thm:pre:algo:main:ver2:pm}
  Let \bg{} be the \RDGraph{} of logic program $\Pi$ and
  let $\col$ be a total coloring of \bg{}.

We have to prove:
\begin{description}
\item[Plus:] 
  $C$ is an admissible coloring of $\bg{}$
  iff 
  there exists a sequence $(\col^i)_{0 \leq i \leq n}$ with the following
  properties: 
  \begin{enumerate}
  \item $\col^0=(\OPprop{}\OPMaxGround)_{\bg{}}^\ast((\emptyset,\emptyset))$\/;
  \item
    \(
    \col^{i+1}
    =
    (\OPprop{}\OPMaxGround)_{\bg{}}^\ast
    (\fOPchoose{\Pi}{\cplus}{\bg{}}{\col^i})    
    \)
    for
    $0\leq i < n$\/;
  \item $\col^n=\col$.
  \end{enumerate}
\item[Minus:] 
  $C$ is an admissible coloring of $\bg{}$
  iff 
  there exists a sequence $(\col^i)_{0 \leq i \leq n}$ with the following
  properties: 
  \begin{enumerate}
  \item $\col^0=(\OPprop{}\OPMaxGround)_{\bg{}}^\ast((\emptyset,\emptyset))$\/;
  \item
    \(
    \col^{i+1}
    =
    (\OPprop{}\OPMaxGround)_{\bg{}}^\ast
    (\fOPchoose{\Pi}{\cminus}{\bg{}}{\col^i})    
    \)
    for
    $0\leq i < n$\/;
  \item $\col^n=\col$.
  \end{enumerate}
\end{description}

\paragraph{'Plus':}
\subparagraph{"$\Leftarrow$"}
  $C$ is an admissible coloring of $\bg{}$
holds by Corollary~\ref{thm:answerset:PS} since $C$ is
closed under 
$\OPprop{\bg{}}$ and closed under $\OPMaxGround_{\bg{}}$.
\subparagraph{"$\Rightarrow$"}
Let $C=(\GR{\Pi}{X},\Pi\setminus \GR{\Pi}{X})$ be an admissible coloring of
  $\bg{}$ for
  answer set $X$  of $\Pi$.
We have to show the existence of a sequence $(\col^i)_{0 \leq i \leq n}$ with
  the given properties. 
Let $\col^0=(\OPprop{}\OPMaxGround)_{\bg{}}^\ast((\emptyset,\emptyset))$.
For $0 < i < n$ let
 \(
    \col^{i+1}
    =
    (\OPprop{}\OPMaxGround)_{\bg{}}^\ast
    (\fOPchoose{\Pi}{\cplus}{\bg{}}{\col^i})    
 \)
where 
$\fOPchoose{\Pi}{\cplus}{\bg{}}{\col^i}=(C^i_\cplus \cup \{r\},C^i_\cminus)$ 
for some $\r \in \GR{\Pi}{X} \cap ( \Pi\setminus (C^i_\cplus \cup
C^i_\cminus))$. 
It remains to show, that $\col^n$ is a total coloring and $C=C^n$ for some $n
  \geq 0$.

Assume, $C^n$ is not total, then $\Pi\setminus (C^n_\cplus \cup C^n_\cminus)
\not\subseteq \GR{\Pi}{X}$.
Otherwise, we could choose an $r\in\GR{\Pi}{X}$ to extend our sequence of
partial colorings.
Then, for all $r\in \Pi\setminus (C^n_\cplus \cup C^n_\cminus)$ we have either
$\pbody{r}\not\subseteq X$ or $\nbody{r}\cap X \not=\emptyset$.
Observe that $X=\head{\colplus^n}=\head{\GR{\Pi}{X}}$ holds by $X$ is an
answer set. 
If there is an $r\in \Pi\setminus (C^n_\cplus \cup C^n_\cminus)$ such that
$\nbody{r}\cap X \not=\emptyset$, then \r\ is blocked by some $\rp\in
C^n_\cplus$ and \r\ had to be colored in $C^n$ by $C^n$ is closed under
$\OPprop{\bg{}}$. 
Thus, $\pbody{r}\not\subseteq X$ for all $r \in \Pi\setminus (C^n_\cplus \cup
C^n_\cminus)$. 
Furthermore, all $r\in \Pi\setminus (C^n_\cplus \cup
C^n_\cminus)$ are in a maximal support graph of $(\bg{},C^n)$ by $C^n$ is
closed under $\OPMaxGround_{\bg{}}$.
Thus, there must exists an $r\in\Pi\setminus  (C^n_\cplus \cup
C^n_\cminus)$ such that $r\in S(\bg{},C^n)\cap \overline{B}(\bg{},C^n)$.
But then, $r$ would be colored in $C^n$ by $C^n$ is closed under
$\OPprop{\bg{}}$.
That's a contradiction.
Thus, $C^n$ is a total coloring.
Furthermore, $C^n=C$ holds by $C^n$ is closed under $\OPprop{\bg{}}$ and by
Theorem~\ref{thm:col:gr}. 

\paragraph{'Minus':}
\subparagraph{"$\Leftarrow$"}
$C$ is an admissible coloring holds by Corollary~\ref{thm:answerset:PS} because $C$ is
closed under 
$\OPprop{\bg{}}$ and closed under $\OPMaxGround_{\bg{}}$.
\subparagraph{"$\Rightarrow$"}
Let $C=(\GR{\Pi}{X},\Pi\setminus \GR{\Pi}{X})$ be an admissible coloring for
  answer set $X$  of $\Pi$.
We have to show the existence of a sequence $(\col^i)_{0 \leq i \leq n}$
satisfying {\bf 'Minus'}. 
Let $\col^0=(\OPprop{}\OPMaxGround)_{\bg{}}^\ast((\emptyset,\emptyset))$.
For $0 < i < n$ let
 \(
    \col^{i+1}
    =
    (\OPprop{}\OPMaxGround)_{\bg{}}^\ast
    (\fOPchoose{\Pi}{\cminus}{\bg{}}{\col^i})    
 \)
where 
$\fOPchoose{\Pi}{\cminus}{\bg{}}{\col^i}=(C^i_\cplus ,C^i_\cminus \cup \{r\})$ 
for some $\r \in (\Pi\setminus\GR{\Pi}{X}) \cap ( \Pi\setminus (C^i_\cplus \cup
C^i_\cminus))$. 
It remains to show, that $\col^n$ is a total coloring and $C^n=C$, but this can be shown
analogous to {\bf 'Plus'}.
\end{proof}
%%% Local Variables: 
%%% mode: latex
%%% TeX-master: "~/tex/Papers/GraphLP/Grundlagen/paper"
%%% End: 

%% file: Proofs/LongVersion/thm_sequence_plus.tex
%---------------------------------------------------------------
\begin{proof}{thm:sequence:plus}
Let \bg{} be the \RDGraph{} of logic program $\Pi$ and $C^X$ be a total
coloring of \bg{}. 
\paragraph{"$\Rightarrow$":}

Let
\(
\col^X
=
(\GR{\Pi}{X},\Pi\setminus\GR{\Pi}{X})
\)
be an admissible coloring of $\bg{}$ for answer set $X$ of $\Pi$.
Clearly, $\col^X$ is a total coloring
and
\( 
\head{\colplus^X}=X
\)
by $X$ is an answer set.
% By Theorem~\ref{thm:XXX}, we have furthermore that
% $\col^X={\fOPprop{\bg{}}{\col^X}}$. 

Let $|\GR{\Pi}{X}|=n-2$ 
and 
$\langle r_i \rangle_{0\leq i < n-2}$ be an enumeration of $\GR{\Pi}{X}$.

Given this,
we define the sequence $(\col^i)_{0 \leq i \leq n}$ such that
\begin{enumerate}
\item $C^0=(\emptyset,\emptyset)$,
\item $C^{i+1}=(C^i_\cplus \cup \{r_i\},C^i_\cminus)$ for $0 \leq i < n-2$,
\item $\col^{n-1}=\fOPpropast{\bg{}}{{\col^{n-2}}}$, and
\item [3.]$\!\!\!\!\!'\;$ $\col^n={\fOPMaxGround{\bg{}}{\col^{n-1}}}$.
\end{enumerate}
Clearly, this sequence satisfies conditions 1, 2, and 3 in
Theorem~\ref{thm:sequence:plus}.

Next, we show that $C^n$ is a total coloring.
More precisely, we show that $\col^n=\col^X$.

By definition, we have
\(
\col^{n-2}=(\GR{\Pi}{X},\emptyset)
\).
By Theorem~\ref{thm:sequence:P:blocked} this implies
\[
\fOPpropast{\bg{}}{{\col^{n-2}}}
=
\col^{n-1}
=
(\GR{\Pi}{X},C^{n-1}_\ominus)
\ ,
\text{ where }
B(\bg{},\col^X)\subseteq \colminus^{n-1}
\]
by $\fOPprop{\bg{}}{C^{n-2}}\sqsubseteq \fOPpropast{\bg{}}{C^{n-2}}$.

By definition of $C^X$ and Theorem~\ref{thm:col:gr},
we have that
\(
\GR{\Pi}{X}=S(\bg{},\col^X)\cap \overline{B}(\bg{},\col^X)
\).
That is, we have
\[
S(\bg{},\col^X)\cap \overline{B}(\bg{},\col^X)
=
\colplus^{n-1}
\ .
\]
Moreover, given that
\(
B(\bg{},\col^X)\subseteq \colminus^{n-1}
\),
we obtain
\[
S(\bg{},\col^X) \cap B(\bg{},\col^X) \subseteq \colminus^{n-1}.
\]
Since $\col^X$ is a total coloring, we have
$C^X={B(\bg{},\col^X)\cup\overline{B}(\bg{},\col^X)}$.
Consequently,
\[
S(\bg{},\col^X)\subseteq \colplus^{n-1} \cup \colminus^{n-1}
\quad\text{ and also }\quad
\Pi\setminus(\colplus^{n-1} \cup \colminus^{n-1})\subseteq\overline{S}(\bg{},\col^X)
\ \text{ holds.}
\]
That is, all rules uncolored in $\col^{n-1}$ are also unsupported in $\col^X$.

Clearly, $(\colplus^X,E)$ is a maximal support graph of $(\bg{},\col^X)$
for some $E\subseteq (\Pi\times \Pi)$.
Given that 
$\col^X_\oplus=\colplus^{n-1}$
and
$\col^X_\ominus\setminus\colminus^{n-1}\subseteq\overline{S}(\bg{},\col^X)$,
$(\colplus^X,E)$ is also a maximal support graph of $(\bg{},\col^{n-1})$.
We get
\begin{eqnarray*}
{\fOPMaxGround{\bg{}}{\col^{n-1}}}
&=&
(\colplus^{n-1},(\Pi\setminus \colplus^X))
\\&=&
(\colplus^X,(\Pi\setminus \colplus^X))
\\&=&
(\colplus^X, \colminus^X)
\\&=&
\col^X
\end{eqnarray*}
That is, we have $C^n=C^X$.

Therefore, we obtain that $C^n$ is a total coloring.
%  and, by
% Corollary~\ref{thm:answerset:PS}, that is satisfies
% $\col^n=\fOPprop{\bg{}}{\col^n}$, that is,
% Condition~4 in Theorem~\ref{thm:sequence:plus}.

\paragraph{"$\Leftarrow$":}
According to Corollary~\ref{thm:answerset:PS},
it is sufficient to show that
\begin{enumerate}
\item $\col=\fOPprop{\bg{}}{\col}$ and
\item $\col=\fOPMaxGround{\bg{}}{\col}$.
\end{enumerate}

\subparagraph{1:}
% The former condition holds in view of Condition~4 in
% Theorem~\ref{thm:sequence:plus}. 
We have to show that
\[
  C
  =
  C \sqcup 
  (
            S (\bg{},\col) \cap \overline{B}(\bg{},\col)
  ,\  
  \overline{S}(\bg{},\col) \cup           B (\bg{},\col)
  )
  \ . 
  \]
For this, it is enough to show, that 
\[
  (
            S (\bg{},\col) \cap \overline{B}(\bg{},\col)
  ,\  
  \overline{S}(\bg{},\col) \cup           B (\bg{},\col)
  )
\sqsubseteq C
\] that is
\begin{description}
\item[(a)] $S (\bg{},\col) \cap \overline{B}(\bg{},\col) \subseteq \colplus$ and
\item[(b)] $\overline{S}(\bg{},\col) \cup B(\bg{},\col) \subseteq \colminus$.
\end{description}

{\bf (a):}
Assume there exists an $\r\in S (\bg{},\col) \cap \overline{B}(\bg{},\col) $
such that $\r\in \colminus$.
We have 2 cases where \r\ had to be colored in the sequence $(C^i)_{0 \leq i
  \leq n}$.
\begin{description}
\item[Case~1:] $\r\in \fOPpropast{\bg{}}{C^{n-2}}_\cminus$ and $r\not\in
  \colminus^{n-2}$  or
\item[Case~2:] $\r\in \fOPMaxGround{\bg{}}{
    \fOPpropast{\bg{}}{C^{n-2}}}_\cminus$ and $\r\not\in
    \fOPpropast{\bg{}}{C^{n-2}}_\cminus$.  
\end{description}
In {\bf Case~1} we have 
$\r\in \fOPpropast{\bg{}}{C^{n-2}}_\cminus= P(C^{n-2})_\cminus$ by
Theorem~\ref{thm:inductive:prop}.
Thus, there exists an $i < \omega$ such that 
$\r\in \overline{S}(\bg{},P^i(C^{n-2}))\cup B(\bg{},P^i(C^{n-2}))$.
By $P^i(C')\sqsubseteq P(C')$  we 
have $\r\in \overline{S}(\bg{},C)\cup B(\bg{},C)$, but this is a
contradiction. to $r\in S(\bg{},C)\cap \overline{B}(\bg{},C)$.\\
In {\bf Case~2} we have $\r\in \fOPMaxGround{\bg{}}{
    \fOPpropast{\bg{}}{C^{n-2}}}_\cminus$.
Let $(V,E)$ be a maximal support graph of
    $(\bg{},\fOPpropast{\bg{}}{C^{n-2}})$ for some $E\subseteq 
    (\Pi\times \Pi)$ 
Then, we have $\r\in \Pi \setminus V$.
By $r\in S (\bg{},\col)$ and 
$\colplus =  \fOPMaxGround{\bg{}}{\fOPpropast{\bg{}}{C^{n-2}}}_\cplus =
    \fOPpropast{\bg{}}{C^{n-2}}_\cplus$,
we have $r\in S(\bg{}, \fOPpropast{\bg{}}{C^{n-2}})$.
But then, we have $\r\in V$ by $V$ being maximal. 
Also, this is a contradiction and thus, we have $\r\in \colplus$.
 
{\bf (b):} 
Let be $\r\in B(\bg{},C)$ and $\r\in \colplus$.
There are 2 cases where \r\ had to be colored with $\cplus$.
\begin{description}
\item[Case~1:]  $\r\in C^{n-2}_\cplus$ or 
\item[Case~2:]  $\r\in \fOPpropast{\bg{}}{C^{n-2}}$ and $\r\not\in
  C^{n-2}_\cplus$. 
\end{description}
In both cases we have $\r\in
B(\bg{},\fOPMaxGround{\bg{}}{\fOPpropast{\bg{}}{C^{n-2}}})$. 
By $\fOPMaxGround{\bg{}}{\fOPpropast{\bg{}}{C^{n-2}}}_\cplus =
    \fOPpropast{\bg{}}{C^{n-2}}_\cplus$  we have
 $\r\in B(\bg{},\fOPpropast{\bg{}}{C^{n-2}})$.
 Hence, we must have $\r\in
\fOPpropast{\bg{}}{C^{n-2}}_\cminus\subseteq \colminus$ by $\OPpropast{\bg{}}$
is closed under $\OPprop{\bg{}}$. 
This is a contradiction.

Let be $\r\in
\overline{S}(\bg{},C)$ and
$\r\in \colplus$.
Then, we have $\r\in
\overline{S}(\bg{},\fOPMaxGround{\bg{}}{\fOPpropast{\bg{}}{C^{n-2}}})$. 
By Theorem~\ref{thm:S:monoton} and $\r\in
\fOPMaxGround{\bg{}}{\fOPMaxGround{\bg{}}{\fOPpropast{\bg{}}{C^{n-2}}}}_\cminus$
we have $\r\in \fOPMaxGround{\bg{}}{\fOPpropast{\bg{}}{C^{n-2}}}_\cminus$.
But, this is a contradiction to $\r\in \colplus$.

\subparagraph{2:}
% We observe that
% \(
% \col={\fOPMaxGround{\bg{}}{C'}}
% \)
% for some partial coloring $C'$.
$\col=\fOPMaxGround{\bg{}}{\col}$ is equivalent to
\(
\fOPMaxGround{\bg{}}{C'}=\fOPMaxGround{\bg{}}{\fOPMaxGround{\bg{}}{C'}}
\)
for $C'=\fOPpropast{\bg{}}{C^{n-2}}$,
which is true by virtue of Theorem~\ref{thm:S:monoton}.
\end{proof}
%---------------------------------------------------------------

%%% Local Variables: 
%%% mode: latex
%%% TeX-master: "~/tex/Papers/GraphLP/Grundlagen/paper"
%%% End: 

%% file: Proofs/LongVersion/thm_sequence_minus.tex
%---------------------------------------------------------------
\begin{proof}{thm:sequence:minus}
  Let \bg{} be the \RDGraph{} of logic program $\Pi$ and 
  let $\col$ be a total coloring of \bg{}.

\paragraph{"$\Rightarrow$":}
Let $C=(\GR{\Pi}{X},\Pi\setminus \GR{\Pi}{X})$ be an admissible coloring of
  $\bg{}$ for
  answer set $X$  of $\Pi$.
We want to construct a sequence $(\col^i)_{0\leq i \leq n}$ with the given
properties.

Let $\langle r_i \rangle_{1 \leq i \leq n-1}$ be an enumeration of $\Pi
\setminus \GR{\Pi}{X}$ where $n-1=|\Pi\setminus \GR{\Pi}{X}|$.
Let be $\col^0=(\emptyset,\emptyset)$.
For $1 \leq i \leq n-1$ let be $\col^{i}=(\colplus^{i-1},\colminus^{i-1}\cup
\{r_i\})$. 
Then, we have $\col^{n-1}=(\emptyset,\Pi\setminus \GR{\Pi}{X})$.
Note that by Theorem~\ref{thm:P:defined} and by $C\in\AC{\Pi}{C^{n-1}}$ we
have that $C^n=\fOPpropast{\bg{}}{\col^{n-1}}$ is a partial coloring.

It remains to show, that
$C= \col^n$ where $C^n=\fOPpropast{\bg{}}{\col^{n-1}}$. 
More precisely, we have to show for all $r\in\Pi$:
\begin{enumerate}
\item If $r\in\GR{\Pi}{X}$ then $r\in
  \colplus^n=\fOPpropast{\bg{}}{\col^{n-1}}_\cplus$.
\item If $r\not\in\GR{\Pi}{X}$ then $r\in
  \colminus^n=\fOPpropast{\bg{}}{\col^{n-1}}_\cminus$.
\end{enumerate}

\subparagraph{1:}
Let be $r\in\GR{\Pi}{X}$, then we have $\nbody{r}\cap X=\emptyset$.
By $X=\head{\GR{\Pi}{X}}$ we have $\nbody{r}\cap\head{\GR{\Pi}{X}}=\emptyset$.
Thus, we have $r\in\overline{B}(\bg{},C^{n-1})$.
By Theorem~\ref{thm:gr:grounded}, there exists an enumeration of $\langle r_i
\rangle_{i\in I}$ of $\GR{\Pi}{X}$ such that for all $i \in I$ we have
$\pbody{r_i} \subseteq \head{\{r_j \mid j <  i\}}$.
Clearly, we have $r_0 \in S(\bg{},C^{n-1})$.
Hence, we have $r_0\in\colplus^n$.
By induction over $i\in I$ we can show that
$r_i\in S(\bg{},(\colplus^{n-1}\cup\{r_j \mid j<i\}, \colminus^{n-1}))$.
Thus, $r\in \colplus^n$ whenever $r\in\GR{\Pi}{X}$.
\subparagraph{2:}
This follows by $\Pi\setminus\GR{\Pi}{X}=\colminus^{n-1}$.

Hence, we conclude that $C^n=C$ holds.

\paragraph{"$\Leftarrow$":}
Let $(\col^i)_{0\leq i \leq n}$ be a sequence with the given properties.
By Corollary~\ref{thm:answerset:PS} we have to show
\begin{enumerate}
\item $\col^n=\fOPprop{\bg{}}{{\col^n}}$ and
\item $\col^n={\fOPMaxGround{\bg{}}{\col^n}}$.
\end{enumerate}

\subparagraph{1:}
This condition is fulfilled by $\col^n=\fOPpropast{\bg{}}{\col^{n-1}}$ and by
$\OPpropast{\bg{}}$ being closed under $\OPprop{\bg{}}$.
\subparagraph{2:}
By $C^n=\fOPpropast{\bg{}}{C^{n-1}}$ and $(\emptyset,\emptyset)$ is a support
graph of $(\bg{},(\emptyset,\Pi\setminus \GR{\Pi}{X}))$ and
Theorem~\ref{thm:weak:support:and:blockage:graph} we have that $(\colplus^n,E)$ is a
support graph of $(\bg{},\col)$ for some $E\subseteq (\Pi \times \Pi)$.
Then,  $(\colplus^n,E)$ is a maximal support graph of $(\bg{},\col)$
and $\col^n={\fOPMaxGround{\bg{}}{\col^n}}$ by $\col^n$ is a total coloring
and $\colminus^n=\Pi\setminus \GR{\Pi}{X}$. 
\end{proof}
%---------------------------------------------------------------

%%% Local Variables: 
%%% mode: latex
%%% TeX-master: "~/tex/Papers/GraphLP/Grundlagen/paper"
%%% End: 

%% file: Proofs/LongVersion/thm_sequence_D_plusminus.tex
\begin{proof}{thm:sequence:D:plusminus}
 Let $\bg{}=(\Pi,\bgezero,\bgeone)$ be the \RDGraph{} of logic program $\Pi$
  and  
  let $\col$ be a total coloring of \bg{}.

\paragraph{"$\Rightarrow$":}
This follows analogous to the proof  "$\Rightarrow$" of
Corollary~\ref{thm:sequence:D:plus}. 

\paragraph{"$\Leftarrow$":}
This follows analogous to the proof of Theorem~\ref{thm:sequence:D:plus} by 
verifying the construction of the maximal support graph of $(\bg{},C)$ as
follows:

Let $R_i=\col^{i+1}_\cplus \setminus \col^i_\cplus$ for $0 \leq i < n-1$.
Let $V^0=\{r_0\}$ and $E^0=\emptyset$.
Assume, $V^i\subseteq \Pi$ and $E^i\subseteq \Pi\times\Pi$  are defined for
some $0 \leq i < n-1$.
Define $V^{i+1}=V^i \cup \{R_i\}$ and $E^{i+1}=E^i \cup E^{R_i}$ where
\[
E^{R_i}= \left\{
  \begin{array}{ll}
    \{(r',r_{i+1})\mid r'\in V^i\} \cap \bgezero
%%   \bigcup_{p\in \pbody{r_{i+1}}} 
%%              \{(\rp,r_{i+1}) \mid \rp\in V^i, \head{\rp}=p\}
  & \text{ if } R_i=\{r_{i+1}\} \text{ for some } r_{i+1}\in\Pi\\
  \emptyset 
  & \text{ if } R_i=\emptyset.
  \end{array}
         \right.
\]
\end{proof}
%%% Local Variables: 
%%% mode: latex
%%% TeX-master: "~/tex/Papers/GraphLP/Grundlagen/paper"
%%% End: 

%% file: Proofs/LongVersion/thm_collection_sequence_D_plusminus.tex
\begin{proof}{thm:collection:sequence:D:plusminus}
These properties follow analogous to
Theorem~\ref{thm:collection:sequence:D:plus} with the modification given in
Theorem~\ref{thm:sequence:D:plusminus} in the construction of the support
graph of $(\bg{},C^i)$ for $0\leq i\leq n$. 
\end{proof}
%%% Local Variables: 
%%% mode: latex
%%% TeX-master: "~/tex/Papers/GraphLP/Grundlagen/paper"
%%% End: 

%% file: Proofs/LongVersion/thm_sequence_D_plus.tex
\begin{proof}{thm:sequence:D:plus}
  Let $\bg{}=(\Pi,\bgezero,\bgeone)$ be the \RDGraph{} of logic program $\Pi$
  and  
  let $\col$ be a total coloring of \bg{}.

\paragraph{"$\Rightarrow$":}
Let $C=(\GR{\Pi}{X},\Pi\setminus \GR{\Pi}{X})$ be an admissible coloring of
  $\bg{}$ for 
  answer set $X$  of $\Pi$.
By Theorem~\ref{thm:gr:grounded} there exists an enumeration $\langle r_i
\rangle_{0 \leq i < n-1}$ of \GR{\Pi}{X} such 
that for all $i \in \{0,\ldots, n-2\}$ we have $\pbody{r_i} \subseteq
\head{\{r_j \mid j <   i\}}$.
We define the sequence $(C^i)_{0 \leq i \leq n}$ as follows:
\begin{enumerate}
\item $\col^0=(\emptyset,\emptyset)$,
\item $\col^{i+1}=(\col^i_\cplus \cup \{r_i\},\col^i_\cminus)$ for $0 \leq i <
  n-1$,
\item $\col^n=\fOPN{\bg{}}{\col^{n-1}}$. 
\end{enumerate}
By construction and by definition of $\OPN{\bg{}}$,  we have that $\col^n$ is
a total coloring, 
$\col^n_\cplus=\GR{\Pi}{X}$, and $\col^n_\cminus=\Pi\setminus \GR{\Pi}{X}$.
By definition of $\col^{i+1}$ for $0 \leq i <  n-1$ we have $\col^{i+1}=
\weakChoose{\bg{}}{\cplus}{\col^i}$ because $r_i \in S(\bg{},C^i)$ by
Theorem~\ref{thm:gr:grounded}. 
Thus, it remains to show that  $\col^n=\fOPprop{\bg{}}{\col^n}$.
$\col^n \sqsubseteq \fOPprop{\bg{}}{\col^n}$ holds by definition of
$\OPprop{\bg{}}$.
Hence, it is enough to show that
\begin{enumerate}
\item $S(\bg{},\col^n) \cap \overline{B}(\bg{},\col^n) \subseteq
  \col^n_\cplus$ and  
\item $\overline{S}(\bg{},\col^n) \cup B(\bg{},\col^n) \subseteq \col^n_\cminus$.
\end{enumerate}
By Theorem~\ref{thm:col:gr} we have,
if $\r\in S(\bg{},\col^n) \cap \overline{B}(\bg{},\col^n)$ then $r\in
\GR{\Pi}{X}=\col^n_\cplus$ and 
if $\r \in \overline{S}(\bg{},\col^n) \cup B(\bg{},\col^n)$ then $r\in
(\Pi \setminus\GR{\Pi}{X})=\col^n_\cminus$.
Thus, $\col^n=\fOPprop{\bg{}}{\col^n}$ and $\col^n=C$.

\paragraph{"$\Leftarrow$":}
By Corollary~\ref{thm:AS:iv} we have to show
\begin{enumerate}
\item $C=\fOPprop{\bg{}}{C}$ and
\item there is a support graph of $(\bg{},C)$.
\end{enumerate}
$C=\fOPprop{\bg{}}{C}$ holds by property~4 and~5 of the defined sequence
$(\col^i)_{0\leq i \leq n}$ in Theorem~\ref{thm:sequence:D:plusminus}.

Now, we want to construct by induction  a support graph $(\colplus^n,E)$ of
$(\bg{},C)$ for 
some $E\subseteq \Pi\times \Pi$.
Let $\{r_i\}=\col^{i+1}_\cplus \setminus \col^i_\cplus$ for $0 \leq i < n-1$.
Let $V^0=\{r_0\}$ and $E^0=\emptyset$.
Assume, $V^i\subseteq \Pi$ and $E^i\subseteq \Pi\times\Pi$  are defined for
some $0 \leq i < n-1$.
Define $V^{i+1}=V^i \cup \{r_{i+1}\}$ and $E^{i+1}=E^i \cup E^{r_{i+1}}$ where
\[
E^{r_{i+1}}=\{(r',r_{i+1})\mid r'\in V^i\}\cap \bgezero.
\]
%% \[
%% E^{r_{i+1}}= \bigcup_{p\in \pbody{r_{i+1}}} 
%%              \{(\rp,r_{i+1}) \mid \rp\in V^i, \head{\rp}=p\}.
%% \]
Let $V=\bigcup_{i < \omega} V^i$ and $E=\bigcup_{i < \omega} E^i$, then
$(V,E)$ is a support graph of $(\bg{},C)$  where $V=\colplus^n$.
Hence, $C$ is an admissible coloring of $\bg{}$.
\end{proof}
%%% Local Variables: 
%%% mode: latex
%%% TeX-master: "~/tex/Papers/GraphLP/Grundlagen/paper"
%%% End: 

%% file: Proofs/LongVersion/thm_collection_sequence_D_plus.tex
\begin{proof}{thm:collection:sequence:D:plus}
  Given the same prerequisites as in Corollary~\ref{thm:sequence:D:plus} with
  $\bg{}=(\Pi,\bgezero,\bgeone)$. 
  Let $(\col^i)_{0 \leq i \leq n}$ be a sequence satisfying conditions~1-5
  in Corollary~\ref{thm:sequence:D:plus}.

\paragraph{1-5:}
Hold analogous to Theorem~\ref{thm:collection:OASC:i}.
%% This holds by definition of $\mathcal{D}_{\bg{}}^{\cplus}$ and $\OPN{\bg{}}$.
%% \paragraph{Show~3:}
%% $\AS{\Pi}{\col^i}\supseteq\AS{\Pi}{\col^{i+1}}$ holds by $\col^i\sqsubseteq
%% \col^{i+1}$. 
%% \paragraph{Show~4:}
%% $\AS{\Pi}{\col^i}\not= \emptyset$ follows analogous to
%% Theorem~\ref{thm:collection:OASC:i}.  
%% \paragraph{Show~5:}
%% The existence of a (maximal) support graph follows by property~4 in this theorem
%% and by Theorem~\ref{thm:max:supported:graph:existence}.
\paragraph{8:}
This follows analogous to the construction of a support graph of
$(\bg{},C)$ in Corollary~\ref{thm:sequence:D:plus}.
\paragraph{9:}
We have to prove
\begin{description}
\item[(9a)] for all $r,\rp\in \colplus^i\cap S(\bg{},C^i) $ we have $(r,\rp)\not\in
  \bgeone|_{S(\bg{},C^i)}$,
\item[(9b)] for all $\r\in S(\bg{},C^i)\cap \colminus^i$ exists an $\rp\in
  \colplus^i\cap S(\bg{},C^i)$ such that $(\rp,\r)\in \bgeone|_{S(\bg{},C^i)}$.
\end{description}
\subparagraph{(9a):}
Let be $r,\rp\in \colplus^i \cap S(\bg{},C^i)$.
Assume that we have $(\r,\rp)\in \bgeone|_{S(\bg{},C^i)}$.
Then, we have $\rp\in B(\bg{},\col^i)$.
By $\col^i\sqsubseteq \col^n$ and by $\col^n$ is closed under $\OPprop{\bg{}}$
we have $\rp\in B(\bg{},\col^n)$ and thus $\rp\in \col^n_\cminus$.
But this is a contradiction to $\rp\in \col^i_\cplus\subseteq \colplus^n$.
\subparagraph{(9b):}
For $0 \leq i \leq n-1$ we have $\col^i_\cminus=\emptyset$ and thus, there is
nothing to show.
Let be $i=n$.
$\col^n$ is closed under $\OPprop{\bg{}}$.
Assume, there is an $\r\in \col^n_\cminus \cap S(\bg{},\col^n)$ such that
$r\in \overline{B}(\bg{},C^n)$. 
Then we have $r\in\colplus^n$  by $C^n$ is closed under $\OPprop{\bg{}}$. 
Thus, for $\r\in \col^n_\cminus \cap S(\bg{},\col^n)$ we have $\r\in
B(\bg{},\col^n)$.
Hence, there must exist an $\rp\in \col^n_\cplus$ such that $(\rp,\r)\in E_1$.
Furthermore, $\rp\in S(\bg{},C^n)$ holds by
$\colplus^n=\fOPprop{\bg{}}{C^n}_\cplus=S(\bg{},C^n)\cap \overline{B}(\bg{},C^n)$.
Thus, we have $(\rp,\r)\in \bgeone|_{(S(\bg{},C^n)}$.
\end{proof}
%%% Local Variables: 
%%% mode: latex
%%% TeX-master: "~/tex/Papers/GraphLP/Grundlagen/paper"
%%% End: 

%% file: Proofs/LongVersion/thm_sequence_DP_plusminus.tex
\begin{proof}{thm:sequence:DP:plusminus}
  Let \bg{} be the \RDGraph{} of logic program $\Pi$ and
  let $\col$ be a total coloring of \bg{}.

\paragraph{"$\Rightarrow$":}
Let $C=(\GR{\Pi}{X},\Pi\setminus\GR{\Pi}{X})$ be an admissible coloring of
$\bg{}$ for answer set $X$ and 
$\langle r_i \rangle_{0\leq i\leq m}$ be an
enumeration of $\GR{\Pi}{X}$ such that for all $i \in \{0,\ldots,m\}$ we
have $\pbody{r_i} \subseteq \head{\{r_j \mid j <   i\}}$.

Let be $\col^0=\fOPpropast{\bg{}}{(\emptyset,\emptyset)}$ and for $0\leq i <
n-1$ let be
\[
    \col^{i+1}
    =
    \fOPpropast{\bg{}}{\weakChoose{\bg{}}{\circ}{\col^i}}
    = \fOPpropast{\bg{}}{(\colplus^i\cup\{r\},\colminus^i)},
\]
where $r=r_j$ for some $0\leq j \leq m$ such that
$\{r_0,\ldots,r_{j-1}\}\subseteq \colplus^i\cup\colminus^i$ and $r_j \not\in
\colplus^i\cup\colminus^i$.
Furthermore, let be  $\col^n=\fOPN{\bg{}}{\col^{n-1}}$
where
$(\Pi\setminus(\colplus^{n-1}\cup\colminus^{n-1}))\cap\GR{\Pi}{X}=\emptyset$. 
Clearly, $C^n$ is a total coloring by definition of $\OPN{\bg{}}$.

Next, we show that $\colplus=\GR{\Pi}{X}$.
Then, we can conclude that $C^n=C$ holds.
Let be $r\in\colplus$.
\begin{description}
\item[Case~1:]  $r$ is the ``choice rule'' selected to obtain $\col^{i+1}$ from
$\col^i$ or 
\item[Case~2:] $r$ was colored with $\cplus$ by $\OPprop{\bg{}}$ at some
iterative step from $C^i$ to $C^{i+1}$ for some $0\leq i < n-1$.
\end{description}
In Case~1 we have $r\in\GR{\Pi}{X}$ by construction.
In Case~2 we have $r\in S(\bg{},C')\cap \overline{B}(\bg{},C')$ for some
$C^i\sqsubseteq C'\sqsubseteq C^{i+1}$.
Then, by Theorem~\ref{thm:nlp:properties} we have $r\in\GR{\Pi}{X}$.

Let be $r\in\GR{\Pi}{X}$ and assume that we have $\r\in\colminus^n$.
Then, we have one of the following cases:
\begin{description}
\item[Case~1:] $r\in\colminus^n\setminus\colminus^{n-1}$ ($r$ is colored by
  $\OPN{\bg{}}$) or 
\item[Case~2:] $r\in\colminus'$ for some $C^i\sqsubset C'\sqsubseteq
  C^{i+1}$ for some $0\leq i < n-1$ ($r$ is colored by $\OPprop{\bg{}}$).
\end{description}
Another case don't exists since  we only have
$\mathcal{D}^{\cplus}_{\bg{}}$ as choice.
In Case~1 we have then a contradiction to
$(\Pi\setminus (\colplus^{n-1}\cup\colminus^{n-1}))\cap
\GR{\Pi}{X}=\emptyset$. 
In Case~2 we have $r\in \overline{S}(\bg{},C'')\cup {B}(\bg{},C'')$ where 
$\col'=\fOPprop{\bg{}}{C''}$ holds for a partial coloring $C''$ such that
$C^i\sqsubseteq C''\sqsubseteq   C^{i+1}$ for some $0\leq i < n-1$.
But then, we get the contradiction $r\not\in\GR{\Pi}{X}$ by
Theorem~\ref{thm:nlp:properties}. 
Thus, we have $\GR{\Pi}{X}=\colplus$ and hence we have $C=C^n$.

It remains to show that $C=\fOPprop{\bg{}}{C}$.
But this holds by Corollary~\ref{thm:AS:iv}.

%% For this, we have to prove
%% \begin{enumerate}
%% \item $S(\bg{},C)\cap \overline{B}(\bg{},C) \subseteq \colplus$ and
%% \item $\overline{S}(\bg{},C)\cup B(\bg{},C) \subseteq \colminus$.
%% \end{enumerate}
%% \subparagraph{Show~1:}
%% Let $r\in S(\bg{},C)\cap \overline{B}(\bg{},C)$.
%% By Theorem~\ref{thm:col:gr} we have $r\in\GR{\Pi}{X}$ and thus, $r\in\colplus$.
%% \subparagraph{Show~2:}
%% Let $r\in \overline{S}(\bg{},C)\cup B(\bg{},C)$.
%% By $C$ is a total coloring and Theorem~\ref{thm:col:gr} we have
%% $r\not\in\GR{\Pi}{X}$ and thus, $r\in \colminus$. 

\paragraph{"$\Leftarrow$":}
By Corollary~\ref{thm:AS:iv}, it remains to show that there exist a support
graph of $(\bg{},C)$. 
That must be $(\colplus,E)$ for some $E\subseteq \Pi\times\Pi$ since $C$ is
total.

Each vertex in the sequence $(\col^i)_{0 \leq i \leq n}$, which is added to
some partial coloring $\col'$ is supported in $(\bg{},\col')$.
Thus, we can construct a support graph of $(\bg{},C)$ by inductive adding of
vertices according to their occurrence in $C'$.
This works similar to proof ``$\leftarrow$'' of
Corollary~\ref{thm:sequence:D:plus}. 
\end{proof}
%%% Local Variables: 
%%% mode: latex
%%% TeX-master: "~/tex/Papers/GraphLP/Grundlagen/paper"
%%% End: 

%% file: Proofs/LongVersion/thm_sequence_DP_plusminus_N.tex
\begin{proof}{thm:sequence:DP:plusminus:N}
  Given the same prerequisites as in Theorem~\ref{thm:sequence:DP:plusminus},
  let $(\col^i)_{0 \leq i \leq n}$ be a sequence satisfying conditions~1-5
  in Theorem~\ref{thm:sequence:DP:plusminus}.
We have to prove 
  \(
  (\fOPN{\bg{}}{\col^{n-1}}_\cminus\setminus\col^{n-1}_\cminus)\subseteq\overline{S}(\bg{},\col)  
  \).

Assume, there exists an $r\in
\fOPN{\bg{}}{\col^{n-1}}_\cminus\setminus\col^{n-1}_\cminus$ such that $r\in
S(\bg{},C)$.
By $r\in \colminus$ and $C$ is closed under $\OPprop{\bg{}}$ we must have
$r\in B(\bg{},C)$.
Thus, there must exists an $\rp\in\colplus$ such that \r\ is blocked by \rp.
By $\fOPN{\bg{}}{\col^{n-1}}_\cplus=\colplus^{n-1}$ we have  $\rp\in
\colplus^{n-1}$. 
Thus, we have $r\in B(\bg{},\col^{n-1})$.
Since $C^{n-1}$ is closed under $\OPprop{\bg{}}$ we must have
$r\in\col^{n-1}_\cminus$.
But this is a contradiction to $r\not\in \colminus^{n-1}$.
Hence, we have 
  \(
  (\fOPN{\bg{}}{\col^{n-1}}_\cminus\setminus\col^{n-1}_\cminus)\subseteq\overline{S}(\bg{},\col)  
  \).
\end{proof}
%%% Local Variables: 
%%% mode: latex
%%% TeX-master: "~/tex/Papers/GraphLP/Grundlagen/paper"
%%% End: 

%% file: Proofs/LongVersion/thm_collection_sequence_DP_plusminus.tex
\begin{proof}{thm:collection:sequence:DP:plusminus}
   Given the same prerequisites as in Theorem~\ref{thm:sequence:D:plusminus},
   let $(\col^i)_{0 \leq i \leq n}$ be a sequence satisfying conditions~1-5
   in Theorem~\ref{thm:sequence:D:plusminus}.

Properties~1.-5. follow analogous to
Theorem~\ref{thm:collection:sequence:D:plusminus} by $C\sqsubseteq
\fOPpropast{\bg{}}{C}$ for a partial coloring $C$.
Property~8 follows analogous to the proof "$\Leftarrow$" of
Theorem~\ref{thm:sequence:DP:plusminus}. 

For $0\leq i < n$ the properties~6 and~7 are fulfilled by $C^i$ being closed
under $\OPprop{\bg{}}$ . 
For $i=n$ the properties~6 and~7 follow by Theorem~\ref{thm:col:gr}, by $C^n$
is total, and $\colplus^n=\GR{\Pi}{X}$. 
\end{proof}
%%% Local Variables: 
%%% mode: latex
%%% TeX-master: "~/tex/Papers/GraphLP/Grundlagen/paper"
%%% End: 

%% file: Proofs/LongVersion/thm_sequence_DP_plus.tex
\begin{proof}{thm:sequence:DP:plus}
This could be proven analogous to the proofs of
Corollary~\ref{thm:sequence:D:plus} and
Theorem~\ref{thm:sequence:DP:plusminus}. 
\end{proof}
%%% Local Variables: 
%%% mode: latex
%%% TeX-master: "~/tex/Papers/GraphLP/Grundlagen/paper"
%%% End: 

%% file: Proofs/LongVersion/thm_collection_sequence_DP_plus.tex
\begin{proof}{thm:collection:sequence:DP:plus}

These properties could be proven analogous to properties given in
theorems~\ref{thm:collection:sequence:D:plus}
and~\ref{thm:collection:sequence:DP:plusminus}. 
\end{proof}
%%% Local Variables: 
%%% mode: latex
%%% TeX-master: "~/tex/Papers/GraphLP/Grundlagen/paper"
%%% End: 

%% file: Proofs/LongVersion/thm_weakP_supportgraph.tex
\begin{proof}{thm:weakP:supportgraph}
  Let $\bg{}=(\Pi,\bgezero,\bgeone)$ be the \RDGraph{} of logic program $\Pi$
  and
  $\col$ be a partial coloring of \bg{}.

Furthermore, let $(\colplus,E)$ be a support graph of $(\bg,\col)$
  for some $E\subseteq (\Pi\times \Pi)$.

By Theorem~\ref{thm:inductive:propT} we have $\weakProp{\bg{}}{\col}=T(C)$.
First, we want to construct a graph  $((\weakProp{\bg{}}{\col})_\oplus,E')$ for
some $E' \subseteq (\Pi \times \Pi)$.
Second, we have to show
\begin{description}
\item[SG] $((\weakProp{\bg{}}{\col})_\oplus,E')$ is a support graph of
  $(\bg{},\col)$ and
\item[M] $((\weakProp{\bg{}}{\col})_\oplus,E')$ is a maximal one .
\end{description}

Let $E^0=E$ and $V^0=T^0(\col)_\cplus =\colplus$.
Assume that we have constructed $V^i\subseteq \Pi$ and $E^i\subseteq \Pi
\times \Pi$ for some $i < \omega$.
Now, we want to construct $V^{i+1}\subseteq \Pi$ and $E^{i+1}\subseteq \Pi
\times \Pi$.
We define
\[
\begin{array}{l}
V^{i+1}=V^i \cup (S(\bg{},T^i(C)) \setminus T^i(C)_\cminus) \text{ and}\\
E^{i+1}= E^i \cup \left(\{(r',r'') \mid r'\in V^i, r''\in V^{i+1}\setminus V^i\}
\cap \bgezero \right).
%% \bigcup \left( \bigcup_{\r_i\in (S(\bg{},T^i(C)) \setminus
%%     T^i(C)_\cminus)\setminus 
%%   V^i} E^{r_i} \right)
\end{array}
\]
We define  $E'=\bigcup_{i<\omega} E^i$.
Clearly, we have $T^{i+1}(C)_\cplus=T^i(C)_\cplus \cup (S(\bg{},T^i(C))
\setminus 
T^i(C)_\cminus)= V^{i+1}$ by $V^i=T^i(C)_\cplus$ for all $i< \omega$.
Furthermore, we have
$\weakProp{\bg{}}{\col}_\oplus =\bigcup_{i<\omega} V^i$ by construction of
each $V^i$ and
by Theorem~\ref{thm:inductive:propT}.

Is remains to show, that we have constructed a maximal support graph of
$(\bg{},\col)$. 
\subparagraph{SG:}
Clearly, $E'$ is a subset of $\bgezero$ and acyclic since $E$ is acyclic and
there are only edges from $V^i$ to $V^{i+1}$.
Furthermore, by construction and by $r\in S(\bg{},T^i(C))$ for all $r\in
V^{i+1}$ we have for
all $r\in  \weakProp{\bg{}}{\col}_\cplus$ that 
\(
\pbody{\r}\subseteq\{\head{\rp}\mid (\rp,\r)\in E'\}
\)
holds.
\subparagraph{M:}
Assume that there exists an $\r\in \Pi \setminus
\weakProp{\bg{}}{\col}_\cplus$ where $\r\not\in \colminus$ such that 
$((\weakProp{\bg{}}{\col})_\cplus\cup\{\r\},E'')$ is a  support graph of
$(\bg{},\col)$ for some $E''\subseteq (\Pi\times \Pi)$.
By definition of a support graph we have $\r\in
S(\bg{},\weakProp{\bg{}}{\col})$  and thus by
Theorem~\ref{thm:inductive:propT} we have $\r\in S(\bg{},T(C))$.
Hence, there exists an $i < \omega$ such that $\r\in S(\bg{},T^i(C))$.
Because $\r\not\in \colminus$ we must have $\r\in T^{i+1}(C)_\cplus$ and thus
$\r\in\weakProp{\bg{}}{\col})_\cplus$ holds. 
That's a contradiction to $r\in\Pi\setminus \fOPpropast{\bg{}}{C}_\cplus$ and thus, 
$((\weakProp{\bg{}}{\col})_\oplus,E')$ is a maximal support graph of
  $(\bg{},\col)$. 
\end{proof}
%%% Local Variables: 
%%% mode: latex
%%% TeX-master: "~/tex/Papers/Preference/Acolorings/paper"
%%% End: 

%% file: Proofs/LongVersion/thm_weakS_S.tex
\begin{proof}{thm:weakS:S}
This follows directly from Theorem~\ref{thm:weakP:supportgraph} and from 
Definition~\ref{def:S} 
and~\ref{def:weakS}. 
\end{proof}
%%% Local Variables: 
%%% mode: latex
%%% TeX-master: "~/tex/Papers/Preference/Acolorings/paper"
%%% End: 

%% file: Proofs/LongVersion/thm_pre_algo_main_ver4.tex
\begin{proof}{thm:pre:algo:main:ver4}
  Let \bg{} be the \RDGraph{} of logic program $\Pi$ and
  let $\col$ be a total coloring of \bg{}.

\paragraph{"$\Leftarrow$"}
By Corollary~\ref{thm:answerset:PS} we have to show
$C=\fOPprop{\bg{}}{C}$ and
$C=\fOPMaxGround{\bg{}}{C}$.
$C=\fOPprop{\bg{}}{C}$ holds since $C=C^n$ is closed under $\OPprop{\bg{}}$.
Analogous to the proof ``$\leftarrow$'' of Theorem~\ref{thm:sequence:DP:plusminus} we can
construct $(\colplus,E)$ as a support graph of $(\bg{},C)$ for some $E\subseteq
\Pi\times\Pi$. 
By Corollary~\ref{thm:weakS:S} and $C=C^n$ is closed under $\fweakS_{\bg{}}$
we conclude $C=\fOPMaxGround{\bg{}}{C}$.

\paragraph{"$\Rightarrow$"}
Let $C=(\GR{\Pi}{X},\Pi\setminus\GR{\Pi}{X})$ be an admissible coloring of
$\bg{}$ for answer set $X$ of $\Pi$. 
Let $\langle r_i \rangle_{0\leq i < n}$ be an enumeration of \GR{\Pi}{X} such
that for all $0\leq i \leq n$ we have $\pbody{r_i} \subseteq \head{\{r_j \mid
  j < 
  i\}}$.
We define $\col^0=(\OPprop{}\fweakS)_{\bg{}}^\ast((\emptyset,\emptyset))$ and
\[
    \col^{i+1}
    =
    (\OPprop{}\fweakS)^\ast_{\bg{}}(\colplus^i\cup\{r\},\colminus')    
\]
where $r=r_j\in\Pi\setminus(\colplus^i\cup\colminus^i)$ and
$\{r_0,\ldots,r_{j-1}\} \subseteq \colplus^i\cup\colminus^i$ for some $0\leq j
\leq n$.
We have to show
\begin{enumerate}
\item $C=C^n$ is a total coloring,
\item $\colplus=\GR{\Pi}{X}$.
\end{enumerate}
But this follows analogous to the proof of
Corollary~\ref{thm:pre:algo:main:ver2:pm}, paragraph {\bf "Plus"}
"$\Rightarrow$".  
\end{proof}
%%% Local Variables: 
%%% mode: latex
%%% TeX-master: "~/tex/Papers/GraphLP/Grundlagen/paper"
%%% End: 

%% file: Proofs/LongVersion/thm_collection_OASC_iv.tex
\begin{proof}{thm:collection:OASC:iv}
Properties~1.--7. follow analogous to proof of
Theorem~\ref{thm:collection:OASC:ii}. 
Property~8 follows from the construction of the support graph $(\colplus,E)$
of $(\bg{},C)$ for some $E\subseteq \Pi\times\Pi$ (see proof of
Theorem~\ref{thm:pre:algo:main:ver4}). 
\end{proof}
%%% Local Variables: 
%%% mode: latex
%%% TeX-master: "~/tex/Papers/GraphLP/Grundlagen/paper"
%%% End: 

%% file: Proofs/LongVersion/thm_pre_algo_main_ver4_pm.tex
\begin{proof}{thm:pre:algo:main:ver4:pm}
Proof is analogous to Corollary~\ref{thm:pre:algo:main:ver2:pm}.
\end{proof}
%%% Local Variables: 
%%% mode: latex
%%% TeX-master: "~/tex/Papers/GraphLP/Grundlagen/paper"
%%% End: 

%% file: Proofs/LongVersion/thm_complexity_operators.tex
\begin{proof}{thm:complexity:operators}
  Let $\bg{}=(\Pi,\bgezero,\bgeone)$ be the \RDGraph{} of logic program $\Pi$. 
  Furthermore, let $n=\mid \Pi\mid$ be the size of $\Pi$ and
  $C$ be a partial coloring of $\bg{}$.

\paragraph{1:}
  Note that checking whether $r\in\Pi$ is supported, unsupported, blocked, or
  unblocked in $(\bg{},C)$ is in $\CO{k}$ where $k= \max_{r\in \Pi}
  |\{(r',r)\in \bgezero\cup\bgeone :\; r'\in \Pi\}|$.
  Hence, it is constant wrt $n$.

  $\fOPprop{\bg{}}{C}$ is defined as
  \(
  C \sqcup 
  (
             S (\bg{},\col) \cap \overline{B}(\bg{},\col)
   ,\  
   \overline{S}(\bg{},\col) \cup           B (\bg{},\col)
  )
  \).
  Hence, for every $r\in \Pi$ it has to be computed whether it belongs to 
  $S (\bg{},\col) \cap \overline{B}(\bg{},\col)$ or $\overline{S}(\bg{},\col)
  \cup  B (\bg{},\col)$. 
  Thus, $\OPprop{\bg{}}$ is computable in $\CO{n}$.
%%   $\fOPprop{\bg{}}{C}$ is defined as
%%   \(
%%   C \sqcup 
%%   (
%%              S (\bg{},\col) \cap \overline{B}(\bg{},\col)
%%    ,\  
%%    \overline{S}(\bg{},\col) \cup           B (\bg{},\col)
%%   )
%%   \).
%%   Hence, for every $r\in \Pi$ it has to be computed whether it belongs to 
%%   $S (\bg{},\col) \cap \overline{B}(\bg{},\col)$ or $\overline{S}(\bg{},\col)
%%   \cup  B (\bg{},\col)$. 
%%   Thus, $\OPprop{\bg{}}$ is computable in $\CO{n}$.

\paragraph{2:}
  We define 
  $k= \max_{r\in \Pi} |\{(r',r)\in \bgezero\cup\bgeone :\; r'\in \Pi\}|$ 
  as the maximal number of predecessors of a vertex and
   $k'= \max_{r\in \Pi} |\{(r,r')\in \bgezero\cup\bgeone :\; r'\in \Pi\}|$ 
  as the maximal number of successors of a vertex.

  Checking whether $r\in\Pi$ is supported, unsupported, blocked, or unblocked in
  $(\bg{},C)$ is in $\CO{k}$.
  Hence, computing whether $r\in S(\bg{},C)\cap\overline{B}(\bg{},\col)$ or
  $r\in  \overline{S}(\bg{},\col) \cup  B (\bg{},\col)$ is in $\CO{k}$.

  \begin{figure}[htbp]
    \centering

\begin{lstlisting}[language=myown, 
                 mathescape=true, 
                 numbers=left]
function propagation(var C, $\Gamma$)
queue := $\emptyset$
for all $r\in\Pi$ do
    $status(r):=false$ 
endfor
for all $r \in  \Pi$ do
    if $r\in(S(\bg{},C)\cap\overline{B}(\bg{},C))\cup\overline{S}(\bg{},C)\cup B(\bg{},C)$
    then push($r$,queue)
         $status(r):=true$
    endif 
endfor  
while queue is not empty do
  r := pop(queue)
  if $r \in S(\bg{}, C) \cap \overline{B}(\bg{}, C)$ and $r \notin \colplus$ 
  then  if $r \in \colminus$ 
	then return false
	else  $\colplus := \colplus \cup \{r\}$
	      for all $r' \in \Pi$ where $status(r')=false, (r, r') \in E_0 \cup E_1$ do
		  if $r'\in(S(\bg{},C)\cap\overline{B}(\bg{},C))\cup\overline{S}(\bg{},C)\cup B(\bg{},C)$
		  then push($r'$,queue)
		       $status(r'):=true$
		  endif		        
	      endfor
	endif
  endif
  if $r \in \overline{S}(\bg{}, C) \cup B(\bg{}, C)$ and $r \notin\colminus$ 
  then if $r \in \colplus$ 
       then return false
       else $\colminus := \colminus \cup \{r\}$
	    for all $r' \in \Pi$ where $status(r')=false, (r, r') \in E_0 \cup E_1$ do
	        if $r'\in(S(\bg{},C)\cap\overline{B}(\bg{},C))\cup\overline{S}(\bg{},C)\cup B(\bg{},C)$
	        then push($r'$,queue)
                     $status(r'):=true$
	        endif		        
            endfor
       endif
  endif
endwhile
\end{lstlisting}
    \caption{Algorithm for computation of $\fOPpropast{\bg{}}{C}$}
    \label{algo:propast}
  \end{figure}

An algorithm for the computation of $\fOPpropast{\bg{}}{C}$ is given in
Figure~\ref{algo:propast}. 
We call a rule $r\in\Pi$ \emph{decided} if the question of the applicability
of $r$ is been resolved that is, we have either 
$r\in S(\bg{},C) \cap \overline{B}(\bg{},C)$
or
$r\in \overline{S}(\bg{},C) \cup B(\bg{},C)$.
We call a rule $r\in\Pi$ \emph{undecided} if $r$ is not decided.
We use $status(r)=true$ for denoting that $r$ is decided and 
$status(r)=false$ for denoting that $r$ is undecided.
%% Undecided rules are uncolored rules and colored rules $r$ where the
%% question of 
%% applicability is not decided, e.g. $r\not\in S(\bg{},C) \cup
%% \overline{S}(\bg{},C)$.  
The main idea of the algorithm is to start from the colored 
vertices and going along the successors of the lastly colored vertices to
look for vertices which can be colored in the next step.

In Step~(2) the queue, which contains all vertices
which can be colored next, is initialized.

Step~(3)--(5) initialize the status of each vertex $r\in\Pi$ as false.
This works in $\CO{n}$.

%% Step~(6)--(19) check whether the given coloring $C$ contains no invalid colored
%% vertices and updates the status of all rules.
%% More precisely, 
%% for all $r\in S(\bg{},C) \cap\overline{B}(\bg{},C)$, Step~(9) returns a
%% failure if $r\in\colminus$ and 
%% Step~(10) sets $status(r)$ as $true$ if $r\in\colplus$.
%% Furthermore,
%% for all $r\in \overline{S}(\bg{},C) \cup B(\bg{},C)$, Step~(15) returns a
%% failure if $r\in\colplus$ and 
%% Step~(16) sets $status(r)$ as $true$ if $r\in\colminus$.
%% Since both {\bf if} conditions work in $\CO{m}$ we have that the {\bf for}
%% loop in Step~(6)--(19) works in $\CO{n}$.

%% Note that after Step~(19), all decided but not yet colored rules have the
%% status false. 
Step~(6--11) are putting all decided (Step~(7)) vertices into the queue.
%% These decided but not yet colored rules are taken into the queue.
Their status is becoming true in Step~(9).
Steps~(6--11) are computable in $\CO{n}$.

The {\bf while} loop in Step~(12)--(38) takes (iterative) a (decided) vertex
$r$ from the queue and considers the cases for coloring this rule.
%% Note that contradictions are detected in Step~(30) and~(42), where
%%  $r$ was colored with the wrong color.
%% After coloring $r$ there may become other rules decided, which are then taken
%% into the queue.

Step~(13) takes a rule from the queue.

Step~(14--25) are handling the case for $\cplus$ coloring of decided rules.
If $r\in\colminus$ then Step~(16) returns a failure and if $r$ is already
colored with $\cplus$ then $r$ is no longer considered.
Else, $r$ is colored with $\cplus$ in Step~(17).
All successors of $r$, which are now decided (Step~(19)) but not yet in the
queue (their status is false), are taken into the queue  and
their status is set to $true$ in Step~(21). 
The {\bf for} loop in Step~(18) works in $\CO{k'*k}$, hence constant wrt $n$.
Thus, Step~(14--25) are computable in constant time wrt $n$.

Analogously, Step~(26--37) consider the case for $\cminus$ coloring of decided
rules. 

By construction of the queue and by adding only rules $r$ with
$status(r)=false$, every rule
is added at most one time in the queue.
Note that after adding rules into the queue their status becomes  $true$.
Hence, the {\bf while} loop in Step~(12--38) is passed through at most $n$
times.
As a conclusion, the function \texttt{propagation} computes
$\fOPpropast{\bg{}}{C}$ in $\CO{n}$.

%%   Since we have to iterate $\OPprop{\bg{}}$ at most $n$ times, where $n=|\Pi|$
%%   and by 
%%   Condition~1 we have that
%%   $\OPpropast{\bg{}}$ is  computable in $\CO{n^2}$.
\paragraph{3:} 
  For computing $\fOPMaxGround{\bg{}}{C}$ we must compute a maximal support
  graph of $(\bg{},C)$.
  This is done by modifying the linear time algorithm of Dowling and
  Gallier~\cite{dowgal84} as follows.

  Let $V$ be the vertex set of a maximal support graph of $(\bg{},C)$ where
  $\bg{}=(\Pi,\bgezero,\bgeone)$.
  Then, $V$ can be computed with the following algorithm:

\[
  \begin{array}{l}
  {\bf function} \text{ MaxSuppGraph}(V,C,\bg{})\\
  V:=\emptyset\\
  queue:=\emptyset\\ 
  {\bf for} \text{ all }r\in \Pi\setminus\colminus\text{ \bf{do}
  }counter(r):=|\pbody{r}|\; {\bf endfor}\\ 
  {\bf for} \text{ all }r\in\Pi\setminus\colminus \text{ \bf{do} }
  \text{{\bf if} }counter(r)=0 \text{ {\bf then} }push(r,queue)\;
  {\bf endif\; endfor}\\
  {\bf while}\; queue \text{ is not empty } {\bf do}\\
  \hspace{5mm} r:=pop(queue)\\
  \hspace{5mm} {\bf for} \text{ all } (r,r')\in \bgezero\; {\bf do}\\
  \hspace{10mm} {\bf if}\; r'\not\in \colminus\; {\bf then }\\
  \hspace{15mm} counter(r'):=counter(r')-1 \\
  \hspace{15mm} {\bf if}\; counter(r')= 0\; {\bf then}\; push(r',queue)\; {\bf
                endif}\\ 
  \hspace{10mm} {\bf endif}\\
  \hspace{5mm} {\bf endfor}\\
  \hspace{5mm} V := V \cup \{r\}\\
  {\bf endwhile}
  \end{array}
\]

This algorithm computes $V$ in linear time in size of $\Pi$, since every
$r\in\Pi$ is entered at most once into the queue and the ``for loop'' in the
``while loop'' is executed at most $m$ times, where $m= \max_{r\in\Pi}
|\{(r,r')\in\bgezero | r'\in\Pi\}|$. 
Hence, $\OPMaxGround_{\bg{}}$ is computable in $\CO{n}$.

\paragraph{4:}
    By Condition~1 and~3, 
    $(\OPprop{}\OPMaxGround{})_{\bg{}}$ is computable in $\CO{n}$.
    Hence, $\PUast{\bg{}}$ is computable in $\CO{n^2}$,
    since we have to iterate $(\OPprop{}\OPMaxGround{})_{\bg{}}$ at most
    $n$ times. 

\paragraph{5:}
For computing $\fOPV{\bg{}}{C}$ we have to compute
$\fOPTast{\bg{}}{C}_\cplus$. 
This is done by a modification of the algorithm given in
Condition~3.
We start with $V=\colplus$ instead of $V=\emptyset$ and consider $\colplus$ while
initializing the counter for every rule $r\in\Pi$.  

\[
  \begin{array}{l}
  {\bf function}\; \mathcal{T_\cplus}(V,C,\bg{})\\
  V:=\colplus\\
  queue:=\emptyset\\
  {\bf for} \text{ all }r\in \Pi\setminus\colminus\text{ \bf{do}
  }counter(r):=|\pbody{r}\setminus \{\head{r'}\mid r'\in \colplus\}|\; {\bf
    endfor}\\  
  {\bf for} \text{ all }r\in\Pi\setminus\colminus \text{ \bf{do} }
  \text{{\bf if} }counter(r)=0 \text{ {\bf then} }push(r,queue)\;
  {\bf endif\; endfor}\\
  {\bf while}\; queue \text{ is not empty } {\bf do}\\
  \hspace{5mm} r:=pop(queue)\\
  \hspace{5mm} {\bf for} \text{ all } (r,r')\in \bgezero\; {\bf do}\\
  \hspace{10mm} {\bf if}\; r'\not\in \colminus\; {\bf then }\\
  \hspace{15mm} counter(r'):=counter(r')-1 \\
  \hspace{15mm} {\bf if}\; counter(r')= 0\; {\bf then}\; push(r',queue)\; {\bf
                endif}\\ 
  \hspace{10mm} {\bf endif}\\
  \hspace{5mm} {\bf endfor}\\
  \hspace{5mm} V := V \cup \{r\}\\
  {\bf endwhile}
  \end{array}
\]
Analogously to Condition~3, $\OPV{\bg{}}$ is computable in $\CO{n}$.

\paragraph{6:}
  This follows directly from Condition~1 and~5.
\paragraph{7:}
  We have   \(
  \fOPN{\bg{}}{C}= (\colplus, \Pi\setminus \colplus)
  \).
  Hence, computing $\OPN{\bg{}}$ is in $\CO{n}$.
\end{proof}
%%% Local Variables: 
%%% mode: latex
%%% TeX-master: "~/TeX/Papers/GraphLP/TPLP/Camera/paper"
%%% End: 

%% file: Proofs/LongVersion/thm_wfs_fitting_P_empty.tex
%---------------------------------------------------------------
\begin{proof}{thm:wfs:fitting:P:empty}
  Let \bg{} be the \RDGraph{} of logic program $\Pi$.
We say that $(X,Y) \subseteq (X',Y')$ if $X\subseteq X'$ and $Y\subseteq Y'$
for any $X,Y,X',Y'\subseteq \atm$.

Let $\col=\fOPpropast{\bg{}}{(\emptyset,\emptyset)}$ be a partial coloring
of \bg{}.
Furthermore, let $\fittingo{\Pi}^\omega\emptyC=( X^\omega,Y^ \omega)$.
We have to prove $(X^\omega,Y^\omega)=(X_C,Y_C)$.
According to Definition~\ref{def:C:XY} and by Theorem~\ref{thm:inductive:prop}
it can be easily seen that 
\[
(X_C,Y_C)=\bigcup_{i < \omega} (X_{C^i},Y_{C^i})
\]
where 
$C^0=(\emptyset,\emptyset)$ and 
$
C^{i+1}=\fOPprop{\bg{}}{C^i}
$.
By definition of Fitting's operator, we obtain
\[
(X^\omega,Y^\omega)=\bigcup_{i < \omega} (X^i,Y^i)
\]
where
$(X^0,Y^0)=(\emptyset,\emptyset)$,
\begin{eqnarray*}
X^i &=& X^{i-1}\cup \{\head{r} \mid r\in\Pi, \pbody{r}\subseteq X^{i-1}, \nbody{r}\subseteq
Y^{i-1}\}, \text{ and} \\
Y^i &= & Y^{i-1} \cup\{ q \mid \text{ for all } r\in \Pi,\text{ if } \head{r}=q
                       \text{, then } \\
&&                       \pbody{r}\cap Y^{i-1}\not=\emptyset \text{ or }
                       \nbody{r}\cap X^{i-1} \not=\emptyset\}.
\end{eqnarray*}

\paragraph{"$(X^\omega,Y^\omega)\subseteq (X_C,Y_C)$":}
We prove by induction over $i$ that $(X^i,Y^i)\subseteq (X_C,Y_C)$ for all $i
< \omega$ hold.

For $i=0$ we have $(X^0,Y^0)=(\emptyset,\emptyset) \subseteq (X_C,Y_C)$.

%% For $i=1$ we have
%% $X^1=\{\head{r} \mid r\in\Pi, \pbody{r}\subseteq \emptyset, \nbody{r}\subseteq
%% \emptyset\}$ and
%% $Y^1=\{q\in \atm\setminus \head{\Pi}\}$.

%% Let $\head{r}=a\in X^1$ then $r\in S(\bg{},\emptyC) \cap
%% \overline{B}(\bg{},\emptyC)$. 
%% Since $C$ is closed under $\OPprop{\bg{}}$ we have $r\in\colplus$ and thus $a\in
%% X_C$. 

%% Let $a\in Y^1$. 
%% Then $a \not\in\head{\Pi}$ and thus
%% $a\in Y_C$ by definition of $Y_C$.

Assume, $(X^k,Y^k) \subseteq (X_C,Y_C)$ holds for all $k \leq i$ for some $i<
\omega$ {\bf (IH)}.
We have to prove $(X^{i+1},Y^{i+1})\subseteq (X_C,Y_C)$.
By induction hypotheses {\bf (IH)}, we have 
\begin{eqnarray*}
X^{i+1}&=& X^i \cup\{\head{r} \mid r\in\Pi, \pbody{r}\subseteq X^i \subseteq X_C,
\nbody{r}\subseteq Y^i\subseteq Y_C\} \text{ and}\\
Y^{i+1}&=&Y^i \cup \{q \mid \text{ for all } r\in \Pi,\text{ if } \head{r}=q
                       \text{, then } \\
&&                       \pbody{r}\cap Y^i\not=\emptyset \text{ or }
                       \nbody{r}\cap X^i \not=\emptyset\}.
\end{eqnarray*}

Let be $\head{r}=a\in X^{i+1}\setminus X^i$, then we have $r\in S(\bg{},C)\cap
\overline{B}(\bg{},C)$. 
Thus, $r\in\colplus$ holds by $C$ is closed under $\OPprop{\bg{}}$ and hence,
we have $a\in X_C$.

Let be $a\in Y^{i+1}\setminus Y^i$ and let be $r\in \Pi$ such that $\head{r}=a$.
If $\pbody{r}\cap Y^i\not=\emptyset$ then $r\in \overline{S}(\bg{},C)$.
If $\nbody{r}\cap X^i \not=\emptyset$ then $r\in B(\bg{},C)$.
Thus, $a\in Y_C$ holds by $C$ is closed under $\OPprop{\bg{}}$.

Hence, we have $(X^{i+1},X^{i+1}) \subseteq (X_C,Y_C)$.
Thus, we have  $(X^\omega,Y^\omega)\subseteq (X_C,Y_C)$.

\paragraph{"$(X^\omega,Y^\omega)\supseteq (X_C,Y_C)$":}
We show by induction over $i$ that $(X_{C^i},Y_{C^i})\subseteq
(X^\omega,Y^\omega)$ holds for all $i < \omega$.
Then, we conclude $(X_C,Y_C)\subseteq (X^\omega,Y^\omega)$.

For $i=0$ we have $(X_{C^0},Y_{C^0})=(\emptyset,\atm\setminus
\head{\Pi})\subseteq (X^1,Y^1)\subseteq (X^\omega,Y^\omega)$.

%% For $i=1$ we have 
%% \begin{eqnarray*}
%%   C^1 & = & \fOPprop{\bg{}}{\emptyC}\\
%%       & = & (S(\bg{},\emptyC)\cap
%%       \overline{B}(\bg{},\emptyC),\overline{S}(\bg{},\emptyC) \cup
%%       B(\bg{},\emptyC)) \\
%%       & = & (S(\bg{},\emptyC)\cap
%%       \overline{B}(\bg{},\emptyC),\overline{S}(\bg{},\emptyC))
%% \end{eqnarray*}

%% Let $a\in X_{C^1}$ then there exists an $r\in\Pi$ such that $\head{r}=a$ and
%% $r \in S(\bg{},\emptyC)\cap \overline{B}(\bg{},\emptyC)$.
%% Thus, $\pbody{r}\subseteq  \emptyset$ and $\nbody{r}\subseteq \atm\setminus
%% \head{\Pi}$.
%% Hence, $a\in X^\omega$ because $\atm\setminus\head{\Pi} \subseteq Y^\omega$.

%% Let $a\in Y_{C^1}$.
%% Then, for all $r\in\Pi$ such that $\head{r}=a$ we have $\pbody{r}\cap
%% (\atm\setminus \head{\Pi})\not=\emptyset$.
%% Hence, $a\in Y^\omega$ by definition of $Y^\omega$.

Assume, $(X_{C^k},Y_{C^k})\subseteq (X^\omega,Y^\omega)$ holds for all $k \leq i$
for some $i < \omega$.
We have to show that $(X_{C^{i+1}},Y_{C^{i+1}})\subseteq (X^\omega,Y^\omega)$
holds where $C^{i+1}=\fOPprop{\bg{}}{C^i}$.

Let be $a\in X_{C^{i+1}}$.
Then there exists an $r\in\Pi$ such that $\head{r}=a$ and $r\in
S(\bg{},C^i)\cap \overline{B}(\bg{},C^i)$.
Hence,  we have $\pbody{r}\subseteq X_{C^i}$ and $\nbody{r}\subseteq Y_{C^i}$.
Thus, we have $a\in X^\omega$ by $(X_{C^i},Y_{C^i})\subseteq (X^\omega,Y^\omega)$.

Let be $a\in Y_{C^{i+1}}$ then for all $r\in \Pi$ such that $\head{r}=a$ we have
$r\in \fOPprop{\bg{}}{C^i}_\cminus$.
Thus, $r\in \colminus^i \cup \overline{S}(\bg{},C^i)\cup B(\bg{},C^i)$.
%% By construction of $\colminus^i$ we have then for all $r\in \Pi, \head{r}=a$
%% that $r\in \overline{S}(\bg{},C^k)$ or $r\in B(\bg{},C^k)$ for some $k \leq
%% i$.
By $C^k \sqsubseteq C^i$ we have $r\in\overline{S}(\bg{},C^i)\cup
B(\bg{},C^i)$.
If $r\in B(\bg{},C^i)$ then $\nbody{r}\cap X_{C^i}\not=\emptyset$.
If $r\in \overline{S}(\bg{},C^i)$ then $\pbody{r}\cap Y_{C^i}\not=\emptyset$.
Hence, we have $a\in Y^\omega$.

Thus, we have proven $(X^\omega,Y^\omega)\supseteq (X_C,Y_C)$.
\end{proof}
%---------------------------------------------------------------

%%% Local Variables: 
%%% mode: latex
%%% TeX-master: "~/tex/Papers/GraphLP/Grundlagen/paper"
%%% End: 

%% file: Proofs/LongVersion/thm_wfs_GUS.tex
% ---------------------------------------------------------------
 \begin{proof}{thm:wfs:GUS}
%%  A set $A\subseteq \atm$ is an unfounded set (according
%%  to~\cite{vangelder91wellfounded}) of $\Pi$ wrt $(X,Y)$ if each $a\in A$ 
%%  satisfies the following condition:
%%  For each $r\in\Pi$ with $\head{r}=a$, one of the following conditions hold:
%%  \begin{enumerate}
%%  \item $\pbody{r}\cap Y \not=\emptyset$ or\\
%%        $\nbody{r}\cap X\not=\emptyset$;
%%  \item there exists a $p\in \pbody{r}$ such that $p\in A$.
%%  \end{enumerate}
%%  The greatest unfounded set of $\Pi$ wrt $(X,Y)$, denoted $\GUS{\Pi}{X,Y}$, is
%%  the  union of all sets that are unfounded wrt $(X,Y)$.

   Let \bg{} be the \RDGraph{} of logic program $\Pi$ and $C$ be a partial
   coloring of \bg{}\ such that $\colminus \subseteq \overline{S}(\bg{},C)\cup
   B(\bg{},C)$. 
   Furthermore, let $(V,E)$ be a maximal support graph of $(\bg{},C)$ for some
   $E\subseteq \Pi\times\Pi$.

   First, we show that $\atm \setminus \head{V}$ is an unfounded set of $\Pi$
   wrt $(X_C,Y_C)$. 
   Let be $a\in \atm \setminus \head{V}$.
   For all $r\in\Pi$ such that $\head{r}=a$,
   we have to show that one of the following conditions hold:
   \begin{description}
   \item[U1:] there exists a $p\in \pbody{r}$ such that all rules, with $p$ as
         head, are in $\colminus$ or\\
         there exists a $q\in \nbody{r}$ where there exists an
         $\rp\in\colplus$ such that $\head{\rp}=q$
         (That is, we have $r\in \overline{S}(\bg{},C)\cup B(\bg{},C)$);
   \item[U2:] there exists a $p\in \pbody{r}$ such that $p\in \atm\setminus
   \head{V}$. 
   \end{description}

   If $a \in \atm\setminus \head{V}$ then either
   $a\not\in\head{\Pi}$ or there are (by
   Definition~\ref{def:acol:maxgrounded}) one of  the following cases for 
   all rules $r\in\Pi$ where $\head{r}=a$:
   \begin{description}
   \item[Case~1:] $\pbody{r}\not\subseteq \head{V}$;
   \item[Case~2:] $r\in\colminus$.
   \end{description}
   If $a\not\in\head{\Pi}$ then {\bf U1} and {\bf U2} are trivially fulfilled
   since $\{r\mid \head{r}=a\}=\emptyset$.

   {\bf Case~1: } If $\pbody{r}\not\subseteq \head{V}$ then there exists a
   $p\in\pbody{r}$ 
   such that $p\in\atm \setminus \head{V}$ and condition {\bf U2} is fulfilled.
   
   {\bf Case~2: } If $r\in\colminus$ then either $r\in\overline{S}(\bg{},C)$
   or $r\in 
   B(\bg{},C)$. 
   If $r\in\overline{S}(\bg{},C)$ then there exists an $p\in\pbody{r}$ such
   that all rules with $p$ as head are in $\colminus$.
   If $r\in B(\bg{},C)$ then there exists an $\rp\in\colplus$  such that
   $\head{\rp}\in \nbody{r}$.
   Thus, condition {\bf U1} is fulfilled.

   Thus, $\atm \setminus \head{V}$ is an unfounded set of $\Pi$ wrt
   $(X_C,Y_C)$. 

   Second, each unfounded set wrt $(X_C,Y_C)$ is in $\atm\setminus
   \head{V}$ by maximality of $(V,E)$ and by Definition~\ref{def:acol:maxgrounded}. 
   Thus, $\atm\setminus \head{V}$ is the greatest unfounded set wrt $(X_C,Y_C)$.
 \end{proof}
% ---------------------------------------------------------------

%%% Local Variables: 
%%% mode: latex
%%% TeX-master: "~/tex/Papers/GraphLP/Grundlagen/paper"
%%% End: 

%% file: Proofs/LongVersion/cor_wfs_GUS.tex
\begin{proof}{cor:wfs:GUS}
   Let \bg{} be the \RDGraph{} of logic program $\Pi$ 
   and $C$ be a partial coloring of \bg{} such that
   $\colminus\subseteq \overline{S}(\bg{},C)\cup B(\bg{},C)$. 

Let  $\fOPMaxGround{\bg{}}{C}=C'$ and let $(V,E)$ be a maximal support graph
of $(\bg{},C)$ for some $E\subseteq \Pi\times\Pi$.
Observe that
\begin{eqnarray*}
  \atm\setminus\head{\Pi\setminus \colminus'} & = &
  \atm\setminus\head{\Pi\setminus (\Pi\setminus V)}\\
  & = & \atm\setminus \head{V}.
\end{eqnarray*}
Thus, by Theorem~\ref{thm:wfs:GUS},  $(\atm\setminus
   \head{\Pi\setminus\colminus'})$ is the greatest unfounded 
   set wrt $(X_C,Y_C)$.
\end{proof}
%%% Local Variables: 
%%% mode: latex
%%% TeX-master: "~/tex/Papers/GraphLP/Grundlagen/paper"
%%% End: 

%% file: Proofs/LongVersion/thm_wfs_main.tex
\begin{proof}{thm:wfs:main}
  Let \bg{} be the \RDGraph{} of logic program $\Pi$. 
  Furthermore, let
  $\col=(\OPprop{}\OPMaxGround{})^\ast_{\bg{}}((\emptyset,\emptyset))$
  be a partial coloring.
  We have to show that $( X_C,Y_C )$ is the well-founded model of $\Pi$.

  First, we show the following Lemma:
  
  {\bf [FP]:}
  Let $C,C'$ be partial colorings of $\bg{}$ such that
  $\colplus \subseteq S(\bg{},C)\cap \overline{B}(\bg{},C)$ and
  $\colminus \subseteq \overline{S}(\bg{},C)\cup B(\bg{},C)$.
  Then, if $C'=\fOPprop{\bg{}}{C}$ then
  $\fittingo{\Pi}(X_C,Y_C)=(X_{C'},Y_{C'})$. 
  
  {\bf Proof of [FP]:}
   Let be $C'=\fOPprop{\bg{}}{C}$.   
   Then, $C'=(\colplus \cup(S(\bg{},C)\cap \overline{B}(\bg{},C)), \colminus
   \cup (\overline{S}(\bg{},C)\cup B(\bg{},C)) )$. 
   By the preconditions we have  
   $C'=(S(\bg{},C)\cap \overline{B}(\bg{},C), \overline{S}(\bg{},C)\cup
   B(\bg{},C))$. 

   We obtain
   \begin{eqnarray*}
     \fittingo{\Pi}^+(X_C,Y_C) & = & \{\head{r}\mid r\in
    \Pi,\pbody{r}\subseteq X_C, \nbody{r}\subseteq Y_C\} \\
     & = & \{\head{r}\mid r\in S(\bg{},C)\cap \overline{B}(\bg{},C)\} \\
     & = & \{\head{r}\mid r\in \colplus'\}\\
     & = & X_{C'},\text{ and }
   \end{eqnarray*}
   \begin{eqnarray*}
      \fittingo{\Pi}^-(X_C,Y_C) & = & 
           \{q \mid \text{ if } \head{r}=q \text{ then } \pbody{r}\cap
           Y_C\not=\emptyset \text{ or } \nbody{r}\cap X_C\not=\emptyset\}\\
      & = & \{q \mid \text{ if } \head{r}=q \text{ then } r\in
           \overline{S}(\bg{},C)\cup B(\bg{},C)\} \\
       & = & \{q \mid \text{ if } \head{r}=q \text{ then } r\in \colminus'\}\\
       & = & Y_{C'}.   
   \end{eqnarray*}
   Hence,  $\fittingo{\Pi}(X_C,Y_C)=(X_{C'},Y_{C'})$.

   Second, we show that $( X_C,Y_C )$ is the well-founded model of $\Pi$.
   For this, we give a definition of the well-founded model~\cite{ross92}.
The mapping $\mathcal{U}_\Pi$, which assigns false to every atom in an
   unfounded set, is defined as follows:
\[
\mathcal{U}_\Pi\langle X,Y \rangle =\langle X',Y'\rangle
\] 
where for all atoms $A$ we have:
\begin{description}
\item[(i)]  $A\in X'$ if $A\in X$,
\item[(ii)] $A\in Y'$ if $A$ is in the greatest unfounded set (wrt $\Pi$
  and $\langle X,Y \rangle$),
\item[(iii)] $A$ is undefined  otherwise.
\end{description}

Then, the well-founded model of $\Pi$, $\mathcal{W}_\Pi^\omega\langle \emptyset,\emptyset\rangle$, is defined as
follows:
  \begin{eqnarray*}
    \mathcal{W}_\Pi^0\langle \emptyset,\emptyset\rangle & = &\langle \emptyset,\emptyset\rangle\\
    \mathcal{W}_\Pi^{i+1}\langle \emptyset,\emptyset\rangle & = &
    \Phi_\Pi(\mathcal{U}_\Pi(\mathcal{W}_\Pi^i\langle
    \emptyset,\emptyset\rangle))\\ 
%%     \mathcal{W}_\Pi^{i+1}\langle \emptyset,\emptyset\rangle & = &
%%    \Phi_\Pi(\mathcal{W}_\Pi^i\langle \emptyset,\emptyset\rangle)\cup 
%%     \mathcal{U}_\Pi(\mathcal{W}_\Pi^i\langle \emptyset,\emptyset\rangle)\\ 
   \mathcal{W}_\Pi^\omega\langle \emptyset,\emptyset\rangle & = & \cup_{i < \omega} \mathcal{W}_\Pi^i \langle \emptyset,\emptyset\rangle,
  \end{eqnarray*}
where $\Phi_\Pi$ denotes Fitting's operator.

Since all atoms $\atm\setminus\head{\Pi}$ are false in the well-founded model,
we have
$\mathcal{W}_\Pi^\omega\langle \emptyset,\emptyset\rangle=
\mathcal{W}_\Pi^\omega\langle \emptyset,\atm\setminus\head{\Pi}\rangle$.
{\bf [FP]} shows the direct correspondence between greatest
unfounded sets and the $\OPMaxGround_{\bg{}}$ operator.
Corollary~\ref{cor:wfs:GUS} shows the direct correspondence between Fitting's
operator and the propagation operator $\OPprop{\bg{}}$.
By induction one can show that 
$( X_C,Y_C )$ is the well-founded model of $\Pi$.
\end{proof}
%%% Local Variables: 
%%% mode: latex
%%% TeX-master: "~/tex/Papers/GraphLP/Grundlagen/paper"
%%% End: 

%% file: Proofs/LongVersion/thm_AS_AC.tex
\begin{proof}{thm:AS:AC}
This follows directly from the definition of $\AS{\Pi}{C}$ and $\AC{\Pi}{C}$.
\end{proof}
%%% Local Variables: 
%%% mode: latex
%%% TeX-master: "~/tex/Papers/GraphLP/Grundlagen/paper"
%%% End: 

%% file: Proofs/LongVersion/thm_as_consGR.tex
\begin{proof}{thm:as:consGR}
 Let $X$ be an answer set and let $\Pi$ be a logic program.
 The definition of answer sets states
 \begin{equation}
   \label{eq:proof:gras:AS1}
 X \text{ is an answer set of }\Pi \text{ iff } \Cn{\Pi^X}=X.
 \end{equation}
 We know that for positive logic programs $\Pi^X$ and $(\GR{\Pi}{X})^\emptyset$
 we have: 
 \begin{equation}
   \label{eq:proof:gras:T:pi}
 \bigcup_{i \geq 0}\TiO{i}{\Pi^X}(\emptyset) = \Cn{\Pi^X}
 \end{equation}
 and
 \begin{equation}
   \label{eq:proof:gras:T:GR}
 \bigcup_{i \geq 0}\TiO{i}{(\GR{\Pi}{X})^\emptyset}(\emptyset) = \Cn{(\GR{\Pi}{X})^\emptyset}.
 \end{equation}

 \paragraph{''$\Rightarrow$''}
 Now let $X$ be an answer set.
 Then we have
 \begin{equation}
   \label{eq:proof:gras:help1}
   \TiO{i}{\Pi^X}(\emptyset) \subseteq X.
 \end{equation}
 for all $i \geq 0$ because of~(\ref{eq:proof:gras:AS1})
 and~(\ref{eq:proof:gras:T:pi}).  
 With (\ref{eq:proof:gras:AS1}) it is sufficient to show
 $\Cn{(\GR{\Pi}{X})^\emptyset}= \Cn{\Pi^X}$.

 We want to prove the equation 
 \[\bigcup_{i \geq 0}\TiO{i}{\Pi^X}(\emptyset)=
 \bigcup_{i \geq 0}\TiO{i}{(\GR{\Pi}{X})^\emptyset}(\emptyset)\]
 by induction over $i$.
 With this equation and using (\ref{eq:proof:gras:T:pi}) and
 (\ref{eq:proof:gras:T:GR}) the statement $X=\Cn{(\GR{\Pi}{X})^\emptyset}$ is proven.
 For $i=0$ we have:
 \[\Ti{0}{\Pi^X}{\emptyset}=\emptyset = \Ti{0}{(\GR{\Pi}{X})^\emptyset}{\emptyset}.\]
 Now let $i=1$.
 Then we have:
\[
\begin{array}{lllr}
 \Ti{1}{\Pi^X}{\emptyset} 
 &= & \T{\Pi^X}{\emptyset} &  \\
 &= & \{\head{\r}: \r \in \Pi^X,\body{\r}\subseteq\emptyset\} &
     \text{(by (\ref{eq:T}))} \\
 &= & \{\head{\r}:\r\in \Pi, & \\
&&\nbody{\r}\cap X=\emptyset,\pbody{\r}\subseteq
   \emptyset\} & \text{(by (\ref{eq:GLreduct}))} \\ 
 &= & \{\head{\r}:\r\in(\GR{\Pi}{X})^\emptyset,\body{\r}=\emptyset\} &
     (\text{by } (\pbody{\r}=\emptyset))\\
 &= & \Ti{1}{(\GR{\Pi}{X})^\emptyset}{\emptyset} & \text{(by (\ref{eq:T}))}
 \end{array}
\]
 Assume that we have the induction hypothesis $(IH)$
 \[\Ti{i}{\Pi^X}{\emptyset} =
 \Ti{i}{(\GR{\Pi}{X})^\emptyset}{\emptyset}. \]
% as assumption of our induction (IA).
% 
 For $i+1$ we have:
 \[\begin{array}{lllr}
\Ti{i+1}{\Pi^X}{\emptyset} 
& = & \T{\Pi^X}{\Ti{i}{\Pi^X}{\emptyset}} &  \\
& = & \{\head{\r}: \r \in \Pi^X,
     \body{\r}\subseteq \Ti{i}{\Pi^X}{\emptyset}\} & 
     \text{(by (\ref{eq:T}))} \\
& = & \{\head{\r}: \r \in \Pi, &\\
&& \nbody{\r}\cap X=\emptyset, \pbody{\r}\subseteq \Ti{i}{\Pi^X}{\emptyset}\} & 
     \text{(by (\ref{eq:GLreduct}))}
 \end{array}\]
 and
 \[\begin{array}{lllr}
\Ti{i+1}{(\GR{\Pi}{X})^\emptyset}{\emptyset} 
& = & \T{(\GR{\Pi}{X})^\emptyset}{\Ti{i}{(\GR{\Pi}{X})^\emptyset}{\emptyset}} & \\  
& = & \{\head{\r}: \r \in
 (\GR{\Pi}{X})^\emptyset,\body{\r}\subseteq\Ti{i}{(\GR{\Pi}{X})^\emptyset}{\emptyset}\} & \text{(by (\ref{eq:T}))} \\ 
& = & \{\head{\r}: \r \in \Pi,\nbody{\r}\cap X=\emptyset, &\\ 
&&    \pbody{\r}\subseteq X,\pbody{\r}\subseteq\Ti{i}{(\GR{\Pi}{X})^\emptyset}{\emptyset}\}
 &   \\ 
& = & \{\head{\r}: \r \in \Pi,\nbody{\r}\cap X=\emptyset, &\\
&&    \pbody{\r}\subseteq X,\pbody{\r}\subseteq\Ti{i}{\Pi^X}{\emptyset}\} & 
     \text{(by (IH))}\\ 
& = & \{\head{\r}: \r \in \Pi,\nbody{\r}\cap X=\emptyset, &\\
&&\pbody{\r}\subseteq\Ti{i}{\Pi^X}{\emptyset}\} & 
     \text{(by (\ref{eq:proof:gras:help1}))}
 \end{array}\]
 Thus we have proven $X=\Cn{(\GR{\Pi}{X})^\emptyset}$.

 \paragraph{''$\Leftarrow$''}

 Now let be $X=\Cn{(\GR{\Pi}{X})^\emptyset}$.
 We have to show that $X$ is an answer set.
 Because of (\ref{eq:proof:gras:T:GR}) we have 
 \begin{equation}
   \label{eq:proof:gras:help2}
   \TiO{i}{(\GR{\Pi}{X})^\emptyset}(\emptyset) \subseteq X
 \end{equation}
 for all $i \geq 0$.
 Using this equation we can show analogously that
 \(\Ti{i}{\Pi^X}{\emptyset} = \Ti{i}{(\GR{\Pi}{X})^\emptyset}{\emptyset}\)
 holds for all $i \geq 0$.
 Therefore, we have 
 \(\Cn{\Pi^X}=\Cn{(\GR{\Pi}{X})^\emptyset}\)
 because of (\ref{eq:proof:gras:T:GR}) and (\ref{eq:proof:gras:T:pi}).
 Finally, (\ref{eq:proof:gras:AS1}) gives us that $X$ is an answer set.
\end{proof}

%%% Local Variables: 
%%% mode: latex
%%% TeX-master: t
%%% End: 

%% file: Proofs/LongVersion/thm_gr_grounded.tex
%------------------------------------------------------
\begin{proof}{thm:gr:grounded}
Let $X$ be an answer set of logic program $\Pi$.

We have to show, that there exists an enumeration $\langle r_i \rangle_{i \in
  I}$ of \GR{\Pi}{X}, such that for all $i\in I$ we have
$\pbody{r_i} \subseteq \head{\{r_j \mid j<i\}}$.

 Since $X$ is an answer set, we know from Theorem~\ref{thm:as:consGR}: 
 $X=\Cn{(\GR{\Pi}{X})^\emptyset}$.
 Furthermore we have 
 \(\Cn{(\GR{\Pi}{X})^\emptyset}=\bigcup_{i \ge 0}
 \Ti{i}{(\GR{\Pi}{X})^\emptyset}{\emptyset}.\) 
Thus, we have that
\(
X = \bigcup_{i \ge 0} \Ti{i}{(\GR{\Pi}{X})^\emptyset}{\emptyset}.
\)
For this reason, we find an enumeration $\langle x_j \rangle_{j \in J}$ of $X$
such that $x_i \in \Ti{k}{(\GR{\Pi}{X})^\emptyset}{\emptyset}$ and
$x_j\in\Ti{l}{(\GR{\Pi}{X})^\emptyset}{\emptyset}$ hold for $i <j$ and some minimal
$k$ and $l$ such that $k < l$.
This enumeration  $\langle x_j \rangle_{j \in J}$ of $X$ gives us an
enumeration $\langle r_i \rangle_{i \in
  I}$ of $\GR{\Pi}{X}$ such that for all $i\in I$ we have
$\pbody{r_i} \subseteq \head{\{r_j \mid j<i\}}$.
%  Now the immediate consequence operator gives the desired enumeration
%  of $\GR{\Pi}{X}^+$ and accordingly for $\GR{\Pi}{X}$. 
% % 
%  It can be easily seen that
%  \[\GR{\Pi}{X}^+=\{\r : \head{\r} \in \bigcup_{i \ge 0}
%  \Ti{i}{\GR{\Pi}{X}^+}{\emptyset}\}\] 
% % 
%  Therefore the rule $\r$, in particular $\head{\r}$, is generated for a
%  certain $i\geq 0$ by way of iterated application of the immediate
%  consequence operator.
% % 
% %  Thus, there exists an enumeration $\langle r_i \rangle_{i \in I}$ of
% %  $\GR{\Pi}{X}^+$, s.t. 
% %  \[j<i \Rightarrow \head{r_i}\in \Ti{k}{\GR{\Pi}{X}^+}{\emptyset}
% %  \text{ and }        \head{r_j}\in \Ti{l}{\GR{\Pi}{X}^+}{\emptyset}
% %  \text{ for } k \not<l \]
% %  for all $i,j \in I$.
% %  %
%  Because of definition of $\TiO{i}{\GR{\Pi}{X}^+}$ we
%  have 
%  \[\Ti{0}{\GR{\Pi}{X}^+}{\emptyset}=\emptyset\]
%  for $i\geq 0$ and
%  \[\begin{array}{ll}
%    & \Ti{i+1}{\GR{\Pi}{X}^+}{\emptyset} \\
%  = & \T{\GR{\Pi}{X}^+}{\Ti{i}{\GR{\Pi}{X}^+}{\emptyset}}\\
%  = & \{\head{\r}: \r \in \GR{\Pi}{X}^+, \body{\r}\in
%  \Ti{i}{\GR{\Pi}{X}^+}{\emptyset} \\  
%  = & \{\head{\r}: \r \in \GR{\Pi}{X}^+, \pbody{\r}\in
%  \Ti{i}{\GR{\Pi}{X}^+}{\emptyset}\}. 
%  \end{array}\]
% % 
%  The last step in the equation is justified, since $\r$ is a positive
%  rule. 
% % 
%  In particular, there exists an $i\in I$ with $\r=r_i$ for $\r^+ \in
%  \GR{\Pi}{X}^+$ and we have  
%  \[\pbody{r_i}\subseteq \head{\{r_j : j<i\}}\]
% % 
%  after constructing the enumeration with iterated application of the
%  immediate consequence operator.
\end{proof}
%------------------------------------------------------
%%% Local Variables: 
%%% mode: latex
%%% TeX-master: t
%%% End: 

%% file: Proofs/LongVersion/thm_X_headcolplus.tex
\begin{proof}{thm:X:headcolplus}
Let $\bg{}$ be the \RDGraph\ of logic program $\Pi$, $\col$ be a partial
coloring of $\bg{}$ and
let $X\in \AS{\Pi}{\col}$ be an answer set of $\Pi$.
By definition of $X$ we have $\colplus\subseteq\GR{\Pi}{X}$, $\colminus\cap
\GR{\Pi}{X}=\emptyset$ and $X=\Cn{(\GR{\Pi}{X})^\emptyset}=\bigcup_{i <
  \omega} \Ti{i}{(\GR{\Pi}{X})^\emptyset}{\emptyset}$.  

By induction we show that for all $\r\in\GR{\Pi}{X}$ we have
  $\head{\r}\in \bigcup_{i < \omega}
  \Ti{i}{(\GR{\Pi}{X})^\emptyset}{\emptyset}$. 
Then by $\head{\colplus}\subseteq \head{\GR{\Pi}{X}}$ we have
  $\head{\colplus}\subseteq X = \bigcup_{i <
  \omega} \Ti{i}{(\GR{\Pi}{X})^\emptyset}{\emptyset}$.

By Theorem~\ref{thm:gr:grounded} we have an enumeration $\langle r_i
  \rangle_{i \in I}$  of $\GR{\Pi}{X}$ such
  that  for all $i \in I$ we have $\pbody{r_i} \subseteq \head{\{r_j \mid j <
  i\}}$.
Let be $I=\{0,\ldots,m\}$ for some $m < \omega$.
Clearly, we have $\pbody{\r_0}=\emptyset \subseteq \bigcup_{i < \omega}
  \Ti{i}{(\GR{\Pi}{X})^\emptyset}{\emptyset}$. 

Let be $\r_k\in \GR{\Pi}{X}$ for $k<m$ and
  $\head{\{r_0,\ldots,r_{k-1}\}}\subseteq\bigcup_{0 \leq i  \leq l} 
  \Ti{i}{(\GR{\Pi}{X})^\emptyset}{\emptyset} $ for some $l<\omega$.
We have to show, that   
$\head{r_k}\subseteq\bigcup_{0 \leq i  \leq l+1} 
  \Ti{i}{(\GR{\Pi}{X})^\emptyset}{\emptyset}$ holds.
Because we have an enumeration of $\GR{\Pi}{X}$ satisfying
  Theorem~\ref{thm:gr:grounded} we have
\(
\pbody{r_k}\subseteq \head{\{r_0,\ldots,r_{k-1}\}}.
\)
By Equation~\ref{eq:T} we have 
\(
\head{r_k} \subseteq \bigcup_{0 \leq i  \leq l+1} 
  \Ti{i}{(\GR{\Pi}{X})^\emptyset}{\emptyset}.
\)
Thus, we have $\head{\colplus}\subseteq \head{\GR{\Pi}{X}}\subseteq \bigcup_{i < \omega} 
  \Ti{i}{(\GR{\Pi}{X})^\emptyset}{\emptyset}=X$.

Assume  \col\ is admissible, then $\colplus=\GR{\Pi}{X}$.
It remains to show that $X\subseteq \head{\colplus}$.
Let $p\in X$ be some atom.
By $X=\Cn{(\GR{\Pi}{X})^\emptyset}$ there must exists an $\r\in
\GR{\Pi}{X}$  such that $p=\head{\r}$.
By $\colplus=\GR{\Pi}{X}$ we have $\r\in \colplus$ and thus we have $p\in
\head{\colplus}$. 
\end{proof}
%%% Local Variables: 
%%% mode: latex
%%% TeX-master: t
%%% End: 

%% file: Proofs/LongVersion/cor_X_headcolplus.tex
\begin{proof}{cor:X:headcolplus}
Let $\bg{}$ be the \RDGraph\ of logic program $\Pi$ and $\col$ be a partial
coloring of $\bg{}$.
Furthermore let $X\in \AS{\Pi}{\col}$.

Let $\col'$ be a total coloring such that $\col \sqsubseteq \col'$,
$\col'_\cplus=\GR{\Pi}{X}$, and $\col'_\cminus\cap\GR{\Pi}{X}=\emptyset$ hold.
Clearly, such $\col'$ must exists because we have an answer set $X$ such that
$\{X\}=\AS{\Pi}{\col'}$.
By Theorem~\ref{thm:X:headcolplus}  we have $\head{\col'_\cplus}=X$.

Let be $p\in \atm$ such that $\{\r\in \Pi \mid \head{\r}=p\}\subseteq
\colminus\subseteq \col'_\cminus$.
Then, $\{\r\in \Pi \mid \head{\r}=p\}\cap \col'_\cplus =\emptyset$ and thus
we have $\head{\r}=p \not\in X$ for all such $r\in \Pi$.
\end{proof}
%%% Local Variables: 
%%% mode: latex
%%% TeX-master: t
%%% End: 

%% file: Proofs/LongVersion/thm_sequence_P_blocked.tex
\begin{proof}{thm:sequence:P:blocked}

  Let $\bg{}=(\Pi,\bgezero,\bgeone)$ be the \RDGraph{} of logic program $\Pi$,
  $X$ be an answer set 
  of $\Pi$
  and $\col$ be a partial coloring of \bg{}
  such that $\colplus=\GR{\Pi}{X}$ and $\colminus=\emptyset$.
  Furthermore, let $\col^X$ be a total coloring of $\bg{}$ such that $\{X\}=
  \AS{\Pi}{\col^X}$. 

We have to show that  $\fOPprop{\bg{}}{\col}=(\colplus,\colminus')$ holds
  where  $B(\bg{},\col^X) \subseteq \colminus'$.

We have
\(
\fOPprop{\bg{}}{C}=(\colplus \cup (S(\bg{},C)\cap \overline{B}(\bg{},C)),
                    \colminus \cup \overline{S}(\bg{},C) \cup B(\bg{},C))
\).
Thus we have to show:
\begin{enumerate}
\item $\colplus =\colplus \cup (S(\bg{},C)\cap \overline{B}(\bg{},C))$ and
\item $B(\bg{},\col^X) \subseteq \colminus \cup \overline{S}(\bg{},C) \cup
  B(\bg{},C)$.
\end{enumerate}
\paragraph{1:}
By Theorem~\ref{thm:col:gr} we have $S(\bg{},C)\cap
\overline{B}(\bg{},C)\subseteq \GR{\Pi}{X}=\colplus$.
\paragraph{2:}
It remains to show that $B(\bg{},\col^X) \subseteq B(\bg{},C)$.
Let be $\r\in B(\bg{},\col^X)$.
Then there exists an $\rp\in \colplus^X$ such that $(\rp,\r)\in \bgeone$.
By $\colplus^X=\GR{\Pi}{X}=\colplus$ we have $\r\in B(\bg{},C)$.
\end{proof}
%%% Local Variables: 
%%% mode: latex
%%% TeX-master: "~/tex/Papers/GraphLP/Grundlagen/paper"
%%% End: 

%% file: Proofs/LongVersion/cor_I_monoton.tex
\begin{proof}{cor:I:monoton}
   Let \bg{} be the \RDGraph{} of logic program $\Pi$ 
   and $C,C'$ be a partial colorings of \bg{} such that $C\sqsubseteq C'$. 

Let be $a\in X_C$, then there exists an $r\in\colplus$ such that $\head{r}=a$.
By $\colplus\subseteq \colplus'$ we have $a\in X_{C'}$.

Let be $a\in Y_C$, then for all $\r\in\Pi$ such that $\head{r}=a$ we have
$r\in\colminus$.
By $\colminus \subseteq \colminus'$ we have $a\in Y_{C'}$.
\end{proof}
%%% Local Variables: 
%%% mode: latex
%%% TeX-master: "~/tex/Papers/GraphLP/Grundlagen/paper"
%%% End: 

%% file: Proofs/LongVersion/thm_inductive_prop.tex
\begin{proof}{thm:inductive:prop}
  Let \bg{} be the \RDGraph{} of logic program $\Pi$ and \col{} be a partial
  coloring of \bg{}.

%% \com{insert $AS(C)=AS(P(C))$}
%% By $\col \sqsubseteq P(C)$ we have $\AS{\Pi}{C} \supseteq
%% \AS{\Pi}{P(C)}$. 
%% Thus, it remains to show that $\AS{\Pi}{C} \subseteq
%% \AS{\Pi}{P(C)}$,
%% which we prove  by induction over $i$.
%% For $i=0$ we have $P^0(C)=C$ and thus $\AS{\Pi}{C} \subseteq
%% \AS{\Pi}{P^0(C)}$.
%% Assume that $\AS{\Pi}{C} \subseteq \AS{\Pi}{P^i(C)}$ for some $i < \omega$.
%% By $\AS{\Pi}{C'} \subseteq
%% \AS{\Pi}{\fOPprop{\bg{}}{C'}}$ for a partial coloring $C'$,
%% we obtain $\AS{\Pi}{P^i(C)} \subseteq
%% \AS{\Pi}{\fOPprop{\bg{}}{P^i(C)}}=\AS{\Pi}{P^{i+1}(C)}$.

\paragraph{1:}
The existence of 
\(
P(\col) = \bigsqcup_{i<\omega} P^i(\col)
\)
follows from Theorem~\ref{thm:P:reflexivity} and from property~3 in this
theorem by induction over $i$.

\paragraph{2:}
By definition of $P(C)$ and Definition~\ref{def:pre:algo:conditions}, $P(C)$
is a partial coloring, because $P(C)$ only operates on partial colorings.

\paragraph{3:}
By Theorem~\ref{thm:AS:AC} it remains to show that $\AS{\Pi}{C}=\AS{\Pi}{P(C)}$.
By $C\sqsubseteq P^i(C)$ we have, if $X\in \AS{\Pi}{P^i(C)}$ then $X\in
\AS{\Pi}{C}$ for all $i< \omega$.
Furthermore, if $X\in\AS{\Pi}{P(C)}$ then $X\in \AS{\Pi}{P^i(C)}$ by
$P^i(C)\sqsubseteq P(C)$ for all $i < w$.
 
Let $X\in \AS{\Pi}{C}$.
We prove by induction over $i$ that 
$X\in\AS{\Pi}{P^i(\col)}$ for all $i < \omega$ and thus $X\in \AS{\Pi}{P(C)}$.
 
For $i=0$ we have $P^0(\col)=\col$ and thus by definition of $\col$,
$X\in\AS{\Pi}{P^0(\col)}$.

Assume, 
$X\in\AS{\Pi}{P^k(\col)}$ for all $0\leq k \leq i$ for some $i < \omega$.
We have to show that $X\in\AS{\Pi}{P^{i+1}(\col)}$.

% By Theorem~\ref{thm:P:reflexivity} we have $P^{i+1}(C)$ exists and thus
% $P^{i+1}(C)$ is a partial coloring.
 Abbreviatory we write $\col'$ instead of $P^i(\col)$.

We have $\col'_\cplus \subseteq \GR{\Pi}{X}$
and $\colminus' \cap \GR{\Pi}{X} =\emptyset$.
By
\begin{eqnarray*}
P^{i+1}(\col) & = & \fOPprop{\bg{}}{\col'}\\
              & = & ( \col'_\cplus \cup (S(\bg{},\col') \cap
              \overline{B}(\bg{},\col') ),
                      \col'_\cminus \cup \overline{S}(\bg{},\col') \cup
              B(\bg{},\col')),
\end{eqnarray*}
it remains to show, that 
\(
S(\bg{},\col') \cap
              \overline{B}(\bg{},\col') \subseteq \GR{\Pi}{X}
\) and 
\(
(\overline{S}(\bg{},\col') \cup B(\bg{},\col')) \cap \GR{\Pi}{X}=\emptyset.
\) 
But this holds by Theorem~\ref{thm:col:gr}.
Thus, $X\in\AS{\Pi}{P^{i+1}(\col)}$.
\paragraph{4:}
$\col=P^0(C)\sqsubseteq P(C)$ holds by definition of $P^0(C)$ and $P^{i+1}(C)$
for all $i < \omega$.
\paragraph{5:}
We have to show, that $\fOPprop{\bg{}}{P(C)}=P(C)$.

By finiteness there exists an $n < \omega$ such that $P(C)=P^n(C)$ and
$P^n(C)=P^{n+1}(C)$.
Thus, we have to show  $\fOPprop{\bg{}}{P^n(C)}=P(C)$.

By definition of $P(C)$ we have $\fOPprop{\bg{}}{P^n(C)}=P^{n+1}(C)
\sqsubseteq P(C)$. 
It remains to show, that $P(C)=P^n(C) \sqsubseteq\fOPprop{\bg{}}{P^n(C)} $.
But this holds by definition of $\OPprop{\bg{}}$.
Thus, $\fOPprop{\bg{}}{P(C)}=P(C)$.
% {\bf Case 1:} Let $\r\in P^n(C)_\cplus$.
% By definition of $P^n(C)$ there exists an $i < n$ such that 
% $\r\in P^{i+1}(C)_\cplus$ and $\r\in P^i(C)_\cplus \cup S(\bg{},P^i(C))\cap
% \overline{B}(\bg{},P^i(C))$.
% By $P^i(C) \sqsubseteq P^n(C)$ we have then 
% $\r\in P^n(C)_\cplus \cup S(\bg{},P^n(C))\cap
% \overline{B}(\bg{},P^n(C))$ and thus $\r\in \fOPprop{\bg{}}{P^n(C)}_\cplus$.

% {\bf Case 2:} Let $\r\in P^n(C)_\cminus$.
% Analogous to case~1 we have $\r\in \fOPprop{\bg{}}{P^n(C)}_\cminus$.

\paragraph{6:}
We have to show that $P(C)$ is the $\sqsubseteq$-smallest partial coloring
  closed under  $\OPprop{\bg{}}$.
Assume there exists a $Q(C)\not=P(C)$ such that $Q(C) \sqsubseteq P(C)$ and
  $Q(C)$   is a partial coloring closed under $\OPprop{\bg{}}$.

Given $Q^0(C)=C=P^0(C)$, there must exists a (minimal) $i < \omega $ such that
\(
Q^{i+1}(C) \not=P^{i+1}(C)
\)
and
\(
Q^j(C) = P^j(C)
\)
for all $j\leq i$.
But then
\[
Q^{i+1}(C)= \fOPprop{\bg{}}{Q^i(C)}  \not=
              \fOPprop{\bg{}}{P^i(C)} = P^{i+1}(C).
\]
By 
\(
Q^i(C) = P^i(C)
\),
we have
\[
\fOPprop{\bg{}}{Q^i(C)} = \fOPprop{\bg{}}{P^i(C)}
\]
and thus
\(
Q^{i+1}(C) = P^{i+1}(C)
\)
by definition of $\OPprop{\bg{}}$.
This is a contradiction.
For this reason, $P(C)$ is the $\sqsubseteq$-smallest partial coloring
  closed under  $\OPprop{\bg{}}$.
\paragraph{7:}
This follows directly from conditions~4--6 and by definition of
$\fOPpropast{\bg{}}{\col}$. 
\end{proof}
%%% Local Variables: 
%%% mode: latex
%%% TeX-master: "~/tex/Papers/Preference/Acolorings/paper"
%%% End: 

%% file: Proofs/LongVersion/thm_inductive_propT.tex
\begin{proof}{thm:inductive:propT}
  Let \bg{} be the \RDGraph{} of logic program $\Pi$ and \col{} be a partial
  coloring of \bg{}.

\paragraph{1:}
We prove by induction over $i$ that $T^i(\col)$ is a partial coloring for all
$i < \omega$.
Then, $T(\col)$ is a partial coloring.

For $i=0$ we have $T^0(\col)=\col$ is a partial coloring.
Assume that $T^k(\col)$ is a partial coloring for all $0\leq k \leq i$ for
some $i < \omega$. 
We have to show that $T^{i+1}(\col)$ is  a partial coloring.
We have
\begin{eqnarray*}
  T^{i+1}(\col) & = & \mathcal{T}_{\bg{}}(T^i(\col)) \\
                & = & ( T^i(C)_\cplus \cup (S(\bg{},T^i(\col)) \setminus
                T^i(\col)_\cminus), T^i(C)_\cminus).
\end{eqnarray*}
But this is clearly a partial coloring because $T^i(\col)$ is a partial
coloring. 
\paragraph{2-5:}
Hold analogous to Theorem~\ref{thm:inductive:prop}.
\end{proof}
%%% Local Variables: 
%%% mode: latex
%%% TeX-master: "~/tex/Papers/GraphLP/Grundlagen/paper"
%%% End: 

%% file: Proofs/LongVersion/thm_inductive_propPU.tex
\begin{proof}{thm:inductive:propPU}
  Let \bg{} be the \RDGraph{} of logic program $\Pi$ and \col{} be a partial
  coloring of \bg{}.

%Furthermore, let $\AS{\Pi}{\col}\not=\emptyset$.

\paragraph{1:}
The existence of 
\(
PU(\col) = \bigcup_{i<\omega} PU^i(\col)
\)
follows inductively by
theorems~\ref{thm:P:reflexivity} and~\ref{thm:max:supported:graph:existence}
(existence of 
maximal support graphs) and by property~3 in this theorem.
\paragraph{2:}
This could be proven analogous to Theorem~\ref{thm:inductive:prop} by the
existence of $\OPMaxGround_{\bg{}}$ in each inductive step for defining $PU(C)$.
\paragraph{3:}
By Theorem~\ref{thm:AS:AC} it remains to show that $\AS{\Pi}{C}=\AS{\Pi}{PU(C)}$.
By $PU(C)\sqsupseteq PU^i(C) \sqsupseteq C$ for all $i < \omega$ we have if
$X\in \AS{\Pi}{PU(C)}$ then $X\in \AS{\Pi}{PU^i(C)}$ and then $X\in
\AS{\Pi}{C}$ for all $i< \omega$.

Let $X\in \AS{\Pi}{C}$.
We prove by induction over $i$ that $X\in \AS{\Pi}{PU^i(\col)}$ for all $i <
\omega$. 
Then, $X\in \AS{\Pi}{PU(C)}$.

For $i=0$ we have $PU^0(C)=C$ and thus
$X\in \AS{\Pi}{PU^0(\col)}$.
Assume,  $X\in \AS{\Pi}{PU^k(\col)}$ for all $0\leq k \leq i$ for some $i <
\omega$. 
Then, $\GR{\Pi}{X}\supseteq PU^i(C)_\cplus$
and $\GR{\Pi}{X}\cap PU^i(C)_\cminus =\emptyset$.
Now, we prove that 
$X\in\AS{\Pi}{PU^{i+1}(\col)}$.
%% \subparagraph{Show~(a):}
%% We have
%% \begin{eqnarray*}
%%   PU^{i+1}(\col) & = & 
%%    PU^i(\col)
%%   \sqcup
%%   \fOPprop{\bg{}}{PU^i(\col)}
%%   \sqcup
%%   \fOPMaxGround{\bg{}}{PU^i(\col)}\\
%%   &=&
%%   \left(
%%    PU^i(C)_\cplus \cup (S(\bg{},PU^i(\col))\cap
%%    \overline{B}(\bg{},PU^i(\col))),  
%%    PU^i(C)_\cminus \cup \overline{S}(\bg{},PU^i(\col))\cup
%%    B(\bg{},PU^i(\col))\cup (\Pi\setminus V)  
%%   \right)
%% \end{eqnarray*}
%% where $(V,E)$ is a maximal support graph of $(\bg{},PU^i(C))$ for some $E
%% \subseteq (\Pi\times \Pi)$.

%% Analogous to Theorem~\ref{thm:inductive:prop} condition~1 we have 
%% \[
%% \left(PU^i(C)_\cplus \cup (S(\bg{},PU^i(\col))\cap
%%    \overline{B}(\bg{},PU^i(\col))) \right) \cap \left(  PU^i(C)_\cminus \cup \overline{S}(\bg{},PU^i(\col))\cup
%%    B(\bg{},PU^i(\col))\right) =\emptyset.
%% \]
%% Thus, it remains to show that
%% \[
%% \left(PU^i(C)_\cplus \cup (S(\bg{},PU^i(\col))\cap
%%    \overline{B}(\bg{},PU^i(\col))) \right)\cap (\Pi \setminus V) =\emptyset.
%% \]

%% {\bf Case 1:} If $\r\in PU^i(C)_\cplus$ then $\r\in V$ by definition of $(V,E)$.
%% {\bf Case 2:} If $\r\in S(\bg{},PU^i(\col))\cap
%%    \overline{B}(\bg{},PU^i(\col))$ then $\r\in V$ by definition of
%% $(V,E)$ and by $\r\not\in PU^i(C)_\cminus$ because 
%% $PU^i(C)_\cminus \cap (S(\bg{},PU^i(\col))\cap
%%    \overline{B}(\bg{},PU^i(\col))) =\emptyset$. 

%% Thus, in both cases $\r\not\in (\Pi\setminus V)$.

We have to show that 
\begin{description}
\item[(a)] $\GR{\Pi}{X} \subseteq PU^{i+1}(C)_\cplus$ and
\item[(b)] $\GR{\Pi}{X} \cap PU^{i+1}(C)_\cminus =\emptyset$.
\end{description}
{\bf (a):}
We have
$PU^{i+1}(C)_\cplus=\fOPMaxGround{\bg{}}{\fOPprop{\bg{}}{PU^i(C)}}_\cplus$.
By Theorem~\ref{thm:inductive:prop} we have $\GR{\Pi}{X}\subseteq
{\fOPprop{\bg{}}{PU^i(C)}}_\cplus=\fOPMaxGround{\bg{}}{\fOPprop{\bg{}}{PU^i(C)}}_\cplus$. 

{\bf (b):}
We have 
\[
PU^{i+1}(C)_\cminus = \fOPMaxGround{\bg{}}{\fOPprop{\bg{}}{PU^i(C)}}_\cminus.
\]
Analogous to Theorem~\ref{thm:inductive:prop} we have
\(
{\fOPprop{\bg{}}{PU^i(C)}}_\cminus \cap \GR{\Pi}{X} =\emptyset
\).
Let $(V,E)$ be a maximal support graph of $(\bg{},\fOPprop{\bg{}}{PU^i(C)})$
for some $E \subseteq (\Pi \times \Pi)$.
To prove
\(
\fOPMaxGround{\bg{}}{\fOPprop{\bg{}}{PU^i(C)}}_\cminus \cap \GR{\Pi}{X}
=\emptyset 
\),
it is enough to show that $(\Pi\setminus V) \cap \GR{\Pi}{X}=\emptyset$.

But this holds by 
\[
\Pi\setminus V \subseteq \{\r\mid \pbody{\r}\not\subseteq X\}\subseteq
\{\r\not\in \GR{\Pi}{X}\}.
\]

\paragraph{4:}
$C=PU^0(C)\sqsubseteq PU(C)$ holds by definition of $PU(C)$.
\paragraph{5:}
We have to show
\begin{enumerate}
\item $\fOPprop{\bg{}}{PU(C)}=PU(C)$ and
\item $\fOPMaxGround{\bg{}}{PU(C)}=PU(C)$.
\end{enumerate}
By finiteness, there exists an $n < \omega$ s.t.\ $PU(C)=PU^n(C)$ and
$PU^n(C)=PU^{n+1}(C)$.   
\subparagraph{1:}
$PU(C) \sqsubseteq \fOPprop{\bg{}}{PU(C)}$ holds by definition of
$\OPprop{\bg{}}$. 
And, we have
$PU^{n+1}(C)=\fOPMaxGround{\bg{}}{\fOPprop{\bg{}}{PU^n(C)}}=PU^n(C)$. 
Thus, we have ${\fOPprop{\bg{}}{PU(C)}} = PU(C)$.
\subparagraph{2:}
By use of Theorem~\ref{thm:S:monoton} we have
\begin{eqnarray*}
  PU(C) & =& PU^n(C)\\
        & = & \fOPMaxGround{\bg{}}{\fOPprop{\bg{}}{PU^{n-1}(C)}}\\
        & = &
    \fOPMaxGround{\bg{}}{\fOPMaxGround{\bg{}}{\fOPprop{\bg{}}{PU^{n-1}(C)}}}\\ 
        & = & \fOPMaxGround{\bg{}}{PU^n(C)}\\
        & = & \fOPMaxGround{\bg{}}{PU(C)}.
\end{eqnarray*}

\paragraph{6:}
 Assume there exists a $Q(C)\not=PU(C)$ such that $Q(C) \sqsubseteq PU(C)$ and
   $Q(C)$  is a partial coloring closed under $\mathcal{P}_{\bg{}}$ and
   $\mathcal{U}_{\bg{}}$. 

 Given $Q^0(C)=C=PU^0(C)$, there must exists a (minimal) $i < \omega $ s.t.\
 \(
 Q^{i+1}(C) \not=PU^{i+1}(C)
 \)
 and
 \(
 Q^j(C) = PU^j(C)
 \)
 for all $j\leq i$.
 But then
 \begin{eqnarray*}
 Q^{i+1}(C) &=& \mathcal{U}_{\bg{}}{Q^i(C)} \sqcup \mathcal{P}_{\bg{}}{Q^i(C)}
               \sqcup Q^i(C) \\
&\not= & 
              \mathcal{U}_{\bg{}}{PU^i(C)} \sqcup \mathcal{P}_{\bg{}}{PU^i(C)}
               \sqcup PU^i(C) = PU^{i+1}(C). 
 \end{eqnarray*}
 But this is a contradiction since $PU^i(C)=Q^i(C)$.
 For this reason, $PU(C)$ is the $\sqsubseteq$-smallest partial coloring
   closed under  $\mathcal{P}_{\bg{}}$ and $\mathcal{U}_{\bg{}}$.

\paragraph{7:}
This follows directly from condition~5 and by definition of
$(\OPprop{}\OPMaxGround{})^\ast_{\bg{}}$. 
\end{proof}
%%% Local Variables: 
%%% mode: latex
%%% TeX-master: "~/tex/Papers/GraphLP/Grundlagen/paper"
%%% End: 

%% file: Proofs/LongVersion/thm_inductive_propPV.tex
\begin{proof}{thm:inductive:propPV}
Holds analogous to Theorem~\ref{thm:inductive:propPU}.
\end{proof}
%%% Local Variables: 
%%% mode: latex
%%% TeX-master: "~/tex/Papers/GraphLP/Grundlagen/paper"
%%% End: 